\documentclass[twoside,11pt]{article}

\PassOptionsToPackage{sort}{natbib}

\usepackage{amsmath}
\usepackage{mathtools}
\usepackage{bbold}
\usepackage{tikz}
\usepackage{xcolor}
\usepackage{subcaption}
\usepackage[ruled,algo2e,linesnumbered]{algorithm2e}

\usepackage[toc,page]{appendix}

\usepackage{etoolbox}
\appto\appendix{\addtocontents{toc}{\protect\setcounter{tocdepth}{2}}}

\usepackage[preprint]{jmlr2e}

\usepackage{tikz}
\usepackage{tikz-qtree}
\usepackage{tikz-3dplot}
\usetikzlibrary{fit}
\usepackage{multirow}
\usetikzlibrary{shadows}
\usepackage{appendix}

\usepackage{longtable}
\usepackage{makecell}
\usepackage{array}

\usepackage{rotating}
\usepackage{enumitem}
\setlist{nolistsep}
\usepackage{amsmath,bm}
\usepackage{bbm}
\usepackage{enumitem}
\usepackage{framed}
\usepackage{pdfpages}

\usepackage{CJK}
\usepackage[utf8]{inputenc}

\usepackage{tikz}
\usepackage{tikz-qtree}
\usepackage{tikz-3dplot}
\usetikzlibrary{fit}
\usetikzlibrary{backgrounds}
\usepackage{pgfplots}
\usepackage{xcolor}
\definecolor{ocrebase}{RGB}{243,102,25}
\definecolor{ocre}{RGB}{0,0,0}
\definecolor{amber}{rgb}{1.0, 0.75, 0.0}
\definecolor{ublue}{rgb}{0.152,0.250,0.545}
\definecolor{ugreen}{rgb}{0,0.5,0}
\definecolor{lgreen}{rgb}{0.9,1,0.8}
\definecolor{lightgreen}{rgb}{0.56, 0.93, 0.56}
\definecolor{kellygreen}{rgb}{0.3, 0.73, 0.09}
\definecolor{xtgreen}{rgb}{0.914,0.945,0.902}
\definecolor{lightgray}{gray}{0.85}
\definecolor{darkblue}{rgb}{0, 0, 0.5}
\definecolor{shadecolor}{rgb}{0.96,0.96,0.93}

\graphicspath{{figs/}}

\makeatletter
    \newtheorem{case@}{Case}
    
\makeatother

\DeclareFontFamily{U}{matha}{\hyphenchar\font45}
\DeclareFontShape{U}{matha}{m}{n}{
  <-6> matha5 <6-7> matha6 <7-8> matha7
  <8-9> matha8 <9-10> matha9
  <10-12> matha10 <12-> matha12
  }{}

\DeclareSymbolFont{matha}{U}{matha}{m}{n}
\DeclareMathSymbol{\Lt}{3}{matha}{"CE}

\DeclareMathOperator*{\argmax}{arg\,max}
\DeclareMathOperator*{\argmin}{arg\,min}

\DeclareMathOperator{\pPr}{Pr}

\renewcommand{\Pr}{\pPr}

\newlength{\myl}
\usepackage{lastpage}

 \ShortHeadings{Introduction to Transformers}
   {Xiao and Zhu}
\firstpageno{1}

\begin{document}
\begin{CJK}{UTF8}{gbsn}
\title{Introduction to Transformers: an NLP Perspective
}

\author{\
      \\
      \\
       \name Tong Xiao \email xiaotong@mail.neu.edu.cn \\
       \addr NLP Lab., Northeastern University, Shenyang, China\\
       NiuTrans Research, Shenyang, China
       \AND
       \name Jingbo Zhu \email zhujingbo@mail.neu.edu.cn \\
       \addr NLP Lab., Northeastern University, Shenyang, China\\
       NiuTrans Research, Shenyang, China
       }

\maketitle

\vspace{2em}

\begin{abstract}%
Transformers have dominated empirical machine learning models of natural language processing. In this paper, we introduce basic concepts of Transformers and present key techniques that form the recent advances of these models. This includes a description of the standard Transformer architecture, a series of model refinements, and common applications. Given that Transformers and related deep learning techniques might be evolving in ways we have never seen, we cannot dive into all the model details or cover all the technical areas. Instead, we focus on just those concepts that are helpful for gaining a good understanding of Transformers and their variants. We also summarize the key ideas that impact this field, thereby yielding some insights into the strengths and limitations of these models.

\vspace{0.5em}
\noindent arxiv: {\color{blue} \url{https://arxiv.org/abs/2311.17633}}

\vspace{0.5em}
\noindent github: {\color{blue} \url{https://github.com/NiuTrans/Introduction-to-Transformers}}
\end{abstract}

\clearpage
\tableofcontents
\clearpage

\section{Background}

\noindent Transformers are a type of neural network \citep{Vaswani-etal:2017Transformer}. They were originally known for their strong performance in machine translation, and are now a de facto standard for building large-scale self-supervised learning systems \citep{devlin-etal:2019bert,brown-etal:2020language}. The past few years have seen the rise of Transformers not only in natural language processing (NLP) but also in several other fields, such as computer vision and multi-modal processing. As Transformers continue to mature, these models are playing an increasingly important role in the research and application of artificial intelligence (AI).

Looking back at the history of neural networks, Transformers have not been around for a long time. While Transformers are ``newcomers'' in NLP, they were developed on top of several ideas, the origins of which can be traced back to earlier work, such as word embedding \citep{Bengio-et-al:2003,Mikolov-et-al:2013distributed} and attention mechanisms \citep{bahdanau2014neural,luong-etal:2015effective}. As a result, Transformers can benefit from the advancements of different sub-fields of deep learning, and provide an elegant way to combine these neural models. On the other hand, Transformers are unique, and differ from previous models in several ways. First, they do not depend on recurrent or convolutional neural networks for modeling sequences of words, but use only attention mechanisms and feed-forward neural networks. Second, the use of self-attention in Transformers makes it easier to deal with global contexts and dependencies among words. Third, Transformers are very flexible architectures and can be easily modified to accommodate different tasks.

The widespread use of Transformers motivates the development of cutting-edge techniques in deep learning. For example, there are significant refinements in self-attention mechanisms, which have been incorporated into many state-of-the-art NLP systems. The resulting techniques, together with the progress in self-supervised learning, have led us to a new era of AI: we are beginning to obtain models of universal language understanding, generation and reasoning. This has been evidenced by recent Transformer-based large language models (LLMs) which demonstrate amazing performance across a broad variety of tasks \citep{bubeck-etal:2023sparks}.

This paper provides an introduction to Transformers while reflecting the recent developments in applying these models to different problems. However, Transformers are so successful that there have been numerous related studies and we cannot give a full description of them. Therefore, we focus this work on the core ideas of Transformers, and present a basic description of the common techniques. We also discuss some recent advances in Transformers, such as model improvements for efficiency and accuracy considerations. Because the field is very active and new techniques are coming out every day, it is impossible to survey all the latest literature and we are not attempting to do so. Instead, we focus on just those concepts and algorithms most relevant to Transformers, aimed at the people who wish to get a general understanding of these models.


\section{The Basic Model}
\label{sec:transformer-the-basic-model}

\noindent Here we consider the model presented in \citet{Vaswani-etal:2017Transformer}'s work. We start by considering the Transformer architecture and then discuss its components in detail.

\subsection{The Transformer Architecture}
\label{sec:transformer-architecture}

\begin{figure}[!htp]
\centering

\begin{tikzpicture}

\def\nodehsep{1.2em} 
\def\whsep{1.5em} 
\def\nodewd{10.5em} 
\def\sep{0.5em} 
\def\wsep{4em} 
\def\glsep{0.05em} 
\def\wordhsep{0.0ex} 

\begin{scope}
\tikzstyle{basenode} = [minimum width=\nodewd,inner sep=0.5pt,rounded corners=1.5pt,draw,thick,font=\footnotesize];

\tikzstyle{Sanode} = [basenode,minimum height=2.6em,fill=orange!30];
\tikzstyle{Resnode} = [basenode,minimum height=1.6em,fill=yellow!30];
\tikzstyle{ffnnode} = [basenode,minimum height=2.6em,fill=blue!20];
\tikzstyle{outputnode} = [basenode,minimum height=1.6em,fill=cyan!20];

\tikzstyle{posnode} = [basenode,minimum height=1.6em];
\tikzstyle{standard} = [thick]
\tikzstyle{gnode} = [minimum width=0.6em,fill=kellygreen!70]

\node [Sanode,anchor=west,align=center] (sa1) at (0,0) {{$\textbf{Self-Attention}$} \\ [\wordhsep]{$\mathrm{Layer}_{\mathrm{self}}(\cdot)$}};
\node [Resnode,anchor=south] (res1) at ([yshift=\nodehsep]sa1.north) {{$\textbf{Add \& LayerNorm}$}};
\node [ffnnode,anchor=south,align=center] (ffn1) at ([yshift=\nodehsep]res1.north) {{$\textbf{Feed-Forward Network}$} \\ [\wordhsep]{$\mathrm{Layer}_{\mathrm{ffn}}(\cdot)$}};
\node [Resnode,anchor=south] (res2) at ([yshift=\nodehsep]ffn1.north) {{$\textbf{Add \& LayerNorm}$}};

\node [posnode,anchor=north] (pos1) at ([yshift=-\whsep,xshift=0em]sa1.south) {};
\node [posnode,minimum width=5.25em,anchor=west,fill=red!30] (pos11) at ([yshift=0em,xshift=0em]pos1.west) {{$\textbf{Word}$}};
\node [posnode,minimum width=5.25em,anchor=east,fill=black!10!white] (pos12) at ([yshift=-0em,xshift=0em]pos1.east) {{$\textbf{Position}$}};

\node [anchor=north] (inputs) at ([yshift=-\whsep]pos1.south) {{$x_1 \ldots x_m$}};

\draw [->,standard] (sa1.north) -- (res1.south);
\draw [->,standard] (res1.north) -- (ffn1.south);
\draw [->,standard] (ffn1.north) -- (res2.south);
\draw [->,standard] ([yshift=-0em]pos1.north) -- (sa1.south);
\draw [->,standard] ([yshift=-0em]inputs.north) -- ([yshift=0em]pos1.south);

\node [Sanode,anchor=west,align=center] (sa2) at ([xshift=2*\wsep]sa1.east) {{$\textbf{Self-Attention}$} \\ [\wordhsep]{$\mathrm{Layer}_{\mathrm{self}}(\cdot)$}};
\node [Resnode,anchor=south] (res3) at ([yshift=\nodehsep]sa2.north) {{$\textbf{Add \& LayerNorm}$}};
\node [Sanode,anchor=south,align=center] (ed1) at ([yshift=\nodehsep]res3.north) {{$\textbf{Encoder-Decoder Attention}$} \\ [\wordhsep]{$\mathrm{Layer}_{\mathrm{cross}}(\cdot)$}};
\node [Resnode,anchor=south] (res4) at ([yshift=\nodehsep]ed1.north) {{$\textbf{Add \& LayerNorm}$}};
\node [ffnnode,anchor=south,align=center] (ffn2) at ([yshift=\nodehsep]res4.north) {{$\textbf{Feed-Forward Network}$} \\ [\wordhsep]{$\mathrm{Layer}_{\mathrm{ffn}}(\cdot)$}};
\node [Resnode,anchor=south] (res5) at ([yshift=\nodehsep]ffn2.north) {{$\textbf{Add \& LayerNorm}$}};
\node [outputnode,anchor=south] (o1) at ([yshift=\whsep]res5.north) {{$\mathrm{Softmax}(\mathbf{S}^{L} \mathbf{W}_{o})$}};

\node [posnode,anchor=north] (pos2) at ([yshift=-\whsep,xshift=0em]sa2.south) {};
\node [posnode,minimum width=5.25em,anchor=west,fill=red!30] (pos21) at ([yshift=0em,xshift=0em]pos2.west) {{$\textbf{Word}$}};
\node [posnode,minimum width=5.25em,anchor=east,fill=black!10!white] (pos22) at ([yshift=-0em,xshift=0em]pos2.east) {{$\textbf{Position}$}};

\node [anchor=north] (outputs) at ([yshift=-\whsep]pos2.south) {{$y_0\ y_1 \ldots y_{n-1}$}};

\draw [->,standard] (sa2.north) -- (res3.south);
\draw [->,standard] (res3.north) -- (ed1.south);
\draw [->,standard] (ed1.north) -- (res4.south);
\draw [->,standard] (res4.north) -- (ffn2.south);
\draw [->,standard] (ffn2.north) -- (res5.south);
\draw [->,standard] (res5.north) -- (o1.south);
\draw [->,standard] (o1.north) -- ([yshift=\whsep]o1.north);
\draw [->,standard] ([yshift=-0em]pos2.north) -- (sa2.south);
\draw [->,standard] ([yshift=-0em]outputs.north) -- ([yshift=0em]pos2.south);

\draw[->,standard] ([yshift=-0.5em]sa1.south) -- ([xshift=-\nodewd*0.5 - \sep,yshift=-0.5em]sa1.south) -- ([xshift=- \sep,yshift=0em]res1.west) -- ([xshift=-0.0em]res1.west);
\draw[->,standard] ([yshift=-0.5em]ffn1.south) -- ([xshift=-\nodewd*0.5 - \sep,yshift=-0.5em]ffn1.south) -- ([xshift=- \sep,yshift=-0em]res2.west) -- ([xshift=-0.0em]res2.west);

\draw[->,standard] ([yshift=-0.5em]sa2.south) -- ([xshift=\nodewd*0.5 + \sep,yshift=-0.5em]sa2.south) -- ([xshift= \sep,yshift=0em]res3.east) -- ([xshift=0.0em]res3.east);
\draw[->,standard] ([yshift=-0.5em]ed1.south) -- ([xshift=\nodewd*0.5 + \sep * 1.3,yshift=-0.5em]ed1.south) -- ([xshift=\sep * 1.2,yshift=0em]res4.east) -- ([xshift=0.0em]res4.east);
\draw[->,standard] ([yshift=-0.5em]ffn2.south) -- ([xshift=\nodewd*0.5 + \sep,yshift=-0.5em]ffn2.south) -- ([xshift= \sep,yshift=0em]res5.east) -- ([xshift=0.0em]res5.east);

\draw[->,standard] (res2.north) -- ([yshift=0.5em]res2.north) -- ([xshift=\nodewd*0.5 + \wsep*1,yshift=0.5em]res2.north) -- ([xshift=-\nodewd*0.318 ,yshift=0em]ed1.west) -- ([xshift=0em,yshift=0em]ed1.west);

\node [rectangle,inner sep=0.9em,rounded corners=1pt,very thick,dotted,draw] [fit = (sa1) (res1) (ffn1) (res2)] (encback) {};
\node [rectangle,inner sep=0.9em,rounded corners=1pt,very thick,dotted,draw] [fit = (sa2) (res3) (res5)] (decback) {};

\node [anchor=east] (counte) at ([xshift=-0em,yshift=2em]encback.west) {$L\times$};
\node [anchor=south west] (encoder) at ([xshift=0.0em,yshift=0.2em]encback.north west) {{\textbf{Encoder}}};
\node [anchor=west] (countd) at ([xshift=0em,yshift=2*2em]decback.east) {$\times L$};
\node [anchor=north east] (decoder) at ([xshift=-0.2em,yshift=0em]decback.north west) {{\textbf{Decoder}}};

\node [anchor=south,gnode,minimum height=1.1em] (g23) at ([xshift=3.5em,yshift=\whsep]o1.north) {};
\node [anchor=south east,gnode,minimum height=1.7em] (g22) at ([xshift=-\glsep,yshift=0em]g23.south west) {};
\node [anchor=south east,gnode,minimum height=1.5em] (g21) at ([xshift=-\glsep,yshift=0em]g22.south west) {};
\node [anchor=south west,gnode,minimum height=2.0em] (g24) at ([xshift=\glsep,yshift=0em]g23.south east) {};
\node [anchor=south west,gnode,minimum height=0.7em] (g25) at ([xshift=\glsep,yshift=0em]g24.south east) {};

\node [anchor=south] (dot) at ([xshift=-0em,yshift=2em]g23.south) {$\vdots$};

\node [anchor=south,gnode,minimum height=2.0em] (g13) at ([xshift=0em,yshift=4em]g23.south) {};
\node [anchor=south east,gnode,minimum height=1.3em] (g12) at ([xshift=-\glsep,yshift=0em]g13.south west) {};
\node [anchor=south east,gnode,minimum height=1.5em] (g11) at ([xshift=-\glsep,yshift=0em]g12.south west) {};
\node [anchor=south west,gnode,minimum height=0.6em] (g14) at ([xshift=\glsep,yshift=0em]g13.south east) {};
\node [anchor=south west,gnode,minimum height=1.0em] (g15) at ([xshift=\glsep,yshift=0em]g14.south east) {};


\node [anchor=east] (pr2) at ([xshift=-0.1em,yshift=0.8em]g21.south west) {$\Pr(\cdot|y_0...y_{n-1},x_1...x_m)$};
\node [anchor=east] (pr1) at ([xshift=-0.1em,yshift=0.8em]g11.south west) {$\Pr(\cdot|y_0,x_1...x_m)$};

\node [anchor=east] (dot2) at ([xshift=-3.5em,yshift=-0.1em]dot.west) {$\vdots$};

\end{scope}
\end{tikzpicture}
\caption{The Transformer architecture~\citep{Vaswani-etal:2017Transformer}. There are $L$ stacked layers on each of the encoder and decoder sides. An encoding layer comprises a self-attention sub-layer and an FFN sub-layer. Both of these sub-layers share the same structure which involves a core function (either $\mathrm{Layer}_{\mathrm{self}}(\cdot)$ or $\mathrm{Layer}_{\mathrm{ffn}}(\cdot)$), followed by a residual connection and a layer normalization unit. Each decoding layer has a similar architecture to the encoding layers, but with an additional encoder-decoder attention sub-layer sandwiched between the self-attention and FFN sub-layers. As with most sequence-to-sequence models, Transformer takes $x_1 \cdots x_m$ and $y_0 \cdots y_{i-1}$ for predicting $y_i$. The representation of an input word comprises a sum of a word embedding and a positional embedding. The distributions $\{\Pr(\cdot|y_0 \cdots y_{i-1},x_1 \cdots x_m)\}$ are generated in sequence by a softmax layer, which operates on a linear transformation of the output from the last decoding layer.}
\label{fig:transformer-architecture}
\end{figure}
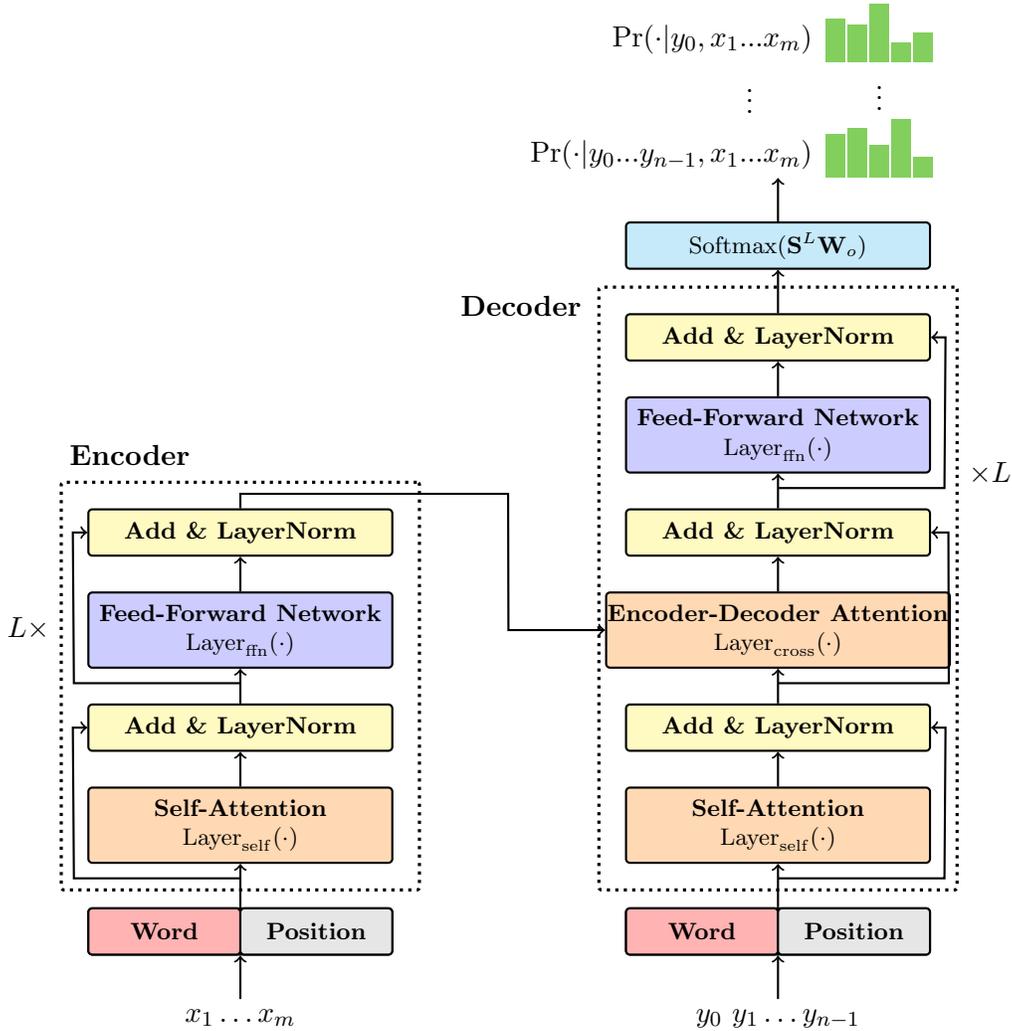

\noindent Figure~\ref{fig:transformer-architecture} shows the standard Transformer model which follows the standard encoder-decoder framework. A Transformer encoder comprises a number of stacked \textbf{encoding layers} (or \textbf{encoding blocks}). Each encoding layer has two different sub-layers (or sub-blocks), called the self-attention sub-layer and the feed-forward neural network (FFN) sub-layer. Suppose we have a source-side sequence $\mathbf{x}=x_1 \cdots x_m$ and a target-side sequence $\mathbf{y}=y_1 \cdots y_n$. The input of an encoding layer is a sequence of $m$ vectors $\mathbf{h}_{1} \cdots \mathbf{h}_{m}$, each having $d_{\mathrm{model}}$ dimensions (or $d$ dimensions for simplicity). Here we use $\mathbf{H} \in \mathbb{R}^{m \times d}$ to denote these input vectors\footnote{Provided $\mathbf{h}_{j} \in \mathbb{R}^{d}$ is a row vector, we have $\mathbf{H} = \begin{bmatrix} \mathbf{h}_{1} \\ \vdots \\ \mathbf{h}_{m}\end{bmatrix}$.}. The self-attention sub-layer first performs a self-attention operation $\mathrm{Att}_{\mathrm{self}}(\cdot)$ on $\mathbf{H}$ to generate an output $\mathbf{C}$:
\begin{eqnarray}
\mathbf{C}& = & \mathrm{Att}_{\mathrm{self}}(\mathbf{H}). \label{eq:self-attention-basic}
\end{eqnarray}

\noindent Here $\mathbf{C}$ is of the same size as $\mathbf{H}$, and can thus be viewed as a new representation of the inputs. Then, a residual connection and a layer normalization unit are added to the output so that the resulting model is easier to optimize.

The original Transformer model employs the \textbf{post-norm} structure where a residual connection is created before layer normalization is performed, as follows:
\begin{eqnarray}
\mathbf{H}_{\mathrm{self}} & = & \mathrm{LNorm}(\mathbf{C} + \mathbf{H}), \label{eq:post-norm-basic}
\end{eqnarray}

\noindent where the addition of $\mathbf{H}$ denotes the residual connection, and $\mathrm{LNorm}(\cdot)$ denotes the layer normalization function. Substituting Eq. (\ref{eq:self-attention-basic}) into Eq. (\ref{eq:post-norm-basic}), we obtain the form of the self-attention sub-layer
\begin{eqnarray}
\mathrm{Layer}_{\mathrm{self}}(\mathbf{H}) & = & \mathbf{H}_{\mathrm{self}} \nonumber \\
                                           & = & \mathrm{LNorm}(\mathrm{Att}_{\mathrm{self}}(\mathbf{H}) + \mathbf{H}). \label{eq:self-attention-layer}
\end{eqnarray}

\noindent The definitions of $\mathrm{LNorm}(\cdot)$ and $\mathrm{Att}_{\mathrm{self}}(\cdot)$ have been given previously, and we will also discuss them later in the section.

The FFN sub-layer takes $\mathbf{H}_{\mathrm{self}}$ and outputs a new representation $\mathbf{H}_{\mathrm{ffn}} \in \mathbb{R}^{m \times d}$. It has the same form as the self-attention sub-layer, with the attention function replaced by the FFN function, given by
\begin{eqnarray}
\mathrm{Layer}_{\mathrm{ffn}}(\mathbf{H}_{\mathrm{self}}) & = & \mathbf{H}_{\mathrm{ffn}} \nonumber \\
                                                          & = & \mathrm{LNorm}(\mathrm{FFN}(\mathbf{H}_{\mathrm{self}}) + \mathbf{H}_{\mathrm{self}}). \label{eq:ffn-layer}
\end{eqnarray}

\noindent Here $\mathrm{FFN}(\cdot)$ can be any feed-forward neural network with non-linear activation functions. The most common structure of $\mathrm{FFN}(\cdot)$ is a two-layer network involving two linear transformations and a ReLU activation function between them.

For deep models, we can stack the above neural networks. Let $\mathbf{H}^{l}$ be the output of layer $l$. Then, we can express $\mathbf{H}^{l}$ as a function of $\mathbf{H}^{l-1}$. We write this as a composition of two sub-layers
\begin{eqnarray}
\mathbf{H}^{l} & = & \mathrm{Layer}_{\mathrm{ffn}}(\mathbf{H}_{\mathrm{self}}^{l}), \\
\mathbf{H}_{\mathrm{self}}^{l} & = & \mathrm{Layer}_{\mathrm{self}}(\mathbf{H}^{l-1}).
\end{eqnarray}

\noindent If there are $L$ encoding layers, then $\mathbf{H}^{L}$ serves as the final output of the encoder. In this case, $\mathbf{H}^{L}$ can be viewed as a learned contextual representation of the input sequence. $\mathbf{H}^{0}$ denotes the initial input to the encoder. In recurrent and convolutional models, $\mathbf{H}^{0}$ typically consists simply of the word embeddings of the input sequence. The Transformer takes a different approach to representing input words and explicitly encodes positional information. In Section~\ref{sec:transformer-pos-encoding}, we will discuss the embedding model used in the Transformer.

The Transformer decoder has a similar structure to the Transformer encoder, comprising $L$ stacked \textbf{decoding layers} (or \textbf{decoding blocks}). Let $\mathbf{S}^{l}$ be the output of the $l$-th decoding layer. We can formulate a decoding layer using the following equations:

\begin{eqnarray}
\mathbf{S}^{l} & = & \mathrm{Layer}_{\mathrm{ffn}}(\mathbf{S}_{\mathrm{cross}}^{l}), \\
\mathbf{S}_{\mathrm{cross}}^{l} & = & \mathrm{Layer}_{\mathrm{cross}}(\mathbf{H}^{L},\mathbf{S}_{\mathrm{self}}^{l-1}) \label{eq:decoder-cross-att-sublayer-simple}, \\
\mathbf{S}_{\mathrm{self}}^{l} & = & \mathrm{Layer}_{\mathrm{self}}(\mathbf{S}^{l-1}). \label{eq:decoder-self-att-sublayer-simple}
\end{eqnarray}

\noindent Here there are three decoder sub-layers. The self-attention and FFN sub-layers are the same as those used in the encoder. $\mathrm{Layer}_{\mathrm{cross}}(\cdot)$ denotes a cross attention sub-layer (or encoder-decoder sub-layer) which models the transformation from the source-side to the target-side. In Section~\ref{sec:transformer-cross-attention}\ we will see that $\mathrm{Layer}_{\mathrm{cross}}(\cdot)$ can be implemented using the same function as $\mathrm{Layer}_{\mathrm{self}}(\cdot)$.

The Transformer decoder outputs a distribution over a vocabulary $V_\mathrm{y}$ at each target-side position. This is achieved by using a softmax layer that normalizes a linear transformation of $\mathbf{S}^{L}$ to distributions of target-side words. To do this, we map $\mathbf{S}^{L}$ to an $n \times |V_{\mathrm{y}}|$ matrix $\mathbf{O}$ by
\begin{eqnarray}
\mathbf{O} & = & \mathbf{S}^{L} \cdot \mathbf{W}_{\mathrm{o}},
\end{eqnarray}

\noindent where $\mathbf{W}_{\mathrm{o}} \in \mathbb{R}^{d \times |V_{\mathrm{y}}|}$ is the parameter matrix of the linear transformation.

Then, the output of the Transformer decoder is given in the form
\begin{eqnarray}
\begin{bmatrix}
\Pr(\cdot|y_0,\mathbf{x})\\
\vdots \\
\Pr(\cdot|y_0 \cdots y_{n-1},\mathbf{x})
\end{bmatrix}
& = &
\mathrm{Softmax}(\mathbf{O}) \nonumber \\
& = &
\begin{bmatrix}
\mathrm{Softmax}(\mathbf{o}_{1})\\
\vdots\\
\mathrm{Softmax}(\mathbf{o}_{n})
\end{bmatrix},
\end{eqnarray}

\noindent where $\mathbf{o}_{i}$ denotes the $i$-th row vector of $\mathbf{O}$, and $y_0$ denotes the start symbol $\langle \mathrm{SOS} \rangle$. Under this model, the probability of $\mathbf{y}$ given $\mathbf{x}$ can be defined as usual,
\begin{eqnarray}
\log \Pr(\mathbf{y}|\mathbf{x}) & = & \sum_{i=1}^{n} \log \Pr(y_{i}|y_0 \cdots y_{i-1},\mathbf{x}).
\end{eqnarray}

This equation resembles the general form of language modeling: we predict the word at position $i$ given all the words up to position $i-1$. Therefore, the target output sequence is shifted one step forward in time relative to the input sequence, that is, the decoder takes $y_0 \cdots y_{n-1}$ as input to predict the output sequence $y_1 \cdots y_{n}$.

The Transformer architecture discussed above has several variants which have been successfully used in different fields of NLP. For example, we can use a Transformer encoder to represent text (referred to as the \textbf{encoder-only architecture}), a Transformer decoder to generate text (referred to as the \textbf{decoder-only architecture}), or a standard encoder-decoder Transformer model to transform an input sequence to an output sequence. In the rest of this chapter, much of the discussion is independent of the particular choice of application, and will be primarily focused on the encoder-decoder architecture. In Section~\ref{sec:transformer-applications}, we will see applications of the encoder-only and decoder-only architectures.

\subsection{Positional Encoding}
\label{sec:transformer-pos-encoding}

\noindent In their original form, both FFNs and attention models used in Transformers ignore an important aspect of sequence modeling, namely that the order of the words plays a crucial role in expressing the meaning of a sequence. This means that the encoder and decoder are insensitive to the positional information of the input words. A simple approach to overcoming this problem is to add positional encodings to the representation of each word in the sequence. More formally, a word $x_j$ can be represented as a $d$-dimensional vector
\begin{eqnarray}
\mathbf{e}_j & = & \mathbf{x}_j + \mathrm{PE}(j).
\end{eqnarray}

\noindent Here $\mathbf{x}_j \in \mathbb{R}^{d}$ is the word embedding, which can be obtained using standard word embedding models. $\mathrm{PE}(j) \in \mathbb{R}^{d}$ is the representation of the position $j$. The vanilla Transformer employs a sinusoidal positional encoding model, which we write in the form
\begin{eqnarray}
\mathrm{PE}(j,2k) & = & \sin(j \cdot \frac{1}{10000^{2k/d}}) \label{eq:pos-sin-encoding-sin-transformer}, \\
\mathrm{PE}(j,2k+1) & = & \cos(j \cdot \frac{1}{10000^{2k/d}}) \label{eq:pos-sin-encoding-cos-transformer},
\end{eqnarray}

\noindent where $\mathrm{PE}(j,k)$ denotes the $k$-th entry of $\mathrm{PE}(j)$. The idea of positional encoding is to distinguish different positions using continuous functions. Here we use the sine and cosine functions with different frequencies. Such a method can be interpreted as a positional numeral system. Because the encoding is based on individual positions, it is also called \textbf{absolute positional encoding}. In Section~\ref{sec:transformer-local-models}, we will see an improvement to this method.

Once we obtain these embeddings, the sequence $\mathbf{e}_1 \cdots \mathbf{e}_m$ is used as input to the Transformer encoder, that is,
\begin{eqnarray}
\mathbf{H}^0 & = & \begin{bmatrix} \mathbf{e}_1 \\ \vdots \\ \mathbf{e}_m \end{bmatrix}.
\end{eqnarray}

\noindent Similarly, we can define the input on the decoder side.

\subsection{Multi-head Self-attention}
\label{sec:multi-head-self-attention}

\noindent The use of self-attention is perhaps one of the most significant advances in sequence-to-sequence models. It attempts to learn and make use of direct interactions between all pairs of inputs. From a representation learning perspective, self-attention models assume that the learned representation at position $i$ (denoted by $\mathbf{c}_{i}$) is a weighted sum of the inputs over the sequence. The output $\mathbf{c}_{i}$ is thus given by
\begin{eqnarray}
\mathbf{c}_i & = & \sum_{j=1}^{m} \alpha_{i,j} \mathbf{h}_{j} \label{eq:transformer-basic-attention-model},
\end{eqnarray}

\noindent where $\alpha_{i,j}$ represents the attention weight assigned to $\mathbf{h}_j$ when computing the representation at position $i$. We can thus view $\mathbf{c}_i$ as a representation of the global context at position $i$. $\alpha_{i,j}$ can be defined in different ways depending on the specific attention model considered. Here we use the scaled dot-product attention function to compute $\alpha_{i,j}$, as follows
\begin{eqnarray}
\alpha_{i,j} & = & \mathrm{Softmax}\left( \mathbf{h}_i \mathbf{h}_j^\top/\beta \right) \nonumber \\
& = & \frac{\exp(\mathbf{h}_i \mathbf{h}_j^\top/\beta)}{\sum_{k=1}^m \exp(\mathbf{h}_i \mathbf{h}_k^\top/\beta)},
\end{eqnarray}

\noindent where the scaling factor  $\beta$ is typically set to $\sqrt{d}$.

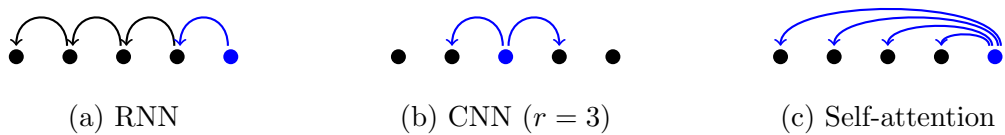
\begin{figure}[!htp]
\centering

\begin{center}

\begin{tikzpicture}

\def\rowsep{0.5cm}
\def\colsep{0.4cm}
\def\ballsize{0.1cm}
\def\stepsize{0.5cm}

\tikzstyle{pnode} = [circle,minimum size=\ballsize,inner sep=2pt,fill=black]

\begin{scope}

\node [anchor=west,pnode] (p1) at (0,0) {};
\node [anchor=west,pnode] (p2) at ([xshift=\stepsize]p1.east) {};
\node [anchor=west,pnode] (p3) at ([xshift=\stepsize]p2.east) {};
\node [anchor=west,pnode] (p4) at ([xshift=\stepsize]p3.east) {};
\node [anchor=west,pnode,fill=blue] (p5) at ([xshift=\stepsize]p4.east) {};

\draw [->,thick] ([yshift=0.1em,xshift=-0.1em]p2.north)  .. controls +(north:0.5cm) and +(north:0.5cm) .. ([yshift=0.1em]p1.north);
\draw [->,thick] ([yshift=0.1em,xshift=-0.1em]p3.north)  .. controls +(north:0.5cm) and +(north:0.5cm) .. ([yshift=0.1em,xshift=0.1em]p2.north);
\draw [->,thick] ([yshift=0.1em,xshift=-0.1em]p4.north)  .. controls +(north:0.5cm) and +(north:0.5cm) .. ([yshift=0.1em,xshift=0.1em]p3.north);
\draw [->,thick,blue] ([yshift=0.1em,xshift=-0.1em]p5.north)  .. controls +(north:0.5cm) and +(north:0.5cm) .. ([yshift=0.1em,xshift=0.1em]p4.north);

\node [anchor=north] (rnnlabel) at ([yshift=-1em]p3.south) {(a) RNN};

\end{scope}

\begin{scope} [xshift=2in]

\node [anchor=west,pnode] (p1) at (0,0) {};
\node [anchor=west,pnode] (p2) at ([xshift=\stepsize]p1.east) {};
\node [anchor=west,pnode,fill=blue] (p3) at ([xshift=\stepsize]p2.east) {};
\node [anchor=west,pnode] (p4) at ([xshift=\stepsize]p3.east) {};
\node [anchor=west,pnode] (p5) at ([xshift=\stepsize]p4.east) {};

\draw [->,thick,blue] ([yshift=0.1em,xshift=-0.1em]p3.north)  .. controls +(north:0.5cm) and +(north:0.5cm) .. ([yshift=0.1em]p2.north);
\draw [->,thick,blue] ([yshift=0.1em,xshift=0.1em]p3.north)  .. controls +(north:0.5cm) and +(north:0.5cm) .. ([yshift=0.1em]p4.north);

\node [anchor=north] (rnnlabel) at ([yshift=-1em]p3.south) {(b) CNN ($r=3$)};

\end{scope}

\begin{scope} [xshift=4in]

\node [anchor=west,pnode] (p1) at (0,0) {};
\node [anchor=west,pnode] (p2) at ([xshift=\stepsize]p1.east) {};
\node [anchor=west,pnode] (p3) at ([xshift=\stepsize]p2.east) {};
\node [anchor=west,pnode] (p4) at ([xshift=\stepsize]p3.east) {};
\node [anchor=west,pnode,fill=blue] (p5) at ([xshift=\stepsize]p4.east) {};

\draw [->,thick,blue] ([yshift=0.1em,xshift=-0.2em]p5.north)  .. controls +(north:0.2cm) and +(north:0.2cm) .. ([yshift=0.1em]p4.north);
\draw [->,thick,blue] ([yshift=0.1em,xshift=-0.1em]p5.north)  .. controls +(north:0.35cm) and +(north:0.35cm) .. ([yshift=0.1em]p3.north);
\draw [->,thick,blue] ([yshift=0.1em,xshift=0.1em]p5.north)  .. controls +(north:0.5cm) and +(north:0.5cm) .. ([yshift=0.1em]p2.north);
\draw [->,thick,blue] ([yshift=0.1em,xshift=0.2em]p5.north)  .. controls +(north:0.65cm) and +(north:0.65cm) .. ([yshift=0.1em]p1.north);

\node [anchor=north] (rnnlabel) at ([yshift=-1em]p3.south) {(c) Self-attention};

\end{scope}

\end{tikzpicture}

\end{center}
\caption{Information flows in recurrent, convolutional, and self-attention models, shown as arrow lines between positions.}
\label{fig:information-flow-rnn-cnn-transformer}
\end{figure}

Compared with conventional recurrent and convolutional models, an advantage of self-attention models is that they shorten the computational ``distance'' between two inputs. Figure~\ref{fig:information-flow-rnn-cnn-transformer} illustrates the information flow in these models. We can see that, given the input at position $i$, self-attention models can directly access any other input. By contrast, recurrent and convolutional models might need two or more jumps to see the entire sequence.

We can adopt a more general perspective on self-attention by using the QKV attention model. Suppose we have a sequence of $\kappa$ queries $\mathbf{Q} = \begin{bmatrix} \mathbf{q}_1 \\ \vdots \\ \mathbf{q}_{\kappa} \end{bmatrix}$, and a sequence of $\psi$ key-value pairs $(\mathbf{K} = \begin{bmatrix} \mathbf{k}_1 \\ \vdots \\ \mathbf{k}_{\psi} \end{bmatrix}, \mathbf{V} = \begin{bmatrix} \mathbf{v}_1 \\ \vdots \\ \mathbf{v}_{\psi} \end{bmatrix})$. The output of the model is a sequence of vectors, each corresponding to a query. The form of the QKV attention is given by
\begin{eqnarray}
\mathrm{Att}_{\mathrm{qkv}}(\mathbf{Q}, \mathbf{K}, \mathbf{V}) & = & \mathrm{Softmax}(\frac{\mathbf{Q} \mathbf{K}^\top}{\sqrt{d}}) \mathbf{V}.
\end{eqnarray}

We can write the output of the QKV attention model as a sequence of row vectors
\begin{eqnarray}
\mathbf{C} & = & \begin{bmatrix} \mathbf{c}_1 \\ \vdots \\ \mathbf{c}_{\kappa} \end{bmatrix} \nonumber \\
           & = & \mathrm{Att}_{\mathrm{qkv}}(\mathbf{Q}, \mathbf{K}, \mathbf{V}) \label{eq:transformer-qkv-attention}.
\end{eqnarray}

To apply this equation to self-attention, we simply have
\begin{eqnarray}
\mathbf{H}^{q} & = & \mathbf{H} \mathbf{W}^{q}, \\
\mathbf{H}^{k} & = & \mathbf{H} \mathbf{W}^{k}, \\
\mathbf{H}^{v} & = & \mathbf{H} \mathbf{W}^{v},
\end{eqnarray}

\noindent where $\mathbf{W}^{q}, \mathbf{W}^{k}, \mathbf{W}^{v} \in \mathbb{R}^{d \times d}$ represent linear transformations of $\mathbf{H}$.

By considering Eq. (\ref{eq:self-attention-basic}), we then obtain
\begin{eqnarray}
\mathbf{C} & = & \mathrm{Att}_{\mathrm{self}}(\mathbf{H}) \nonumber \\
           & = & \mathrm{Att}_{\mathrm{qkv}}(\mathbf{H}^{q}, \mathbf{H}^{k}, \mathbf{H}^{v}) \nonumber \\
           & = & \mathrm{Softmax}(\frac{\mathbf{H}^{q} [\mathbf{H}^{k}]^\top}{\sqrt{d}}) \mathbf{H}^{v}.
\end{eqnarray}

\noindent Here $\mathrm{Softmax}(\frac{\mathbf{H}^{q} [\mathbf{H}^{k}]^\top}{\sqrt{d}})$ is an $m \times m$ matrix in which each row represents a distribution over all input positions, that is
\begin{eqnarray}
\textrm{row $i$} & = & \begin{bmatrix} \alpha_{i,1} & \cdots & \alpha_{i,m} \end{bmatrix}.
\end{eqnarray}

We can improve the above self-attention model by using a technique called \textbf{multi-head attention}. This method can be motivated from the perspective of learning from multiple lower-dimensional feature subspaces. It projects the input into multiple subspaces and learns separate representations in each. Specifically, we project the entire input space into $\tau$ subspaces (referred to as \textbf{heads}). For example, we transform $\mathbf{H} \in \mathbb{R}^{m \times d}$ into $\tau$ matrices of size $m \times \frac{d}{\tau}$, denoted by $\{\mathbf{H}_{1}^{\mathrm{head}}, \dots, \mathbf{H}_{\tau}^{\mathrm{head}}\}$. The attention model is then applied $\tau$ times, once for each head. Finally, the outputs of these model runs are concatenated and transformed by a linear projection. This procedure can be expressed by
\begin{eqnarray}
\mathbf{C} & = & \mathrm{Merge}(\mathbf{C}_{1}^{\mathrm{head}},\cdots,\mathbf{C}_{\tau}^{\mathrm{head}}) \mathbf{W}_{c}.
\end{eqnarray}

\noindent For each head $h$,
\begin{eqnarray}
\mathbf{C}_{h}^{\mathrm{head}} & = & \mathrm{Softmax} \left( \frac{\mathbf{H}_{h}^{q} [\mathbf{H}_{h}^{k}]^\top}{\sqrt{d}} \right) \mathbf{H}_{h}^{v}, \\
\mathbf{H}_{h}^{q} & = & \mathbf{H} \mathbf{W}_{h}^{q}, \\
\mathbf{H}_{h}^{k} & = & \mathbf{H} \mathbf{W}_{h}^{k}, \\
\mathbf{H}_{h}^{v} & = & \mathbf{H} \mathbf{W}_{h}^{v}.
\end{eqnarray}

\noindent Here $\mathrm{Merge}(\cdot)$ is the concatenation function, and $\mathrm{Att}_{\mathrm{qkv}}(\cdot)$ is the attention function described in Eq. (\ref{eq:transformer-qkv-attention}). $\mathbf{W}_{h}^{q}, \mathbf{W}_{h}^{k}, \mathbf{W}_{h}^{v} \in \mathbb{R}^{d \times \frac{d}{\tau}}$ are the parameters of the projections from a $d$-dimensional space to a $\frac{d}{\tau}$-dimensional space for the queries, keys, and values. Thus, $\mathbf{H}_{h}^{q}$, $\mathbf{H}_{h}^{k}$, $\mathbf{H}_{h}^{v}$, and $\mathbf{C}_{h}^{\mathrm{head}}$ are all $m \times \frac{d}{\tau}$ matrices. $\mathrm{Merge}(\mathbf{C}_{1}^{\mathrm{head}}, \cdots ,\mathbf{C}_{\tau}^{\mathrm{head}})$ produces an $m \times d$ matrix. It is then transformed by a linear mapping $\mathbf{W}_{c} \in \mathbb{R}^{d \times d}$, leading to the final result $\mathbf{C} \in \mathbb{R}^{m \times d}$.

Although the notation may seem tedious, implementing multi-head models is straightforward using modern deep learning toolkits. A common method in Transformer-based systems is to store inputs from all the heads in data structures called tensors, enabling us to leverage parallel computing resources for greater efficiency. A more general discussion of the QKV attention and multi-head attention models can be found in the literature on encoder-decoder models.

\subsection{Layer Normalization}
\label{sec:transformer-layer-normalization}

\noindent Layer normalization provides a simple and effective means to make the training of neural networks more stable by standardizing the activations of the hidden layers in a layer-wise manner. As introduced by \citet{Ba-etal:2016layer}, given a layer's output $\mathbf{h} \in \mathbb{R}^{d}$, the layer normalization method computes a standardized output $\mathrm{LNorm}(\mathbf{h}) \in \mathbb{R}^{d}$ by
\begin{eqnarray}
\mathrm{LNorm}(\mathbf{h}) & = & \boldsymbol{\alpha} \odot \frac{\mathbf{h} - \mu}{\sigma + \epsilon} + \boldsymbol{\beta}. \label{eq:layer-normalization}
\end{eqnarray}

\noindent Here $\mu \in \mathbb{R}$ and $\sigma \in \mathbb{R}$ are the scalar mean and standard deviation of the activations, respectively. Let $h_k$ be the $k$-th dimension of $\mathbf{h}$. The scalars $\mu$ and $\sigma$ are defined as
\begin{eqnarray}
\mu & = & \frac{1}{d} \sum_{k=1}^{d} h_k,  \\
\sigma & = & \sqrt{\frac{1}{d} \sum_{k=1}^{d} (h_k-\mu)^2 }.
\end{eqnarray}

\noindent Here $\boldsymbol{\alpha} \in \mathbb{R}^{d}$ and $\boldsymbol{\beta} \in \mathbb{R}^{d}$ are the scale and shift parameters. They can be treated as parameters of layer normalization, whose values are learned jointly with other parameters of the Transformer model. The addition of $\epsilon$ to $\sigma$ is used for numerical stability. In general, $\epsilon$ is chosen to be a very small number.

We illustrate the layer normalization method for the hidden states of an encoder in the following example (assume that $m=4$, $d=3$, $\boldsymbol{\alpha}=\mathbf{1}$, $\boldsymbol{\beta}=\mathbf{0}$, and $\epsilon=0.1$).
\begin{equation}
\begin{matrix}
\mathbf{h}_1 \\
\mathbf{h}_2 \\
\mathbf{h}_3 \\
\mathbf{h}_4
\end{matrix}
\begin{bmatrix}
1 & 1 & 2 \\
0.9 & 0.9 & 0 \\
0.7 & 0.8 & 0 \\
3 & 1 & 7
\end{bmatrix}
\ \ \
\begin{matrix}
\mu = 1.3,\ \sigma=0.5 \\
\mu = 0.6,\ \sigma=0.4 \\
\mu = 0.5,\ \sigma=0.4 \\
\mu = 3.7,\ \sigma=2.5
\end{matrix}
\ \ \ \implies\ \ \
\begin{bmatrix}
\frac{{\color{red} 1}-1.3}{0.5 + 0.1} & \frac{{\color{red} 1}-1.3}{0.5 + 0.1} & \frac{{\color{red} 2}-1.3}{0.5 + 0.1} \\
\frac{{\color{red} 0.9}-0.6}{0.4 + 0.1} & \frac{{\color{red} 0.9}-0.6}{0.4 + 0.1} & \frac{{\color{red} 0}-0.6}{0.4 + 0.1} \\
\frac{{\color{red} 0.7}-0.5}{0.4 + 0.1} & \frac{{\color{red} 0.8}-0.5}{0.4 + 0.1} & \frac{{\color{red} 0}-0.5}{0.4 + 0.1} \\
\frac{{\color{red} 3}-3.7}{2.5 + 0.1} & \frac{{\color{red} 1}-3.7}{2.5 + 0.1} & \frac{{\color{red} 7}-3.7}{2.5 + 0.1}
\end{bmatrix}
\nonumber
\end{equation}

As discussed in Section~\ref{sec:transformer-architecture}, the layer normalization unit in each sub-layer is used to standardize the output of a residual block. Here we describe a more general formulation for this structure. Suppose that $F(\cdot)$ is a neural network sub-layer. Then, the post-norm structure of $F(\cdot)$ is given by:
\begin{eqnarray}
\mathbf{H}_{\mathrm{out}} & = & \mathrm{LNorm}(F(\mathbf{H}_{\mathrm{in}}) + \mathbf{H}_{\mathrm{in}}), \label{eq:post-norm-simple}
\end{eqnarray}

\noindent where $\mathbf{H}_{\mathrm{in}}$ and $\mathbf{H}_{\mathrm{out}}$ are the input and output of this sub-layer. Clearly, Eq. (\ref{eq:self-attention-layer}) is an instance of this equation.

An alternative approach to incorporating layer normalization and residual connections into the modeling is to execute the $\mathrm{LNorm}(\cdot)$ function right before the $F(\cdot)$ function, and to establish an identity mapping from the input to the output of the entire sub-layer. This structure, known as the \textbf{pre-norm} structure, can be expressed in the form
\begin{eqnarray}
\mathbf{H}_{\mathrm{out}} & = & F(\mathrm{LNorm}(\mathbf{H}_{\mathrm{in}})) + \mathbf{H}_{\mathrm{in}}.
\end{eqnarray}

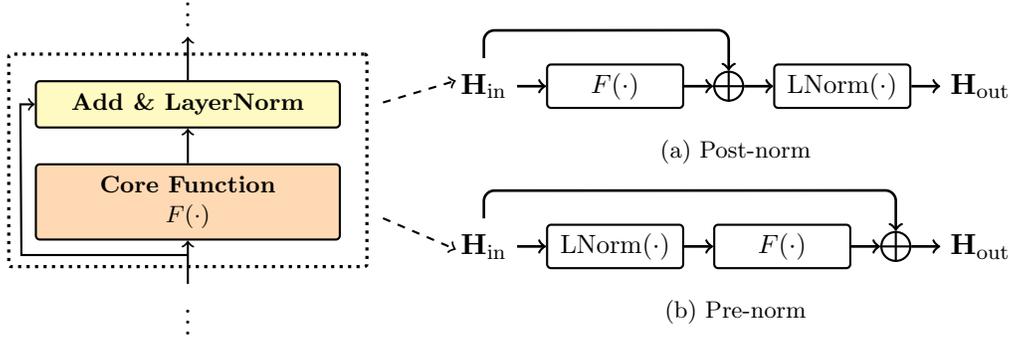
\begin{figure}[!htp]
\centering

\begin{tikzpicture}

\def\nodehsep{1.2em} 
\def\whsep{1.5em} 
\def\nodewd{10.5em} 
\def\sep{0.5em} 
\def\normsep{1.0em} 
\def\drawsep{1.16em}
\def\wordhsep{0.0ex} 

\tikzstyle{basenode} = [minimum width=\nodewd,inner sep=0.5pt,rounded corners=1.5pt,draw,thick,font=\footnotesize];
\tikzstyle{Sanode} = [basenode,minimum height=2.6em,fill=orange!30];
\tikzstyle{Resnode} = [basenode,minimum height=1.6em,fill=yellow!30];

\tikzstyle{normbasenode} = [inner sep=0.5pt,rounded corners=2pt,thick,minimum width=4.7em,minimum height=1.5em,inner sep=0.5pt,rounded corners=1.5pt,draw,thick,font=\footnotesize];
\tikzstyle{lnnode} = [normbasenode];
\tikzstyle{fnode} = [normbasenode];

\tikzstyle{standard} = [thick]


\begin{scope}

\node [Sanode,anchor=west,align=center] (sa1) at (0,0) {{$\textbf{Core Function}$} \\ [\wordhsep]{$F(\cdot)$}};
\node [Resnode,anchor=south] (res1) at ([yshift=\nodehsep]sa1.north) {{$\textbf{Add \& LayerNorm}$}};

\node [anchor=north] (inputs) at ([yshift=-\whsep]sa1.south) {{$\vdots$}};
\node [anchor=south] (outputs) at ([yshift=\whsep]res1.north) {{$\vdots$}};

\draw [->,standard] ([yshift=-\whsep]sa1.south) -- (sa1.south);
\draw [->,standard] (sa1.north) -- (res1.south);
\draw [->,standard] (res1.north) -- ([yshift=\whsep]res1.north);

\draw[->,standard] ([yshift=-0.5em]sa1.south) -- ([xshift=-\nodewd*0.5 - \sep,yshift=-0.5em]sa1.south) -- ([xshift=- \sep,yshift=0em]res1.west) -- ([xshift=-0.0em]res1.west);

\node [rectangle,inner sep=0.9em,rounded corners=1pt,very thick,dotted,draw] [fit = (sa1) (res1)] (blockback) {};

\end{scope}

\begin{scope}[xshift=2.6in,yshift=4.0em]

\node [anchor=east] (x1) at (-0.5em, 0) {$\mathbf{H}_{\mathrm{in}}$};
\node [anchor=west,fnode] (F1) at ([xshift=\normsep]x1.east){\small{$F(\cdot)$}};
\node [anchor=west,circle,draw,minimum size=1em,thick] (n1) at ([xshift=\normsep]F1.east) {};
\node [anchor=west,lnnode] (ln1) at ([xshift=\normsep]n1.east){\small{$\mathrm{LNorm}(\cdot)$}};
\node [anchor=west] (x2) at ([xshift=\normsep]ln1.east) {$\mathbf{H}_{\mathrm{out}}$};

\node [anchor=north] (x3) at ([yshift=-4em]x1.south) {$\mathbf{H}_{\mathrm{in}}$};
\node [anchor=west,lnnode] (F2) at ([xshift=\normsep]x3.east){\small{$\mathrm{LNorm}(\cdot)$}};
\node [anchor=west,fnode] (ln2) at ([xshift=\normsep]F2.east){\small{$F(\cdot)$}};
\node [anchor=west,circle,draw,,minimum size=1em,thick] (n2) at ([xshift=\normsep]ln2.east){};
\node [anchor=west] (x4) at ([xshift=\normsep]n2.east) {$\mathbf{H}_{\mathrm{out}}$};

\draw[->, line width=1pt] (x1.east)--(F1.west);
\draw[->, line width=1pt] (F1.east)--(n1.west);
\draw[->, line width=1pt] (n1.east)--(ln1.west);
\draw[->, line width=1pt] (ln1.east)--(x2.west);
\draw[->, line width=1pt] (x3.east)--(F2.west);
\draw[->, line width=1pt] (F2.east)--(ln2.west);
\draw[->, line width=1pt] (ln2.east)--(n2.west);

\draw[->, line width=1pt] (n2.east)--(x4.west);
\draw[->,rounded corners,line width=1pt] ([yshift=0.2em]x1.north) -- ([yshift=\drawsep]x1.north) -- ([yshift=1.4em]n1.north) -- (n1.north);
\draw[->,rounded corners,line width=1pt] ([yshift=0.2em]x3.north) -- ([yshift=\drawsep]x3.north) -- ([yshift=1.4em]n2.north) -- (n2.north);

\draw[-,thick] (n1.west)--(n1.east);
\draw[-,thick] (n1.north)--(n1.south);
\draw[-,thick] (n2.west)--(n2.east);
\draw[-,thick] (n2.north)--(n2.south);

\node [anchor=south] (k1) at ([yshift=-0.1em]x1.north){};
\node [anchor=south] (k2) at ([yshift=-0.1em]x3.north){};
\begin{pgfonlayer}{background}
\node [rectangle,inner sep=0.3em,fill=white] [fit = (x1) (F1) (n1) (ln1) (x2) (k1)] (box0) {};
\node [rectangle,inner sep=0.3em,fill=white] [fit = (x3) (F2) (n2) (ln2) (x4) (k2)] (box1) {};
\end{pgfonlayer}

\node [anchor=north] (c1) at ([yshift=-0.4em]box0.south){\footnotesize {(a) Post-norm}};
\node [anchor=north] (c2) at ([yshift=-0.4em]box1.south){\footnotesize {(b) Pre-norm}};
\end{scope}

\draw[->,dashed,thick] ([xshift=0.5em,yshift=2em]blockback.east)--([xshift=0.5em,yshift=-1.5em]box0.north west);
\draw[->,dashed,thick] ([xshift=0.5em,yshift=-2em]blockback.east)--([xshift=0.5em,yshift=1em]box1.south west);

\end{tikzpicture}
\caption{The post-norm and pre-norm structures. $F(\cdot) = $ core function, $\mathrm{LNorm}(\cdot) =$ layer normalization, and $\oplus=$ residual connection.}
\label{fig:post-norm-and-pre-norm}
\end{figure}

Both post-norm and pre-norm Transformer models are widely used in NLP systems. See Figure~\ref{fig:post-norm-and-pre-norm} for a comparison of these two structures. In general, residual connections are considered an effective means to make the training of multi-layer neural networks easier. In this sense, the pre-norm Transformer seems promising because it follows the convention that a residual connection is created to bypass the whole network and that the identity mapping from the input to the output leads to easier optimization of deep models. However, when considering the expressive power of a model, there may be modeling advantages to using the post-norm Transformer, because it does not rely as heavily on residual connections and enforces more sophisticated modeling for representation learning. In Section~\ref{sec:transformer-deep-models}, we will present a further discussion on this issue.

\subsection{Feed-forward Neural Networks}
\label{sec:transformer-fnn}

\noindent The use of FFNs in Transformers is inspired in part by the fact that complex outputs can be formed by transforming the inputs through non-linearities. While the self-attention model itself has some non-linearity (in $\mathrm{Softmax}(\cdot)$), a more common way to introduce non-linearity is to consider additional layers with non-linear activation functions and linear transformations. Given an input $\mathbf{H}_{\mathrm{in}} \in \mathbb{R}^{m \times d}$ and an output $\mathbf{H}_{\mathrm{out}} \in \mathbb{R}^{m \times d}$, the $\mathbf{H}_{\mathrm{out}} = \mathrm{FFN}(\mathbf{H}_{\mathrm{in}})$ function in the Transformer has the following form
\begin{eqnarray}
\mathbf{H}_{\mathrm{out}} & = &  \mathbf{H}_{\mathrm{hidden}} \mathbf{W}_f + \mathbf{b}_f, \\
\mathbf{H}_{\mathrm{hidden}} & = & \mathrm{ReLU}(\mathbf{H}_{\mathrm{in}}\mathbf{W}_h + \mathbf{b}_h),
\end{eqnarray}

\noindent where $\mathbf{H}_{\mathrm{hidden}} \in \mathbb{R}^{m \times d_{\mathrm{ffn}}}$ represents the hidden states, and $\mathbf{W}_h \in \mathbb{R}^{d \times d_{\mathrm{ffn}}}$, $\mathbf{b}_h \in \mathbb{R}^{d_{\mathrm{ffn}}}$, $\mathbf{W}_f \in \mathbb{R}^{d_{\mathrm{ffn}} \times d}$ and $\mathbf{b}_f \in \mathbb{R}^{d}$ are the parameters. This is a two-layer FFN in which the first layer (or hidden layer) introduces a non-linearity through $\mathrm{ReLU}(\cdot)$\footnote{$\mathrm{ReLU}(x) = \max\{0,x\}$.} and the second layer involves only a linear transformation. It is common practice in the Transformer to use a larger hidden layer size. For example, a common choice is $d_{\mathrm{ffn}} = 4d$, that is, the size of each hidden representation is 4 times that of the input.

Note that using a wide FFN sub-layer has proven to be of great practical value in many state-of-the-art systems. However, a consequence of this is that the model's parameter count is dominated by the FFN. Table~\ref{tab:transformer-parameter-num-and-time} shows the number of parameters and time complexities for different modules of a standard Transformer-based system. We see that FFNs dominate the model size when $d_{\mathrm{ffn}}$ is large, though they are not the most time-consuming components. In the case of very large Transformer models, we therefore wish to address this problem to build more efficient systems.

\begin{table}[!t]
\centering
\begingroup
\begin{tabular}{r | r | c | c | r }
\multicolumn{2}{c|}{Sub-model} & \# of Parameters & Time Complexity & $\times$ \\ \hline
\multirow{3}{*}{Encoder} & Multi-head Self-attention & $4d^2$ & $O(m^2 \cdot d)$ & $L$ \\
                         & Feed-forward Network & $2d \cdot d_{\mathrm{ffn}} + d + d_{\mathrm{ffn}}$ & $O(m \cdot d \cdot d_{\mathrm{ffn}})$ & $L$ \\
                         & Layer Normalization & $2d$ & $O(m \cdot d)$ & $2L$ \\ \hline
\multirow{4}{*}{Decoder} & Multi-head Self-attention & $4d^2$ & $O(n^2 \cdot d)$ & $L$ \\
                         & Multi-head Cross-attention & $4d^2$ & $O(m \cdot n \cdot d)$ & $L$ \\
                         & Feed-forward Network & $2d \cdot d_{\mathrm{ffn}} + d + d_{\mathrm{ffn}}$ & $O(n \cdot d \cdot d_{\mathrm{ffn}})$ &  $L$ \\
                         & Layer Normalization & $2d$ & $O(n \cdot d)$ & $3L$
\end{tabular}
\endgroup
\caption{Number of parameters and time complexities of different Transformer modules under various setups. $m =$ source-sequence length, $n=$ target-sequence length, $d=$ default number of dimensions of a hidden layer, $d_{\mathrm{ffn}}=$ number of dimensions of the FFN hidden layer, $\tau=$ number of heads in the attention models, and $L=$ number of encoding or decoding layers. The column $\times$ indicates the number of times a sub-model is applied on the encoder or decoder side. The time complexities are estimated by counting the number of floating-point multiplications.}
\label{tab:transformer-parameter-num-and-time}
\end{table}

\subsection{Decoder-side Attention}
\label{sec:transformer-cross-attention}

\noindent A decoder layer involves two attention sub-layers: the first is a self-attention sub-layer, and the second is a cross-attention sub-layer. These sub-layers are based on either the post-norm or the pre-norm structure, but differ in how the attention functions are defined. Consider, for example, the post-norm structure described in Eq. (\ref{eq:post-norm-simple}). We can define the cross-attention and self-attention sub-layers for a decoding layer to be
\begin{eqnarray}
\mathbf{S}_{\mathrm{cross}} & = & \mathrm{Layer}_{\mathrm{cross}}(\mathbf{H}_{\mathrm{enc}},\mathbf{S}_{\mathrm{self}}) \nonumber \\
& = & \mathrm{LNorm}(\mathrm{Att}_{\mathrm{cross}}(\mathbf{H}_{\mathrm{enc}},\mathbf{S}_{\mathrm{self}}) + \mathbf{S}_{\mathrm{self}}), \\
\mathbf{S}_{\mathrm{self}} & = & \mathrm{Layer}_{\mathrm{self}}(\mathbf{S}) \nonumber \\
& = & \mathrm{LNorm}(\mathrm{Att}_{\mathrm{self}}(\mathbf{S}) + \mathbf{S}),
\end{eqnarray}

\noindent where $\mathbf{S} \in \mathbb{R}^{n \times d}$ is the input to the self-attention sub-layer, $\mathbf{S}_{\mathrm{cross}} \in \mathbb{R}^{n \times d}$ and $\mathbf{S}_{\mathrm{self}} \in \mathbb{R}^{n \times d}$ are the outputs of the sub-layers, and $\mathbf{H}_{\mathrm{enc}} \in \mathbb{R}^{m \times d}$ is the output of the encoder \footnote{For an encoder having $L$ encoder layers, $\mathbf{H}_{\mathrm{enc}} = \mathbf{H}^{L}$.}.

As with conventional attention models, cross-attention is primarily used to model the correspondence between the source-side and target-side sequences. The $\mathrm{Att}_{\mathrm{cross}}(\cdot)$ function is based on the QKV attention model, which generates an output by querying a collection of key-value pairs. More specifically, we define the queries, keys, and values as linear mappings of $\mathbf{S}_{\mathrm{self}}$ and $\mathbf{H}_{\mathrm{enc}}$, as follows
\begin{eqnarray}
\mathbf{S}_{\mathrm{self}}^{q} & = & \mathbf{S}_{\mathrm{self}} \mathbf{W}_{\mathrm{cross}}^{q}, \\
\mathbf{H}_{\mathrm{enc}}^{k} & = & \mathbf{H}_{\mathrm{enc}} \mathbf{W}_{\mathrm{enc}}^{k}, \\
\mathbf{H}_{\mathrm{enc}}^{v} & = & \mathbf{H}_{\mathrm{enc}} \mathbf{W}_{\mathrm{enc}}^{v},
\end{eqnarray}

\noindent where $\mathbf{W}_{\mathrm{cross}}^{q}, \mathbf{W}_{\mathrm{enc}}^{k}, \mathbf{W}_{\mathrm{enc}}^{v} \in \mathbb{R}^{d \times d}$ are the parameters of the mappings. In other words, the queries are defined based on $\mathbf{S}_{\mathrm{self}}$, while the keys and values are defined based on $\mathbf{H}_{\mathrm{enc}}$.

$\mathrm{Att}_{\mathrm{cross}}(\cdot)$ is then defined as
\begin{eqnarray}
\mathrm{Att}_{\mathrm{cross}}(\mathbf{H}_{\mathrm{enc}},\mathbf{S}_{\mathrm{self}}) & = & \mathrm{Att}_{\mathrm{qkv}}(\mathbf{S}_{\mathrm{self}}^{q}, \mathbf{H}_{\mathrm{enc}}^{k}, \mathbf{H}_{\mathrm{enc}}^{v}) \nonumber \\
                                                                                             & = & \mathrm{Softmax} \left(\frac{\mathbf{S}_{\mathrm{self}}^{q} [\mathbf{H}_{\mathrm{enc}}^{k}]^\top}{\sqrt{d}} \right) \mathbf{H}_{\mathrm{enc}}^{v}.
\end{eqnarray}

The $\mathrm{Att}_{\mathrm{self}}(\cdot)$ function has a similar form to $\mathrm{Att}_{\mathrm{cross}}(\cdot)$, with linear mappings of $\mathbf{S}$ taken as the queries, keys, and values, like this
\begin{eqnarray}
\mathrm{Att}_{\mathrm{self}}(\mathbf{S}) & = & \mathrm{Att}_{\mathrm{qkv}}(\mathbf{S}^{q},\mathbf{S}^{k},\mathbf{S}^{v}) \nonumber \\
                                         & = & \mathrm{Softmax}(\frac{\mathbf{S}^{q} [\mathbf{S}^{k}]^\top}{\sqrt{d}} + \mathbf{M}) \mathbf{S}^{v},
\end{eqnarray}

\noindent where $\mathbf{S}^{q} = \mathbf{S} \mathbf{W}_{\mathrm{dec}}^{q}$, $\mathbf{S}^{k} = \mathbf{S} \mathbf{W}_{\mathrm{dec}}^{k}$, and $\mathbf{S}^{v} = \mathbf{S} \mathbf{W}_{\mathrm{dec}}^{v}$ are linear mappings of $\mathbf{S}$ with parameters $\mathbf{W}_{\mathrm{dec}}^{q}, \mathbf{W}_{\mathrm{dec}}^{k}, \mathbf{W}_{\mathrm{dec}}^{v} \in \mathbb{R}^{d \times d}$.

This form is similar to that of Eq. (\ref{eq:transformer-qkv-attention}). However, a key difference compared to encoder self-attention is that the model here must follow the rule of left-to-right generation (see Figure~\ref{fig:self-attention-encoder-vs-decoder}). That is, given a target-side word at position $i$, the model can only see the target-side words in the left context $y_1, \dots, y_{i-1}$. To enforce this, we add a masking variable $\mathbf{M}$ to the unnormalized weight matrix $\frac{\mathbf{S}^{q} [\mathbf{S}^{k}]^\top}{\sqrt{d}}$. Both $\mathbf{M}$ and the weight matrix are of size $n \times n$, so large negative values in $\mathbf{M}$ suppress the corresponding attention scores. To prevent attention to the right context (future words) at step $i$, $\mathbf{M}$ is defined as:
\begin{eqnarray}
M(i,j) & = &
\begin{cases}
0 & i \ge j \\
-\infty & i < j
\end{cases},
\end{eqnarray}

\begin{figure}[!htp]
\centering

\begin{center}

\begin{tikzpicture}

\def\rowsep{0.5cm}
\def\colsep{0.4cm}
\def\ballsize{0.23cm}
\def\stepsize{0.9cm}

\tikzstyle{pnode} = [circle,minimum size=\ballsize,inner sep=0pt,fill=black,thick]

\begin{scope}

\node [anchor=west,pnode] (p11) at (0,0) {};
\node [anchor=west,pnode] (p12) at ([xshift=\stepsize]p11.east) {};
\node [anchor=west,pnode] (p13) at ([xshift=\stepsize]p12.east) {};
\node [anchor=west,pnode] (p14) at ([xshift=\stepsize]p13.east) {};
\node [anchor=west,pnode] (p15) at ([xshift=\stepsize]p14.east) {};

\node [anchor=west,pnode,fill=black!20] (p21) at ([yshift=1.8cm]p11.west) {};
\node [anchor=west,pnode,fill=black!20] (p22) at ([xshift=\stepsize]p21.east) {};
\node [anchor=west,pnode,fill=black!20] (p23) at ([xshift=\stepsize]p22.east) {};
\node [anchor=west,pnode] (p24) at ([xshift=\stepsize]p23.east) {};
\node [anchor=west,pnode,fill=black!20] (p25) at ([xshift=\stepsize]p24.east) {};

\node [anchor=north] (i1) at (p11.south) {\footnotesize{$i-3$}};
\node [anchor=north] (i2) at (p12.south) {\footnotesize{$i-2$}};
\node [anchor=north] (i3) at (p13.south) {\footnotesize{$i-1$}};
\node [anchor=north] (i4) at (p14.south) {\footnotesize{$i$}};
\node [anchor=north] (i5) at (p15.south) {\footnotesize{$i+1$}};

\draw [->,black!20] (p11.90) -- (p21.-90);
\draw [->,black!20] (p12.110) -- (p21.-70);
\draw [->,black!20] (p13.130) -- (p21.-50);
\draw [->,black!20] (p14.150) -- (p21.-30);
\draw [->,black!20] (p15.170) -- (p21.-10);

\draw [->,black!20] (p11.70) -- (p22.-110);
\draw [->,black!20] (p12.90) -- (p22.-90);
\draw [->,black!20] (p13.110) -- (p22.-70);
\draw [->,black!20] (p14.130) -- (p22.-50);
\draw [->,black!20] (p15.150) -- (p22.-30);

\draw [->,black!20] (p11.50) -- (p23.-130);
\draw [->,black!20] (p12.70) -- (p23.-110);
\draw [->,black!20] (p13.90) -- (p23.-90);
\draw [->,black!20] (p14.110) -- (p23.-70);
\draw [->,black!20] (p15.130) -- (p23.-50);

\draw [->,thick] (p11.30) -- (p24.-150);
\draw [->,thick] (p12.50) -- (p24.-130);
\draw [->,thick] (p13.70) -- (p24.-110);
\draw [->,thick] (p14.90) -- (p24.-90);
\draw [->,thick] (p15.110) -- (p24.-70);

\draw [->,black!20] (p11.10) -- (p25.-170);
\draw [->,black!20] (p12.30) -- (p25.-150);
\draw [->,black!20] (p13.50) -- (p25.-130);
\draw [->,black!20] (p14.70) -- (p25.-110);
\draw [->,black!20] (p15.90) -- (p25.-90);

\node [anchor=north] (rnnlabel) at ([yshift=-2em]p13.south) {(a) Encoder-side Self-attention};

\end{scope}

\begin{scope} [xshift=3in]

\node [anchor=west,pnode] (p11) at (0,0) {};
\node [anchor=west,pnode] (p12) at ([xshift=\stepsize]p11.east) {};
\node [anchor=west,pnode] (p13) at ([xshift=\stepsize]p12.east) {};
\node [anchor=west,pnode] (p14) at ([xshift=\stepsize]p13.east) {};
\node [anchor=west,pnode] (p15) at ([xshift=\stepsize]p14.east) {};

\node [anchor=west,pnode,fill=black!20] (p21) at ([yshift=1.8cm]p11.west) {};
\node [anchor=west,pnode,fill=black!20] (p22) at ([xshift=\stepsize]p21.east) {};
\node [anchor=west,pnode,fill=black!20] (p23) at ([xshift=\stepsize]p22.east) {};
\node [anchor=west,pnode] (p24) at ([xshift=\stepsize]p23.east) {};
\node [anchor=west,pnode,fill=black!20] (p25) at ([xshift=\stepsize]p24.east) {};

\node [anchor=north] (i1) at (p11.south) {\footnotesize{$i-3$}};
\node [anchor=north] (i2) at (p12.south) {\footnotesize{$i-2$}};
\node [anchor=north] (i3) at (p13.south) {\footnotesize{$i-1$}};
\node [anchor=north] (i4) at (p14.south) {\footnotesize{$i$}};
\node [anchor=north] (i5) at (p15.south) {\footnotesize{$i+1$}};

\draw [->,black!20] (p11.90) -- (p21.-90);

\draw [->,black!20] (p11.70) -- (p22.-110);
\draw [->,black!20] (p12.90) -- (p22.-90);

\draw [->,black!20] (p11.50) -- (p23.-130);
\draw [->,black!20] (p12.70) -- (p23.-110);
\draw [->,black!20] (p13.90) -- (p23.-90);

\draw [->,thick] (p11.30) -- (p24.-150);
\draw [->,thick] (p12.50) -- (p24.-130);
\draw [->,thick] (p13.70) -- (p24.-110);
\draw [->,thick] (p14.90) -- (p24.-90);

\draw [->,black!20] (p11.10) -- (p25.-170);
\draw [->,black!20] (p12.30) -- (p25.-150);
\draw [->,black!20] (p13.50) -- (p25.-130);
\draw [->,black!20] (p14.70) -- (p25.-110);
\draw [->,black!20] (p15.90) -- (p25.-90);

\node [anchor=north] (rnnlabel) at ([yshift=-2em]p13.south) {(b) Decoder-side Self-attention};

\end{scope}

\end{tikzpicture}

\end{center}
\caption{Self-attention on the encoder and decoder sides. Each line connects an input and an output of the self-attention model, indicating a dependency of an output state on an input state. For encoder self-attention, the output at any position is computed by having access to the entire sequence. By contrast, for decoder self-attention, the output at position $i$ is computed by seeing only inputs at positions up to $i$.}
\label{fig:self-attention-encoder-vs-decoder}
\end{figure}

\noindent where $M(i,j)$ indicates a bias term for the alignment score between positions $i$ and $j$. Below we show an example of how the masking variable is applied (assume $n=4$).
\begin{eqnarray}
&   & \mathrm{Softmax}(\frac{\mathbf{S}^{q} [\mathbf{S}^{k}]^\top}{\sqrt{d}} + \mathbf{M}) \nonumber \\
& = &
\mathrm{Softmax}(
\begin{bmatrix}
2 & 0.1 & 1 & 1 \\
0 & 0.9 & 0.9 & 0.9 \\
0.2 & 0.8 & 0.7 & 2 \\
0.3 & 1 & 0.3 & 3
\end{bmatrix}
+
\begin{bmatrix}
0 & {\color{red} -\infty} & {\color{red} -\infty} & {\color{red} -\infty} \\
0 & 0 & {\color{red} -\infty} & {\color{red} -\infty} \\
0 & 0 & 0 & {\color{red} -\infty} \\
0 & 0 & 0 & 0 \\
\end{bmatrix}
)
\nonumber \\
& = & \mathrm{Softmax}(
\begin{bmatrix}
2 & {\color{red} -\infty} & {\color{red} -\infty} & {\color{red} -\infty} \\
0 & 0.9 & {\color{red} -\infty} & {\color{red} -\infty} \\
0.2 & 0.8 & 0.7 & {\color{red} -\infty} \\
0.3 & 1 & 0.3 & 3 \\
\end{bmatrix}
)
\nonumber \\
& = &
\begin{bmatrix}
1 & {\color{red} 0} & {\color{red} 0} & {\color{red} 0} \\
0.3 & 0.7 & {\color{red} 0} & {\color{red} 0} \\
0.2 & 0.4 & 0.4 & {\color{red} 0} \\
0.05 & 0.1 & 0.05 & 0.8 \\
\end{bmatrix}.
\end{eqnarray}

As noted in Section~\ref{sec:multi-head-self-attention}, it is straightforward to improve these models by using the multi-head attention mechanism. In addition, because decoders are typically the most time-consuming components in practical applications, the overall computational efficiency of these systems depends heavily on optimizing decoder-side attention.

\subsection{Training and Inference}

\noindent Transformers can be trained and used in a standard manner. For example, we can train a Transformer model by performing gradient descent to minimize a loss function on the training data, and test the trained model by performing beam search on unseen data. Below we present some of the techniques that are typically used in the training and inference of Transformer models.

\begin{itemize}
\item \vspace{0.5em} \textbf{Learning Rate Scheduling}. Like standard neural networks, Transformers can be directly trained using back-propagation. The training process is generally iterated many times to make the model fit the training data well. In each training step, we update the weights of the neural networks by moving them a small step in the direction of the negative gradient of the loss function. There are many ways to design the update rule for training. A popular choice is the Adam optimization method~\citep{kingma-ba:2014adam}. To adjust the learning rate during training, \citet{Vaswani-etal:2017Transformer} present a learning rate scheduling strategy that increases the learning rate linearly for a number of steps and then decays it gradually. They design a learning rate of the form
    \begin{eqnarray}
    \eta & = & \eta_{0} \cdot \min \left\{ n_{\mathrm{step}}^{-0.5},\ n_{\mathrm{step}} \cdot (n_{\rm{warmup}})^{-1.5} \right\}, \label{eq:transformer-learning-rate}
    \end{eqnarray}

    \noindent where $\eta_{0}$ denotes the initial learning rate, $n_{\mathrm{step}}$ denotes the number of training steps we have executed, and $n_{\mathrm{warmup}}$ denotes the number of warmup steps. In the first $n_{\mathrm{warmup}}$ steps, the learning rate $\eta$ grows larger as training proceeds. It reaches its highest value at the point where $n_{\mathrm{step}} = n_{\mathrm{warmup}}$, and then decreases as an inverse square root function (i.e., $\eta_{0} \cdot n_{\mathrm{step}}^{-0.5}$).

\item \vspace{0.3em} \textbf{Batching and Padding}. To make a trade-off between global optimization and training convergence, it is common practice to update the weights using a relatively small collection of samples, called a \textbf{minibatch}. Therefore, we can consider a batch version of the forward and backward computation processes in which the entire minibatch is used together to obtain the gradient information. One advantage of batching is that it allows the system to leverage efficient tensor operations to process multiple sequences in a single run. This requires all input sequences in a minibatch to be stored in a single memory block so they can be read and processed simultaneously. To illustrate this idea, consider a minibatch containing four samples whose source-sides are:

    \vspace{0.5em}
    \begin{center}
    \begingroup
    \begin{tabular}{l l l l l l}
    A & B & C & D & E & F \\
    M & N \\
    R & S & T \\
    W & X & Y & Z
    \end{tabular}
    \endgroup
    \end{center}

    We can store these sequences in a $4 \times 6$ continuous block where each ``row'' represents a sequence, like this

    \vspace{0.5em}
    \begin{center}
    \begingroup
    \begin{tabular}{l l l l l l}
    A & B & C & D & E & F \\
    M & N & $\square$ & $\square$ & $\square$ & $\square$ \\
    R & S & T & $\square$ & $\square$ & $\square$\\
    W & X & Y & Z & $\square$ & $\square$
    \end{tabular}
    \endgroup
    \end{center}

    Here, padding words $\square$ are appended to the shorter sequences so that all sequences are properly aligned in memory. Typically, we do not want padding to affect the operation of the system, so we can simply define $\square$ as a zero vector (referred to as \textbf{zero padding}). On the other hand, in some cases we are interested in using padding to describe something that is not covered by the input sequences. For example, we can replace padding words with context words from the left (or right) of a sequence, though this may require modifications to the system to ensure that the newly added context words do not cause unintended content to appear in the output.

\item \vspace{0.3em} \textbf{Search and Caching}. At test time, we need to search the space of candidate hypotheses (or candidate target-side sequences) to identify the hypothesis with the highest score
    \begin{eqnarray}
    \hat{\mathbf{y}} & = & \argmax_{\mathbf{y}}\ \mathrm{score}(\mathbf{x}, \mathbf{y}),
    \end{eqnarray}

    \noindent where $\mathrm{score}(\mathbf{x}, \mathbf{y})$ is the model score of the target-side sequence $\mathbf{y}$ given the source-side sequence $\mathbf{x}$. While there are many search algorithms designed to achieve this, most share a similar structure: the search program operates by extending a pool of candidate target-side sequences one token at a time. In this way, the resulting algorithm can be viewed as a left-to-right generation procedure. For a more detailed discussion of search algorithms and model scores of general sequence-to-sequence models, we refer the reader to the literature. Note that all designs for $\mathrm{score}(\mathbf{x}, \mathbf{y})$, no matter how complex, are based on computing $\Pr(\mathbf{y}|\mathbf{x})$. Because the attention models used in Transformers require computing the dot-product of each pair of input vectors in a layer, the time complexity of the search algorithm is a quadratic function of the length of $\mathbf{y}$. It is therefore inefficient to repeatedly compute the outputs of the attention models for previously processed positions. This problem can be addressed by caching the states of each layer for the words we have already seen. Figure~\ref{fig:transformer-decoder-cache} illustrates the use of the caching mechanism in a search step. All states for positions $< i$ are maintained and easily accessed in a cache. At position $i$, all we need is to compute the states for the newly added word and then update the cache.

    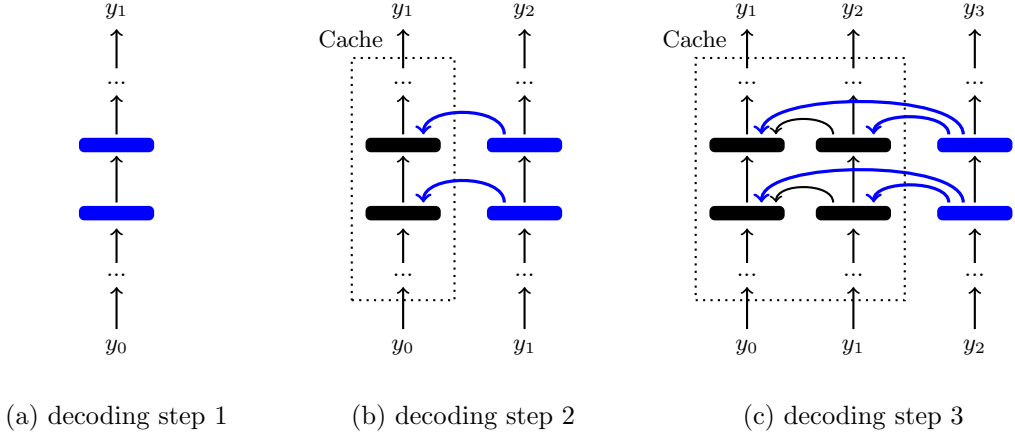
\begin{figure}[!htp]
    \centering

\begin{center}

\begin{tikzpicture}

\def\rowsep{0.7cm}
\def\colsep{0.4cm}

\tikzstyle{lnode} = [minimum width=1cm,minimum height=0.5em,inner sep=0pt,fill=black,rounded corners=2]

\begin{scope}

\node [lnode,anchor=west] (layer11) at (0,0) {};
\node [lnode,anchor=south] (layer21) at ([yshift=\rowsep]layer11.north) {};
\draw [->,thick] ([yshift=-0.8*\rowsep]layer11.south) -- ([yshift=-0.1cm]layer11.south) node [pos=0,below] (s1) {\footnotesize{...}};
\draw [->,thick] ([yshift=-0.8*\rowsep]s1.south) -- ([yshift=0.0cm]s1.south) node [pos=0,below] (w1) {\footnotesize{$y_0$}};
\draw [->,thick] ([yshift=0.1em]layer11.north) -- ([yshift=-0.1em]layer21.south);
\draw [->,thick] ([yshift=0.1em]layer21.north) -- ([yshift=0.8*\rowsep]layer21.north) node [pos=1,above] (t1) {\footnotesize{...}};
\draw [->,thick] ([yshift=0.1em]t1.north) -- ([yshift=0.8*\rowsep]t1.north) node [pos=1,above] (p1) {\footnotesize{$y_1$}};

\begin{pgfonlayer}{background}
\node [draw,black,thick,dotted,inner sep=5pt,fill=white]  [fit = (s1) (layer11) (t1)] (cache) {};
\end{pgfonlayer}

\node [anchor=south] (clabel) at (cache.north west) {\footnotesize{Cache}};

\node [lnode,anchor=west,fill=blue] (layer12) at ([xshift=1.5*\colsep]layer11.east) {};
\node [lnode,anchor=south,fill=blue] (layer22) at ([yshift=\rowsep]layer12.north) {};
\draw [->,thick] ([yshift=-0.8*\rowsep]layer12.south) -- ([yshift=-0.1cm]layer12.south) node [pos=0,below] (s2) {\footnotesize{...}};
\draw [->,thick] ([yshift=-0.8*\rowsep]s2.south) -- ([yshift=0.0cm]s2.south) node [pos=0,below] (w2) {\footnotesize{$y_1$}};
\draw [->,thick] ([yshift=0.1em]layer12.north) -- ([yshift=-0.1em]layer22.south);
\draw [->,thick] ([yshift=0.1em]layer22.north) -- ([yshift=0.8*\rowsep]layer22.north) node [pos=1,above] (t2) {\footnotesize{...}};
\draw [->,thick] ([yshift=0.1em]t2.north) -- ([yshift=0.8*\rowsep]t2.north) node [pos=1,above] (p2) {\footnotesize{$y_2$}};

\draw [->,very thick,blue] ([xshift=-0.7em,yshift=0.1em]layer12.north) .. controls +(north:1em) and +(north:1em) .. ([xshift=0.7em,yshift=0.1em]layer11.north);
\draw [->,very thick,blue] ([xshift=-0.7em,yshift=0.1em]layer22.north) .. controls +(north:1em) and +(north:1em) .. ([xshift=0.7em,yshift=0.1em]layer21.north);

\node [anchor=north] (caption) at ([yshift=-1em,xshift=2*\colsep]w1.south) {\small{(b) decoding step 2}};

\end{scope}

\begin{scope}[xshift=-1.5in]

\node [lnode,anchor=west,fill=blue] (layer11) at (0,0) {};
\node [lnode,anchor=south,fill=blue] (layer21) at ([yshift=\rowsep]layer11.north) {};
\draw [->,thick] ([yshift=-0.8*\rowsep]layer11.south) -- ([yshift=-0.1cm]layer11.south) node [pos=0,below] (s1) {\footnotesize{...}};
\draw [->,thick] ([yshift=-0.8*\rowsep]s1.south) -- ([yshift=0.0cm]s1.south) node [pos=0,below] (w1) {\footnotesize{$y_0$}};
\draw [->,thick] ([yshift=0.1em]layer11.north) -- ([yshift=-0.1em]layer21.south);
\draw [->,thick] ([yshift=0.1em]layer21.north) -- ([yshift=0.8*\rowsep]layer21.north) node [pos=1,above] (t1) {\footnotesize{...}};
\draw [->,thick] ([yshift=0.1em]t1.north) -- ([yshift=0.8*\rowsep]t1.north) node [pos=1,above] (p1) {\footnotesize{$y_1$}};

\node [anchor=north] (caption) at ([yshift=-1em,xshift=0]w1.south) {\small{(a) decoding step 1}};

\end{scope}

\begin{scope}[xshift=1.8in]

\node [lnode,anchor=west] (layer11) at (0,0) {};
\node [lnode,anchor=south] (layer21) at ([yshift=\rowsep]layer11.north) {};
\draw [->,thick] ([yshift=-0.8*\rowsep]layer11.south) -- ([yshift=-0.1cm]layer11.south) node [pos=0,below] (s1) {\footnotesize{...}};
\draw [->,thick] ([yshift=-0.8*\rowsep]s1.south) -- ([yshift=0.0cm]s1.south) node [pos=0,below] (w1) {\footnotesize{$y_0$}};
\draw [->,thick] ([yshift=0.1em]layer11.north) -- ([yshift=-0.1em]layer21.south);
\draw [->,thick] ([yshift=0.1em]layer21.north) -- ([yshift=0.8*\rowsep]layer21.north) node [pos=1,above] (t1) {\footnotesize{...}};
\draw [->,thick] ([yshift=0.1em]t1.north) -- ([yshift=0.8*\rowsep]t1.north) node [pos=1,above] (p1) {\footnotesize{$y_1$}};

\node [lnode,anchor=west] (layer12) at ([xshift=1*\colsep]layer11.east) {};
\node [lnode,anchor=south] (layer22) at ([yshift=\rowsep]layer12.north) {};
\draw [->,thick] ([yshift=-0.8*\rowsep]layer12.south) -- ([yshift=-0.1cm]layer12.south) node [pos=0,below] (s2) {\footnotesize{...}};
\draw [->,thick] ([yshift=-0.8*\rowsep]s2.south) -- ([yshift=0.0cm]s2.south) node [pos=0,below] (w2) {\footnotesize{$y_1$}};
\draw [->,thick] ([yshift=0.1em]layer12.north) -- ([yshift=-0.1em]layer22.south);
\draw [->,thick] ([yshift=0.1em]layer22.north) -- ([yshift=0.8*\rowsep]layer22.north) node [pos=1,above] (t2) {\footnotesize{...}};
\draw [->,thick] ([yshift=0.1em]t2.north) -- ([yshift=0.8*\rowsep]t2.north) node [pos=1,above] (p2) {\footnotesize{$y_2$}};

\draw [->,thick] ([xshift=-0.7em,yshift=0.1em]layer12.north) .. controls +(north:0.7em) and +(north:0.7em) .. ([xshift=1em,yshift=0.1em]layer11.north);
\draw [->,thick] ([xshift=-0.7em,yshift=0.1em]layer22.north) .. controls +(north:0.7em) and +(north:0.7em) .. ([xshift=1em,yshift=0.1em]layer21.north);

\begin{pgfonlayer}{background}
\node [draw,black,thick,dotted,inner sep=5pt,fill=white]  [fit = (s1) (layer11) (layer12) (t1)] (cache) {};
\end{pgfonlayer}

\node [anchor=south] (clabel) at (cache.north west) {\footnotesize{Cache}};

\node [lnode,anchor=west,fill=blue] (layer13) at ([xshift=1.5*\colsep]layer12.east) {};
\node [lnode,anchor=south,fill=blue] (layer23) at ([yshift=\rowsep]layer13.north) {};
\draw [->,thick] ([yshift=-0.8*\rowsep]layer13.south) -- ([yshift=-0.1cm]layer13.south) node [pos=0,below] (s3) {\footnotesize{...}};
\draw [->,thick] ([yshift=-0.8*\rowsep]s3.south) -- ([yshift=0.0cm]s3.south) node [pos=0,below] (w3) {\footnotesize{$y_2$}};
\draw [->,thick] ([yshift=0.1em]layer13.north) -- ([yshift=-0.1em]layer23.south);
\draw [->,thick] ([yshift=0.1em]layer23.north) -- ([yshift=0.8*\rowsep]layer23.north) node [pos=1,above] (t3) {\footnotesize{...}};
\draw [->,thick] ([yshift=0.1em]t3.north) -- ([yshift=0.8*\rowsep]t3.north) node [pos=1,above] (p3) {\footnotesize{$y_3$}};

\draw [->,very thick,blue] ([xshift=-0.9em,yshift=0.1em]layer13.north) .. controls +(north:0.8em) and +(north:0.8em) .. ([xshift=0.7em,yshift=0.1em]layer12.north);
\draw [->,very thick,blue] ([xshift=-0.9em,yshift=0.1em]layer23.north) .. controls +(north:0.8em) and +(north:0.8em) .. ([xshift=0.7em,yshift=0.1em]layer22.north);
\draw [->,very thick,blue] ([xshift=-0.4em,yshift=0.1em]layer13.north) .. controls +(north:1.5em) and +(north:1.5em) .. ([xshift=0.5em,yshift=0.1em]layer11.north);
\draw [->,very thick,blue] ([xshift=-0.4em,yshift=0.1em]layer23.north) .. controls +(north:1.5em) and +(north:1.5em) .. ([xshift=0.5em,yshift=0.1em]layer21.north);

\node [anchor=north] (caption) at ([yshift=-1em,xshift=0]w2.south) {\small{(c) decoding step 3}};

\end{scope}

\end{tikzpicture}

\end{center}
    \caption{Illustration of the caching mechanism in Transformer decoders. Rectangles indicate the states of decoding layers or sub-layers. At step $i$, all the states from previous steps are stored in a cache (see dotted boxes), and we only need to compute the states for the current step (see blue rectangles and arrows). We then add the newly generated states to the cache and move on to step $i+1$.}
    \label{fig:transformer-decoder-cache}
    \end{figure}

\end{itemize}
\vspace{0.3em}


\section{Syntax-aware Models}

\noindent Although Transformers are deep learning models that do not explicitly make use of linguistic structures or assumptions, it may be necessary to incorporate our prior knowledge into such systems. This is in part because NLP researchers have long believed that a higher level of abstraction in data representation is needed to develop ideal NLP systems, and there have been many systems that use structural priors. However, structure is a broad topic. For a discussion of the various types of structure one might consider, see the work of \citet{see:2018deep}. For example, the inductive biases used in our model design can be thought of as some structural prior, and NLP models can also learn the underlying structure of problems by themselves. In this subsection we will discuss some of these issues. We will focus on the methods of introducing linguistic structure into Transformer models. As Transformers can be applied to many NLP tasks, which differ much in their input and output formats, we will primarily discuss modifications to Transformer encoders (call them \textbf{syntax-aware Transformer encoders}). Our discussion, however, is general, and these methods can easily be extended to Transformer decoders.

\subsection{Syntax-aware Input and Output}
\label{sec:syntax-input-and-output}

\noindent One of the simplest methods of incorporating structure into NLP systems is to modify the input sequence, leaving the system unchanged. As a simple example, consider a sentence where each word $x_j$ is assigned a set of $\kappa$ syntactic labels $\{\mathrm{tag}_j^{1}, \cdots ,\mathrm{tag}_j^{\kappa}\}$ (e.g., POS labels and dependency labels). We can concatenate these symbols to define a new ``word''
\begin{equation}
x_j/\mathrm{tag}_j^{1}/\cdots/\mathrm{tag}_j^{\kappa} \nonumber
\end{equation}

\noindent Then, the embedding of this word is given by
\begin{eqnarray}
\mathbf{e}_j & = & e(x_j/\mathrm{tag}_j^{1}/\cdots/\mathrm{tag}_j^{\kappa}) + \mathrm{PE}(j),
\end{eqnarray}

\noindent where $e(x_j/\mathrm{tag}_j^{1}/ \cdots /\mathrm{tag}_j^{\kappa}) \in \mathbb{R}^{d}$ is the embedding of $x_j/\mathrm{tag}_j^{1}/\cdots/\mathrm{tag}_j^{\kappa}$. Since $x_j/\mathrm{tag}_j^{1}/ \cdots /\mathrm{tag}_j^{\kappa}$ is a complex symbol, we decompose the learning problem of $e(x_j/\mathrm{tag}_j^{1}/ \cdots /\mathrm{tag}_j^{\kappa})$ into easier problems. For example, we can develop $\kappa$ embedding models, each producing an embedding for a given tag. Then, we write $e(x_j/\mathrm{tag}_j^{1}/ \cdots /\mathrm{tag}_j^{\kappa})$ as a sum of the word embedding and tag embeddings
\begin{eqnarray}
e(x_j/\mathrm{tag}_j^{1}/ \cdots /\mathrm{tag}_j^{\kappa}) & = & \mathbf{x}_j + e(\mathrm{tag}_j^{1}) + \cdots + e(\mathrm{tag}_j^{\kappa}),
\end{eqnarray}

\noindent where $\{e(\mathrm{tag}_j^{1}), \cdots ,e(\mathrm{tag}_j^{\kappa})\}$ are the embeddings of the tags. Alternatively, we can combine these embeddings via a neural network in the form
\begin{eqnarray}
e(x_j/\mathrm{tag}_j^{1}/ \cdots /\mathrm{tag}_j^{\kappa}) & = & \mathrm{FFN}_{\mathrm{embed}}(\mathbf{x}_j,e(\mathrm{tag}_j^{1}), \cdots ,e(\mathrm{tag}_j^{\kappa})),
\end{eqnarray}

\noindent where $\mathrm{FFN}_{\mathrm{embed}}(\cdot)$ is a feed-forward neural network that has one or two layers.

We can apply a similar technique on the decoder side, treating $y_i/\mathrm{tag}_i^{1}/\cdots/\mathrm{tag}_i^{\kappa}$ as a syntax-augmented target word. However, doing so may drastically increase the size of the target-side vocabulary, posing a computational challenge for both training and inference.

Another common method for representing a sentence is via a syntax tree. In linguistics, the syntax of a sentence can be interpreted in many different ways, resulting in various grammars and corresponding tree-based (or graph-based) representations. While these representations differ in their exact syntactic forms, a general approach for utilizing them in sequence modeling is \textbf{tree linearization}. Consider the following sentence annotated with a constituency-based parse tree:

\begin{center}
\begin{tikzpicture}
    \begin{scope}[scale=1.0, sibling distance=20pt, level distance=30pt]
    \Tree[.\node(r){\large S};
            [.\node(n11){\large NP};
                [.\node(n111){\large PRP};  \node(l1){\normalsize It};]
            ]
            [.\node(n12){\large VP};
                [.\node(n121){\large VBZ}; \node(l2){\normalsize 's};]
                [.\node(n121){\large ADJP};
                    [.\node(n1211){\large JJ}; \node(l3){\normalsize interesting};]
                ]
            ]
            [.\node(n13){\large .};  \node(n4){\normalsize !};]
        ]
    \end{scope}
\end{tikzpicture}
\end{center}

\noindent We can write this tree structure as a sequence of words, syntactic labels and brackets via a tree traversal algorithm, as follows

\vspace{0.8em}
\begingroup
\renewcommand{\arraystretch}{1.2}
\begin{tabular}{l l l l l l l l l l l l}
(S & (NP & (PRP & {\color{red} It} & )$_{\textrm{PRP}}$ & )$_{\textrm{NP}}$ & (VP & (VBZ & {\color{red} 's} & )$_{\textrm{VBZ}}$ & (ADJP & (JJ \\
\multicolumn{2}{l}{{\color{red} interesting}} & )$_{\textrm{JJ}}$ & )$_{\textrm{ADJP}}$ & )$_{\textrm{VP}}$ & (. & {\color{red} !} & )$_{\textrm{.}}$ & )$_{\textrm{S}}$
\end{tabular}
\endgroup
\vspace{0.8em}

This sequence of syntactic tokens can be used directly as input to the encoder, that is, each token is represented by the sum of its word and positional embeddings, and this combined embedding is treated as a standard input to the encoder. One application of linearized trees is tree-to-string machine translation, in which a syntax tree in one language is translated into a string in another language~\citep{li-etal:2017modeling,currey-heafield:2018multi}. Linearized trees can also be utilized for tree generation. For example, parsing tasks can be framed as sequence-to-sequence problems, mapping an input text to a sequential representation of its corresponding syntax tree~\citep{vinyals-etal:2015grammar,choe-charniak:2016parsing}. Figure~\ref{fig:tree-linearization-for-t2s-and-parsing} illustrates these models. It should be noted that the methods described here are not specific to the Transformer and can be applied to many sequence-based architectures, such as RNN-based models.

\begin{figure}[!htp]
\centering

\begin{center}

\begin{tikzpicture}

\def\arrowsep{2.0em}
\def\encdecsep{2.5em}
\def\ewsep{0.5em}
\def\ewbias{0.25em}
\def\dwsep{0.5em}
\def\dwbias{1.5em}
\def\drawsep{0.5em}
\def\labelsep{2.5in}

\tikzstyle{pnode} = [rectangle,inner sep=0.2em,minimum height=2.5em,draw,very thick]
\tikzstyle{wnode} = [inner sep=0.2em,minimum width=2.2em,minimum height=1.2em]
\tikzstyle{pdwnode} = [minimum width=2.8em]


\begin{scope}[xshift=0in]


\node [pnode,anchor=west,draw=black,minimum width=20em] (encoder) at (0,0) {\large{Encoder}};
\node [pnode,anchor=west,draw=black,minimum width=7.5em] (decoder) at ([xshift=\encdecsep]encoder.east) {\large{Decoder}};

\draw [->,very thick] ([xshift=0em,yshift=0em]encoder.east) -- ([xshift=0em,yshift=0em]decoder.west);

{\footnotesize
\node [anchor=north west,wnode] (ew1) at ([xshift=\ewbias,yshift=-\arrowsep*1.2]encoder.south west) {(ADJP};
\node [anchor=west,wnode] (ew2) at ([xshift=\ewsep,yshift=0em]ew1.east) {(JJ};
\node [anchor=west,wnode] (ew3) at ([xshift=\ewsep,yshift=0em]ew2.east) {Great};
\node [anchor=west,wnode] (ew4) at ([xshift=\ewsep,yshift=0em]ew3.east) {)$_{\textrm{JJ}}$};
\node [anchor=west,wnode] (ew5) at ([xshift=\ewsep,yshift=0em]ew4.east) {(.};
\node [anchor=west,wnode] (ew6) at ([xshift=\ewsep,yshift=0em]ew5.east) {!};
\node [anchor=west,wnode] (ew7) at ([xshift=\ewsep,yshift=0em]ew6.east) {)$_{\textrm{.}}$};
\node [anchor=west,wnode] (ew8) at ([xshift=\ewsep,yshift=0em]ew7.east) {)$_{\textrm{ADJP}}$};


\node [anchor=north,wnode] (dw1) at ([xshift=\dwbias,yshift=-\arrowsep*1.2]decoder.south west) {$\langle \mathrm{SOS} \rangle$};
\node [anchor=west,wnode] (dw2) at ([xshift=\ewsep,yshift=0em]dw1.east) {很};
\node [anchor=west,wnode] (dw3) at ([xshift=\ewsep,yshift=0em]dw2.east) {好};

\node [anchor=south,wnode] (dwu1) at ([xshift=\dwbias,yshift=\arrowsep*1.2]decoder.north west) {很};
\node [anchor=west,wnode] (dwu2) at ([xshift=\ewsep,yshift=0em]dwu1.east) {好};
\node [anchor=west,wnode] (dwu3) at ([xshift=\ewsep,yshift=0em]dwu2.east) {！};
}

\draw [->,thick] ([xshift=0em,yshift=\drawsep]ew1.north) -- ([xshift=0em,yshift=\arrowsep-\drawsep]ew1.north);
\draw [->,thick] ([xshift=0em,yshift=\drawsep]ew2.north) -- ([xshift=0em,yshift=\arrowsep-\drawsep]ew2.north);
\draw [->,thick] ([xshift=0em,yshift=\drawsep]ew3.north) -- ([xshift=0em,yshift=\arrowsep-\drawsep]ew3.north);
\draw [->,thick] ([xshift=0em,yshift=\drawsep]ew4.north) -- ([xshift=0em,yshift=\arrowsep-\drawsep]ew4.north);
\draw [->,thick] ([xshift=0em,yshift=\drawsep]ew5.north) -- ([xshift=0em,yshift=\arrowsep-\drawsep]ew5.north);
\draw [->,thick] ([xshift=0em,yshift=\drawsep]ew6.north) -- ([xshift=0em,yshift=\arrowsep-\drawsep]ew6.north);
\draw [->,thick] ([xshift=0em,yshift=\drawsep]ew7.north) -- ([xshift=0em,yshift=\arrowsep-\drawsep]ew7.north);
\draw [->,thick] ([xshift=0em,yshift=\drawsep]ew8.north) -- ([xshift=0em,yshift=\arrowsep-\drawsep]ew8.north);

\draw [->,thick] ([xshift=0em,yshift=\drawsep]dw1.north) -- ([xshift=0em,yshift=\arrowsep-\drawsep]dw1.north);
\draw [->,thick] ([xshift=0em,yshift=\drawsep]dw2.north) -- ([xshift=0em,yshift=\arrowsep-\drawsep]dw2.north);
\draw [->,thick] ([xshift=0em,yshift=\drawsep]dw3.north) -- ([xshift=0em,yshift=\arrowsep-\drawsep]dw3.north);

\draw [->,thick] ([xshift=0em,yshift=-\arrowsep+\drawsep]dwu1.south) -- ([xshift=0em,yshift=-\drawsep]dwu1.south);
\draw [->,thick] ([xshift=0em,yshift=-\arrowsep+\drawsep]dwu2.south) -- ([xshift=0em,yshift=-\drawsep]dwu2.south);
\draw [->,thick] ([xshift=0em,yshift=-\arrowsep+\drawsep]dwu3.south) -- ([xshift=0em,yshift=-\drawsep]dwu3.south);

\node [anchor=north] (label) at ([xshift=\labelsep,yshift=-\arrowsep*2.2]encoder.south west) {\small{(a) Tree-to-String Machine Translation}};

\end{scope}


\begin{scope}[yshift=-2.1in]


\node [pnode,anchor=west,draw=black,minimum width=5em] (encoder) at (0,0) {\large{Encoder}};
\node [pnode,anchor=west,draw=black,minimum width=22.5em] (decoder) at ([xshift=\encdecsep]encoder.east) {\large{Decoder}};

\draw [->,very thick] ([xshift=0em,yshift=0em]encoder.east) -- ([xshift=0em,yshift=0em]decoder.west);

{\footnotesize
\node [anchor=north west,wnode] (ew1) at ([xshift=\ewbias,yshift=-\arrowsep*1.2]encoder.south west) {Great};
\node [anchor=west,wnode] (ew2) at ([xshift=\ewsep,yshift=0em]ew1.east) {!};


\node [anchor=north,wnode,pdwnode] (dw1) at ([xshift=\dwbias,yshift=-\arrowsep*1.2]decoder.south west) {$\langle \mathrm{SOS} \rangle$};
\node [anchor=west,wnode,pdwnode] (dw2) at ([xshift=\ewsep,yshift=0em]dw1.east) {(ADJP};
\node [anchor=west,wnode,pdwnode] (dw3) at ([xshift=\ewsep,yshift=0em]dw2.east) {(JJ};
\node [anchor=west,wnode,pdwnode] (dw4) at ([xshift=\ewsep,yshift=0em]dw3.east) {Great};
\node [anchor=west,wnode,pdwnode] (dw5) at ([xshift=\ewsep,yshift=0em]dw4.east) {)$_{\textrm{JJ}}$};
\node [anchor=west,wnode,pdwnode] (dw6) at ([xshift=\ewsep,yshift=0em]dw5.east) {(.};
\node [anchor=west,wnode,pdwnode] (dw7) at ([xshift=\ewsep,yshift=0em]dw6.east) {!};
\node [anchor=west,wnode,pdwnode] (dw8) at ([xshift=\ewsep,yshift=0em]dw7.east) {)$_{\textrm{.}}$};

\node [anchor=south,wnode,pdwnode] (dwu1) at ([xshift=\dwbias,yshift=\arrowsep*1.2]decoder.north west) {(ADJP};
\node [anchor=west,wnode,pdwnode] (dwu2) at ([xshift=\ewsep,yshift=0em]dwu1.east) {(JJ};
\node [anchor=west,wnode,pdwnode] (dwu3) at ([xshift=\ewsep,yshift=0em]dwu2.east) {Great};
\node [anchor=west,wnode,pdwnode] (dwu4) at ([xshift=\ewsep,yshift=0em]dwu3.east) {)$_{\textrm{JJ}}$};
\node [anchor=west,wnode,pdwnode] (dwu5) at ([xshift=\ewsep,yshift=0em]dwu4.east) {(.};
\node [anchor=west,wnode,pdwnode] (dwu6) at ([xshift=\ewsep,yshift=0em]dwu5.east) {!};
\node [anchor=west,wnode,pdwnode] (dwu7) at ([xshift=\ewsep,yshift=0em]dwu6.east) {)$_{\textrm{.}}$};
\node [anchor=west,wnode,pdwnode] (dwu8) at ([xshift=\ewsep,yshift=0em]dwu7.east) {)$_{\textrm{ADJP}}$};
}

\draw [->,thick] ([xshift=0em,yshift=\drawsep]ew1.north) -- ([xshift=0em,yshift=\arrowsep-\drawsep]ew1.north);
\draw [->,thick] ([xshift=0em,yshift=\drawsep]ew2.north) -- ([xshift=0em,yshift=\arrowsep-\drawsep]ew2.north);

\draw [->,thick] ([xshift=0em,yshift=\drawsep]dw1.north) -- ([xshift=0em,yshift=\arrowsep-\drawsep]dw1.north);
\draw [->,thick] ([xshift=0em,yshift=\drawsep]dw2.north) -- ([xshift=0em,yshift=\arrowsep-\drawsep]dw2.north);
\draw [->,thick] ([xshift=0em,yshift=\drawsep]dw3.north) -- ([xshift=0em,yshift=\arrowsep-\drawsep]dw3.north);
\draw [->,thick] ([xshift=0em,yshift=\drawsep]dw4.north) -- ([xshift=0em,yshift=\arrowsep-\drawsep]dw4.north);
\draw [->,thick] ([xshift=0em,yshift=\drawsep]dw5.north) -- ([xshift=0em,yshift=\arrowsep-\drawsep]dw5.north);
\draw [->,thick] ([xshift=0em,yshift=\drawsep]dw6.north) -- ([xshift=0em,yshift=\arrowsep-\drawsep]dw6.north);
\draw [->,thick] ([xshift=0em,yshift=\drawsep]dw7.north) -- ([xshift=0em,yshift=\arrowsep-\drawsep]dw7.north);
\draw [->,thick] ([xshift=0em,yshift=\drawsep]dw8.north) -- ([xshift=0em,yshift=\arrowsep-\drawsep]dw8.north);

\draw [->,thick] ([xshift=0em,yshift=-\arrowsep+\drawsep]dwu1.south) -- ([xshift=0em,yshift=-\drawsep]dwu1.south);
\draw [->,thick] ([xshift=0em,yshift=-\arrowsep+\drawsep]dwu2.south) -- ([xshift=0em,yshift=-\drawsep]dwu2.south);
\draw [->,thick] ([xshift=0em,yshift=-\arrowsep+\drawsep]dwu3.south) -- ([xshift=0em,yshift=-\drawsep]dwu3.south);
\draw [->,thick] ([xshift=0em,yshift=-\arrowsep+\drawsep]dwu4.south) -- ([xshift=0em,yshift=-\drawsep]dwu4.south);
\draw [->,thick] ([xshift=0em,yshift=-\arrowsep+\drawsep]dwu5.south) -- ([xshift=0em,yshift=-\drawsep]dwu5.south);
\draw [->,thick] ([xshift=0em,yshift=-\arrowsep+\drawsep]dwu6.south) -- ([xshift=0em,yshift=-\drawsep]dwu6.south);
\draw [->,thick] ([xshift=0em,yshift=-\arrowsep+\drawsep]dwu7.south) -- ([xshift=0em,yshift=-\drawsep]dwu7.south);
\draw [->,thick] ([xshift=0em,yshift=-\arrowsep+\drawsep]dwu8.south) -- ([xshift=0em,yshift=-\drawsep]dwu8.south);

\node [anchor=north] (label) at ([xshift=\labelsep,yshift=-\arrowsep*2.2]encoder.south west) {\small{(b) Constituency Parsing}};

\end{scope}

\end{tikzpicture}

\end{center}
\caption{Illustration of tree linearization on either the encoder or decoder side. For tree-to-string machine translation, the encoder takes a sequential representation of an input parse tree, and the decoder outputs the corresponding translation. For parsing, the encoder takes a sentence, and the decoder outputs the corresponding syntax tree.}
\label{fig:tree-linearization-for-t2s-and-parsing}
\end{figure}

\subsection{Syntax-aware Attention Models}
\label{sec:syntax-aware-attention}

\noindent For Transformer models, it is also natural to make use of syntax trees to guide the process of learning sequence representations. In the previous section we saw how representations of a sequence can be computed by relating different positions within that sequence. This allows us to impose structure on the attention distributions over positions. To do this, we use the encoder self-attention with an additive mask
\begin{eqnarray}
\mathrm{AttSyn}_{\mathrm{self}}(\mathbf{H}) & = & \mathrm{Softmax}(\frac{\mathbf{H}^{q} [\mathbf{H}^{k}]^\top}{\sqrt{d}} + \mathbf{M}) \mathbf{H}^{v} \label{eq:self-attention-syn-add-masking}
\end{eqnarray}

\noindent or alternatively with a multiplicative mask
\begin{eqnarray}
\mathrm{AttSyn}_{\mathrm{self}}(\mathbf{H}) & = & \mathrm{Softmax}(\frac{\mathbf{H}^{q} [\mathbf{H}^{k}]^\top}{\sqrt{d}} \odot \mathbf{M}) \mathbf{H}^{v}, \label{eq:self-attention-syn-mul-masking}
\end{eqnarray}

\noindent where $\mathbf{M} \in \mathbb{R}^{m \times m}$ is a matrix of masking variables in which a larger value of $M(i,j)$ indicates a stronger syntactic correlation between positions $i$ and $j$. In the following description we choose Eq. (\ref{eq:self-attention-syn-mul-masking}) as the basic form.

One common way to design $\mathbf{M}$ is to project syntactic relations of the input tree structure into constraints over the sequence. Here we consider constituency parse trees and dependency parse trees for illustration. Generally, two types of masking methods are employed.

\begin{figure}[!htp]
\centering

\begin{tikzpicture}
\def\wsep{0.5em}
\def\shsep{1.5em}
\def\bhsep{2.5em}
\def\bsep{1.5em}
 
\tikzstyle{wnode} = [rectangle,inner sep=0.2em,minimum width=2.0em,minimum height=1.5em]
\tikzstyle{bnode} = [fill=blue!40,inner sep=1pt, minimum size=18pt]
\tikzstyle{llnode} = [rectangle,inner sep=0.2em,minimum width=2.0em,minimum height=1.5em]
\tikzstyle{lunode} = [rectangle,inner sep=0.2em,minimum width=2.0em,minimum height=1.5em,rotate=45]

\begin{scope}[]

\node [anchor=south west,wnode] (w1) at (0,0) {The};
\node [anchor=west,wnode] (w2) at ([xshift=\wsep,yshift=0.0em]w1.east) {concert};
\node [anchor=west,wnode] (w3) at ([xshift=\wsep,yshift=0.0em]w2.east) {was};
\node [anchor=west,wnode] (w4) at ([xshift=\wsep,yshift=0.0em]w3.east) {wonderful};
\node [anchor=west,wnode] (w5) at ([xshift=\wsep,yshift=0.0em]w4.east) {!};

\draw [->,very thick] ([xshift=-0.3em]w2.north) .. controls +(north:\shsep) and +(north:\shsep) .. ([xshift=-0.0em]w1.north);
\draw [->,very thick] ([xshift=-0.3em]w4.north) .. controls +(north:\shsep) and +(north:\shsep) .. ([xshift=-0.0em]w3.north);
\draw [->,very thick] ([xshift=0.0em]w4.north) .. controls +(north:\bhsep) and +(north:\bhsep) .. ([xshift=0.3em]w2.north);
\draw [->,very thick] ([xshift=0.3em]w4.north) .. controls +(north:\shsep) and +(north:\shsep) .. ([xshift=-0.0em]w5.north);

\node [anchor=north] (l1) at ([yshift=-0.55in,xshift=-0.0em]w3.south) {\small{(a) Dependency Parse Tree}};
\node [anchor=west] (l2) at ([yshift=-0.0em,xshift=5.4em]l1.east) {\small{(b) Mask $\mathbf{M}$ (darker color means larger value)}};
\end{scope}

\begin{scope}[xshift=4in,yshift=0.8in,scale=1.2]

\foreach \i / \j / \k in 
    {0/0/3.38,0/1/2.62,0/2/1.08,0/3/1.85,0/4/1.08,
     1/0/2.24,1/1/3.43,1/2/1.05,1/3/2.24,1/4/1.05,
     2/0/0.77,2/1/1.65,2/2/3.40,2/3/2.53,2/4/1.65,
     3/0/0.57,3/1/2.00,3/2/2.00,3/3/3.43,3/4/2.00,
     4/0/0.77,4/1/1.65,4/2/1.65,4/3/2.53,4/4/3.40}
    \pgfmathsetmacro\c{exp(\k*0.5)*0.14*70}
    \node[anchor=center,bnode,fill=blue!\c] (b\j\i) at (\i*\bsep,-\j*\bsep) {};

\node [anchor=east,llnode] (ll1) at ([xshift=-\wsep,yshift=0.0em]b00.west) {The};
\node [anchor=east,llnode] (ll2) at ([xshift=-\wsep,yshift=0.0em]b10.west) {concert};
\node [anchor=east,llnode] (ll3) at ([xshift=-\wsep,yshift=0.0em]b20.west) {was};
\node [anchor=east,llnode] (ll4) at ([xshift=-\wsep,yshift=0.0em]b30.west) {wonderful};
\node [anchor=east,llnode] (ll5) at ([xshift=-\wsep,yshift=0.0em]b40.west) {!};

\node [anchor=south west,lunode] (lu1) at ([xshift=0.0em,yshift=\wsep*2]b00.center) {The};
\node [anchor=south west,lunode] (lu2) at ([xshift=0.0em,yshift=\wsep*2]b01.center) {concert};
\node [anchor=south west,lunode] (lu3) at ([xshift=0.0em,yshift=\wsep*2]b02.center) {was};
\node [anchor=south west,lunode] (lu4) at ([xshift=0.0em,yshift=\wsep*2]b03.center) {wonderful};
\node [anchor=south west,lunode] (lu5) at ([xshift=0.0em,yshift=\wsep*2]b04.center) {!};

\end{scope}

\end{tikzpicture}
\caption{Priors induced from a dependency parse tree. The row $i$ of the matrix $\mathbf{M}$ represents a distribution that describes how much weight we can give to $M(i,j)$ in terms of the syntactic distance between $i$ and $j$.}
\label{fig:dependency-attention-prior}
\end{figure}
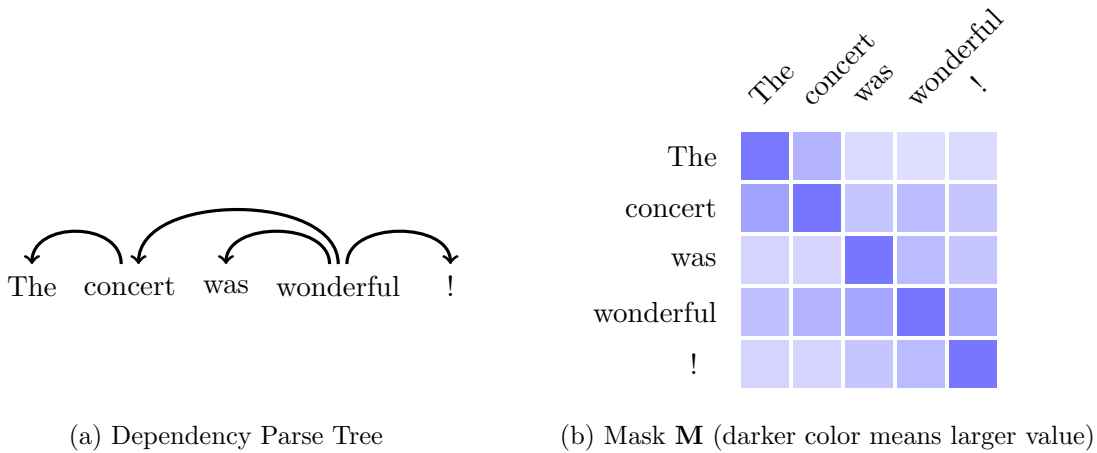

\begin{itemize}
\item \vspace{0.5em} \textbf{0-1 Masking}. This method assigns $M(i,j)=1$ if the words at positions $i$ and $j$ are considered syntactically correlated and $M(i,j)=0$ otherwise~\citep{zhang-etal:2020sg,bai-etal:2021syntax}. To model the relation between two words in a syntax tree, we can consider the distance between their corresponding nodes. One of the simplest forms is given by
    \begin{eqnarray}
    M(i,j) & = &
    \begin{cases}
    1 & \omega(i,j) \le \omega_{\mathrm{max}} \\
    0 & \textrm{otherwise}
    \end{cases}, \label{eq:0-1-masking-syntactic-attention}
    \end{eqnarray}

    \noindent where $\omega(i,j)$ is the shortest path length between the nodes corresponding to the words at positions $i$ and $j$. For example, given a dependency parse tree, $\omega(i,j)$ is the number of dependency edges in the path between the two words. For a constituency parse tree, all words are leaf nodes. Thus, $\omega(i,j)$ represents the tree distance between the two leaves within the same branch of the tree. $\omega_{\mathrm{max}}$ is a parameter used to control the maximum distance between two nodes that can still be considered syntactically correlated. For example, assuming a dependency parse tree with $\omega_{\mathrm{max}}=1$, Eq. (\ref{eq:0-1-masking-syntactic-attention}) enforces the constraint that the attention score between positions $i$ and $j$ is computed only if they have a parent-dependent relation\footnote{For multiplicative masks, $M(i,j) = 0$ does not mean that the final attention weight between $j$ and $i$ is zero, because an input of zero to the softmax function does not yield a zero probability. A method to strictly ``mask out'' an entry of $\mathrm{Softmax}\left(\frac{\mathbf{H}^{q} [\mathbf{H}^{k}]^\top}{\sqrt{d}}\right)$ is to use an additive mask and set $M(i,j) = -\infty$ if $\omega(i,j) > \omega_{\mathrm{max}}$.}.
\item \vspace{0.3em} \textbf{Soft Masking}. Instead of treating $\mathbf{M}$ as a hard constraint, we can use it as a soft constraint that scales the attention weight between positions $i$ and $j$ based on the degree to which the corresponding words are correlated. One approach is to reduce the attention weight as $\omega(i,j)$ becomes larger. A simple method is to transform $\omega(i,j)$ such that $M(i,j)$ is negatively correlated with $\omega(i,j)$ and falls within the interval $[0,1]$:
    \begin{eqnarray}
    M(i,j) & = & \mathrm{DNorm}(\omega(i,j)).
    \end{eqnarray}

    There are several alternative designs for $\mathrm{DNorm}(\cdot)$. For example, one can compute a standardized score of $-\omega(i,j)$ by subtracting its mean and dividing by its standard deviation~\citep{chen-etal:2018syntax}, or normalize $1/\omega(i,j)$ over all possible $j$ in the sequence~\citep{xu-etal:2021syntax}. In cases where parsers can output a score between positions $i$ and $j$, this score can also be used to compute $M(i,j)$. For example, a dependency parser can produce the probability of the word at position $i$ being the parent of the word at position $j$~\citep{strubell-etal:2018linguistically}. We can then write $M(i,j)$ as
    \begin{eqnarray}
    M(i,j) & = & \mathrm{Pr}_{\mathrm{parent}}(i|j)
    \end{eqnarray}

    or alternatively
    \begin{eqnarray}
    M(i,j) & = & \max\{\mathrm{Pr}_{\mathrm{parent}}(i|j), \mathrm{Pr}_{\mathrm{parent}}(j|i)\},
    \end{eqnarray}

    \noindent where $\mathrm{Pr}_{\mathrm{parent}}(i|j)$ and $\mathrm{Pr}_{\mathrm{parent}}(j|i)$ are the probabilities given by the parser. See Figure~\ref{fig:dependency-attention-prior} for an example of inducing a soft masking variable from a dependency parse tree.

\end{itemize}
\vspace{0.5em}

\subsection{Multi-branch Models}
\label{sec:transformer-multi-branch-model}

\noindent Integrating syntax into NLP systems presents several challenges. This is partly because automatic parse trees may contain errors, and partly because the use of syntax may lead to strong assumptions about the underlying structure of a sentence. Rather than combining syntactic and word information into a single monolithic model, it may be more flexible and effective to build one model to encode syntactic information and another to encode word sequences. One way to achieve this is through the use of multiple neural networks (referred to as \textbf{branches} or \textbf{paths}), each dealing with one type of input. The outputs of these branches are then combined to produce a final output~\citep{xie-etal:2017aggregated,fan-etal:2020multi,lin-etal:2022multi}. Various methods have therefore been used to combine different types of input for neural models like Transformers.

\begin{figure}[!htp]
\centering

\begin{center}

\begin{tikzpicture}

\def\rowsep{0.7cm}
\def\colsep{0.4cm}
\def\nodesize{1cm}

\tikzstyle{enode} = [inner sep=0pt,fill=black,rounded corners=2]

\begin{scope}

\node [anchor=south west,minimum width=\nodesize,minimum height=2.5*\nodesize,draw,very thick,fill=green!30] (tencoder) at (0,0) {};
\node [anchor=west,minimum width=\nodesize,minimum height=2.5*\nodesize,draw,very thick,fill=red!30] (xencoder) at ([xshift=0.3*\nodesize]tencoder.east) {};
\node [rotate=90] (tenclabel) at (tencoder) {$\small{\mathrm{Encode}_{\mathrm{syn}}(\cdot)}$};
\node [rotate=90] (xenclabel) at (xencoder) {$\small{\mathrm{Encode}_{\mathrm{text}}(\cdot)}$};

\draw [<-,very thick] ([yshift=-0.1cm]tencoder.south) -- ([yshift=-1.35em]tencoder.south) node [pos=1,below] {\small{$\mathbf{t}$}};
\draw [<-,very thick] ([yshift=-0.1cm]xencoder.south) -- ([yshift=-1.5em]xencoder.south) node [pos=1,below] {\small{$\mathbf{x}$}};

\node [anchor=south west,minimum width=(2 + 0.35)*\nodesize,minimum height=0.7*\nodesize,draw,very thick] (comb) at ([yshift=0.4*\nodesize]tencoder.north west) {\small{$\mathrm{Combine}(\cdot)$}};
\draw [->,very thick] ([yshift=0.1em]tencoder.north) -- ([yshift=0.4*\nodesize-0.1em]tencoder.north);
\draw [->,very thick] ([yshift=0.1em]xencoder.north) -- ([yshift=0.4*\nodesize-0.1em]xencoder.north);

\draw [->,very thick] ([yshift=0.1em]comb.north) -- ([yshift=1.5em]comb.north) node [pos=1,above] {\small{...}};

\node [anchor=north] (caption) at ([yshift=-3em,xshift=0.15*\nodesize]tencoder.south east) {\small{(a) Multi-encoder}};

\end{scope}

\begin{scope}[xshift=2in]

\node [anchor=south west,minimum width=3*\nodesize,minimum height=(0.7+2.5+0.4+0.1)*\nodesize,draw,very thick] (model) at (0,0) {};

\draw [<-,very thick] ([yshift=-0.1cm]model.south) -- ([yshift=-1.5em]model.south) node [pos=1,below] {\small{$\mathbf{x}$}};
\draw [->,very thick] ([yshift=0.1em]model.north) -- ([yshift=1.5em]model.north) node [pos=1,above] {\small{...}};

\node [anchor=south,minimum width=2.2*\nodesize,minimum height=0.4*\nodesize,draw,thick] (layer1) at ([yshift=0.3*\nodesize]model.south) {};
\node [anchor=north,minimum width=2.2*\nodesize,minimum height=0.4*\nodesize,draw,thick] (layern) at ([yshift=-0.3*\nodesize]model.north) {};

\node [anchor=west,minimum width=0.8*\nodesize,minimum height=1.2*\nodesize,draw,thick,fill=green!30] (tlayer) at ([xshift=0.4*\nodesize]model.west) {};
\node [anchor=east,minimum width=0.8*\nodesize,minimum height=1.2*\nodesize,draw,thick,fill=red!30] (xlayer) at ([xshift=-0.4*\nodesize]model.east) {};
\node [rotate=90] (tbranch) at (tlayer) {\scriptsize{$\mathrm{Branch}_1$}};
\node [rotate=90] (xbranch) at (xlayer) {\scriptsize{$\mathrm{Branch}_2$}};

\draw [->,thick] ([yshift=0.1em,xshift=-0.5em]layer1.north) -- ([yshift=-0.1em]tlayer.south);
\draw [->,thick] ([yshift=0.1em,xshift=0.5em]layer1.north) -- ([yshift=-0.1em]xlayer.south);
\draw [<-,thick] ([yshift=-0.1em,xshift=-0.5em]layern.south) -- ([yshift=0.1em]tlayer.north);
\draw [<-,thick] ([yshift=-0.1em,xshift=0.5em]layern.south) -- ([yshift=0.1em]xlayer.north);

\draw [<-,very thick,dotted] ([yshift=-0.8em,xshift=-0.1em]tlayer.north west) -- ([yshift=-0.8em,xshift=-2em]tlayer.north west) node [pos=1,left] {\small{$\mathbf{t}$}};

\node [anchor=north] (caption) at ([yshift=-3em,xshift=0]model.south) {\small{(b) Multi-branch as a Sub-model}};

\end{scope}

\begin{scope}[xshift=4.2in]

\node [anchor=south west,minimum width=3*\nodesize,minimum height=(0.7+2.5+0.4+0.1)*\nodesize,draw,very thick] (model) at (0,0) {};

\draw [<-,very thick] ([yshift=-0.1cm]model.south) -- ([yshift=-1.5em]model.south) node [pos=1,below] {\small{$\mathbf{x}$}};
\draw [->,very thick] ([yshift=0.1em]model.north) -- ([yshift=1.5em]model.north) node [pos=1,above] {\small{...}};

\node [anchor=south,minimum width=2.2*\nodesize,minimum height=0.4*\nodesize,draw,thick] (layer1) at ([yshift=0.3*\nodesize]model.south) {};
\node [anchor=north,minimum width=2.2*\nodesize,minimum height=0.4*\nodesize,draw,thick] (layern) at ([yshift=-0.3*\nodesize]model.north) {};

\node [anchor=center,minimum width=1.5*\nodesize,minimum height=0.8*\nodesize,draw,thick,fill=red!30] (head1) at ([xshift=0.6em,yshift=0.6em]model.center) {};
\node [anchor=center,minimum width=1.5*\nodesize,minimum height=0.8*\nodesize,draw,thick,fill=red!30] (head2) at ([xshift=0.2em,yshift=0.2em]model.center) {};
\node [anchor=center,minimum width=1.5*\nodesize,minimum height=0.8*\nodesize,draw,thick,fill=red!30] (head3) at ([xshift=-0.2em,yshift=-0.2em]model.center) {};
\node [anchor=center,minimum width=1.5*\nodesize,minimum height=0.8*\nodesize,draw,thick,fill=green!30] (head4) at ([xshift=-0.6em,yshift=-0.6em]model.center) {$\scriptsize{\mathrm{Head}_1}$};

\draw [<-,very thick,dotted] ([yshift=-0.8em,xshift=-0.1em]head4.north west) -- ([yshift=-0.8em,xshift=-2em]head4.north west) node [pos=1,left] {\small{$\mathbf{t}$}};

\draw [->,thick] ([yshift=0.1em]layer1.north) -- ([yshift=0.4*\nodesize]layer1.north);
\draw [<-,thick] ([yshift=-0.1em]layern.south) -- ([yshift=-0.4*\nodesize]layern.south);

\node [anchor=north] (caption) at ([yshift=-3em,xshift=0]model.south) {\small{(c) Multi-head Attention}};

\end{scope}

\end{tikzpicture}

\end{center}
\caption{Multi-branch architectures. There are two inputs: a sentence (denoted by $\mathbf{x}$) and the syntax tree of the sentence (denoted by $\mathbf{t}$). In the multi-encoder architecture (see sub-figure (a)), two encoders are constructed to encode $\mathbf{x}$ and $\mathbf{t}$, respectively. A combination model then takes the outputs of the encoders and produces a combined representation of $\mathbf{x}$ and $\mathbf{t}$. The idea of multi-branch networks can be used for designing sub-models of the encoder. A simple example is that we create multiple paths in parallel for some layers of the encoder (see sub-figure (b)). Another example is multi-head attention (see sub-figure (c)) where we use different heads to learn different representations. }
\label{fig:multi-branch-transformer}
\end{figure}
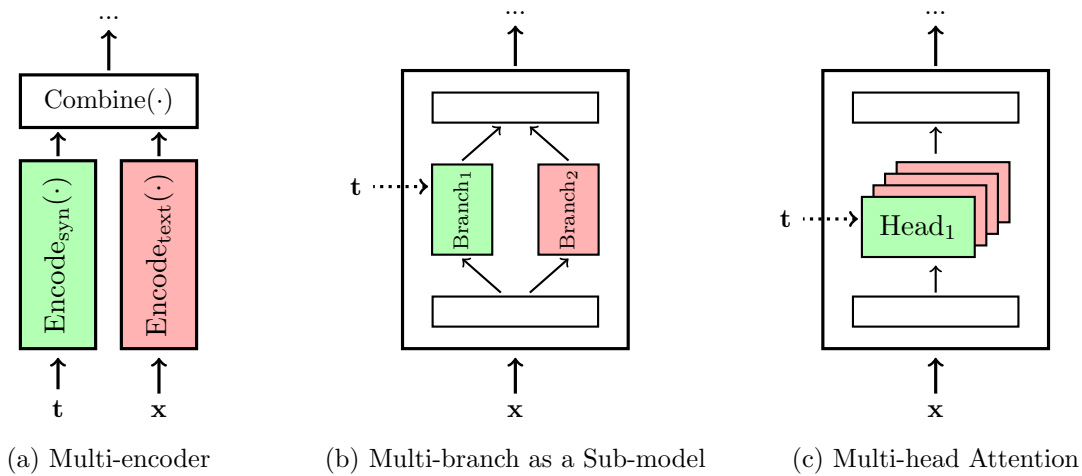

One commonly used approach is to build two separate encoders, in which one model is trained to encode the syntactic input (denoted by $\mathbf{t}$), and the other is trained to encode the standard textual input (denoted by $\mathbf{x}$). Figure~\ref{fig:multi-branch-transformer}(a) illustrates this multi-encoder architecture. The syntactic encoder $\mathrm{Encode}_{\mathrm{syn}}(\mathbf{t})$ is based on models presented in Sections~\ref{sec:syntax-input-and-output} and~\ref{sec:syntax-aware-attention}, and the text encoder $\mathrm{Encode}_{\mathrm{text}}(\mathbf{x})$ is a standard Transformer encoder. The representations generated by these encoders are then fed into the combination model as input and combined into a hybrid representation, given by
\begin{eqnarray}
\mathbf{H}_{\mathrm{hybrid}} & = & \mathrm{Combine}(\mathbf{H}_{\mathrm{syn}},\mathbf{H}_{\mathrm{text}}) \nonumber \\
                             & = & \mathrm{Combine}(\mathrm{Encode}_{\mathrm{syn}}(\mathbf{t}),\mathrm{Encode}_{\mathrm{text}}(\mathbf{x})). \label{eq:transformer-multi-encoders}
\end{eqnarray}

There are several designs for $\mathrm{Combine}(\cdot)$, depending on the tasks to which the encoders are applied. For example, if we want to develop a text classifier, $\mathrm{Combine}(\cdot)$ can be a simple pooling network. For more complicated tasks, such as machine translation, $\mathrm{Combine}(\cdot)$ can also be a Transformer encoder, and we can fuse information from different sources by performing self-attention on $[\mathbf{H}_{\mathrm{syn}},\mathbf{H}_{\mathrm{text}}]$.

Although we focus primarily on syntactic models in this section, the general multi-encoder architecture can be used in many tasks where inputs from additional sources are required. For example, one can use one encoder to represent a sentence and another to represent the previous sentence in the same document. Combining the two encoders thus yields a context-aware model~\citep{voita-etal:2018context,li-etal:2020does}. Furthermore, the architectures of the encoders do not need to be restricted to Transformers. We can choose different models for different branches. For instance, in a widely used two-branch encoding architecture, we can employ a CNN-based encoder to model local context and a Transformer encoder to model global context~\citep{wu-etal:2020lite}.

Sub-models of Transformers can also be multi-branch neural networks. See Figure~\ref{fig:multi-branch-transformer} (b) for an example involving two self-attention branches. One is the standard self-attention network $\mathrm{Att}_{\mathrm{self}}(\mathbf{H})$, and the other is the syntax-aware self-attention network $\mathrm{AttSyn}_{\mathrm{self}}(\mathbf{H})$. The output of the self-attention model is a linear combination of the outputs of these two branches~\citep{xu-etal:2021syntax}, given by
\begin{eqnarray}
\mathbf{H}_{\mathrm{self}} & = & \alpha \cdot \mathrm{Att}_{\mathrm{self}}(\mathbf{H}) + (1 - \alpha) \cdot \mathrm{AttSyn}_{\mathrm{self}}(\mathbf{H}),
\end{eqnarray}

\noindent where $\alpha$ is a combination coefficient. $\mathbf{H}_{\mathrm{self}}$ can be used as usual by applying a layer normalization function and adding a residual connection, thus keeping the overall architecture identical to standard Transformer models.

Multi-head attention networks can also be viewed as a form of multi-branch models. Therefore, we can provide guidance from syntax to only some of the heads while keeping the rest unchanged~\citep{strubell-etal:2018linguistically}. This approach is illustrated in Figure~\ref{fig:multi-branch-transformer} (c) where only one head of the self-attention sub-layer makes use of syntax trees for computing attention weights.

\subsection{Multi-scale Models}
\label{sec:transformer-multi-scale-models}

\noindent In linguistics, syntax studies how sentences are built from smaller constituents. Different levels of these constituents are generally organized in a hierarchical structure, called a \textbf{syntactic hierarchy}. It is therefore possible to use multiple levels of syntactic constituents to explain the same sentence. For example, words capture how the sentence is constructed from small meaningful units, and phrases capture how the sentence is constructed from larger linguistic units.

Multi-scale Transformers leverage different levels of abstraction in data to represent a sentence using features at multiple scales. A common approach is to write a sentence in multiple different forms and then combine them using a multi-branch network~\citep{hao-etal:2019multi}. For example, consider the following sentence:

\vspace{0.5em}
\begin{center}
The oldest beer-making facility was discovered in China.
\end{center}
\vspace{0.5em}

\noindent We can tokenize it into a sequence of words, denoted by

\vspace{0.5em}
\begin{center}
$\mathbf{x}_{\mathrm{words}}=$ The oldest beer-making facility was discovered in China .
\end{center}
\vspace{0.5em}

\noindent Alternatively, we can write it as a sequence of phrases by using a parser, denoted by

\vspace{0.5em}
\begin{center}
$\mathbf{x}_{\mathrm{phrases}}=$ [The oldest beer-making facility]$_{\textrm{NP}}$ [was discovered in China]$_{\textrm{VP}}$ [.]$_{\textrm{.}}$
\end{center}
\vspace{0.5em}

The simplest way to build a multi-scale model is to encode $\mathbf{x}_{\mathrm{words}}$ and $\mathbf{x}_{\mathrm{phrases}}$ using two separate Transformer encoders. The outputs of these encoders are then combined in some way. This leads to the same form as Eq. (\ref{eq:transformer-multi-encoders}), and we can view this model as an instance of the general multi-encoder architecture.

Both $\mathbf{x}_{\mathrm{words}}$ and $\mathbf{x}_{\mathrm{phrases}}$ can be viewed as sequences of tokens. For example, $\mathbf{x}_{\mathrm{words}}$ has nine word-based tokens, and $\mathbf{x}_{\mathrm{phrases}}$ has three phrase-based tokens\footnote{$\mathbf{x}_{\mathrm{phrases}}$ comprises three tokens: \textit{The oldest beer-making facility}, \textit{was discovered in China}, and \textit{.}}. However, enumerating all possible phrases would result in a huge vocabulary. We therefore need a method to represent each phrase as an embedding in a computationally efficient manner. By treating phrase embedding as a sequence modeling problem, it becomes straightforward to learn sub-sequence representations by employing various sequence models. We now have a two-stage learning process. In the first stage, we learn the embeddings of input units at different scales using separate models. In the second stage, we learn to encode sequences at different scales using a multi-branch model.

More generally, we do not need to restrict ourselves to linguistically meaningful units in multi-scale representation learning. For example, we can learn sub-word segmentations from data and represent an input sentence as a sequence of sub-words. This results in a hierarchical representation of the sentence (e.g., sub-words $\to$ words $\to$ phrases). While the learned sub-words may not have linguistic meanings, they provide new insights into modeling words and phrases, as well as a new scale of features. Also, we do not strictly need to develop multiple encoders for multi-scale modeling. An alternative approach is to incorporate representations at different scales directly into the multi-head self-attention modules, which makes it easier to model the interactions among different scales~\citep{guo-etal:2020multi,li-etal:2022learningmultiscale}.

A problem with the approaches described above, however, is that the representations (or attention weight matrices) learned at different scales are of different sizes. For example, in the scenarios above, the representation learned from $\mathbf{x}_{\mathrm{words}}$ is a $9 \times d$ matrix, while the representation learned from $\mathbf{x}_{\mathrm{phrases}}$ is a $3 \times d$ matrix. A simple solution to this problem is to perform upsampling on the phrase-based representation to expand it to a $9 \times d$ matrix. Likewise, we can perform downsampling on the word-based representation to shrink it to a $3 \times d$ matrix. The combination model $\mathrm{Combine}(\cdot)$ can then be implemented as described in Section~\ref{sec:transformer-multi-branch-model}.

It is worth noting that multi-scale modeling is widely discussed across several fields. In computer vision, for example, multi-scale modeling is often referred to as the process of learning a series of feature maps from the input image~\citep{fan-etal:2021multiscale,li-etal:2022mvitv2}. Unlike the multi-branch models presented here, multi-scale vision Transformer models make use of the hierarchical nature of features when representing images. Systems of this kind are typically based on a stack of layers in which each layer learns features at a larger scale (e.g., a higher channel capacity) from the features produced by the previous layer.

\subsection{Transformers as Syntax Learners}

\noindent So far we have discussed syntax trees as constraints or priors on the encoding process so that we can make use of linguistic representations in learning neural networks. It is natural to wonder whether these neural models can learn some knowledge of linguistic structure directly from data, without human-designed linguistic annotations. This reflects a fundamental goal in NLP: to learn linguistic knowledge from data and encode it implicitly within models.

To explore the linguistic properties learned by NLP systems, a straightforward method is to examine the syntactic behaviors of the systems' outputs. For example, we can examine whether the outputs of language generation systems contain grammatical errors. Another approach is to ask these systems to accomplish tasks that are linguistically meaningful, even though they were not explicitly trained to do so~\citep{brown-etal:2020language}. However, merely examining how model predictions exhibit syntactic abilities is not sufficient to fully answer the question. It is also possible that neural networks learn linguistic knowledge but do not utilize it during prediction~\citep{clark-etal:2019does}. Therefore, we need to investigate what is modeled and learned inside the neural networks themselves.

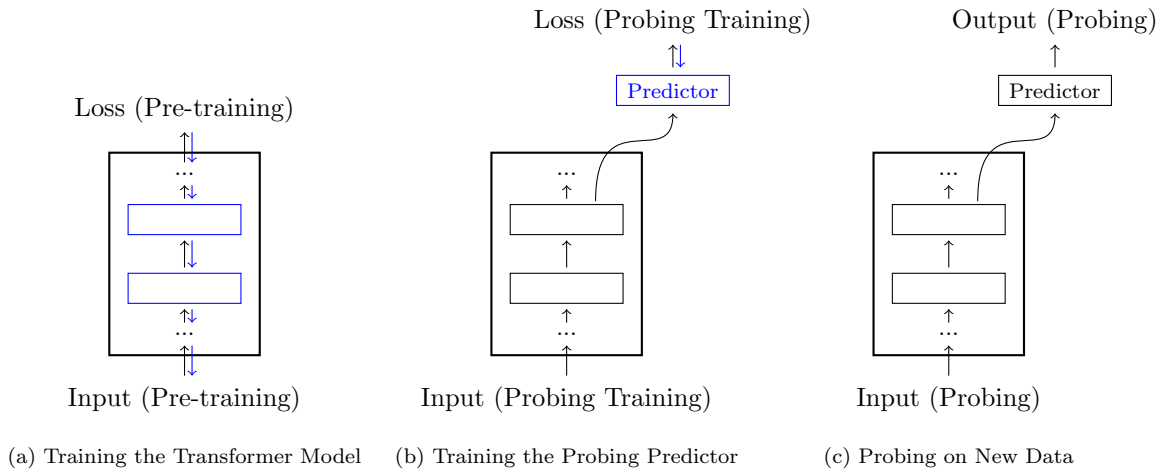
\begin{figure}[!htp]
\centering

\begin{center}

\begin{tikzpicture}

\def\rowsep{0.5cm}
\def\colsep{0.4cm}
\def\nodesize{1cm}

\tikzstyle{layer} = [inner sep=2pt,minimum width=1.5cm,minimum height=0.4cm,draw]

\begin{scope}

\node [anchor=west,layer,blue] (layer1) at (0,0) {};
\node [anchor=south,layer,blue] (layer2) at ([yshift=\rowsep]layer1.north) {};
\node [anchor=north] (low) at ([yshift=-0.5*\rowsep]layer1.south) {\footnotesize{...}};
\node [anchor=south] (high) at ([yshift=0.5*\rowsep]layer2.north) {\footnotesize{...}};
\node [anchor=north] (input) at ([yshift=-0.8*\rowsep]low.south) {\small{Input (Pre-training)}};
\node [anchor=south] (output) at ([yshift=0.8*\rowsep]high.north) {\small{Loss (Pre-training)}};

\begin{pgfonlayer}{background}
\node [draw,thick,minimum width=2cm] [fit=(layer1) (low) (high)] (model) {};
\end{pgfonlayer}

\draw [->] (input.north) -- (low.south);
\draw [->] (low.north) -- ([yshift=-2pt]layer1.south);
\draw [->] ([yshift=2pt]layer1.north) -- ([yshift=-2pt]layer2.south);
\draw [->] ([yshift=2pt]layer2.north) -- (high.south);
\draw [->] (high.north) -- (output.south);

\draw [<-,blue] ([xshift=0.1cm]input.north) -- ([xshift=0.1cm]low.south);
\draw [<-,blue] ([xshift=0.1cm]low.north) -- ([yshift=-2pt,xshift=0.1cm]layer1.south);
\draw [<-,blue] ([yshift=2pt,xshift=0.1cm]layer1.north) -- ([yshift=-2pt,xshift=0.1cm]layer2.south);
\draw [<-,blue] ([yshift=2pt,xshift=0.1cm]layer2.north) -- ([xshift=0.1cm]high.south);
\draw [<-,blue] ([xshift=0.1cm]high.north) -- ([xshift=0.1cm]output.south);

\node [anchor=north] (caption) at ([yshift=-0.5em]input.south) {\scriptsize{(a) Training the Transformer Model}};

\end{scope}

\begin{scope}[xshift=2in]

\node [anchor=west,layer] (layer1) at (0,0) {};
\node [anchor=south,layer] (layer2) at ([yshift=\rowsep]layer1.north) {};
\node [anchor=north] (low) at ([yshift=-0.5*\rowsep]layer1.south) {\footnotesize{...}};
\node [anchor=south] (high) at ([yshift=0.5*\rowsep]layer2.north) {\footnotesize{...}};
\node [anchor=north] (input) at ([yshift=-0.8*\rowsep]low.south) {\small{Input (Probing Training)}};

\begin{pgfonlayer}{background}
\node [draw,thick,minimum width=2cm] [fit=(layer1) (low) (high)] (model) {};
\end{pgfonlayer}

\draw [->] (input.north) -- (low.south);
\draw [->] (low.north) -- ([yshift=-2pt]layer1.south);
\draw [->] ([yshift=2pt]layer1.north) -- ([yshift=-2pt]layer2.south);
\draw [->] ([yshift=2pt]layer2.north) -- (high.south);

\node [anchor=south west,layer,blue] (probinglayer) at ([yshift=1.5*\rowsep,xshift=1em]high.north east) {\scriptsize{Predictor}};
\node [anchor=south] (output) at ([yshift=0.8*\rowsep]probinglayer.north) {\small{Loss (Probing Training)}};

\draw [->] ([yshift=0.1cm]probinglayer.north) -- ([yshift=0]output.south);
\draw [<-,blue] ([xshift=0.1cm,yshift=0.1cm]probinglayer.north) -- ([xshift=0.1cm]output.south);
\draw [->] ([xshift=1em,yshift=0.1em]layer2.north) .. controls +(north:3.5em) and +(south:2em) .. ([yshift=-0.1cm]probinglayer.south);

\node [anchor=north] (caption) at ([yshift=-0.5em]input.south) {\scriptsize{(b) Training the Probing Predictor}};

\end{scope}

\begin{scope}[xshift=4in]

\node [anchor=west,layer] (layer1) at (0,0) {};
\node [anchor=south,layer] (layer2) at ([yshift=\rowsep]layer1.north) {};
\node [anchor=north] (low) at ([yshift=-0.5*\rowsep]layer1.south) {\footnotesize{...}};
\node [anchor=south] (high) at ([yshift=0.5*\rowsep]layer2.north) {\footnotesize{...}};
\node [anchor=north] (input) at ([yshift=-0.8*\rowsep]low.south) {\small{Input (Probing)}};

\begin{pgfonlayer}{background}
\node [draw,thick,minimum width=2cm] [fit=(layer1) (low) (high)] (model) {};
\end{pgfonlayer}

\draw [->] (input.north) -- (low.south);
\draw [->] (low.north) -- ([yshift=-2pt]layer1.south);
\draw [->] ([yshift=2pt]layer1.north) -- ([yshift=-2pt]layer2.south);
\draw [->] ([yshift=2pt]layer2.north) -- (high.south);

\node [anchor=south west,layer] (probinglayer) at ([yshift=1.5*\rowsep,xshift=1em]high.north east) {\scriptsize{Predictor}};
\node [anchor=south] (output) at ([yshift=0.8*\rowsep]probinglayer.north) {\small{Output (Probing)}};

\draw [->] ([yshift=0.1cm]probinglayer.north) -- ([yshift=0]output.south);
\draw [->] ([xshift=1em,yshift=0.1em]layer2.north) .. controls +(north:3.5em) and +(south:2em) .. ([yshift=-0.1cm]probinglayer.south);

\node [anchor=north] (caption) at ([yshift=-0.5em]input.south) {\scriptsize{(c) Probing on New Data}};

\end{scope}

\end{tikzpicture}

\end{center}
\caption{An overview of probing for Transformer-based models. Given a Transformer model (e.g., a Transformer-based language model), we first optimize the model parameters on some unlabeled data. Then, we develop a predictor which takes the states of a hidden layer of the Transformer model and generates outputs for a probing task (see sub-figure (a)). The predictor can be trained as usual in which only the parameters of the predictor are optimized and the parameters of the Transformer model are fixed (see sub-figure (b)). The Transformer model and the predictor are used together to make predictions on new data for probing (see sub-figure (c)).}
\label{fig:transformer-probing}
\end{figure}

One approach to examining the latent linguistic structure in Transformer models is to develop \textbf{probes} to determine whether and to what extent these models capture linguistic notions such as dependency relations and parts-of-speech. A general probing approach involves extracting the internal representations of the models and probing them for linguistic phenomena. For Transformers, this is usually achieved by examining the attention maps and/or the outputs of attention layers. We then construct a \textbf{probing predictor} (or \textbf{probing classifier}) that takes these internal representations as input and outputs the corresponding linguistic notions~\citep{belinkov:2022probing}. The probing predictor can be based on either simple heuristics or parameterized models optimized specifically for the probing task. Recent work has shown that large-scale Transformer-based language models exhibit strong capabilities, referred to as \textbf{emergent abilities}, across various probing tasks. Figure~\ref{fig:transformer-probing} illustrates this probing process.

Many probing methods have been used in recent work on analyzing and understanding what is learned in neural encoders. Here we describe some of the popular ones.

\begin{itemize}
\item \vspace{0.5em} \textbf{Trees}. Given a trained Transformer encoder, it is easy to determine how likely it is that two words in a sentence share a linguistic relationship by computing the attention weight between them. We can use this quantity to define a metric measuring the syntactic distance between two words at positions $i$ and $j$
    \begin{eqnarray}
    d_{s}(i,j) & = & 1 - \alpha(i,j).
    \end{eqnarray}
    By using this metric, it is straightforward to construct the \textbf{minimum-spanning tree} for the sentence. That is, we connect all the words to form a tree structure with the minimum total distance. This tree structure can be viewed as a latent tree representation of the sentence induced from the neural network. While this dependency-tree-like structure can serve as a source of learned syntactic information for downstream tasks, it says little about the model's actual knowledge of syntax. An approach to aligning the representations in the encoder with linguistic structure is to learn to produce syntax trees that are consistent with human annotations. To achieve this, we need to develop a probing predictor trained on tree-annotated data. Suppose we have a human-annotated dependency tree for a given sentence. For each pair of words, we can obtain a distance $\omega(i,j)$ by counting the number of edges between them. Then, we can learn a distance metric based on the internal representations of the encoder to approximate $\omega(i,j)$. A simple form of such a metric relies on the Euclidean distance~\citep{manning-etal:2020emergent}. Let $\mathbf{A} \in \mathbb{R}^{d \times k_s}$ be a parameter matrix. The squared Euclidean distance is given by
    \begin{eqnarray}
    d_{s}^2(i,j) & = & \left\| (\mathbf{h}_i - \mathbf{h}_j) \mathbf{A} \right\|_2^{2},
    \end{eqnarray}
    \noindent where $\mathbf{h}_i$ and $\mathbf{h}_j$ are the representations produced by an encoding layer at positions $i$ and $j$\footnote{In general, $\mathbf{h}_i$ and $\mathbf{h}_j$ are the outputs of the last layer of the encoder. Alternatively, they can be weighted sums of the outputs of all the layers.}. Given a set of tree-annotated sentences $S$, we can optimize the model by
    \begin{eqnarray}
    \hat{\mathbf{A}} & = & \argmin_{\mathbf{A}} \sum_{s \in S} \frac{1}{|s|^2} \sum_{i \in s, j \in s} |\omega(i,j) - d_{s}^2(i,j)|,
    \end{eqnarray}

    \noindent where $|s|$ is the length of the sentence $s$, and $(i,j)$ indicates a pair of words in $s$. The optimized model is then used to parse test sentences via the minimum-spanning tree algorithm, and we can compare the parse trees against the human-annotated trees. To obtain directed trees, which are standard forms of dependency syntax, one can update the above model by considering the relative distance of a word to the root. More details can be found in \citet{manning-etal:2020emergent}'s work. Here the probing predictor functions similarly to a neural parser, trained to predict a syntax tree based on a representation of the input sentence.  This idea can be extended to other forms of syntactic structure, such as phrase structure trees~\citep{shi-etal:2016does}.

\item \vspace{0.3em} \textbf{Syntactic and Semantic Labels}. Many syntactic and semantic parsing tasks can be framed as problems of predicting linguistic labels given a sentence or its segments. A simple example is part-of-speech tagging in which each word of a sentence is labeled with a word class. A probe for part-of-speech tagging can be a classifier that takes a representation $\mathbf{h}_j$ each time and outputs the corresponding word class. One general probing approach to these problems is \textbf{edge probing}~\citep{tenney-etal:2019what,tenney-etal:2019bert}. Given a sentence, a labeled edge is defined as a tuple
    \begin{eqnarray}
    (\mathrm{span}_1,\mathrm{span}_2,\mathrm{label}) \nonumber,
    \end{eqnarray}

    \noindent where $\mathrm{span}_1$ is a span $[i_1,j_1]$, and $\mathrm{span}_2$ is an optional second span $[i_2,j_2]$, and $\mathrm{label}$ is the corresponding label. Our goal is to learn a probe to predict $\mathrm{label}$ given $\mathrm{span}_1$ and $\mathrm{span}_2$. For example, for part-of-speech tagging, $\mathrm{span}_1$ is a unit span $[j,j]$ for each position $j$, $\mathrm{span}_2$ is an empty span, and $\mathrm{label}$ is the part-of-speech tag corresponding to the $j$-th word of the sentence. For dependency parsing and coreference resolution, $\mathrm{span}_1$ and $\mathrm{span}_2$ are two words or entities, and $\mathrm{label}$ is the relationship between them. For constituency parsing, $\mathrm{span}_1$ is a span of words, $\mathrm{span}_2$ is an empty span, and $\mathrm{label}$ is the syntactic category of the tree node yielding $\mathrm{span}_1$. In simple cases, the probing model can be a multi-layer feed-forward neural network with a softmax output layer. As usual, this model is trained on labeled data, and then tested on new data.

\item \vspace{0.3em} \textbf{Surface Forms of Words and Sentences}. Probing tasks can also be designed to examine whether the representations encode the surface information of sentences or words~\citep{adifine-etal:2016fine,conneau-etal:2018you}. A simple sentence-level probing task is \textbf{sentence length prediction}. To do this, we first represent the sentence as a single vector $\mathbf{h}$\footnote{$\mathbf{h}$ can be computed by performing a pooling operation on $\{\mathbf{h}_1, \cdots ,\mathbf{h}_m\}$}, and then build a classifier to categorize $\mathbf{h}$ into the corresponding length bin. Similarly, probes can be built to predict whether two words at positions $i$ and $j$ are reordered in the sentence given $\mathbf{h}_i$ and $\mathbf{h}_j$. Also, we can develop probes to address conventional problems in morphology. For example, we reconstruct the word at position $j$ or predict its sense with the representation $\mathbf{h}_j$. In addition, probing tasks can be focused on particular linguistic problems, for example, numeracy~\citep{wallace-etal:2019nlp} and function words~\citep{kim-etal:2019probing}.

\item \vspace{0.3em} \textbf{Cloze}. Of course, we can probe neural models for problems beyond syntax and morphology. One perspective on large-scale pre-trained Transformer models is to view them as knowledge bases containing facts about the world. It is therefore tempting to see if we can apply them to test factual knowledge. A simple method is to ask a probe to recover the missing item of a sentence~\citep{petroni-etal:2019language}. For example, if we have a cloze test

    \vspace{0.5em}
    \begin{center}
    Shiji was written by \underline{\hspace{2em}}.
    \end{center}
    \vspace{0.5em}

    we wish the probe to give an answer \textit{Sima Qian} because there is a subject-relation-object fact (Shiji, written-by, Sima Qian). This probe can simply be a \textbf{masked language model} that is widely used in self-supervised learning of Transformer encoders.

\end{itemize}
\vspace{0.5em}

In NLP, probing is closely related to the pre-training of large language models. In general, we can view probing tasks as downstream applications of these pre-trained models, though probing is ordinarily used as an analytical tool to quickly test the capabilities of the models. Ideally, we want to develop a probe that makes the best use of the internal representations to solve a given task. However, when a probe is complex and sufficiently well-trained, it becomes difficult to determine whether the problem was solved by the rich information in the representations or by the modeling capacity of the probe itself. A common way to isolate the contribution of the representations is to compare the probes against reasonable baselines or to conduct comparisons using control tasks~\citep{hewitt-etal:2019designing,belinkov:2022probing}.


\section{Improved Architectures}

\noindent In this section we present several improvements to the vanilla Transformer model. Unlike the previous section, most of the improvements are from the perspective of machine learning, rather than linguistics.

\subsection{Locally Attentive Models}
\label{sec:transformer-local-models}

\noindent The self-attention methods discussed in Section~\ref{sec:multi-head-self-attention} can be viewed as learning representations over the entire input sequence. While this global attention mechanism provides a strong ability to capture long-distance dependencies, it does not explicitly model local information. Here, we consider a few techniques that attempt to explicitly capture locality in representations.

\subsubsection{Priors for Local Modeling}

\noindent One of the simplest ways to introduce local context modeling into Transformers is to add a penalty term to the attention function, thereby discouraging large attention weights between distant positions. On the encoder side, this leads to a form that we have already encountered several times in this chapter:
\begin{eqnarray}
\mathrm{AttLocal}_{\mathrm{self}}(\mathbf{H}) & = & \mathrm{Softmax} \left(\frac{\mathbf{H}^{q} [\mathbf{H}^{k}]^\top}{\sqrt{d}} - \gamma \cdot \mathbf{G} \right) \mathbf{H}^{v}, \label{eq:self-attention-local}
\end{eqnarray}

\noindent where $\gamma$ is the weight (or temperature) of the penalty term, and $\mathbf{G} \in \mathbb{R}^{m \times m}$ is the penalty matrix. Each entry $G(i,j)$ indicates how heavily we penalize the model for attending to position $j$ from position $i$. A simple choice for $G(i,j)$ is the absolute distance between $i$ and $j$, for example
\begin{eqnarray}
G(i,j) & = & |i - j|.
\end{eqnarray}

\noindent Alternatively, $G(i,j)$ can be defined using a Gaussian penalty function~\citep{yang-etal:2018modeling}
\begin{eqnarray}
G(i,j) & = & \frac{(i - j)^2}{2 \sigma_i^2},
\end{eqnarray}

\noindent where $\sigma_i$ is the standard deviation of the Gaussian distribution. For a given $i$, both penalty terms increase (either linearly or exponentially) as the distance $|i-j|$ grows.

This method can be extended to the cross-attention model, as follows
\begin{eqnarray}
\mathrm{AttLocal}_{\mathrm{cross}}(\mathbf{H},\mathbf{S}) & = & \mathrm{Softmax} \left(\frac{\mathbf{S}^{q} [\mathbf{H}^{k}]^\top}{\sqrt{d}} - \gamma \cdot \mathbf{G} \right) \mathbf{H}^{v}, \label{eq:self-attention-cross}
\end{eqnarray}

\noindent where $\mathbf{G}$ is an $n \times m$ matrix. Each entry of $\mathbf{G}$ can be defined as
\begin{eqnarray}
G(i,j) & = & \frac{(\mu_i - j)^2}{2 \sigma_i^2},
\end{eqnarray}

\noindent where $\mu_i$ is the mean of the Gaussian distribution over the source-side positions. Both $\mu_i$ and $\sigma_i$ can be determined using heuristics. Alternatively, we can develop additional neural networks to model them, learning their parameters jointly with the rest of the model. For example, we could use a feed-forward neural network to predict $\mu_i$ given $\mathbf{s}_i$.

An alternative to Eq. (\ref{eq:self-attention-local}) (or Eq. (\ref{eq:self-attention-cross})) is to treat the penalty term as a separate distribution and linearly combine it with the original attention model. For example, we can define the self-attention model as
{\small
\begin{eqnarray}
\mathrm{AttLocal}_{\mathrm{self}}(\mathbf{H}) & = & \left((1-\beta) \cdot \mathrm{Softmax} \left(\frac{\mathbf{H}^{q} [\mathbf{H}^{k}]^\top}{\sqrt{d}} \right) + \beta \cdot \mathrm{Softmax} \left( - \gamma \cdot \mathbf{G} \right) \right) \mathbf{H}^{v},
\end{eqnarray}
}

\noindent where $\beta \in [0,1]$ is the linear combination coefficient. Note that, to avoid manually tuning the hyperparameter $\beta$, we can use a gating network to dynamically predict $\beta$ and train it end-to-end.

Another alternative is to use a multiplicative mask to incorporate the prior into the modeling, similar to Eq. (\ref{eq:self-attention-syn-mul-masking}). This is given by
\begin{eqnarray}
\mathrm{AttLocal}_{\mathrm{self}}(\mathbf{H}) & = & \mathrm{Softmax} \left(\frac{\mathbf{H}^{q} [\mathbf{H}^{k}]^\top}{\sqrt{d}} \odot \mathbf{G}' \right) \mathbf{H}^{v}. \label{eq:self-attention-local-multiplicative}
\end{eqnarray}

\noindent Here, $\mathbf{G}' \in [0,1]^{m \times m}$ is a matrix with entries in $[0,1]$. The scalar $G'(i,j)$ takes a maximum value of 1 when $i=j$ and decays as $j$ moves further away from $i$. $G'(i,j)$ can be obtained by normalizing $-G(i,j)$ over all $j$, or by using other decaying functions.

\subsubsection{Local Attention}

\noindent The term \textit{local attention} has been used broadly to describe a wide range of problems and to refer to many different models in the NLP literature. The methods discussed above are those that impose soft constraints on attention models. In fact, local attention has its origins in attempts to restrict the scope of attention models for modeling and computational considerations~\citep{luong-etal:2015effective}. Research in this area often focuses on introducing hard constraints, so that the resulting models can focus on parts of the input and ignore the rest. For example, we can predict a span of source-side positions to attend to, given a target-side position~\citep{sperber-etal:2018self,yang-etal:2018modeling,sukhbaatar-etal:2019adaptive}. Also, attention spans can be induced from syntax trees, for example, syntactic subtree structures can help narrow the attention scope. Thus, many syntax-constrained models can be viewed as instances of local attention (see Section~\ref{sec:transformer-multi-scale-models}). In addition, the concept of local attention can be generalized to a rich set of models, such as \textbf{sparse attention models}, although these models are often discussed in the context of efficient machine learning methods. We will see a few examples in Section~\ref{sec:transformer-efficient-models}.

In deep learning, one of the most widely used architectures for learning features from a restricted region of the input is CNNs. It is thus interesting to consider methods of combining CNNs and Transformer models to obtain the benefits of both approaches, for example, CNNs deal with short-term dependencies, and self-attention models deal with long-term dependencies. One approach is to build a two-branch sequence model where one branch is based on CNNs and the other is based on self-attention models~\citep{wu-etal:2020lite}. Another approach is to incorporate CNN layers into Transformer blocks in a way that enables the model to learn both local and global representations~\citep{wu-etal:2019pay,gulati-etal:2020conformer}.

\subsubsection{Relative Positional Embedding}

\noindent Relative positional embedding, also known as \textbf{relative positional representation} (\textbf{RPR}), is an improvement to the absolute positional embedding method used in standard Transformers\citep{shaw-etal:2018self,huang-etal:2018music}. The idea of RPR is that we model the distance between two positions in a sequence rather than giving each position a fixed representation. As a result, we have a pair-wise representation $\mathrm{PE}(i,j)$ for two positions $i$ and $j$. One simple way to define $\mathrm{PE}(i,j)$ is to consider it as a lookup table for all pairs of $i$ and $j$. More specifically, let $\mathbf{u}_\pi$ be a $d$-dimensional representation for a given distance $\pi$. The form of $\mathrm{PE}(i,j)$ in the vanilla RPR method is given by
\begin{eqnarray}
\mathrm{PE}(i,j) & = & \mathbf{u}_{\mathrm{clip}(j-i,k_{\mathrm{rpr}})},
\end{eqnarray}

\noindent where $\mathrm{clip}(x,k_{\mathrm{rpr}})$ is a function that clips $x$ in the interval $[-k_{\mathrm{rpr}},k_{\mathrm{rpr}}]$
\begin{eqnarray}
\mathrm{clip}(x,k_{\mathrm{rpr}}) & = & \max\{-k_{\mathrm{rpr}},\min\{x,k_{\mathrm{rpr}}\}\}. \label{eq:transformer-rpr-clip}
\end{eqnarray}

Thus, we have a model with parameters
\begin{eqnarray}
\mathbf{U}_{\mathrm{rpr}} & = & \begin{bmatrix} \mathbf{u}_{-k_{\mathrm{rpr}}} \\ \vdots \\ \mathbf{u}_{0} \\ \vdots \\ \mathbf{u}_{k_{\mathrm{rpr}}} \end{bmatrix}.
\end{eqnarray}

\noindent While this notation is somewhat informal, we can view $\mathbf{U}_{\mathrm{rpr}}$ as a matrix $\in \mathbb{R}^{(2k_{\mathrm{rpr}}+1) \times d}$, and select a row corresponding to $\mathrm{clip}(j-i,k_{\mathrm{rpr}})$ when computing RPR for given $i$ and $j$.

Using the above method, we can define three RPR models $\mathrm{PE}^{q}(i,j)$, $\mathrm{PE}^{k}(i,j)$ and $\mathrm{PE}^{v}(i,j)$ for queries, keys, and values, respectively. Then, following the form of Eq. (\ref{eq:transformer-basic-attention-model}), the output of the self-attention model at position $i$ can be written as
\begin{eqnarray}
\mathbf{c}_i & = & \sum_{j=1}^{m} \alpha_{i,j} \left[\mathbf{h}_{j}^{v} + \mathrm{PE}^{v}(i,j) \right] \nonumber \\
             & = & \sum_{j=1}^{m} \alpha_{i,j} \mathbf{h}_{j}^{v} + \sum_{j=1}^{m} \alpha_{i,j} \mathrm{PE}^{v}(i,j), \label{eq:transformer-rpr-representation}
\end{eqnarray}

\noindent where $\mathbf{h}_{j}^{v}$ is the $j$-th row vector of $\mathbf{H}^{v}$. This representation comprises two components: $\sum_{j=1}^{m} \alpha_{i,j} \mathbf{h}_{j}^{v}$ is the basic representation, and $\sum_{j=1}^{m} \alpha_{i,j} \mathrm{PE}^{v}(i,j)$ is the positional representation.

The attention weight $\alpha_{i,j}$ is computed in a regular way, but with additional terms $\mathrm{PE}^{q}(i,j)$ and $\mathrm{PE}^{k}(i,j)$ added to each query and key:
\begin{eqnarray}
\alpha_{i,j} & = & \mathrm{Softmax}(\frac{[\mathbf{h}_{i}^{q} + \mathrm{PE}^{q}(i,j)] [\mathbf{h}_{j}^{k} + \mathrm{PE}^{k}(i,j)]^\top}{\sqrt{d}}). \label{eq:transformer-rpr-alpha}
\end{eqnarray}

\begin{figure}[!htp]
\centering
\input{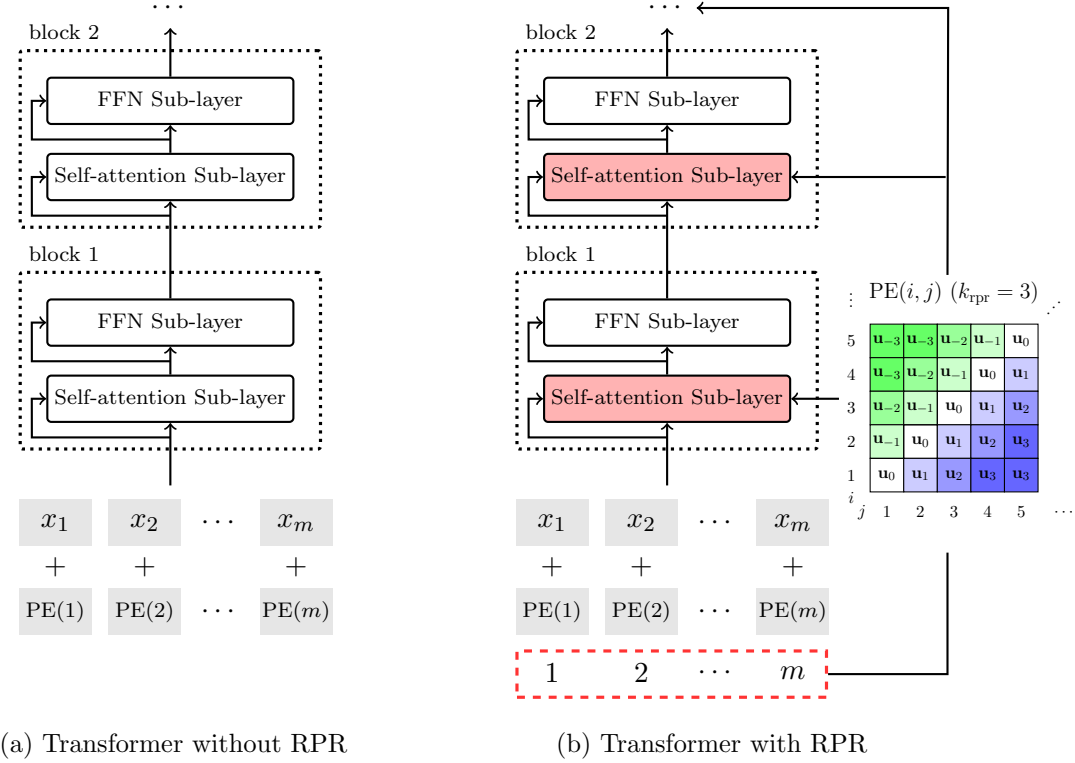}
\caption{Transformer encoders without and with relative positional representation (RPR). In RPR, each pair of positions is represented as a vector $\mathrm{PE}(i,j)$ using a model parameterized by $\mathbf{U}_{\mathrm{rpr}}$. $\mathrm{PE}(i,j)$ is fed into each self-attention sub-layer so that we can make use of the positional information in intermediate steps of learning representations.}
\label{fig:transformer-rpr}
\end{figure}

Figure~\ref{fig:transformer-rpr} shows the Transformer encoder architectures with and without RPR. When RPR is adopted, $\mathrm{PE}^{q}(i,j)$, $\mathrm{PE}^{k}(i,j)$, $\mathrm{PE}^{v}(i,j)$ are directly fed to each self-attention sub-layer, and so we can make better use of positional information for sequence modeling. Note that the use of the clipping function (see Eq. (\ref{eq:transformer-rpr-clip})) makes the modeling simple because we do not need to distinguish the relative distances for cases $|j-i| \ge k_{\mathrm{rpr}}$. This clipped distance-based model can lead to better modeling in local context windows.

Eqs. (\ref{eq:transformer-rpr-representation}) and (\ref{eq:transformer-rpr-alpha}) provide a general approach to position-sensitive sequence modeling. There are many variants of this model. In the early work on RPR by \citet{shaw-etal:2018self}, the positional representations for queries are removed, and the model works only with $\mathrm{PE}^{k}(i,j)$ and $\mathrm{PE}^{v}(i,j)$, like this
\begin{eqnarray}
\alpha_{i,j} & = & \mathrm{Softmax}(\frac{\mathbf{h}_{i}^{q} [\mathbf{h}_{j}^{k} + \mathrm{PE}^{k}(i,j)]^\top}{\sqrt{d}}). \label{eq:transformer-rpr-alpha-simple}
\end{eqnarray}

By contrast, some variants attempt to improve the RPR model when computing attention weights but ignore $\mathrm{PE}^{v}(i,j)$ when computing value representations~\citep{dai-etal:2019transformer,he-etal:2021deberta}. Instead of treating RPR as an additive term to each representation, researchers have also explored other ways of introducing RPR into the Transformer~\citep{huang-etal:2020improve,raffel-etal:2020exploring}. We refer interested readers to these papers for more details.

\subsection{Deep Models}
\label{sec:transformer-deep-models}

\noindent Many state-of-the-art NLP systems are based on deep Transformer models. For example, recent large language models generally comprise tens of Transformer layers (or more precisely, hundreds of layers), demonstrating strong performance on many tasks~\citep{ouyang-etal:2022training,touvron-etal:2023llama}. By stacking Transformer layers, it is straightforward to construct a deep model. However, as is often the case, training very deep neural networks is challenging. One difficulty arises from the fact that the error surfaces of deep neural networks are highly non-convex and contain many local optima, making the training process likely to get stuck. While there are optimization algorithms that can help alleviate this problem, most practical approaches rely on gradient-based methods for optimizing deep neural networks. As a result, training a model with many Transformer layers becomes challenging due to vanishing and exploding gradients during back-propagation. Here, we consider several techniques for training deep Transformer models.

\subsubsection{Re-thinking the Pre-Norm and Post-Norm Architectures}

\noindent As introduced previously, a Transformer sub-layer is a residual network where a shortcut is created to add the input of the network directly to the output of the sub-layer. This allows gradients to flow more directly from the output back to the input, mitigating the vanishing gradient problem. In general, a residual connection in a Transformer is used together with a layer normalization unit to form a sub-layer. This leads to two types of architectures, called post-norm and pre-norm. Specifically, recall from Section~\ref{sec:transformer-layer-normalization} that the post-norm architecture can be expressed as:
\begin{eqnarray}
\mathbf{z}^{l} & = & \mathrm{LNorm}(F^{l}(\mathbf{z}^{l-1}) + \mathbf{z}^{l-1}), \label{eq:post-norm-general}
\end{eqnarray}

\noindent where $\mathbf{z}^{l}$ and $\mathbf{z}^{l-1}$ are the output and input of the sub-layer $l$, respectively, and $F^{l}(\cdot)$ is the core function of this sub-layer. The pre-norm architecture takes the identity mapping $\mathbf{z}^{l-1}$ outside the layer normalization function, given in the form
\begin{eqnarray}
\mathbf{z}^{l} & = & F^{l}(\mathrm{LNorm}(\mathbf{z}^{l-1})) + \mathbf{z}^{l-1}. \label{eq:pre-norm-general}
\end{eqnarray}

Consider the difference between the information flow in these two architectures:

\begin{itemize}
\item \vspace{0.5em} The post-norm architecture prevents the identity mapping of the input from being directly added to the final output of the sub-layer. This is not a true residual network because all the information is passed through a non-linear function (i.e., the layer normalization unit). Thus, the post-norm architecture is less effective for back-propagation. \citet{wang-etal:2019learning} show that the gradient of the loss of an $L$-sub-layer Transformer network with respect to $\mathbf{z}^{l}$ is given by
    \begin{eqnarray}
    \frac{\partial E}{\partial \mathbf{z}^{l}} & = & \frac{\partial E}{\partial \mathbf{z}^{L}} \cdot \prod_{k=l}^{L-1} \frac{\partial \mathrm{LNorm}(\mathbf{v}^k)}{\partial \mathbf{v}^k} \cdot \prod_{k=l}^{L-1} \left( 1 + \frac{\partial F^{k}(\mathbf{z}^{k})}{\partial \mathbf{z}^{k}} \right),
    \end{eqnarray}

    \noindent where $\mathbf{z}^{L}$ is the output of the last layer, $\mathbf{v}^k$ is short for $F^{k}(\mathbf{z}^{k-1})$, and $E$ is the error measured by some loss function. $\frac{\partial \mathrm{LNorm}(\mathbf{v}^k)}{\partial \mathbf{v}^k}$ and $\frac{\partial F^{k}(\mathbf{z}^{k})}{\partial \mathbf{z}^{k}}$ are the gradients of the layer normalization function and the core function, respectively.  Although the equation here appears complex, we see that $\prod_{k=l}^{L-1} \frac{\partial \mathrm{LNorm}(\mathbf{v}^k)}{\partial \mathbf{v}^k}$ is simply a product of $L-l$ factors. This means that the error gradient will be rescaled repeatedly as $L$ becomes larger, leading to a higher risk of vanishing or exploding gradients in deeper models.

\item \vspace{0.3em} The pre-norm architecture describes a standard residual neural network where the input of the entire network is added directly to its output. We can write the gradient of the error at $\mathbf{z}^{l}$ as:
    \begin{eqnarray}
    \frac{\partial E}{\partial \mathbf{z}^{l}} & = & \frac{\partial E}{\partial \mathbf{z}^{L}} \cdot \left( 1 + \sum_{k=l}^{L-1} \frac{\partial F^{k}(\mathrm{LNorm}(\mathbf{z}^{k}))}{\partial \mathbf{z}^{k}} \right) \nonumber \\
    & = & \frac{\partial E}{\partial \mathbf{z}^{L}} + \frac{\partial E}{\partial \mathbf{z}^{L}} \cdot \sum_{k=l}^{L-1} \frac{\partial F^{k}(\mathrm{LNorm}(\mathbf{z}^{k}))}{\partial \mathbf{z}^{k}}.
    \end{eqnarray}

    It is easy to see that $\frac{\partial E}{\partial \mathbf{z}^{l}}$ receives direct feedback regarding the error, because the first term on the right-hand side (i.e., $\frac{\partial E}{\partial \mathbf{z}^{L}}$) is the gradient of the model output, which is independent of the network depth.
\end{itemize}
\vspace{0.5em}

The use of the pre-norm architecture also helps optimization during early gradient descent steps. For example, it has been found that pre-norm Transformer models can be trained using a larger learning rate in the early stages of training, instead of requiring a gradual learning rate warmup from a small value~\citep{xiong-etal:2020layer}.

While the pre-norm architecture facilitates easier optimization of deep Transformer models, we cannot simply conclude that it is a universally better choice compared to the post-norm architecture. In fact, both post-norm and pre-norm Transformer models have been successfully used in many applications. For example, the post-norm architecture is widely used in BERT-like models, while the pre-norm architecture is the more popular choice for recent generative large language models. Broadly, these two architectures provide different paradigms for designing deep Transformer models, each with distinct advantages and disadvantages. The post-norm architecture forces the representation to be learned through more non-linear functions, resulting in a highly expressive but complicated model that is relatively hard to train. By contrast, the pre-norm architecture makes the training of Transformer models easier, but it may be less expressive than its post-norm counterpart if the learned model becomes overly dependent on the shortcut paths.

An improvement to these architectures is to control the extent to which we want to ``skip'' a sub-layer. A simple way to do this is to weight different paths rather than treating them equally. For example, a scalar factor of a residual connection can be introduced to determine how heavily we weight this residual connection relative to the path of the core function~\citep{he-etal:2016identity,liu-etal:2020rethinking,liu-eta:2020understanding}. A more general form of this model is given by
\begin{eqnarray}
\mathbf{z}^{l} & = & \mathrm{LNorm}(F^{l}(\mathbf{z}^{l-1}) + \beta \cdot \mathbf{z}^{l-1}) + \gamma \cdot \mathbf{z}^{l-1},
\end{eqnarray}

\noindent where $\beta$ is the weight of the identity mapping inside the layer normalization function, and $\gamma$ is the weight of the identity mapping outside the layer normalization function. Clearly, both the post-norm and pre-norm architectures can be seen as special cases of this equation. That is, if $\beta=1$ and $\gamma=0$, then it will become Eq. (\ref{eq:post-norm-general}); if $\beta=0$ and $\gamma=1$, it will become Eq. (\ref{eq:pre-norm-general}). This model provides a multi-branch view of building residual blocks. The input to this block can be computed through multiple paths with different modeling complexities. When $\beta$ and $\gamma$ are small, the representation is forced to be learned through a ``deep'' model with multiple layers of cascaded non-linear units.  In contrast, when $\beta$ and $\gamma$ are large, the representation is more likely to be learned using a ``shallow'' model with fewer layers. To determine the optimal choices of $\beta$ and $\gamma$, one can give them fixed values by considering some theoretical properties or system performance on validation sets, or compute these values by using additional functions that can be trained to do so~\citep{srivastava-etal:2015highway}. It should be emphasized that many other types of architecture can be considered in the design of a Transformer sub-layer. It is possible, for instance, to introduce more layer normalization units into a sub-layer~\citep{ding-etal:2021cogview,wang-etal:2022foundation}, or, conversely, to simply remove them from a sub-layer~\citep{bachlechner-etal:2021rezero}.

\subsubsection{Parameter Initialization}

\noindent As with other deep neural networks, there is considerable interest in developing parameter initialization methods for deep Transformer models in order to facilitate optimization in a more favorable region of the parameter space. However, initialization is a broad topic in the optimization of machine learning models, and a general discussion of this topic lies beyond the scope of this section. Here, we will focus on specific parameter initialization methods used in Transformer-based systems rather than general optimization problems.

While the parameters of a neural network can be initialized in various ways, most practical systems adopt simple techniques to provide appropriate initial values. Consider, for example, the Xavier initialization for a parameter matrix $\mathbf{W} \in \mathbb{R}^{d_{\mathrm{in}} \times d_{\mathrm{out}}}$~\citep{glorot-bengio:2010understanding}. We define a variable $\eta$ as follows
\begin{eqnarray}
\eta & = & \mathrm{gain} \cdot \sqrt{\frac{6}{d_{\rm{in}} + d_{\rm{out}}}},
\end{eqnarray}

\noindent where $\mathrm{gain}$ is a hyper-parameter which equals 1 by default. Then, each entry of $\mathbf{W}$ can be initialized by using a uniform distribution
\begin{eqnarray}
W & \sim & U \left( -\eta,\eta \right)
\end{eqnarray}

\noindent or, alternatively, using a Gaussian distribution
\begin{eqnarray}
W & \sim & \mathcal{N} \left(0, \eta^2 \right).
\end{eqnarray}

This method can be easily adapted to initialize Transformer models with a large number of layers. One common approach is to find a more suitable value for $\mathrm{gain}$ by taking into account the fact that the optimal initial conditions for optimization might differ across neural networks of varying depths. For example, one can increase the value of $\mathrm{gain}$ as the depth of the model grows. Thus, $\mathrm{gain}$ can be defined as a function of the network depth in the following form
\begin{eqnarray}
\mathrm{gain} & = & a \cdot L^{b}, \label{eq:transformer-init-gain-general}
\end{eqnarray}

\noindent where $a$ is a scalar, and $L^{b}$ is the network depth raised to the power of $b$. Typically, both $a$ and $b$ are positive numbers, indicating a preference for larger initial parameter values in deeper models. For example, \citet{wang-etal:2022deepnet} demonstrate that by choosing appropriate values for $a$ and $b$, a very deep Transformer model can be successfully trained.

Eq. (\ref{eq:transformer-init-gain-general}) assigns $\mathrm{gain}$ the same value for all of the sub-layers. However, it is found that the norm of gradients becomes smaller as a sub-layer is located farther from the output layer. This consistent application of $\mathrm{gain}$ across the entire model could result in under-training of the lower layers due to the gradient vanishing problem. For this reason, one can develop methods that are sensitive to the position of a sub-layer in the neural network. The general form of such methods is given by
\begin{eqnarray}
\mathrm{gain} & = & \frac{a}{l^{b}}. \label{eq:transformer-init-gain-sub-layer-depth}
\end{eqnarray}

\noindent Here $l$ denotes the depth of a sub-layer. If $l$ is larger (i.e., the sub-layer is closer to the output), $\mathrm{gain}$ will be smaller and the corresponding parameters will be set to smaller values. An example of this method can be found in the work of \citet{zhang-etal:2019improving}.

It is also, of course, straightforward to apply general methods of initializing deep multi-layer neural networks to Transformer models. An example is to consider the \textbf{Lipschitz constant} in parameter initialization, which has been shown to help improve the stability of training deep models~\citep{szegedy-etal:2014intriguing,xu-etal:2020lipschitz}. Another approach is to use second-order methods to estimate the proper values of the parameters. For example, one can compute the Hessian of each parameter matrix to model its curvature~\citep{skorski-etal:2021revisiting}.

For models with a large number of layers, it is also possible to pre-train some of the layers via smaller models and use their trained parameters to initialize bigger models~\citep{chen-etal:2015net2net}. That is, we first obtain a rough estimation of the parameters in a cheap way, and then continue the training process on the whole model as usual. These methods fall into a class of training methods, called \textbf{model growth} or \textbf{depth growth}.

As a simple example, consider a Transformer model (e.g., a Transformer encoder) of  $2L$ sub-layers. We can train this model by using the \textbf{shallow-to-deep training} method~\citep{li-etal:2020shallow}. First, we train an $L$-sub-layer model (call it the shallow model) in a regular way. Then, we create a $2L$-sub-layer model (call it the deep model) by stacking the shallow model twice, and further train this deep model. To construct deeper models, this procedure can be repeated multiple times, say, we start with a model of $L$ sub-layers, and obtain a model of $L \cdot 2^{I}$ after $I$ iterations. Note that many pre-trained models are used in a similar manner. For example, for BERT-like methods, a Transformer encoder is trained on large-scale data, and the optimized parameters are then used to initialize downstream systems.

\subsubsection{Layer Fusion}

\noindent Another problem with training a deep Transformer model is that the final prediction is conditioned on the last layer of the neural network. While the use of residual connections enables direct access to lower-level layers from a higher-level layer, there is still a ``long'' path for passing information from the bottom to the top. One simple way to address this is to create residual connections that skip more layers. For example, consider a group of $L$ Transformer sub-layers. For the sub-layer at depth $l$, we can introduce $l-1$ residual connections, each connecting this sub-layer to a previous sub-layer. In this way, we develop a densely connected network where each sub-layer takes as input the outputs of all previous sub-layers~\citep{huang-etal:2017densely}. The output of the last sub-layer can be seen as a combination of representations at different levels.

Following the notation used in the previous subsections, we denote the output of the sub-layer at depth $l$ by $\mathbf{z}^{l}$, and denote the function of the sub-layer by $\mathrm{Layer}^{l}(\cdot)$. Then, $\mathbf{z}^{l}$ can be expressed as
\begin{eqnarray}
\mathbf{z}^{l} & = & \mathrm{Layer}^{l}(\mathbf{z}^{1}, \cdots ,\mathbf{z}^{l-1}).
\end{eqnarray}

We can simply view $\mathrm{Layer}^{l}(\cdot)$ as a function that fuses the information from $\{\mathbf{z}^{1}, \cdots ,\mathbf{z}^{l-1}\}$.  There are many possible choices for $\mathrm{Layer}^{l}(\cdot)$. For example, a simple form of $\mathrm{Layer}^{l}(\cdot)$ is given by
\begin{eqnarray}
\mathrm{Layer}^{l}(\mathbf{z}^{1},\cdots,\mathbf{z}^{l-1}) & = & \mathrm{LNorm}(F^{l}(\mathbf{Z}^{l})), \\
\mathbf{Z}^{l} & = & \phi(\mathbf{z}^{1},\cdots,\mathbf{z}^{l-1}).
\end{eqnarray}

\noindent Here $\phi(\cdot)$ takes the layer outputs $\{\mathbf{z}^{1},\cdots,\mathbf{z}^{l-1}\}$ and fuses them into a single representation $\mathbf{Z}^{l}$. A simple instance of $\phi(\cdot)$ is average pooling which computes the sum of $\{\mathbf{z}^{1},\cdots,\mathbf{z}^{l-1}\}$ divided by $l-1$. See Table~\ref{tab:transformer-layer-fusion-functions} for more examples of $\phi(\cdot)$.

\begin{table}[!t]
\centering
\begingroup
\renewcommand{\arraystretch}{1.2}
\begin{tabular}{r | l}
Entry & Function \\ \hline
Average Pooling & $\phi(\mathbf{z}^{1},\cdots,\mathbf{z}^{l-1})=\frac{1}{l-1} \sum_{k=1}^{l-1}\mathbf{z}^k$ \\
Weighted Sum & $\phi(\mathbf{z}^{1},\cdots,\mathbf{z}^{l-1})=\sum_{k=1}^{l-1} \mathrm{weight}_k \cdot \mathbf{z}^k$ \\
Feedforward Network & $\phi(\mathbf{z}^{1},\cdots,\mathbf{z}^{l-1})=\mathrm{FFN}([\mathbf{z}^1,\cdots,\mathbf{z}^{l-1}])$ \\
Self-attention & $\phi(\mathbf{z}^{1},\cdots,\mathbf{z}^{l-1})=\mathrm{FFN}([\mathrm{Att}_{\mathrm{self}}(\mathbf{z}^1,\cdots,\mathbf{z}^{l-1})])$
\end{tabular}
\endgroup
\caption{Fusion functions. $\mathrm{FFN}(\cdot)$ = feedforward neural network, $[\cdot]$ = concatenating the input vectors, and $\mathrm{Att}_{\mathrm{self}}(\cdot)$ = self-attention function. All fusion functions can be followed by a layer normalization function, for example, we can write the weighted sum of $\{\mathbf{z}^{1},\cdots,\mathbf{z}^{l-1}\}$ as $\phi(\mathbf{z}^{1},\cdots,\mathbf{z}^{l-1})=\mathrm{LNorm}(\sum_{k=1}^{l-1} \mathrm{weight}_k \cdot \mathbf{z}^k)$.}
\label{tab:transformer-layer-fusion-functions}
\end{table}

Adopting a similar architecture for a Transformer sub-layer, we can also consider a post-norm form
\begin{eqnarray}
\mathrm{Layer}^{l}(\mathbf{z}^{1},\cdots,\mathbf{z}^{l-1}) & = & \mathrm{LNorm}(\mathbf{Z}^{l}), \label{eq:layer-fusion-post-norm-layer} \\
\mathbf{Z}^{l} & = & \phi(F^{l}(\mathbf{z}^{l-1}),\mathbf{z}^{1},\cdots,\mathbf{z}^{l-1}), \label{eq:layer-fusion-post-norm-fusion}
\end{eqnarray}

\noindent or a pre-norm form
\begin{eqnarray}
\mathrm{Layer}^{l}(\mathbf{z}^{1},\cdots,\mathbf{z}^{l-1}) & = & \mathbf{Z}^{l} \label{eq:layer-fusion-pre-norm-layer}, \\
\mathbf{Z}^{l} & = & \phi(\mathrm{LNorm}(F^{l}(\mathbf{z}^{l-1})),\mathbf{z}^{1},\cdots,\mathbf{z}^{l-1}). \label{eq:layer-fusion-pre-norm-fusion}
\end{eqnarray}

These models are quite general. For example, a standard post-norm encoder sub-layer can be recovered as a special case of Eqs. (\ref{eq:layer-fusion-post-norm-layer}-\ref{eq:layer-fusion-post-norm-fusion}), if we remove the dependencies on sub-layers $1$ to $l-2$, and define $\phi(\cdot)$ to be
\begin{eqnarray}
\phi(F^{l}(\mathbf{z}^{l-1}),\mathbf{z}^{1},\cdots,\mathbf{z}^{l-1}) & = & F^{l}(\mathbf{z}^{l-1}) + \mathbf{z}^{l-1}.
\end{eqnarray}

Densely connected networks make it easier for information to flow through direct connections between sub-layers, but the resulting models are slightly more complex, especially when we use parameterized fusion functions. In practice, we typically add dense connections only to a subset of the sub-layers, so the overall network is not overly dense. For example, we might only add connections from lower sub-layers to the final few sub-layers. Thus, predictions can be made with direct access to different levels of representation~\citep{wang-etal:2018multi}.

\subsubsection{Regularization}

\noindent In machine learning, regularization is used to mitigate overfitting when training deep neural networks. It is therefore straightforward to apply regularization techniques to Transformer models. Since the topic of regularization is well studied, here we consider several methods that are particularly useful for training deep Transformer models.

One approach to regularizing deep Transformer models is to randomly skip sub-layers or layers during training~\citep{huang-etal:2016deep,pham-etal:2019very}. In each run of the model, we retain each sub-layer with a probability $\rho$ and stack the selected sub-layers to form a new model. Thus, we essentially train an ensemble of different neural networks with shared architectures and parameters on the same dataset. In this way, each sub-layer learns to operate somewhat independently, and overfitting is reduced by preventing the co-adaptation of sub-layers. In fact, dropping out sub-layers (or layers) and dropping out neurons are essentially variations on the same theme. Sometimes, the method described here is referred to as \textbf{sub-layer dropout} or \textbf{layer dropout}.

At test time, we need to combine all possible networks to make predictions. A simple method to achieve this is to rescale the outputs of the stochastic components of the model~\citep{li-etal:2021learning}. As an example, suppose each sub-layer has a pre-norm architecture. Then, the output of the sub-layer at depth $l$ is given by
\begin{eqnarray}
\mathbf{z}^{l} & = & \rho \cdot \mathrm{LNorm}(F^{l}(\mathbf{z}^{l-1})) + \mathbf{z}^{l-1}.
\end{eqnarray}

Another idea is to share parameters across sub-layers. One of the simplest methods is to use the exact same parameters for all corresponding sub-layers~\citep{dehghani-etal:2018universal}. For example, all the FFN sub-layers could be based on the same feed-forward network. This approach has a regularizing effect similar to methods that add the norms of parameter matrices to the loss function to penalize model complexity. For practical systems, there can be significant benefits to adopting a shared architecture, as it allows us to reuse the same sub-model to build a deep neural network, thereby substantially reducing the overall memory footprint. We will see further discussion regarding efficiency issues in Section~\ref{sec:efficient-methods-parameter-sharing}.

\subsection{Numerical Method-Inspired Models}
\label{sec:transformer-numerical-methods}

\noindent A residual network computes its output as the sum of the identity mapping and a transformation of the input. Such a model can be interpreted as an Euler discretization of \textbf{ordinary differential equations} (\textbf{ODEs})~\citep{ee-etal:proposal2017,haber-ruthotto:2017stable}. To illustrate this idea, we consider a general form of residual networks
\begin{eqnarray}
\mathbf{z}^{l} & = & f^l \big(\mathbf{z}^{l-1} \big) + \mathbf{z}^{l-1}, \label{eq:pre-norm-ode-form}
\end{eqnarray}

\noindent where $f^l(\mathbf{z}^{l-1})$ denotes a function that takes an input variable $\mathbf{z}^{l-1}$ and produces an output variable in the same space. Clearly, a Transformer sub-layer is a special case of this equation. For example, for a pre-norm Transformer, we have $f^l(\cdot)=\mathrm{LNorm}(F^{l}(\cdot))$.

For notational simplicity, we rewrite the above equation in an equivalent form
\begin{eqnarray}
\mathbf{z}(l) & = & f \big(\mathbf{z}(l-1),l \big) + \mathbf{z}(l-1). \label{eq:pre-norm-ode-form-t}
\end{eqnarray}

\noindent We use the notations $\mathbf{z}(l)$ and $f(\mathbf{z}(\cdot,l))$ to emphasize that $\mathbf{z}(\cdot)$ and $f(\cdot)$ are functions of $l$. Here we assume that $l$ is a discrete variable. If we relax $l$ to a continuous variable and $\mathbf{z}(l)$ to a continuous function of $l$, then we can express Eq. (\ref{eq:pre-norm-ode-form-t}) as
\begin{eqnarray}
\mathbf{z}(l) & = & \triangle l \cdot f \big(\mathbf{z}(l - \triangle l), l \big) + \mathbf{z}(l - \triangle l). \label{eq:pre-norm-ode-form-delta-f}
\end{eqnarray}

\noindent This can be further written as
\begin{eqnarray}
\frac{\mathbf{z}(l) - \mathbf{z}(l - \triangle l)}{\triangle l } & = & f \big( \mathbf{z}(l - \triangle l),l \big). \label{eq:pre-norm-ode-form-delta-f-2}
\end{eqnarray}

\noindent Taking the limit $\triangle l \rightarrow 0$, we have an ODE
\begin{eqnarray}
\frac{\mathrm{d} \mathbf{z}(l)}{\mathrm{d} l} & = & f \big(\mathbf{z}(l),l \big). \label{eq:transformer-to-ode}
\end{eqnarray}

We can therefore say that a pre-norm Transformer sub-layer (i.e., Eqs. (\ref{eq:pre-norm-ode-form-t}) and (\ref{eq:pre-norm-ode-form})) is an Euler discretization of the above ODE. This is an interesting result! A sub-layer essentially serves as an ODE solver.

Eqs. (\ref{eq:pre-norm-ode-form-t}) and (\ref{eq:pre-norm-ode-form}) represent standard forms of the \textbf{Euler method}. This method computes a new estimate of the solution by moving one step forward along $l$ from the previous estimation. In general, two aspects can be considered in the design of numerical methods for ODEs.

\begin{itemize}
\item \vspace{0.5em} \textbf{Linear Multi-step Methods}. A linear multi-step method computes the current estimate of the solution by taking the estimations and derivative information from multiple previous steps. A general formulation of $p$-step methods can be expressed as
    \begin{eqnarray}
    \mathbf{z}(l) & = & \sum_{i=1}^{p} a_{i} \cdot \mathbf{z}(l-i) + h \sum_{i=1}^{p+1} b_{i} \cdot f \big(\mathbf{z}(l-i),\ l-i+1 \big), \label{eq:linear-multi-step-methods-general}
    \end{eqnarray}
    \noindent where $h$ is the size of the step we move each time\footnote{Let $\{t_0,\cdots,t_i\}$ denote the values of the variable $l$ at steps $\{0,\cdots,i\}$. In linear multi-step methods, it is assumed that $t_i=t_0+ih$.}, that is, $\triangle l$ in Eqs. (\ref{eq:pre-norm-ode-form-delta-f}) and (\ref{eq:pre-norm-ode-form-delta-f-2}). $\{a_i\}$ and $\{b_i\}$ are coefficients of the solution points and derivatives in the linear combination. Given this definition, we can think of the Euler method as a single-step, low-order method of solving ODEs\footnote{In numerical analysis, the \textbf{local truncation error} of a method of solving ODEs at a step is defined to be the difference between the approximated solution computed by the method and the true solution. The method is called order $p$ if it has a local truncation error $O(h^{p+1})$.}.

\item \vspace{0.3em} \textbf{(Higher-order) Runge-Kutta Methods}. Runge-Kutta (RK) methods and their variants provide ways to compute the next step solution by taking intermediate results in solving an ODE. As a result, we obtain higher-order methods but still follow the form of single-step methods, that is, the estimated solution is dependent only on $\mathbf{z}(l-1)$ rather than on the outputs at multiple previous steps.
\end{itemize}
\vspace{0.5em}

In fact, linear multi-step methods, though not explicitly mentioned, have been used in the layer fusion methods discussed in Section~\ref{sec:transformer-deep-models}. For example, taking Eqs. (\ref{eq:layer-fusion-pre-norm-layer}) and (\ref{eq:layer-fusion-pre-norm-fusion}) and a linear fusion function, a pre-norm sub-layer with dense connections to all previous sub-layers can be expressed as
\begin{eqnarray}
\mathrm{Layer}^{l}(\mathbf{z}^{1},\cdots,\mathbf{z}^{l-1}) & = & a_1 \cdot \mathbf{z}^{l-1} + \cdots + a_{l-1} \cdot \mathbf{z}^{1} + b_1 \cdot \mathrm{LNorm}(F^{l}(\mathbf{z}^{l-1})).
\end{eqnarray}

\noindent This equation is an instance of Eq. (\ref{eq:linear-multi-step-methods-general}) where we set $h=1$ and remove some of the terms on the right-hand side.

It is also straightforward to apply Runge-Kutta methods to Transformer models~\citep{li-etal:2022ode}. Given an ODE as described in Eq. (\ref{eq:transformer-to-ode}), an explicit $p$-th order Runge-Kutta solution is given by
\begin{eqnarray}
\mathbf{z}(l) & = & \mathbf{z}(l-1) + \sum_{i=1}^{p} \gamma_i \cdot \mathbf{g}_i, \\
\mathbf{g}_i & = & h \cdot f \big( \mathbf{z}(l-1) + \sum_{j=1}^{i-1} \beta_{i,j} \cdot \mathbf{g}_j,\ l - 1 + \lambda_i \cdot h \big). \label{eq:runge-kutta-general-g}
\end{eqnarray}

\noindent Here $\mathbf{g}_i$ represents an intermediate step which is present only during the above process. $\{\gamma_i\}$, $\{\beta_{i,j}\}$ and $\{\lambda_i\}$ are coefficients that are determined by using the Taylor series of $\mathbf{z}(l)$. To simplify the model, we assume that the same function $f$ is used for all $\{\mathbf{g}_i\}$. Then, we remove the dependency on the term $l - 1 + \lambda_i \cdot h$ in $f$, and rewrite Eq. (\ref{eq:runge-kutta-general-g}) as
\begin{eqnarray}
\mathbf{g}_i & = & h \cdot f \big( \mathbf{z}(l-1) + \sum_{j=1}^{i-1} \beta_{i,j} \cdot \mathbf{g}_j\big), \label{eq:runge-kutta-general-g-new}
\end{eqnarray}

\noindent where $f(\cdot)$ is a function that is independent of $i$.

As an example, consider the 4th-order Runge-Kutta (RK4) solution
\begin{eqnarray}
\mathbf{z}(l) & = & \mathbf{z}(l-1) + \frac{1}{6}(\mathbf{g}_1 + 2 \mathbf{g}_2 + 2 \mathbf{g}_3 + \mathbf{g}_4), \\
\mathbf{g}_1 & = & h \cdot f(\mathbf{z}(l-1)), \\
\mathbf{g}_2 & = & h \cdot f(\mathbf{z}(l-1) + \frac{1}{2} \mathbf{g}_1), \\
\mathbf{g}_3 & = & h \cdot f(\mathbf{z}(l-1) + \frac{1}{2} \mathbf{g}_2), \\
\mathbf{g}_4 & = & h \cdot f(\mathbf{z}(l-1) + \mathbf{g}_3).
\end{eqnarray}

\begin{figure}[!htp]
\centering
\input{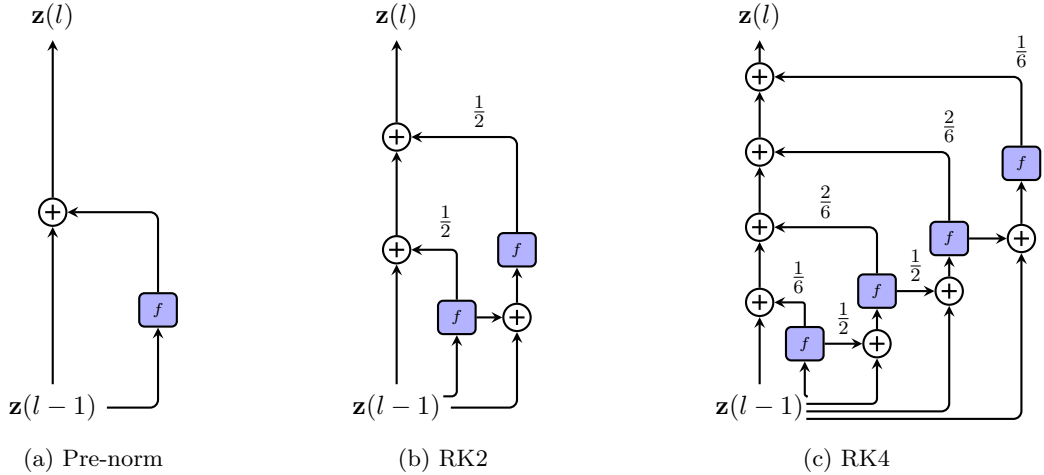}
\caption{Pre-norm (a) and Runge-Kutta (b and c) sub-layer architectures. $\mathbf{z}(l-1)$ denotes the input of a sub-layer at depth $l$, $\mathbf{z}(l)$ denotes the output of the sub-layer, and $f$ (in blue boxes) denotes the function $f(\cdot)=\mathrm{LNorm}(F^{l}(\cdot))$.}
\label{fig:rk4-pre-norm-architecture}
\end{figure}

\noindent These equations define a new architecture of sub-layer. For example, by setting $h=1$ and $f(\cdot) = \mathrm{LNorm}(F^{l}(\cdot))$, we obtain an RK4 Transformer sub-layer, as shown in Figure~\ref{fig:rk4-pre-norm-architecture}. This method leads to a deep model because each sub-layer involves four evaluations of $f(\cdot)$ in sequence. On the other hand, the resulting model is parameter efficient because we reuse the same function $f(\cdot)$ within the sub-layer, without introducing additional parameters.

So far in this subsection, our discussion has focused on designing Transformer architectures inspired by dynamical systems. While the basic ODE model is continuous with respect to the depth $l$, these methods still follow the general framework of neural networks in which $l$ is treated as a discrete variable, and the representational power of the models is largely determined by this hyper-parameter. An alternative approach is to use neural ODEs to relax the ``depth'' into a continuous variable. In this way, we can obtain a continuous-depth model for computing the solution to ODEs. However, as a detailed discussion of neural ODEs lies beyond the scope of this chapter, we refer interested readers to related papers for more details~\citep{chen-etal:2018neural,kidger:2022neural,xiao-etal:2026ordinary}.

\subsection{Wide Models}
\label{sec:transformer-wide-models}

\noindent Most of the methods that we have studied so far in this section are examples of learning and using deep models. Another design choice we generally face is determining the width of a neural network. Typically, the \textbf{width} of a Transformer model is defined as the dimensionality of a representation at some position in the input sequence, that is, the parameter $d$. Increasing this width is a common method for obtaining a more complex and powerful model. For example, in the work of \citet{Vaswani-etal:2017Transformer}, a wide model (referred to as Transformer-Big) leads to significant improvements in translation quality for machine translation systems. Such wide models have been widely used to boost system performance on large-scale tasks~\citep{lepikhing-etal:2021shard,fedus-etal:2022switch}.

However, developing very wide Transformer models is difficult. One major challenge is that training such systems is computationally expensive. While the number of model parameters (the model size) grows linearly with $d$, the time complexity of the model grows quadratically with $d$ (see Table~\ref{tab:transformer-parameter-num-and-time}). In some NLP tasks, it has been found empirically that the training effort required to obtain satisfactory performance follows a scaling law~\citep{kaplan-etal:2020scaling}. These results emphasize the need for ways to improve training efficiency when enlarging $d$.

One simple method is to incrementally grow the model along the dimension of $d$, rather than training the model from scratch. Suppose we have an initial model with a $d_1 \times d_1$ parameter matrix $\mathbf{W}_1$, for example, the linear transformation of each query or key in some layer. We can train this model to obtain optimized $\mathbf{W}_1$ in a regular way. Then, we want to extend this model to a wider model where $\mathbf{W}_1$ is replaced by a $d_2 \times d_2$ parameter matrix $\mathbf{W}_2$. Let us assume for simplicity that $d_2 = k d_1$. There are several ways to expand a $d_1 \times d_1$ matrix to a $k d_1 \times k d_1$ matrix. The simplest of these may be to use $\mathbf{W}_1$ to fill $\mathbf{W}_2$. We can write $\mathbf{W}_2$ in the form
\begin{eqnarray}
\mathbf{W}_2 & = &
\begin{matrix}
k\ \textrm{times} & \ \\
\begin{bmatrix}
\frac{\mathbf{W}_1}{\rho} & \cdots & \frac{\mathbf{W}_1}{\rho} \\
\vdots & & \vdots \\
\frac{\mathbf{W}_1}{\rho} & \cdots & \frac{\mathbf{W}_1}{\rho}
\end{bmatrix}
&
\begin{rotate}{90}\textrm{\hspace{-1.3em} $k$ times}\end{rotate}
\\
\end{matrix},
\end{eqnarray}

\noindent where $\rho$ is a hyper-parameter that is used to control the norm of $\mathbf{W}_2$. For example, if $\rho = k$, $\mathbf{W}_2$ will have the same $l_1$ norm as $\mathbf{W}_1$. The above equation provides a good starting point for training the wide model, and we can train $\mathbf{W}_2$ as usual after initialization. The procedure can be repeated a number of times for constructing a model with arbitrary width. Both this method and the depth growth method described in Section~\ref{sec:transformer-deep-models} are instances of the general method of model growth. In other words, we can obtain a larger model by extending a small model either vertically or horizontally, or both. Alternative methods for transforming $\mathbf{W}_1$ to $\mathbf{W}_2$ involve those considering other mathematical properties of the transformation~\citep{chen-etal:2015net2net}. These models can fall under the reusable neural networks where we are concerned with models and algorithms for transferring parameters from small models to (significantly) larger models~\citep{wang-etal:2023learning}.

A second difficulty in building a wide Transformer model is the large memory requirement. Since the feedforward network generally has a larger hidden layer than other parts of the model, it demands relatively more memory as the model becomes wider. Consider the feedforward network described in Section~\ref{sec:transformer-fnn}
\begin{eqnarray}
\mathbf{H}_{\mathrm{out}} & = &  \mathrm{FFN}(\mathbf{H}_{\mathrm{in}}) \nonumber \\
& = & \mathrm{ReLU}(\mathbf{H}_{\mathrm{in}} \cdot \mathbf{W}_h + \mathbf{b}_h) \cdot  \mathbf{W}_f + \mathbf{b}_f, \label{eq:transformer-ffn-moe}
\end{eqnarray}

\noindent where $\mathbf{W}_h \in \mathbb{R}^{d \times d_{\mathrm{ffn}}}$ and $\mathbf{W}_f \in \mathbb{R}^{d_{\mathrm{ffn}} \times d}$ are the parameters of the linear transformations. $d_{\mathrm{ffn}}$ is typically several times larger than $d$. Therefore, $\mathbf{W}_h$ and $\mathbf{W}_f$ will dominate the model size if $d$ and $d_{\mathrm{ffn}}$ have very large values.

In some cases, the size of the feedforward network may exceed the memory capacity of a single device. This problem can be addressed by using the \textbf{mixture-of-experts} (\textbf{MoE}) models~\citep{shazeer-etal:2017outrageously}. An MoE model consists of $M$ expert models $\{e_1(\cdot),\cdots,e_M(\cdot)\}$. Given an input $\mathbf{h}_{\mathrm{in}} \in \mathbb{R}^{d}$, each expert model produces an output $e_k(\mathbf{h}_{\mathrm{in}})$. The output of the MoE model is a linear combination of $\{e_1(\mathbf{h}_{\mathrm{in}}),\cdots,e_M(\mathbf{h}_{\mathrm{in}})\}$, given by
\begin{eqnarray}
\mathbf{h}_{\mathrm{out}} & = & \sum_{i=1}^{M} g_i(\mathbf{h}_{\mathrm{in}}) \cdot e_i(\mathbf{h}_{\mathrm{in}}),
\end{eqnarray}

\noindent where $g(\cdot)$ is a gating model (also called \textbf{routing model}). Its output is a vector $g(\mathbf{h}_{\mathrm{in}}) = \begin{bmatrix} g_1(\mathbf{h}_{\mathrm{in}}) & \cdots & g_M(\mathbf{h}_{\mathrm{in}})\end{bmatrix}$ in which each entry $g_i(\mathbf{h}_{\mathrm{in}})$ indicates the weight of the corresponding expert model. In many applications, it is assumed that $g(\mathbf{h}_{\mathrm{in}})$ is a sparse vector. This means that only a small number of expert models are involved in computing the output. A widely-used form of $g(\mathbf{h}_{\mathrm{in}})$ is given by using the softmax layer
\begin{eqnarray}
g(\mathbf{h}_{\mathrm{in}}) & = & \mathrm{Softmax}(\mathbf{h}_{\mathrm{in}} \cdot \mathbf{W}_{g}),
\end{eqnarray}

\noindent where $\mathbf{W}_{g} \in \mathbb{R}^{d \times M}$ is the parameter matrix of the layer. To enforce sparsity on $g(\mathbf{h}_{\mathrm{in}})$, we can select the top-$k$ entries of $g(\mathbf{h}_{\mathrm{in}})$, that is, we set non-top-$k$ entries to 0. An alternative method is to first perform top-$k$ selection on $\mathbf{h}_{\mathrm{in}} \cdot \mathbf{W}_{g}$ and then normalize the top-$k$ entries using the softmax function.

Let $\pi$ be the set of the indices of the top-$k$ expert models. The MoE model with top-$k$ routing has the following form
\begin{eqnarray}
\mathbf{h}_{\mathrm{out}} & = & \sum_{i \in \pi} g_i(\mathbf{h}_{\mathrm{in}}) \cdot e_i(\mathbf{h}_{\mathrm{in}}).
\end{eqnarray}

An advantage of this approach is that we can distribute different expert models to different processors, making it possible to execute these models on parallel computing machines. In each run of the MoE model, either during training or inference, we only need to activate and use $k$ expert models rather than all of the expert models. In this way, the MoE approach is automatically learning a sparse model by limiting the number of active expert models each time in training and inference. The sparsity is determined by the hyperparameter $k$, e.g., a small value of $k$ leads to a sparse model, and a large value of $k$ leads to a dense model.

Let us return to the discussion of Eq. (\ref{eq:transformer-ffn-moe}). It is straightforward to apply the MoE approach to feedforward neural networks. To simplify the discussion, consider the linear transformation of the first layer as shown in Eq. (\ref{eq:transformer-ffn-moe}), that is, $\mathbf{H}_{\mathrm{in}} \cdot \mathbf{W}_h$. We can approximate $\mathbf{H}_{\mathrm{in}} \cdot \mathbf{W}_h$ in an MoE form
\begin{eqnarray}
\mathbf{H}_{\mathrm{in}} \cdot \mathbf{W}_h & \approx & \sum_{i \in \pi} g_i(\mathbf{H}_{\mathrm{in}}) \cdot e_i(\mathbf{H}_{\mathrm{in}}) \nonumber \\
& = & \sum_{i \in \pi} g_i(\mathbf{H}_{\mathrm{in}}) \cdot [\mathbf{H}_{\mathrm{in}} \cdot \mathbf{W}_h^{i}]. \label{eq:moe-ffn-w-h}
\end{eqnarray}

\noindent Here $\mathbf{W}_h$ is divided into $M$ slices (or sub-matrices) $\{\mathbf{W}_h^{1},\cdots,\mathbf{W}_h^{M}\}$, written as
\begin{eqnarray}
\mathbf{W}_h & = & \begin{bmatrix} \mathbf{W}_h^{1} & \cdots & \mathbf{W}_h^{M} \end{bmatrix}.
\end{eqnarray}

\noindent Hence each expert model $e_i(\mathbf{H}_{\mathrm{in}}) = \mathbf{H}_{\mathrm{in}} \cdot \mathbf{W}_h^{i}$ solves a sub-problem of the original linear mapping, and Eq. (\ref{eq:moe-ffn-w-h}) can be thought of as a divide-and-conquer solution to the matrix multiplication problem.

\begin{figure}[!htp]
\centering

\begin{tikzpicture}

\def\nodehsep{1.2em} 
\def\whsep{1.5em} 
\def\nodewd{10.5em} 
\def\wordhsep{0.0ex} 
\def\sep{0.5em} 
\def\gatemodelwd{4.3em} 
\def\sumhsep{5.0em} 
\def\gatewd{1.9em} 
\def\ffnwd{3.4em} 
\def\ffnsep{0.9em} 
\def\ffnhsep{0.3em} 
\def\wffnhsep{0.4em} 
\def\csize{11pt} 

\tikzstyle{basenode} = [minimum width=\nodewd,inner sep=0.5pt,rounded corners=1.5pt,draw,thick,font=\footnotesize];
\tikzstyle{Sanode} = [basenode,minimum height=2.0em];
\tikzstyle{Resnode} = [basenode,minimum height=1.6em];

\tikzstyle{ffnnode} = [basenode,minimum height=2.0em,minimum width=\ffnwd];
\tikzstyle{gatemodelnode} = [basenode,minimum height=2.0em,minimum width=\gatemodelwd];
\tikzstyle{sumnode} = [basenode,minimum height=2.0em]
\tikzstyle{wnode} = [minimum height=2.0em];

\tikzstyle{standard} = [thick]

\begin{scope}

\node [Sanode,anchor=west,align=center,fill=orange!30] (sa1) at (0,0) {{$\textbf{Feed-Forward Network}$}};
\node [Resnode,anchor=south,fill=yellow!30] (res1) at ([yshift=\nodehsep]sa1.north) {{$\textbf{Add \& LayerNorm}$}};

\node [anchor=north] (inputs) at ([yshift=-\whsep*1.5]sa1.south) {{$\vdots$}};
\node [anchor=south] (outputs) at ([yshift=\whsep*1.5]res1.north) {{$\vdots$}};

\draw [->,standard] ([yshift=-\whsep*1.5]sa1.south) -- (sa1.south);
\draw [->,standard] (sa1.north) -- (res1.south);
\draw [->,standard] (res1.north) -- ([yshift=\whsep*1.5]res1.north);

\draw[->,standard] ([yshift=-0.5em]sa1.south) -- ([xshift=-\nodewd*0.5 - \sep,yshift=-0.5em]sa1.south) -- ([xshift=- \sep,yshift=0em]res1.west) -- ([xshift=-0.0em]res1.west);

\node [rectangle,inner sep=0.9em,rounded corners=1pt,very thick,dotted,draw] [fit = (sa1) (res1)] (blockback) {};

\node [anchor=south west,font=\footnotesize] (inputs) at ([yshift=0em]blockback.north west) {{FFN sub-layer}};

\node [anchor=west,rectangle,inner sep=0.9em,rounded corners=1pt,thick,draw,minimum width=3.15in,minimum height=2in] (rightback) at ([xshift=2em]blockback.east) {};

\draw[->,standard] ([yshift=-0em]sa1.east)..controls + (east:2.5em) and + (west:2.5em)..([xshift=-0.0em]rightback.west);

\node [gatemodelnode,anchor=south west,align=center] (gate) at ([xshift=\ffnsep*0.5,yshift=\ffnhsep*6.5]rightback.south west) {Gating \\ [\wordhsep]Model};

\node [ffnnode,anchor=west] (ffn1) at ([xshift=\ffnsep,yshift=0em]gate.east) {\scriptsize{$\mathrm{FFN}_1(\cdot)$}};
\node [ffnnode,anchor=west] (ffn2) at ([xshift=\ffnsep,yshift=0em]ffn1.east) {\scriptsize{$\mathrm{FFN}_2(\cdot)$}};
\node [anchor=west,minimum height=2.0em] (ffn3) at ([xshift=\ffnsep,yshift=0em]ffn2.east) {\scriptsize{$\cdots$}};
\node [ffnnode,anchor=west] (ffn4) at ([xshift=\ffnsep,yshift=0em]ffn3.east) {\scriptsize{$\mathrm{FFN}_M(\cdot)$}};

\node [wnode,anchor=north] (w1) at ([xshift=0em,yshift=-\ffnhsep]gate.south) {\scriptsize{$\mathbf{H}_{\mathrm{in}}$}};
\node [wnode,anchor=north] (w2) at ([xshift=0em,yshift=-\ffnhsep]ffn1.south) {\scriptsize{$\mathbf{H}_{\mathrm{in}}$}};
\node [wnode,anchor=north] (w3) at ([xshift=0em,yshift=-\ffnhsep]ffn2.south) {\scriptsize{$\mathbf{H}_{\mathrm{in}}$}};
\node [wnode,anchor=north] (w4) at ([xshift=0em,yshift=-\ffnhsep]ffn3.south) {\scriptsize{$\cdots$}};
\node [wnode,anchor=north] (w5) at ([xshift=0em,yshift=-\ffnhsep]ffn4.south) {\scriptsize{$\mathbf{H}_{\mathrm{in}}$}};

\draw [->,standard] ([yshift=-\wffnhsep]w1.north) -- (gate.south);
\draw [->,standard] ([yshift=-\wffnhsep]w2.north) -- (ffn1.south);
\draw [->,standard] ([yshift=-\wffnhsep]w3.north) -- (ffn2.south);
\draw [->,standard] ([yshift=-\wffnhsep]w5.north) -- (ffn4.south);

\node [sumnode,anchor=south west,minimum width=\ffnwd*4.32] (suml) at ([xshift=0em,yshift=\sumhsep]ffn1.north west) {\scriptsize{$\sum_{i \in \pi} g_i(\mathbf{H}_{\mathrm{in}}) \cdot \mathrm{FFN}_i(\mathbf{H}_{\mathrm{in}})$}};

\node [wnode,anchor=south] (wo1) at ([xshift=0em,yshift=\ffnhsep]suml.north) {\scriptsize{$\mathbf{H}_{\mathrm{out}}$}};
\draw [->,standard] (suml.north) -- ([yshift=\wffnhsep]wo1.south);

\node [circle,anchor=south,minimum size=\csize,draw,thick,inner sep=0.5pt] (dot1) at ([xshift=0em,yshift=\sumhsep*0.6]ffn1.north) {};
\node [circle,anchor=south,minimum size=\csize,draw,thick,inner sep=0.5pt] (dot2) at ([xshift=0em,yshift=\sumhsep*0.4]ffn2.north) {};
\node [circle,anchor=south,minimum size=\csize,draw,thick,inner sep=0.5pt] (dot3) at ([xshift=0em,yshift=\sumhsep*0.2]ffn4.north) {};

\draw [-] (dot1.north west) -- (dot1.south east);
\draw [-] (dot1.north east) -- (dot1.south west);
\draw [-] (dot2.north west) -- (dot2.south east);
\draw [-] (dot2.north east) -- (dot2.south west);
\draw [-] (dot3.north west) -- (dot3.south east);
\draw [-] (dot3.north east) -- (dot3.south west);

\draw [->,standard] (ffn1.north) -- (dot1.south);
\draw [->,standard] (ffn2.north) -- (dot2.south);
\draw [->,standard] (ffn4.north) -- (dot3.south);

\draw [->,standard] (dot1.north) -- ([yshift=\sumhsep*0.18]dot1.north);
\draw [->,standard] (dot2.north) -- ([yshift=\sumhsep*0.38]dot2.north);
\draw [->,standard] (dot3.north) -- ([yshift=\sumhsep*0.58]dot3.north);


\draw [->,standard,rounded corners=1.5pt,dashed] ([xshift=-\gatemodelwd*0.3]gate.north)..controls + (north:3.5em) and + (west:0.5em)..(dot1.west);
\draw [->,standard,rounded corners=1.5pt,dashed] (gate.north)..controls + (north:2.5em) and + (west:0.5em)..(dot2.west);
\draw [->,standard,rounded corners=1.5pt,dashed] ([xshift=\gatemodelwd*0.3]gate.north)..controls + (north:1.5em) and + (west:0.5em)..(dot3.west);
\end{scope}

\end{tikzpicture}

\caption{An illustration of the MoE model applied to an FFN sub-layer. There are $M$ FFNs (call them expert models) and a gating model. Each FFN is weighted by the gating model. The output of the model is the sum of the weighted outputs of the top-$k$ FFNs (denoted by $\pi$). Because these FFNs work independently and can be placed on different computing devices, the model can be easily scaled up as $M$ is larger.}
\label{fig:transformer-ffn-moe}
\end{figure}
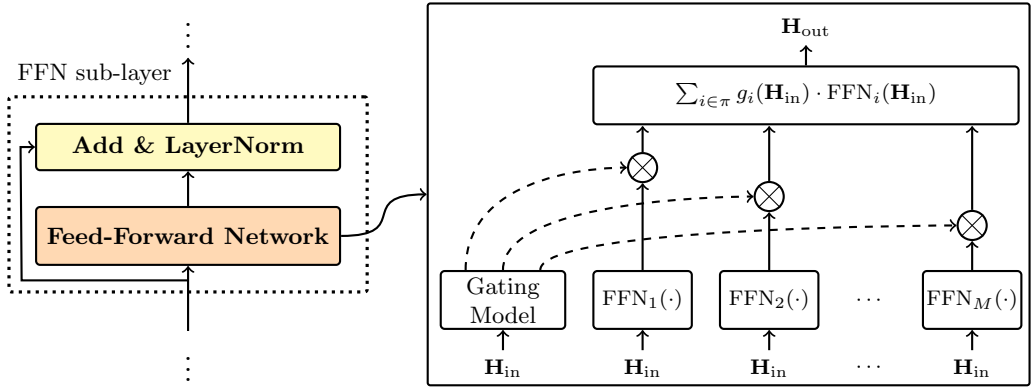

We can, of course, treat any feedforward neural network as an expert model, resulting in the following model
\begin{eqnarray}
\mathbf{H}_{\mathrm{out}} & = & \sum_{i \in \pi} g_i(\mathbf{H}_{\mathrm{in}}) \cdot \mathrm{FFN}_i(\mathbf{H}_{\mathrm{in}}),
\end{eqnarray}

\noindent where $\mathrm{FFN}_i(\cdot)$ is a ``small'' feedforward neural network that has the same form as Eq. (\ref{eq:transformer-ffn-moe}). This model is illustrated with an example in Figure~\ref{fig:transformer-ffn-moe}. In practical implementations, all these expert models can be run in parallel on different devices, and so the resulting system is efficient.

Note that, from a perspective of machine learning, MoE is a general approach to combining different neural networks, each of which is developed to address a different aspect of the problem~\citep{yuksel-etal:2012twenty,masoudnia-ebrahimpour:2014mixture}. The application here is just a special instance of the general framework of MoE. The approach is also often used to improve the overall performance of predictors, which can be discussed in the field of ensemble learning~\citep{Zhou-2012}.

Another difficulty in developing massive Transformer models is training instability. As with many other exceptionally large neural networks, the straightforward optimization of a Transformer model with billions of parameters may lead to the model getting trapped in local minima and, occasionally, to large spikes in the loss during training~\citep{lepikhing-etal:2021shard,fedus-etal:2022switch,chowdhery-etal:2022palm}. Even with careful tuning of hyperparameters, training strategies, and initial parameter values, it is common to encounter situations where the training must be restarted from an earlier checkpoint to escape unstable optimization regions. One of the reasons for this training difficulty is that the standard implementations of linear algebra operations, such as matrix multiplication, can become numerically unstable when they operate on extremely large vectors and matrices. It is therefore crucial to improve training stability by employing numerically robust methods and mixed-precision techniques.


\section{Efficient Models}
\label{sec:transformer-efficient-models}

\noindent Efficiency is an important consideration for many practical applications of Transformer models. For example, we may wish to run and/or train a Transformer model given memory and time constraints. Efficiency is not a single issue, but covers a wide range of problems. While these problems can be categorized in several different ways, there are two fundamental aspects one may consider in an efficiency problem.
\begin{itemize}
\item \vspace{0.5em} \textbf{Time and Space Efficiencies}. For a given problem, we wish the model to be small and fast, while remaining as accurate as possible in solving the problem. For example, in some machine translation applications, we may develop a model with a small number of parameters to fit into limited memory, and may develop a fast search algorithm to achieve low-latency translation. A practical difficulty here is that improving efficiency often leads to worse predictions. In many cases, we need to seek a trade-off between efficiency and accuracy.
\item \vspace{0.3em} \textbf{Scalability}. When the problem is scaled up, we wish that the additional computational overhead required is minimized. For example, the training of a neural network is called efficient if it takes a reasonably short time to optimize it as the volume of training data increases. Another metric of efficiency is the amount of resources consumed in processing larger inputs. For instance, a machine translation system is inefficient in translating long sentences if the memory footprint and latency grow exponentially with the number of input words.
\end{itemize}
\vspace{0.5em}

In this section, we will not discuss all the issues related to efficiency, which is a very broad topic. We instead consider the widely-used efficient approaches to Transformer-based sequence modeling and generation, some of which are refinements of model architectures, and some of which are architecture-agnostic approaches and could be used in other systems as well. Most of the discussions here are focused on developing lightweight and fast Transformer models that are relatively robust to long input and output sequences.

In general, the same optimization method can be applied to different modules of a Transformer-based system. To simplify the discussion, we will mostly consider self-attention sub-layers and FFN sub-layers in this section. Our discussion, however, is general and the methods presented here can be applied to other parts of a Transformer model, for example, cross-attention sub-layers.

\subsection{Sparse Attention}
\label{sec:transformer-sparse-attention}

\noindent In practice, the attention mechanisms used in Transformers are time-consuming, especially when the input sequences are long. To illustrate, consider a Transformer decoder that predicts a distribution over the vocabulary at each time step given the previous words. Suppose the sequence generated by the decoder has length $n$ and the input to a self-attention sub-layer is an $n \times d$ matrix $\mathbf{S}$. First, $\mathbf{S}$ is linearly transformed to obtain the queries $\mathbf{S}^{q} \in \mathbb{R}^{n \times d}$, keys $\mathbf{S}^{k} \in \mathbb{R}^{n \times d}$, and values $\mathbf{S}^{v} \in \mathbb{R}^{n \times d}$. To simplify the notation in this subsection, we use $\mathbf{Q}$, $\mathbf{K}$ and $\mathbf{V}$ to represent $\mathbf{S}^{q}$, $\mathbf{S}^{k}$, and $\mathbf{S}^{v}$, respectively.

The output of the self-attention sub-layer can then be computed using
\begin{eqnarray}
\mathrm{Att}_{\mathrm{self}}(\mathbf{S}) & = & \mathbf{A} \mathbf{V}, \label{eq:efficient-self-attetention-decoder-as}
\end{eqnarray}

\noindent where $\mathbf{A}$ is an $n \times n$ attention matrix or attention map
\begin{eqnarray}
\mathbf{A} & = & \mathrm{Softmax}(\frac{\mathbf{Q} \mathbf{K}^\top}{\sqrt{d}} + \mathbf{M}). \label{eq:efficient-self-attetention-decoder-softmax}
\end{eqnarray}

\noindent Here, $\mathbf{M}$ is a masking matrix used to prevent the model from attending to future context at each position. Specifically, for a position $i$, $M(i,j) = 0$ if $j \le i$, and $M(i,j) = -\infty$ otherwise. Both the time and space complexity of the self-attention sub-layer scale quadratically with $n$\footnote{More precisely, the memory footprint is $n^2 + n \cdot d$, which is dominated by the $n^2$ term when $n \gg d$.}. As a result, standard self-attention becomes computationally prohibitive for large values of $n$.

The standard implementation of the above model relies on dense matrix computations, such as the multiplications in Eqs. (\ref{eq:efficient-self-attetention-decoder-as}) and (\ref{eq:efficient-self-attetention-decoder-softmax}). One common approach to reducing both memory footprint and floating-point operations is sparsification. By assuming $\mathbf{A}$ is a sparse matrix, only a fraction $\varrho \cdot n^2$ of its entries remain non-zero, where $\varrho$ is the \textbf{sparsity ratio}. Using sparse matrix representations drastically reduces memory requirements. Furthermore, computing $\frac{\mathbf{Q} \mathbf{K}^\top}{\sqrt{d}}$ and $\mathbf{A} \mathbf{V}$ becomes more efficient, as the model only processes a small subset of relevant positions.

Given a position $i$, we define the \textbf{attention field} $\pi_i$ to be the set of positions that are considered in computing the representation at this position. We therefore only need to compute the dot-product attention between the given position $i$ and each position $j \in \pi_i$. This results in a sparse attention matrix $\mathbf{A}'$ where
\begin{eqnarray}
A'(i,j) & = &
\begin{cases}
a_{i,j} & j \in \pi_i\ \textrm{and}\ j \le i \\
0 & \textrm{otherwise}
\end{cases},
\end{eqnarray}

\noindent where $a_{i,j}$ is a non-zero weight. A simple implementation of this model involves a slight modification to $\mathbf{M}$, leading to a new masking variable $\mathbf{M}'$
\begin{eqnarray}
M'(i,j) & = &
\begin{cases}
0 & j \in \pi_i\ \textrm{and}\ j \le i \\
-\infty & \textrm{otherwise}
\end{cases}.
\end{eqnarray}

\noindent In practical implementations, a more efficient approach is to employ sparse operations for $\mathbf{Q} \mathbf{K}^\top$ and $\mathbf{A}' \mathbf{V}$ by considering $\mathbf{M}'$ and $\mathbf{A}'$, respectively. That is, we compute only for pairs with non-zero attention weights and skip the rest.

Several approaches to sparse self-attention can be considered.

\begin{figure}[!htp]
\centering
\input{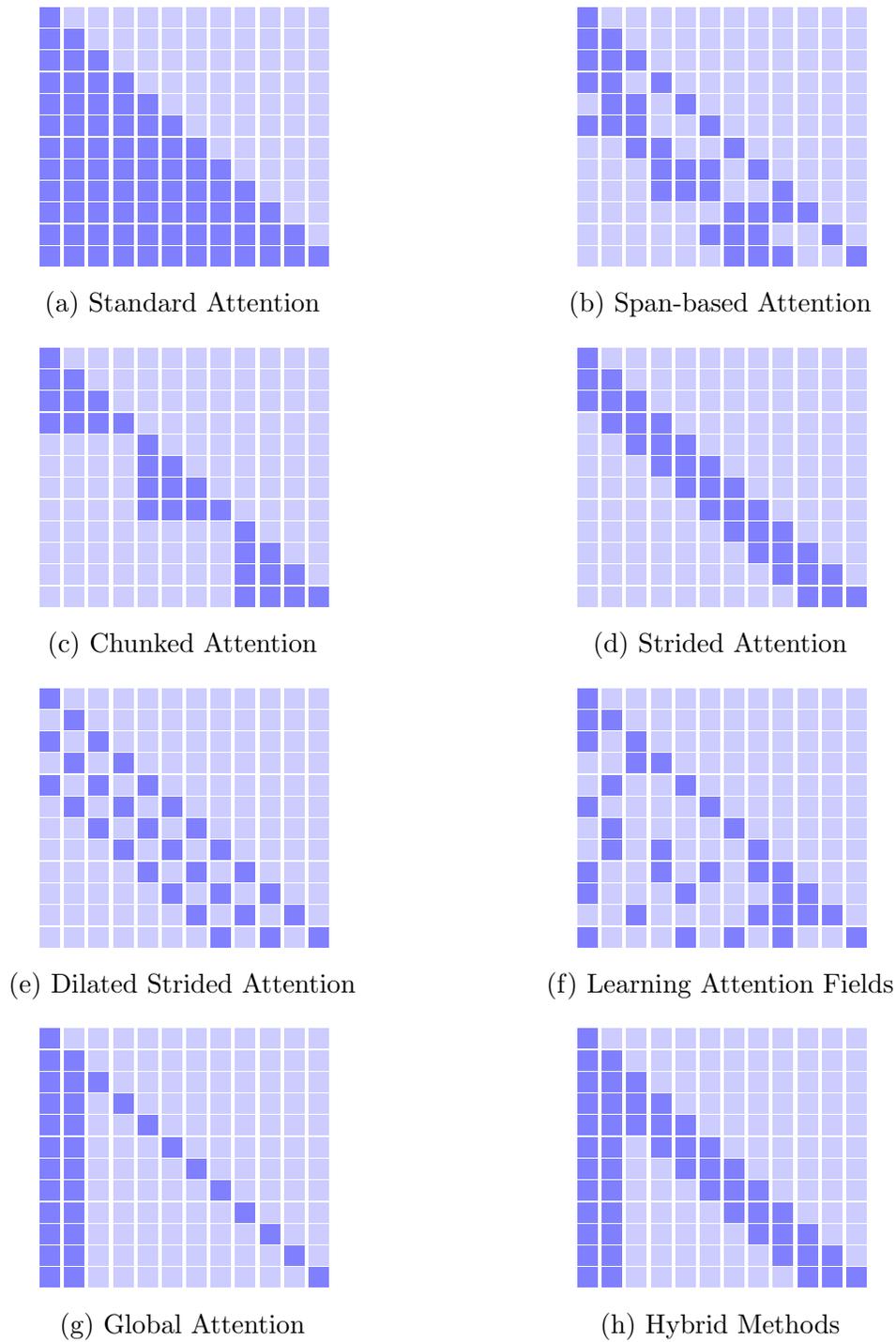}
\caption{Illustration of the attention maps of different models (self-attention on the decoder side). Dark cells mean $A'(i,j) \ne 0$ (i.e., $i$ attends to $j$), and light cells mean $A'(i,j) = 0$ (i.e., $i$ does not attend to $j$). In all these attention maps, we assume that every position attends to itself by default (see diagonals). }
\label{fig:sparse-attention-map}
\end{figure}

\begin{itemize}

\item \vspace{0.5em} \textbf{Span-based Attention}/\textbf{Local Attention}. As discussed in Section~\ref{sec:transformer-local-models}, the use of context in sequence modeling is local in many cases. The basic idea of local attention is to restrict attention to a local region of the input sequence. We can then write $\pi_i$ as
    \begin{eqnarray}
    \pi_i & = & [a_i^l,a_i^r],
    \end{eqnarray}

\noindent where $a_i^l$ and $a_i^r$ are the left and right ends of $\pi_i$. $a_i^r - a_i^l + 1$ determines the size of the region, and so we can use it to control the sparsity of the attention model, for example, if $a_i^r - a_i^l + 1 \ll n$, the model would be very sparse. $a_i^l$ and $a_i^r$ can be obtained by using either heuristics or machine learning methods. The reader may refer to related papers for more details~\citep{luong-etal:2015effective,sperber-etal:2018self,yang-etal:2018modeling,sukhbaatar-etal:2019adaptive}. See Figure~\ref{fig:sparse-attention-map} (b) for an illustration of local attention.

\item \vspace{0.3em} \textbf{Chunked Attention}. In this method, we segment a sequence into chunks and run the attention model on each of them~\citep{parmar-etal:2018image,qiu-etal:2020blockwise}. Given a sequence $\{1,\cdots,n\}$, we define $\{\mathrm{chunk}_1,\cdots,\mathrm{chunk}_q\}$ to be a segmentation of the sequence. A chunk can be expressed as a span
    \begin{eqnarray}
    \mathrm{chunk}_k & = & [c_k^l,c_k^r].
    \end{eqnarray}

    In the attention step, we treat each chunk as a sequence and perform self-attention on it as usual. In other words, the representation at position $i$ is computed by using only the context in the chunk that $i$ belongs to. In this sense, this model can be thought of as a variant of local attention. Figure~\ref{fig:sparse-attention-map} (c) shows an illustration of this model. A remaining issue is how to segment the sequence. There are several ways to do this. For example, as discussed in Section~\ref{sec:transformer-multi-scale-models}, we can do segmentation from a linguistic perspective, and segment the sequence into linguistically motivated units. In practical systems, it is sometimes more convenient to segment the sequence into chunks that are of equal length. Thus, the sparsity of the model is controlled by the size of these chunks, for example, the use of smaller chunks would lead to a more sparse attention model.

\item \vspace{0.3em} \textbf{Strided Attention}. Since the chunked attention approach enforces a hard segmentation on the input sequence, it may lose the ability to learn representations from inputs in different chunks. An alternative way to achieve chunk-wise attention is to allow overlap between chunks~\citep{child-etal:2019generating,Beltagy-etal:2020Longformer,ainslie-etal:2020etc}. This approach is analogous to the family of approaches that are commonly used to apply a local model to 1D or 2D data to generate outputs of the same shape. Like CNNs, we use a context window to represent the field of input of the attention model. The context window slides along the sequence, each time moving forward a step of size $\mathrm{stride}$. As a special case, if $\mathrm{stride}$ equals the size of the context window, this model reduces to the chunked attention model mentioned above. If $\mathrm{stride}$ is smaller than the size of the context window, the attention model will become denser. Figure~\ref{fig:sparse-attention-map} (d) shows the case of $\mathrm{stride}=1$ where the chunk overlapping is maximized. A way to achieve relatively sparser attention is to use a \textbf{dilated context window}. Figure~\ref{fig:sparse-attention-map} (e) shows an example of the dilated strided attention model, where the context window is discontinuous, with gaps of size 1.

\item \vspace{0.3em} \textbf{Learning Attention Fields}. Because the attention field $\pi_i$ can be any subset of $\{1,\cdots,n\}$, we can develop more general sparse attention models by considering attention maps beyond chunk-based patterns. The only question is how to determine which positions the model attends to for a given position. One simple approach is to use a computationally cheaper model to estimate the ``importance'' of each position. Then, attention weights are computed only for some of the positions which are thought to be most important~\citep{zhou-etal:2021informer}. A second approach is grouping: positions are grouped, and then the attention weights are computed only for positions in the same group. This can be achieved by clustering keys and queries. For example, we can cluster keys and queries via $k$-means clustering. The centroids of the clusters can be treated as additional parameters of the attention model, and so can be learned during optimization~\citep{roy-etal:2021efficient}. One benefit of learning attention fields is that the model can spread its attention more broadly over the sequence. This is a useful property for many NLP problems because word dependencies are sometimes long-range, not restricted to a local context window. See Figure~\ref{fig:sparse-attention-map} (f) for an example of the attention map learned through this model. Alternative approaches to learning to attend are to use sorting or hashing functions to group similar key and query vectors~\citep{kitaev-etal:2020reformer,tay-etal:2020sparse}. These functions can be either heuristically designed functions or neural networks with learnable parameters. By using these functions, we can reorder the sequence so that the inputs in the same group are adjacent in the reordered sequence. In this way, the resulting attention map follows a chunk-wise pattern, and the model is computationally efficient through the use of the chunked attention approach.

\item \vspace{0.3em} \textbf{Hybrid Methods}. Above, we have discussed a range of different sparse attention models. It is natural to explore methods that combine multiple models together to make use of their benefits in some way. A simple way to do this is to combine the attention fields of different models. For example, in \citet{zaheer-etal:2020big}'s system, the attention map is generated by considering three different sparse models, including local attention (chunked attention), global attention, and random attention\footnote{Here the global attention model attends each word only to a special word which accounts for the entire sequence and is often placed at the beginning of the sequence. The random attention model attends each word to a random set of the words of the sequence.}. The resulting model is still a sparse model, but is somewhat more robust as it involves multiple patterns from different perspectives of attention modeling. Another way of combining multiple attention models is to use different models for different heads in multi-head attention~\citep{child-etal:2019generating,Beltagy-etal:2020Longformer}. For example, one can use one head as a local attention model, and use another head as a global attention model (see Figure~\ref{fig:sparse-attention-map} (g-h)).

\end{itemize}
\vspace{0.5em}

One disadvantage of sparse models compared to dense models is that they are often less efficient on GPUs/CPUs due to hardware constraints. While sparse models can ideally reduce both the memory and computation requirements, the computational throughput of sparse models is much slower than that of dense models. In practice, it is difficult for sparse models to approach the peak FLOPS of a GPU or CPU\footnote{FLOPS = floating point operations per second.}. Therefore, they are primarily used to improve memory efficiency rather than achieving raw computational speedups on current hardware. On the other hand, sparse models are still of great use to NLP practitioners in the context of memory-efficient Transformer, especially when Transformers are used to deal with extremely long sequences.

\subsection{Recurrent and Memory Models}
\label{sec:transformer-recurrent-and-memory-models}

\noindent For sequence generation problems, Transformers can also be thought of as memory systems. Consider again the general autoregressive setting, in which we are given the states of the previous $i-1$ positions and wish to predict the next state. In self-attention, this is achieved by using the query at position $i$ (i.e., $\mathbf{q}_{i}$) to access the key-value pairs of the previous positions (i.e., $\{(\mathbf{k}_{1},\mathbf{v}_{1}),\cdots,(\mathbf{k}_{i-1},\mathbf{v}_{i-1})\}$). Then, we move to position $i+1$ and append $(\mathbf{k}_{i},\mathbf{v}_{i})$ to the collection of key-value pairs. This procedure can be intuitively interpreted using memory mechanisms. A Transformer model maintains a memory bank that retains past information. As the model processes the sequence, it repeats the same operation: generating outputs by reading from the memory, and then updating the memory to incorporate new information. This is illustrated in Figure~\ref{fig:transformer-as-memory}.

\begin{figure}[!htp]
\centering

\begin{tikzpicture}

\def\nodehsep{0.8cm}

\tikzstyle{bnode} = [minimum width=0.4cm,minimum height=0.4cm,inner sep=2pt];

\begin{scope}

\node [bnode,anchor=west] (p1) at (0,0) {$1$};
\node [bnode,anchor=west] (p2) at ([xshift=\nodehsep]p1.east) {$2$};
\node [bnode,anchor=west] (p3) at ([xshift=\nodehsep]p2.east) {$3$};
\node [bnode,anchor=west] (pdots) at ([xshift=\nodehsep]p3.east) {...};
\node [bnode,anchor=west] (pi2) at ([xshift=\nodehsep]pdots.east) {$i-2$};
\node [bnode,anchor=west] (pi1) at ([xshift=\nodehsep]pi2.east) {$i-1$};
\node [bnode,anchor=west] (pi0) at ([xshift=2*\nodehsep]pi1.east) {$i$};
\node [bnode,anchor=west] (pin) at ([xshift=\nodehsep]pi0.east) {$i+1$};

\draw[decoration={brace,amplitude=9pt}, decorate,thick] ([yshift=0.2cm,xshift=3pt]p1.north west) -- ([yshift=0.2cm,xshift=-3pt]pi1.north east);

\node [anchor=center,minimum width=6cm,minimum height=0.8cm,draw,thick] (memory) at ([yshift=1.5cm]pdots.north west) {Memory ($\{1,...,i-1\}$)};

\node [anchor=center] (state) at ([yshift=1.5cm]pi0.north) {State};

\draw [<-,thick] ([xshift=-2pt,yshift=3pt]state.west) .. controls +(130:0.7cm) and +(50:0.5cm) .. ([xshift=2pt,yshift=3pt]memory.east) node [pos=0.5,above] {\scriptsize{Read (self-attention)}};
\draw [->,thick] ([xshift=-2pt,yshift=-3pt]state.west) .. controls +(-130:0.7cm) and +(-50:0.5cm) .. ([xshift=2pt,yshift=-3pt]memory.east) node [pos=0.5,below] {\scriptsize{Update}};
\draw [->] ([yshift=0.2cm]pi0.north) -- ([yshift=-0.2cm]state.south);

\begin{pgfonlayer}{background}
\node [inner sep=2pt,fill=red!20]  [fit = (p1) (pi1)] (hisotrywords) {};
\node [inner sep=2pt,fill=blue!20]  [fit = (pi0) ] (currentwords) {};
\end{pgfonlayer}

\draw [->, thick] ([yshift=-0.5cm,xshift=-0.3cm]p1.south west) -- ([yshift=-0.5cm,xshift=12cm]p1.south west);
\node [anchor=north] (label) at ([yshift=-0.6cm]pi0.south) {\small{Position $i$}};
\node [inner sep=0,circle,fill=black,minimum size=4pt] (dot) at ([yshift=0.1cm]label.north) {};

\end{scope}

\begin{scope}[yshift=-5cm]

\node [bnode,anchor=west] (p1) at (0,0) {$1$};
\node [bnode,anchor=west] (p2) at ([xshift=\nodehsep]p1.east) {$2$};
\node [bnode,anchor=west] (p3) at ([xshift=\nodehsep]p2.east) {$3$};
\node [bnode,anchor=west] (pdots) at ([xshift=\nodehsep]p3.east) {...};
\node [bnode,anchor=west] (pi2) at ([xshift=\nodehsep]pdots.east) {$i-2$};
\node [bnode,anchor=west] (pi1) at ([xshift=\nodehsep]pi2.east) {$i-1$};
\node [bnode,anchor=west] (pi0) at ([xshift=\nodehsep]pi1.east) {$i$};
\node [bnode,anchor=west] (pin) at ([xshift=2*\nodehsep]pi0.east) {$i+1$};

\draw[decoration={brace,amplitude=9pt}, decorate,thick] ([yshift=0.2cm,xshift=3pt]p1.north west) -- ([yshift=0.2cm,xshift=-3pt]pi0.north east);

\node [anchor=center,minimum width=6cm,minimum height=0.8cm,draw,thick] (memory) at ([yshift=1.5cm]pdots.north west) {Memory ($\{1,...,i-1,i\}$)};

\node [anchor=center] (state) at ([yshift=1.5cm]pin.north) {State};

\draw [<-,thick] ([xshift=-2pt,yshift=3pt]state.west) .. controls +(130:0.7cm) and +(50:0.5cm) .. ([xshift=2pt,yshift=3pt]memory.east) node [pos=0.5,above] {\scriptsize{Read (self-attention)}};
\draw [->,thick] ([xshift=-2pt,yshift=-3pt]state.west) .. controls +(-130:0.7cm) and +(-50:0.5cm) .. ([xshift=2pt,yshift=-3pt]memory.east) node [pos=0.5,below] {\scriptsize{Update}};
\draw [->] ([yshift=0.2cm]pin.north) -- ([yshift=-0.2cm]state.south);

\begin{pgfonlayer}{background}
\node [inner sep=2pt,fill=red!20]  [fit = (p1) (pi0)] (hisotrywords) {};
\node [inner sep=2pt,fill=blue!20]  [fit = (pin) ] (currentwords) {};
\end{pgfonlayer}

\draw [->, thick] ([yshift=-0.5cm,xshift=-0.3cm]p1.south west) -- ([yshift=-0.5cm,xshift=12cm]p1.south west);
\node [anchor=north] (label) at ([yshift=-0.6cm]pin.south) {\small{Position $i+1$}};
\node [inner sep=0,circle,fill=black,minimum size=4pt] (dot) at ([yshift=0.1cm]label.north) {};

\end{scope}

\end{tikzpicture}
\caption{The Transformer as a memory system. At position $i$, the collection of key-value pairs from positions $\{1,\cdots,i-1\}$ serves as a memory of past context. The model accesses this memory to generate an output, and subsequently adds the key-value pair of position $i$ to the memory. Moving to the next position, this cycle of memory access and update is repeated.}
\label{fig:transformer-as-memory}
\end{figure}

\subsubsection{Cache-based Memory}

\noindent The memory here can be viewed as a datastore of vectors. From a machine learning perspective, this is a non-parametric model, and the cost of accessing the model grows as the sequence becomes longer. Clearly, such a variable-length memory will generally be infeasible if the model deals with extremely long sequences. For modeling sequences of arbitrary length, it is common to use a fixed-length memory instead. As in many NLP problems, one of the simplest ways to do this is to maintain a cache of recent information, that is, we restrict the modeling to a context window. Let $n_c$ be the size of the context window. The model keeps track of the $n_c - 1$ latest states preceding the current position, so that its closest predecessors can be considered at each step. This means that, for each position, a self-attention sub-layer attends to $n_c - 1$ previous positions, as follows

\vspace{0.3em}
\begin{center}

\begin{tikzpicture}

\def\nodehsep{1.0cm}
\def\nodevsep{1.0cm}

\tikzstyle{snode} = [circle,minimum size=0.2cm,inner sep=0pt,draw,white,fill=black];

\begin{scope}

\node [snode,anchor=west] (s11) at (0,0) {};
\node [snode,anchor=east] (s12) at ([xshift=-\nodehsep]s11.west) {};
\node [snode,anchor=east] (s13) at ([xshift=-\nodehsep]s12.west) {};
\node [snode,anchor=east] (s14) at ([xshift=-\nodehsep]s13.west) {};
\node [snode,anchor=east] (s15) at ([xshift=-\nodehsep]s14.west) {};
\node [snode,anchor=east] (s16) at ([xshift=-\nodehsep]s15.west) {};

\node [snode,anchor=south] (s21) at ([yshift=\nodevsep]s11.north) {};
\node [snode,anchor=south,white,fill=gray!30] (s22) at ([yshift=\nodevsep]s12.north) {};
\node [snode,anchor=south,white,fill=gray!30] (s23) at ([yshift=\nodevsep]s13.north) {};
\node [snode,anchor=south,white,fill=gray!30] (s24) at ([yshift=\nodevsep]s14.north) {};
\node [snode,anchor=south,white,fill=gray!30] (s25) at ([yshift=\nodevsep]s15.north) {};
\node [snode,anchor=south,white,fill=gray!30] (s26) at ([yshift=\nodevsep]s16.north) {};

\node [anchor=west] (input1) at ([xshift=0.5cm]s11.east) {\small{Input}};
\node [anchor=west] (output) at ([xshift=0.5cm]s21.east) {\small{Output}};

\node [anchor=north] (p1) at ([yshift=-0.1cm]s11.south) {\footnotesize{$i$}};
\node [anchor=north] (p2) at ([yshift=-0.1cm]s12.south) {\footnotesize{$i - 1$}};
\node [anchor=north] (p3) at ([yshift=-0.1cm]s13.south) {\footnotesize{$i - 2$}};
\node [anchor=north] (p4) at ([yshift=-0.1cm]s14.south) {\footnotesize{$i - 3$}};
\node [anchor=north] (p5) at ([yshift=-0.1cm]s15.south) {\footnotesize{$i - 4$}};
\node [anchor=north] (p6) at ([yshift=-0.1cm]s16.south) {\footnotesize{$i - 5$}};

\draw [->,thick] (s21.210) -- (s13.60);
\draw [->,thick] (s21.240) -- (s12.75);
\draw [->,thick] (s21.270) -- (s11.90);

\end{scope}

\end{tikzpicture}
\end{center}
\vspace{0.3em}

\noindent If we stack multiple self-attention sub-layers, a larger context window would be considered. For example, a model involving two self-attention sub-layers has a context window of size $2n_c - 1$, as follows

\vspace{0.3em}
\begin{center}

\begin{tikzpicture}

\def\nodehsep{1.0cm}
\def\nodevsep{1.0cm}

\tikzstyle{snode} = [circle,minimum size=0.2cm,inner sep=0pt,draw,white,fill=black];

\begin{scope}

\node [snode,anchor=west] (s11) at (0,0) {};
\node [snode,anchor=east] (s12) at ([xshift=-\nodehsep]s11.west) {};
\node [snode,anchor=east] (s13) at ([xshift=-\nodehsep]s12.west) {};
\node [snode,anchor=east] (s14) at ([xshift=-\nodehsep]s13.west) {};
\node [snode,anchor=east] (s15) at ([xshift=-\nodehsep]s14.west) {};
\node [snode,anchor=east] (s16) at ([xshift=-\nodehsep]s15.west) {};

\node [snode,anchor=south] (s21) at ([yshift=\nodevsep]s11.north) {};
\node [snode,anchor=south] (s22) at ([yshift=\nodevsep]s12.north) {};
\node [snode,anchor=south] (s23) at ([yshift=\nodevsep]s13.north) {};
\node [snode,anchor=south,white,fill=gray!30] (s24) at ([yshift=\nodevsep]s14.north) {};
\node [snode,anchor=south,white,fill=gray!30] (s25) at ([yshift=\nodevsep]s15.north) {};
\node [snode,anchor=south,white,fill=gray!30] (s26) at ([yshift=\nodevsep]s16.north) {};

\node [snode,anchor=south] (s31) at ([yshift=\nodevsep]s21.north) {};
\node [snode,anchor=south,white,fill=gray!30] (s32) at ([yshift=\nodevsep]s22.north) {};
\node [snode,anchor=south,white,fill=gray!30] (s33) at ([yshift=\nodevsep]s23.north) {};
\node [snode,anchor=south,white,fill=gray!30] (s34) at ([yshift=\nodevsep]s24.north) {};
\node [snode,anchor=south,white,fill=gray!30] (s35) at ([yshift=\nodevsep]s25.north) {};
\node [snode,anchor=south,white,fill=gray!30] (s36) at ([yshift=\nodevsep]s26.north) {};

\node [anchor=west] (input1) at ([xshift=0.5cm]s11.east) {\small{Layer 1}};
\node [anchor=west] (input2) at ([xshift=0.5cm]s21.east) {\small{Layer 2}};
\node [anchor=west] (output) at ([xshift=0.5cm]s31.east) {\small{Output}};

\node [anchor=north] (p1) at ([yshift=-0.1cm]s11.south) {\footnotesize{$i$}};
\node [anchor=north] (p2) at ([yshift=-0.1cm]s12.south) {\footnotesize{$i - 1$}};
\node [anchor=north] (p3) at ([yshift=-0.1cm]s13.south) {\footnotesize{$i - 2$}};
\node [anchor=north] (p4) at ([yshift=-0.1cm]s14.south) {\footnotesize{$i - 3$}};
\node [anchor=north] (p5) at ([yshift=-0.1cm]s15.south) {\footnotesize{$i - 4$}};
\node [anchor=north] (p6) at ([yshift=-0.1cm]s16.south) {\footnotesize{$i - 5$}};

\draw [->,thick] (s21.210) -- (s13.60);
\draw [->,thick] (s21.240) -- (s12.75);
\draw [->,thick] (s21.270) -- (s11.90);

\draw [->,thick] (s22.210) -- (s14.60);
\draw [->,thick] (s22.240) -- (s13.75);
\draw [->,thick] (s22.270) -- (s12.90);

\draw [->,thick] (s23.210) -- (s15.60);
\draw [->,thick] (s23.240) -- (s14.75);
\draw [->,thick] (s23.270) -- (s13.90);

\draw [->,thick] (s31.210) -- (s23.60);
\draw [->,thick] (s31.240) -- (s22.75);
\draw [->,thick] (s31.270) -- (s21.90);

\end{scope}

\end{tikzpicture}
\end{center}
\vspace{0.3em}

Therefore, we can capture a sufficiently large context by using a multi-layer Transformer model. Note that the context window model here is essentially the same as the strided attention model presented in the preceding section. Systems of this type are often easy to implement: we slide a window along the sequence, and, in each move, we make predictions at the last position of the window (for inference), or back-propagate errors (for training).

An alternative strategy for training a context window model is chunked attention. The sequence is explicitly partitioned into sub-sequences (or chunks) of a fixed length $n_c$. These chunks are then treated as independent training samples. This approach, however, completely ignores the relationship between inputs in different chunks. One way to address this issue is to introduce dependence between chunks. For example, the \textbf{Transformer-XL} model allows every chunk to access one or more preceding chunks~\citep{dai-etal:2019transformer}. In the simplest case, consider an example in which $\mathrm{chunk}_k$ can see its predecessor $\mathrm{chunk}_{k-1}$. Each position in $\mathrm{chunk}_k$ can attend to all its preceding positions in both $\mathrm{chunk}_k$ and $\mathrm{chunk}_{k-1}$.

In Transformer-XL, this approach is implemented in a simplified form. First, each position is constrained to attend to $n_c - 1$ previous positions so that the size of the attention field of a position is the same in the training and inference stages. Such a method turns the problem back to strided attention, making the implementation of the attention model straightforward. However, unlike standard strided attention, Transformer-XL performs training in a chunk-wise manner. Once we finish the training on a chunk, we directly move to the next chunk, rather than sliding the context window a small step forward. Second, while this approach allows for connections between chunks, the parameters of the sub-network corresponding to $\mathrm{chunk}_{k-1}$ are fixed, and we only update the parameters of the sub-network on $\mathrm{chunk}_{k}$ in the $k$-th step. See Figure~\ref{fig:transformer-xl-illustration} for an illustration.

\begin{figure}[!htp]
\centering

\begin{tikzpicture}

\def\nodehsep{1.0cm}
\def\nodevsep{1.0cm}

\tikzstyle{snode} = [circle,minimum size=0.2cm,inner sep=0pt,draw,white,fill=black];


\begin{scope}

\node [snode,anchor=west,white,fill=gray!40] (s11) at (0,0) {};
\node [snode,anchor=east,white,fill=gray!40] (s12) at ([xshift=-\nodehsep]s11.west) {};
\node [snode,anchor=east,white,fill=gray!40] (s13) at ([xshift=-\nodehsep]s12.west) {};
\node [snode,anchor=east] (s14) at ([xshift=-\nodehsep]s13.west) {};
\node [snode,anchor=east] (s15) at ([xshift=-\nodehsep]s14.west) {};
\node [snode,anchor=east] (s16) at ([xshift=-\nodehsep]s15.west) {};
\node [snode,anchor=east] (s17) at ([xshift=-\nodehsep]s16.west) {};
\node [snode,anchor=east] (s18) at ([xshift=-\nodehsep]s17.west) {};
\node [snode,anchor=east] (s19) at ([xshift=-\nodehsep]s18.west) {};

\node [snode,anchor=south,white,fill=gray!40] (s21) at ([yshift=\nodevsep]s11.north) {};
\node [snode,anchor=south,white,fill=gray!40] (s22) at ([yshift=\nodevsep]s12.north) {};
\node [snode,anchor=south,white,fill=gray!40] (s23) at ([yshift=\nodevsep]s13.north) {};
\node [snode,anchor=south] (s24) at ([yshift=\nodevsep]s14.north) {};
\node [snode,anchor=south] (s25) at ([yshift=\nodevsep]s15.north) {};
\node [snode,anchor=south] (s26) at ([yshift=\nodevsep]s16.north) {};
\node [snode,anchor=south] (s27) at ([yshift=\nodevsep]s17.north) {};
\node [snode,anchor=south] (s28) at ([yshift=\nodevsep]s18.north) {};
\node [snode,anchor=south,white,fill=gray!40] (s29) at ([yshift=\nodevsep]s19.north) {};

\node [snode,anchor=south,white,fill=gray!40] (s31) at ([yshift=\nodevsep]s21.north) {};
\node [snode,anchor=south,white,fill=gray!40] (s32) at ([yshift=\nodevsep]s22.north) {};
\node [snode,anchor=south,white,fill=gray!40] (s33) at ([yshift=\nodevsep]s23.north) {};
\node [snode,anchor=south] (s34) at ([yshift=\nodevsep]s24.north) {};
\node [snode,anchor=south] (s35) at ([yshift=\nodevsep]s25.north) {};
\node [snode,anchor=south] (s36) at ([yshift=\nodevsep]s26.north) {};
\node [snode,anchor=south,white,fill=gray!40] (s37) at ([yshift=\nodevsep]s27.north) {};
\node [snode,anchor=south,white,fill=gray!40] (s38) at ([yshift=\nodevsep]s28.north) {};
\node [snode,anchor=south,white,fill=gray!40] (s39) at ([yshift=\nodevsep]s29.north) {};

\node [anchor=west] (input1) at ([xshift=0.5cm]s11.east) {\small{Layer 1}};
\node [anchor=west] (input2) at ([xshift=0.5cm]s21.east) {\small{Layer 2}};
\node [anchor=west] (output) at ([xshift=0.5cm]s31.east) {\small{Output}};

\node [anchor=north] (p1) at ([yshift=-0.3cm]s11.south) {\footnotesize{$i+3$}};
\node [anchor=north] (p2) at ([yshift=-0.3cm]s12.south) {\footnotesize{$i+2$}};
\node [anchor=north] (p3) at ([yshift=-0.3cm]s13.south) {\footnotesize{$i+1$}};
\node [anchor=north] (p4) at ([yshift=-0.3cm]s14.south) {\footnotesize{$i$}};
\node [anchor=north] (p5) at ([yshift=-0.3cm]s15.south) {\footnotesize{$i-1$}};
\node [anchor=north] (p6) at ([yshift=-0.3cm]s16.south) {\footnotesize{$i-2$}};
\node [anchor=north] (p7) at ([yshift=-0.3cm]s17.south) {\footnotesize{$i-3$}};
\node [anchor=north] (p8) at ([yshift=-0.3cm]s18.south) {\footnotesize{$i-4$}};
\node [anchor=north] (p9) at ([yshift=-0.3cm]s19.south) {\footnotesize{$i-5$}};

\node [anchor=north] (caption) at ([yshift=-0.3cm]p4.south) {\small{(a) Step $k$ of chunk-wise training}};

\begin{pgfonlayer}{background}
\node [draw,thick,dotted,inner sep=6pt] [fit = (s11) (s13) (s31)] (chunkbox1) {};
\node [draw,thick,dotted,inner sep=6pt] [fit = (s14) (s16) (s34)] (chunkbox2) {};
\node [draw,thick,dotted,inner sep=6pt] [fit = (s17) (s19) (s37)] (chunkbox3) {};
\end{pgfonlayer}

\node [anchor=south] (chunk1) at (chunkbox1.north) {\small{$\mathrm{chunk}_{k+1}$}};
\node [anchor=south] (chunk2) at (chunkbox2.north) {\small{$\mathrm{chunk}_{k}$}};
\node [anchor=south] (chunk3) at (chunkbox3.north) {\small{$\mathrm{chunk}_{k-1}$}};

\draw [->,thick,blue] (s24.210) -- (s16.60);
\draw [->,thick,blue] (s24.240) -- (s15.75);
\draw [->,thick,blue] (s24.270) -- (s14.90);

\draw [->,dashed,thick,blue] (s25.210) -- (s17.60);
\draw [->,thick,blue] (s25.240) -- (s16.75);
\draw [->,thick,blue] (s25.270) -- (s15.90);

\draw [->,dashed,thick,blue] (s26.210) -- (s18.60);
\draw [->,dashed,thick,blue] (s26.240) -- (s17.75);
\draw [->,thick,blue] (s26.270) -- (s16.90);

\draw [->,dashed] (s27.210) -- (s19.60);
\draw [->,dashed] (s27.240) -- (s18.75);
\draw [->,dashed] (s27.270) -- (s17.90);

\draw [->,dashed] (s28.240) -- (s19.75);
\draw [->,dashed] (s28.270) -- (s18.90);

\draw [->,thick,blue] (s34.210) -- (s26.60);
\draw [->,thick,blue] (s34.240) -- (s25.75);
\draw [->,thick,blue] (s34.270) -- (s24.90);

\draw [->,dashed,thick,blue] (s35.210) -- (s27.60);
\draw [->,thick,blue] (s35.240) -- (s26.75);
\draw [->,thick,blue] (s35.270) -- (s25.90);

\draw [->,dashed,thick,blue] (s36.210) -- (s28.60);
\draw [->,dashed,thick,blue] (s36.240) -- (s27.75);
\draw [->,thick,blue] (s36.270) -- (s26.90);

\end{scope}


\begin{scope}[yshift=-6cm]

\node [snode,anchor=west] (s11) at (0,0) {};
\node [snode,anchor=east] (s12) at ([xshift=-\nodehsep]s11.west) {};
\node [snode,anchor=east] (s13) at ([xshift=-\nodehsep]s12.west) {};
\node [snode,anchor=east] (s14) at ([xshift=-\nodehsep]s13.west) {};
\node [snode,anchor=east] (s15) at ([xshift=-\nodehsep]s14.west) {};
\node [snode,anchor=east] (s16) at ([xshift=-\nodehsep]s15.west) {};
\node [snode,anchor=east] (s17) at ([xshift=-\nodehsep]s16.west) {};
\node [snode,anchor=east,white,fill=gray!40] (s18) at ([xshift=-\nodehsep]s17.west) {};
\node [snode,anchor=east,white,fill=gray!40] (s19) at ([xshift=-\nodehsep]s18.west) {};

\node [snode,anchor=south] (s21) at ([yshift=\nodevsep]s11.north) {};
\node [snode,anchor=south] (s22) at ([yshift=\nodevsep]s12.north) {};
\node [snode,anchor=south] (s23) at ([yshift=\nodevsep]s13.north) {};
\node [snode,anchor=south] (s24) at ([yshift=\nodevsep]s14.north) {};
\node [snode,anchor=south] (s25) at ([yshift=\nodevsep]s15.north) {};
\node [snode,anchor=south,white,fill=gray!40] (s26) at ([yshift=\nodevsep]s16.north) {};
\node [snode,anchor=south,white,fill=gray!40] (s27) at ([yshift=\nodevsep]s17.north) {};
\node [snode,anchor=south,white,fill=gray!40] (s28) at ([yshift=\nodevsep]s18.north) {};
\node [snode,anchor=south,white,fill=gray!40] (s29) at ([yshift=\nodevsep]s19.north) {};

\node [snode,anchor=south] (s31) at ([yshift=\nodevsep]s21.north) {};
\node [snode,anchor=south] (s32) at ([yshift=\nodevsep]s22.north) {};
\node [snode,anchor=south] (s33) at ([yshift=\nodevsep]s23.north) {};
\node [snode,anchor=south,white,fill=gray!40] (s34) at ([yshift=\nodevsep]s24.north) {};
\node [snode,anchor=south,white,fill=gray!40] (s35) at ([yshift=\nodevsep]s25.north) {};
\node [snode,anchor=south,white,fill=gray!40] (s36) at ([yshift=\nodevsep]s26.north) {};
\node [snode,anchor=south,white,fill=gray!40] (s37) at ([yshift=\nodevsep]s27.north) {};
\node [snode,anchor=south,white,fill=gray!40] (s38) at ([yshift=\nodevsep]s28.north) {};
\node [snode,anchor=south,white,fill=gray!40] (s39) at ([yshift=\nodevsep]s29.north) {};

\node [anchor=west] (input1) at ([xshift=0.5cm]s11.east) {\small{Layer 1}};
\node [anchor=west] (input2) at ([xshift=0.5cm]s21.east) {\small{Layer 2}};
\node [anchor=west] (output) at ([xshift=0.5cm]s31.east) {\small{Output}};

\node [anchor=north] (p1) at ([yshift=-0.3cm]s11.south) {\footnotesize{$i+3$}};
\node [anchor=north] (p2) at ([yshift=-0.3cm]s12.south) {\footnotesize{$i+2$}};
\node [anchor=north] (p3) at ([yshift=-0.3cm]s13.south) {\footnotesize{$i+1$}};
\node [anchor=north] (p4) at ([yshift=-0.3cm]s14.south) {\footnotesize{$i$}};
\node [anchor=north] (p5) at ([yshift=-0.3cm]s15.south) {\footnotesize{$i-1$}};
\node [anchor=north] (p6) at ([yshift=-0.3cm]s16.south) {\footnotesize{$i-2$}};
\node [anchor=north] (p7) at ([yshift=-0.3cm]s17.south) {\footnotesize{$i-3$}};
\node [anchor=north] (p8) at ([yshift=-0.3cm]s18.south) {\footnotesize{$i-4$}};
\node [anchor=north] (p9) at ([yshift=-0.3cm]s19.south) {\footnotesize{$i-5$}};

\node [anchor=north] (caption) at ([yshift=-0.3cm]p4.south) {\small{(b) Step $k+1$ of chunk-wise training}};

\begin{pgfonlayer}{background}
\node [draw,thick,dotted,inner sep=6pt] [fit = (s11) (s13) (s31)] (chunkbox1) {};
\node [draw,thick,dotted,inner sep=6pt] [fit = (s14) (s16) (s34)] (chunkbox2) {};
\node [draw,thick,dotted,inner sep=6pt] [fit = (s17) (s19) (s37)] (chunkbox3) {};
\end{pgfonlayer}

\node [anchor=south] (chunk1) at (chunkbox1.north) {\small{$\mathrm{chunk}_{k+1}$}};
\node [anchor=south] (chunk2) at (chunkbox2.north) {\small{$\mathrm{chunk}_{k}$}};
\node [anchor=south] (chunk3) at (chunkbox3.north) {\small{$\mathrm{chunk}_{k-1}$}};

\draw [->,thick,blue] (s21.210) -- (s13.60);
\draw [->,thick,blue] (s21.240) -- (s12.75);
\draw [->,thick,blue] (s21.270) -- (s11.90);

\draw [->,dashed,thick,blue] (s22.210) -- (s14.60);
\draw [->,thick,blue] (s22.240) -- (s13.75);
\draw [->,thick,blue] (s22.270) -- (s12.90);

\draw [->,dashed,thick,blue] (s23.210) -- (s15.60);
\draw [->,dashed,thick,blue] (s23.240) -- (s14.75);
\draw [->,thick,blue] (s23.270) -- (s13.90);

\draw [->,dashed] (s24.210) -- (s16.60);
\draw [->,dashed] (s24.240) -- (s15.75);
\draw [->,dashed] (s24.270) -- (s14.90);

\draw [->,dashed] (s25.210) -- (s17.60);
\draw [->,dashed] (s25.240) -- (s16.75);
\draw [->,dashed] (s25.270) -- (s15.90);

\draw [->,thick,blue] (s31.210) -- (s23.60);
\draw [->,thick,blue] (s31.240) -- (s22.75);
\draw [->,thick,blue] (s31.270) -- (s21.90);

\draw [->,dashed,thick,blue] (s32.210) -- (s24.60);
\draw [->,thick,blue] (s32.240) -- (s23.75);
\draw [->,thick,blue] (s32.270) -- (s22.90);

\draw [->,dashed,thick,blue] (s33.210) -- (s25.60);
\draw [->,dashed,thick,blue] (s33.240) -- (s24.75);
\draw [->,thick,blue] (s33.270) -- (s23.90);

\end{scope}

\end{tikzpicture}
\caption{Illustration of chunk-wise training in Transformer-XL~\citep{dai-etal:2019transformer}. The input sequence is divided into non-overlapping chunks of length $n_c$. Training proceeds one chunk at a time. Within $\mathrm{chunk}_k$, the attention span for every token is restricted to a left-context window of size $n_c$. This structure permits cross-chunk attention. For example, position $i - 2$ in $\mathrm{chunk}_k$ can attend to positions $i - 3$ and $i - 4$ residing in $\mathrm{chunk}_{k-1}$ (see sub-figure (a)). During back-propagation, gradients are computed for the current sub-network ($\mathrm{chunk}_k$), treating cached activations from previous chunks as fixed constants. Dashed lines denote forward-pass information flows that do not receive gradients. After completing the training step for $\mathrm{chunk}_k$, the process moves to the next chunk and repeats.}
\label{fig:transformer-xl-illustration}
\end{figure}

The above model is similar in spirit to recurrent models because both of them require the computation in one step to depend on the states of the preceding steps. However, it is not in the standard form of a recurrent model, in which the output of a recurrent unit in one step is the input in the next step. Instead, the ``recurrence'' is expressed by involving connections across layers and chunks, that is, the output of one layer in $\mathrm{chunk}_{k-1}$ is used as the input of a higher-level layer in $\mathrm{chunk}_{k}$.

\subsubsection{Encoding Long-term Memory}

\noindent Another idea for representing the states of a sequence is to frame the task as an encoding problem. Instead of storing all key-value vectors during left-to-right generation, we compress the memory of the entire ``history'' into a fixed number of encoded key-value vectors. These encoded vectors can be either a small subset of the original sequence $\{(\mathbf{k}_{1},\mathbf{v}_{1}),\cdots,(\mathbf{k}_{i-1},\mathbf{v}_{i-1})\}$, or a small set of newly generated vectors that encode this entire history.

One way to achieve this encoding is to apply a pooling operation over $\{(\mathbf{k}_{1},\mathbf{v}_{1}),\cdots,(\mathbf{k}_{i-1},\mathbf{v}_{i-1})\}$~\citep{rae-etal:2019compressive}. For example, by using average pooling, the memory is reduced to a single key-value pair $(\bar{\mathbf{k}},\bar{\mathbf{v}})$:
\begin{eqnarray}
\bar{\mathbf{k}} & = & \frac{1}{i-1} \sum_{j=1}^{i-1} \mathbf{k}_{j} \label{eq:average-attention-k}, \\
\bar{\mathbf{v}} & = & \frac{1}{i-1} \sum_{j=1}^{i-1} \mathbf{v}_{j} \label{eq:average-attention-v}.
\end{eqnarray}

\noindent This leads to a very efficient model, as we only need to update the aggregated vectors $(\bar{\mathbf{k}},\bar{\mathbf{v}})$ incrementally~\citep{zhang-etal:2018accelerating}. Let $(\bar{\mathbf{k}}[i],\bar{\mathbf{v}}[i])$ represent the state of the memory at position $i$. A more general formulation is given recursively:
\begin{eqnarray}
\bar{\mathbf{k}}[i] & = & \mathrm{KMem}(\bar{\mathbf{k}}[i-1],\mathbf{k}_{i-1}) \label{eq:memory-attention-k}, \\
\bar{\mathbf{v}}[i] & = & \mathrm{VMem}(\bar{\mathbf{v}}[i-1],\mathbf{v}_{i-1}) \label{eq:memory-attention-v}.
\end{eqnarray}

\noindent where $\mathrm{KMem}(\cdot)$ and $\mathrm{VMem}(\cdot)$ are functions that update the memory by taking both the states of the memory at the previous position (i.e., $\bar{\mathbf{k}}[i-1]$ and $\bar{\mathbf{v}}[i-1]$) and the new states (i.e., $\mathbf{k}_{i-1}$ and $\mathbf{v}_{i-1}$). There are many possible choices for functions $\mathrm{KMem}(\cdot)$ and $\mathrm{VMem}(\cdot)$. For example, if $\mathrm{KMem}(\cdot)$ and $\mathrm{VMem}(\cdot)$ are weighted sum functions, we can derive the same forms as Eqs. (\ref{eq:average-attention-k}) and (\ref{eq:average-attention-v}). If $\mathrm{KMem}(\cdot)$ and $\mathrm{VMem}(\cdot)$ are recurrent cells (e.g., standard RNN or LSTM cells), we obtain a recurrent model of memory.

Extending the above concept to memories with more than one key-value pair is straightforward. One approach is chunk-level representation. Let $\{(\bar{\mathbf{k}}_{1},\bar{\mathbf{v}}_{1}),\cdots,(\bar{\mathbf{k}}_{\kappa},\bar{\mathbf{v}}_{\kappa})\}$ be a memory of size $\kappa$, where each $(\bar{\mathbf{k}}_{j},\bar{\mathbf{v}}_{j})$ is a representation of a chunk of length $n_c$. This memory can thus encode a sequence of maximum length $\kappa \cdot n_c$, computing each $(\bar{\mathbf{k}}_{j},\bar{\mathbf{v}}_{j})$ via Eqs. (\ref{eq:memory-attention-k}) and (\ref{eq:memory-attention-v}) over its corresponding chunk. A second approach is to organize the memory into a priority queue. A learned scoring function, which is often an auxiliary neural network, evaluates and assigns a priority score to each generated key-value pair. High-scoring pairs are inserted into the queue via a push operation. In this way, the memory actively maintains a dynamic collection of only the most valuable key-value pairs across the input sequence.

Although representing memory as a set of vectors is an intuitive design choice for Transformers, its capacity is bounded by the number of vectors. An alternative paradigm is \textbf{continuous memory}. This approach leverages function approximation, treating the sets of keys $\{\mathbf{k}_{1},\cdots,\mathbf{k}_{i-1}\}$ or values $\{\mathbf{v}_{1},\cdots,\mathbf{v}_{i-1}\}$ as a series of data points to be fitted by a continuous function. Instead of explicitly storing these vectors, the memory is parameterized by the functions themselves. A common method is to use a linear combination of simple basis functions to approximate the complex function defined by the discrete key or value vectors~\citep{martins-etal:2022former}.

It is also straightforward to use a short-term memory and a long-term memory simultaneously so that we can combine the merits of both. For example, we use a cache-based memory to capture local context, and use an efficient long-term memory that encodes the entire history to model long-range dependencies. This idea is also similar to that used in combining different sparse attention models as discussed in the previous subsection.

\subsubsection{Retrieval-based Methods}

\noindent So far in this subsection, we have discussed approaches based on fixed-length models. It is also possible to develop efficient memory models by improving the efficiency of accessing memory, instead of strictly reducing memory capacity. One way to achieve this is to store the past key-value pairs in a database (call it a \textbf{vector database}), and to find the most similar key-value pairs when querying the database. To be more precise, given a query $\mathbf{q}$, we use the database to find a set of top-$p$ relevant key-value pairs (denoted by $\Omega_p$) by performing a similarity search based on the dot-product between the query and key vectors. The query $\mathbf{q}$ then attends to $\Omega_p$ as in standard self-attention models. The idea behind this method is to consider only a small number of elements that contribute most to the attention result. Therefore, the model is essentially a sparse attention model which is computationally efficient. Another advantage of this method is that it allows for fast similarity search over a very large set of vectors because of the highly optimized implementation of vector databases. Framing memory as a retrieval system falls under the broader paradigm of \textbf{retrieval-augmented generation} (\textbf{RAG}) or retrieval-augmented approaches. It provides a scalable framework for incorporating external memory into neural architectures like Transformers~\citep{guu-etal:2020retrieval,lewis-etal:2020retrieval,wu-etal:2021memorizing}.

\subsection{Low-dimensional Models}
\label{sec:transformer-low-dim-models}

\noindent In many practical applications, Transformers are ``high-dimensional'' models. This is not only because the input and/or output data is in high-dimensional spaces, but also because some of the intermediate representations of the data in the model are high-dimensional.  As discussed in Section~\ref{sec:transformer-sparse-attention}, this high dimensionality arises in part from computing the weighted sum of value vectors, as in Eq. (\ref{eq:low-dim-self-attetention-decoder-as}) (repeated here for convenience)
\begin{eqnarray}
\mathrm{Att}_{\mathrm{self}}(\mathbf{S}) & = & \mathbf{A} \mathbf{V} \label{eq:low-dim-self-attetention-decoder-as}
\end{eqnarray}

\noindent and computing the attention matrix itself, as in Eq. (\ref{eq:low-dim-self-attetention-decoder-softmax})
\begin{eqnarray}
\mathbf{A} & = & \mathrm{Softmax}(\frac{\mathbf{Q} \mathbf{K}^\top}{\sqrt{d}} + \mathbf{M}), \label{eq:low-dim-self-attetention-decoder-softmax}
\end{eqnarray}

\noindent which involves large matrix multiplications $\mathbf{Q} \mathbf{K}^\top$ and $\mathbf{A} \mathbf{V}$ when the length $n$ and the hidden dimensionality $d$ are large.

The $\mathbf{A} \mathbf{V}$ and $\mathbf{Q} \mathbf{K}^\top$ operations scale with a time complexity of $O(n^2 \cdot d)$ and a space complexity of $O(n^2 + n \cdot d)$. While previous approaches have mitigated this complexity using sparse models, this subsection focuses on methods that approximate these operations via dense computation. One straightforward idea is to project $\mathbf{Q}$, $\mathbf{K}$, and $\mathbf{V}$ into smaller matrices, thereby reducing the computational burden of matrix multiplication. Since $\mathbf{Q}$, $\mathbf{K}$, and $\mathbf{V}$ all reside in $\mathbb{R}^{n \times d}$, this can be achieved by reducing either the length dimension $n$, the hidden dimension $d$, or both.

\subsubsection{Reducing $n$}

\noindent Note that the output $\mathrm{Att}_{\mathrm{self}}(\mathbf{S})$ is required to be an $n \times d$ matrix, and so we cannot reduce the number of queries. We instead consider reducing the number of keys and values. Suppose $n'$ is an integer with $n' < n$, and $\mathbf{K}$ and $\mathbf{V}$ can be transformed into $n' \times d$ matrices $\mathbf{K}'$ and $\mathbf{V}'$ in some way. We can obtain a ``smaller'' model simply by replacing $\mathbf{K}$ and $\mathbf{V}$ with $\mathbf{K}'$ and $\mathbf{V}'$, giving
\begin{eqnarray}
\mathrm{Att}_{\mathrm{self}}(\mathbf{S}) & = & \mathbf{A} {\color{red} \mathbf{V}'} \label{eq:low-dim-self-attetention-decoder-smaller-matrix-as} \\
\mathbf{A} & = & \mathrm{Softmax}(\frac{\mathbf{Q} [{\color{red} \mathbf{K}'}]^\top}{\sqrt{d}} + \mathbf{M}). \label{eq:low-dim-self-attetention-decoder-smaller-matrix-softmax}
\end{eqnarray}

\noindent This model is in the standard form of self-attention, but has lower time and space complexities, that is, $O(n' \cdot n \cdot d) < O(n^2 \cdot d)$ and $O(n' \cdot n + n' \cdot d) < O(n^2 + n \cdot d)$. If $n' \ll n$, the resulting model will be linear with $n$.

The key problem here is how to obtain $\mathbf{K}'$ and $\mathbf{V}'$ in a way that retains much of the information in $\mathbf{K}$ and $\mathbf{V}$. There are several ways to do so. One simple method is to select the keys and values that are thought to be important. The importance of a key (or value) can be computed in terms of some computationally inexpensive measure. For example, we can sample a small number of query-key dot-products and estimate the importance of a key by collecting these dot-product results.

Although straightforward, the above method still relies on sparse operations like sampling and aggregation. Alternatively, dense computation can be used to project $\mathbf{K}$ and $\mathbf{V}$ down to $\mathbf{K}'$ and $\mathbf{V}'$. A common approach employs CNNs~\citep{liu-etal:2018generating}. Let $\mathrm{Conv}(\cdot)$ denote a 1D convolution operation sliding along the $n$ dimension. $\mathbf{K}'$ is then computed as:
\begin{eqnarray}
\mathbf{K}' & = & \mathrm{Conv}(\mathbf{K},\mathbf{W}_{c},\mathrm{size}_r,\mathrm{stride}),
\end{eqnarray}

\noindent where $\mathbf{W}_{c}$ is the parameter matrix of the filters, $\mathrm{size}_r$ is the size of the receptive field, and $\mathrm{stride}$ is the step size of the convolution. In general, we can achieve a high compression rate by choosing large values for $\mathrm{size}_r$ and $\mathrm{stride}$. Likewise, we can compute $\mathbf{V}'$ using another convolutional function. It is worth noting that, if the parameter $n'$ is fixed for all samples, compression of $\mathbf{K}$ and $\mathbf{V}$ along the length dimension is essentially the same as the fixed-length memory model as described in the preceding subsection. The methods presented here are more general and could be applied to variable-length memories.

We might also be tempted to model the attention function by considering the attention matrix $\mathbf{A}$ as a high-dimensional representation of data and then applying conventional dimensionality reduction methods. Empirically, in many problems, $\mathbf{A}$ (or more precisely $\mathbf{Q} \mathbf{K}^\top$) is found to be a low-rank matrix. In this case, we can compress $\mathbf{A}$ while retaining as much information as possible. There are many ways to do so. For example, we might use a product of smaller matrices as an approximation to $\mathbf{A}$ via the SVD technique. However, this introduces computational overhead in using SVD compared with the standard attention model. A simpler idea is to directly transform $\mathbf{K}$ and $\mathbf{V}$ into smaller-sized matrices via linear mappings, given by
\begin{eqnarray}
\mathbf{K}' & = & \mathbf{U}^{k} \mathbf{K}, \\
\mathbf{V}' & = & \mathbf{U}^{v} \mathbf{V},
\end{eqnarray}

\noindent where $\mathbf{U}^{k} \in \mathbb{R}^{n' \times n}$ and $\mathbf{U}^{v} \in \mathbb{R}^{n' \times n}$ are parameter matrices. Clearly, this leads to a model which is equivalent to that described in Eqs. (\ref{eq:low-dim-self-attetention-decoder-smaller-matrix-as}) and (\ref{eq:low-dim-self-attetention-decoder-smaller-matrix-softmax}). While highly intuitive, this approach has been proven to yield a sufficiently small approximation error $\epsilon$, provided that $n'$ scales linearly with $d/\epsilon^2$~\citep{wang-etal:2020linformer}.

\subsubsection{Reducing $d$}

\noindent Another approach to reducing computational complexity is to reduce the dimensionality $d$. One of the simplest methods is to project all queries and keys onto a $d'$-dimensional space ($d' < d$), and to compute the dot-product between query-key pairs in the new space. For modeling, we only need to replace $\mathbf{Q} \in \mathbb{R}^{n \times d}$ and $\mathbf{K} \in \mathbb{R}^{n \times d}$ with new representations $\mathbf{Q}' \in \mathbb{R}^{n \times d'}$ and $\mathbf{K}' \in \mathbb{R}^{n \times d'}$. We can easily modify Eq. (\ref{eq:low-dim-self-attetention-decoder-softmax}) to use $\mathbf{Q}'$ and $\mathbf{K}'$ in computing the attention matrix
\begin{eqnarray}
\mathbf{A} & = & \mathrm{Softmax}(\frac{{\color{red} \mathbf{Q}'} [{\color{red} \mathbf{K}'}]^\top}{\sqrt{d'}} + \mathbf{M}).
\end{eqnarray}

\noindent $\mathbf{Q}'$ and $\mathbf{K}'$ are given by
\begin{eqnarray}
\mathbf{Q}' & = & \mathbf{Q} \mathbf{U}^{q}, \\
\mathbf{K}' & = & \mathbf{K} \mathbf{U}^{k},
\end{eqnarray}

\noindent where $\mathbf{U}^{q} \in \mathbb{R}^{d \times d'}$ and $\mathbf{U}^{k} \in \mathbb{R}^{d \times d'}$ are parameter matrices of linear transformations.

It is also possible to exploit \textbf{kernel methods} to obtain an efficient dot-product attention model. The basic idea is to map data points (represented as vectors) from one space to another, transforming a problem that is difficult in the original space into one that is more tractable in the new space. The ``kernel trick'' allows us to compute such inner products implicitly via a kernel function, without explicitly constructing the mapping $\phi(\cdot)$\footnote{In mathematical analysis, the inner product is a generalized notion of the dot-product, typically denoted by $\langle \cdot,\cdot \rangle$. A formal definition requires that $\langle \cdot,\cdot \rangle$ satisfies several properties in a vector space. Although the inner product takes different forms in different contexts, in Euclidean space $\mathbb{R}^{d}$, it is equivalent to the dot-product. That is, given two vectors $\mathbf{a} \in \mathbb{R}^{d}$ and $\mathbf{b} \in \mathbb{R}^{d}$, we have
\begin{eqnarray}
\langle \mathbf{a}, \mathbf{b}\rangle & = & \mathbf{a} \cdot \mathbf{b} \nonumber \\
& = & \sum_{i=1}^{d} a_i \cdot b_i
\end{eqnarray}
}. Such inner products in feature space are commonly represented by kernel functions, denoted as $K(\cdot,\cdot)$.

It is interesting to approximate $\mathbf{A}$ in a fashion analogous to $K(\cdot,\cdot)$ in kernel methods. To illustrate, note in Eq. (\ref{eq:low-dim-self-attetention-decoder-softmax}) $\mathbf{A}$ represents normalized attention weights. The numerator can be written in the form
\begin{eqnarray}
\widetilde{\mathbf{A}} & = & \mathrm{Mask}(\exp(\frac{\mathbf{Q} \mathbf{K}^\top}{\sqrt{d}})).
\end{eqnarray}

\noindent Here $\mathrm{Mask}(\cdot)$ is a function which has the same effect as using the additive masking variable $\mathbf{M}$. Then, $\mathbf{A}$ can be expressed as
\begin{eqnarray}
\mathbf{A} & = & \mathbf{D}^{-1} \widetilde{\mathbf{A}},
\end{eqnarray}

\noindent where $\mathbf{D}$ is an $n \times n$ diagonal matrix. Each entry of the main diagonal is the sum of the entries of the corresponding row in $\widetilde{\mathbf{A}}$, denoting the normalization factor of softmax. Substituting this equation into Eq. (\ref{eq:low-dim-self-attetention-decoder-softmax}), we have
\begin{eqnarray}
\mathrm{Att}_{\mathrm{self}}(\mathbf{S}) & = & \mathbf{D}^{-1} \widetilde{\mathbf{A}} \mathbf{V}. \label{eq:attention-das-form}
\end{eqnarray}

In this model, $\widetilde{A}(i,j)$ can be viewed as a similarity function over all query-key pairs in a $d$-dimensional space. Here we assume that this function, which is in the form of the dot-product of vectors, can be approximated by a kernel function
\begin{eqnarray}
\widetilde{A}(i,j) & = & K(\mathbf{q}_i,\mathbf{k}_j) \nonumber \\
                   & = & \langle \phi(\mathbf{q}_i), \phi(\mathbf{k}_j) \rangle \nonumber .
\end{eqnarray}

\noindent $\phi(\cdot)$ is a mapping from $\mathbb{R}^{d}$ to $\mathbb{R}^{d'}$. We can represent the queries and keys in the following form
\begin{eqnarray}
\mathbf{Q}' & = & \phi(\mathbf{Q}) \nonumber \\
& = & \begin{bmatrix} \phi(\mathbf{q}_1) \\ \vdots \\ \phi(\mathbf{q}_n) \end{bmatrix}, \\
\mathbf{K}' & = & \phi(\mathbf{K}) \nonumber \\
& = & \begin{bmatrix} \phi(\mathbf{k}_1) \\ \vdots \\ \phi(\mathbf{k}_n) \end{bmatrix}.
\end{eqnarray}

Then, we develop a kernelized attention model by approximating the attention weight $\alpha_{i,j}$ in the form
\begin{eqnarray}
\alpha_{i,j} & \approx & \frac{\phi(\mathbf{q}_i) \phi(\mathbf{k}_j)^\top}{\sum_{j'=1}^{n}\phi(\mathbf{q}_i) \phi(\mathbf{k}_{j'})^\top}.
\end{eqnarray}

The key idea behind this kernelized attention model is that we can remove the softmax function if the queries and keys are mapped to a new space. Using this approximation, the $i$-th output vector of the attention model (i.e., the $i$-th row vector of $\mathrm{Att}_{\mathrm{self}}(\mathbf{S})$) is given by
\begin{eqnarray}
\mathbf{c}_{i} & = & \sum_{j=1}^{n} \alpha_{i,j} \cdot \mathbf{v}_j \nonumber \\
               & \approx & \sum_{j=1}^{n} \Big( \frac{\phi(\mathbf{q}_i) \phi(\mathbf{k}_j)^\top}{\sum_{j'=1}^{n}\phi(\mathbf{q}_i) \phi(\mathbf{k}_{j'})^\top} \cdot \mathbf{v}_j \Big) \nonumber \\
               & = & \frac{\sum_{j=1}^{n} \phi(\mathbf{q}_i) \phi(\mathbf{k}_j)^\top \mathbf{v}_j}{\sum_{j'=1}^{n}\phi(\mathbf{q}_i) \phi(\mathbf{k}_{j'})^\top} \nonumber \\
               & = & \frac{\phi(\mathbf{q}_i) {\color{red} (} \sum_{j=1}^{n} \phi(\mathbf{k}_j)^\top \mathbf{v}_j {\color{red} )}}{\phi(\mathbf{q}_i) {\color{red} (} \sum_{j'=1}^{n} \phi(\mathbf{k}_{j'})^\top {\color{red} )}}. \label{eq:context-vector-kernel-methods}
\end{eqnarray}

\noindent Here $\phi(\mathbf{q}_i) \in \mathbb{R}^{1 \times d'}$, while $\phi(\mathbf{k}_j)^\top \mathbf{v}_j \in \mathbb{R}^{d' \times d}$. Therefore, the inner term $\sum_{j=1}^n \phi(\mathbf{k}_j)^\top \mathbf{v}_j$ is a $d' \times d$ matrix, and the final product yields a vector in $\mathbb{R}^{1 \times d}$.

Although the equation appears a bit complicated, the idea is simple: instead of attending the query to all keys to obtain the attention weight $\alpha_{i,j}$, we can compute the sum of the multiplications $\sum_{j=1}^{n} \phi(\mathbf{k}_j)^\top \mathbf{v}_j \in \mathbb{R}^{d' \times d}$ and then multiply it with the kernelized query $\phi(\mathbf{q}_i)$. Returning to the notation used in Eq. (\ref{eq:attention-das-form}), we define the $i$-th diagonal entry of $\mathbf{D}$ to be $\phi(\mathbf{q}_i) \sum_{j'=1}^{n} \phi(\mathbf{k}_{j'})^\top$. Then, the attention model can be re-expressed in the form
\begin{eqnarray}
\mathrm{Att}_{\mathrm{self}}(\mathbf{S}) & = & \mathbf{D}^{-1} \phi(\mathbf{Q}) \phi(\mathbf{K})^\top \mathbf{V} \nonumber \\
& = & \mathbf{D}^{-1} \mathbf{Q}' \mathbf{K}'^\top \mathbf{V} \nonumber \\
& = & \mathbf{D}^{-1} {\color{red} \big(} \mathbf{Q}' {\color{red}(}\mathbf{K}'^\top \mathbf{V} {\color{red})} {\color{red}\big)}.
\end{eqnarray}

\noindent Here we change the order of computation from left-to-right to right-to-left using parentheses. Given that $\mathbf{Q}' \in \mathbb{R}^{n \times d'}$ and $\mathbf{K}' \in \mathbb{R}^{n \times d'}$, this model has time and space complexities of $O(n \cdot d \cdot d')$ and $O(n \cdot d + n \cdot d' + d \cdot d')$, respectively. Therefore, the model is linear with respect to the sequence length $n$, and is sometimes called the \textbf{linear attention model}. One computational advantage of this model is that we need only compute the multiplication $\mathbf{K}'^\top \mathbf{V}$ (i.e., $\sum_{j=1}^{n} \phi(\mathbf{k}_j)^\top \mathbf{v}_j$) and the corresponding normalization factor (i.e., $\sum_{j'=1}^{n} \phi(\mathbf{k}_{j'})^\top$) once. The results can then be used for any query~\citep{katharopoulos-etal:2020transformers}. The model needs to maintain $\sum_{j=1}^{n} \phi(\mathbf{k}_j)^\top \mathbf{v}_j$ and $\sum_{j'=1}^{n} \phi(\mathbf{k}_{j'})^\top$ and update them when new key and value vectors arrive.

Still, there are several limitations regarding this kernelized model, for example, how to develop the feature map $\phi(\cdot)$ to obtain a good approximation to the standard attention model. Interested readers may refer to \citet{choromanski-etal:2020rethinking}'s work for more details.

A second idea for reducing $d$ is to take sub-space models, in which a problem in a $d$-dimensional space is transformed into sub-problems in lower-dimensional spaces, and the solution to the original problem is approximated by some combination of the solutions to these sub-problems. In a general sub-space model, a $d$-dimensional key vector $\mathbf{k}$ can be mapped into a set of $d'$-dimensional vectors $\{ \mathbf{K}'_{1},\cdots,\mathbf{K}'_{\eta} \}$. To simplify modeling, we can do this by vector segmentation, that is, we segment $\mathbf{k}$ into $\eta$ sub-vectors, each having $d' = \frac{d}{\eta}$ dimensions. We can transform all query and value vectors in the same way. Then, the attention model is applied in each of these sub-spaces.

This method, however, does not reduce the total amount of computation. As presented in \citet{lample-etal:2019large}'s work, we can instead approximate the dot-product attention over a set of key-value pairs by considering top-$p$ candidates in each sub-space. More precisely, we find $p$-best key-value pairs in each sub-space, which is computationally cheaper. The Cartesian product of these $p$-best key sets consists of $p^\eta$ product keys. Likewise, we obtain $p^\eta$ product values. The remaining work is simple: the $d$-dimensional queries attend to these $d$-dimensional product keys and values. An interesting difference between this sub-space model and the $d$-dimensional space model is that the generated product keys and values may be different from any of the original key-values $\{(\mathbf{k}_{1},\mathbf{v}_{1}),\cdots,(\mathbf{k}_{i-1},\mathbf{v}_{i-1})\}$. This provides a way for learning new representations of the past information.

So far we have discussed approaches to dimensionality reduction along either the $n$ or $d$ dimension. It is straightforward to combine them to develop a ``lower-dimensional'' model. As an example, suppose that we have the $n \to n'$ reduction for keys and values, and the $d \to d'$ reduction for queries and keys. The model takes the form
\begin{eqnarray}
\mathrm{Att}_{\mathrm{self}}(\mathbf{S}) & = & \mathbf{A} {\color{red} \mathbf{V}'} \nonumber , \\
\mathbf{A} & = & \mathrm{Softmax}(\frac{{\color{red} \mathbf{Q}'} {\color{red} \mathbf{K}'}^\top}{\sqrt{d'}} + \mathbf{M}),
\end{eqnarray}

\noindent where $\mathbf{Q}' \in \mathbb{R}^{n \times d'}$, $\mathbf{K}' \in \mathbb{R}^{n' \times d'}$, and $\mathbf{V}' \in \mathbb{R}^{n' \times d'}$ are low-dimensional representations for queries, keys and values. As usual, we can easily obtain these representations through the linear mappings of $\mathbf{Q}$, $\mathbf{K}$ and $\mathbf{V}$. The time and space complexities of this model are $O(n' \cdot n \cdot d')$ and $O(n' \cdot n + n' \cdot d')$.

\subsection{Parameter and Activation Sharing}
\label{sec:efficient-methods-parameter-sharing}

\noindent Redundancy is common to most large-scale neural networks. As a result, many of these models are over-parameterized, making training and inference less efficient. One common approach to reducing redundancy is to simplify the model by removing unnecessary components. For example, we can either prune a complex model or share sub-modules across different parts of the model to obtain a more compact model. In this subsection, we discuss methods for sharing parameters and intermediate states in Transformer models. We leave the discussion of model transfer and pruning to Section~\ref{sec:transformer-model-transfer-and-pruning}.

Shared-parameter architectures are widely used in neural network-based systems. Well-known examples include CNNs and RNNs, where the same set of parameters (or layers) is applied across different regions of the input. This produces a large neural network, parts of which have the same architecture and the same shared parameters. For Transformers as well as other sequence models, the sharing mechanism can be applied to different levels of modeling. A simple example is embedding sharing. In machine translation, a typical strategy for dealing with words in two languages is to develop two separate embedding models. Alternatively, one can use a single embedding model for both languages. The parameters of the model are then learned during the training of both the source-side and target-side networks. Such a strategy is also often adopted in multilingual sequence models, such as language models that are able to deal with text in many different languages.

For multi-layer neural networks, a popular method is layer-wise sharing. Suppose there is a stack of layers, all of which have the same form
\begin{eqnarray}
\mathbf{S}^{l} & = & \mathrm{Layer}(\mathbf{S}^{l-1};\theta^{l}).
\end{eqnarray}

\noindent We can tie the parameters for some or all of these layers. For example, given a set of layers $\{l_1, l_2,\cdots,l_n\}$, we enforce the constraint $\theta^{l_1} = \theta^{l_2} = \cdots = \theta^{l_n}$, so that we can obtain a smaller model and the optimization of the model can be easier. In practice, this shared-layer model is highly advantageous if many layers are involved, because we can repeat the same process many times to construct a very deep neural network~\citep{dehghani-etal:2018universal}. For example, sharing a single FFN sub-layer across all encoder layers is found to be effective in reducing the redundancy in machine translation systems~\citep{pires-etal:2023one}.

For Transformers, sharing can also be performed in multi-head attention. An example of this is \textbf{multi-query attention}~\citep{shazeer:2019fast}. Recall from Section~\ref{sec:multi-head-self-attention} that the output of a head $h$ in standard multi-head self-attention can be written as
\begin{eqnarray}
\mathbf{C}_{h}^{\mathrm{head}} & = & \mathrm{Att}_{\mathrm{qkv}}(\mathbf{S}_{h}^{q}, \mathbf{S}_{h}^{k}, \mathbf{S}_{h}^{v}) \nonumber \\
                               & = & \mathrm{Att}_{\mathrm{qkv}}(\mathbf{S} \mathbf{W}_{h}^{q}, \mathbf{S} \mathbf{W}_{h}^{k}, \mathbf{S} \mathbf{W}_{h}^{v}).
\end{eqnarray}

\noindent Here $\mathbf{S}_{h}^{q} = \mathbf{S} \mathbf{W}_{h}^{q}$, $\mathbf{S}_{h}^{k} = \mathbf{S} \mathbf{W}_{h}^{k}$, and $\mathbf{S}_{h}^{v} = \mathbf{S} \mathbf{W}_{h}^{v}$ are the queries, keys, and values, which are obtained by linearly transforming the input $\mathbf{S}$ with distinct parameter matrices $\mathbf{W}_{h}^{q}$, $\mathbf{W}_{h}^{k}$, and $\mathbf{W}_{h}^{v}$. In multi-query attention, we share the same keys and values across all the heads, but use different queries for different heads. The form of this model is given by
\begin{eqnarray}
\mathbf{C}_{h}^{\mathrm{head}} & = & \mathrm{Att}_{\mathrm{qkv}}(\mathbf{S} \mathbf{W}_{h}^{q}, \mathbf{S} \mathbf{W}_{0}^{k}, \mathbf{S} \mathbf{W}_{0}^{v}).
\end{eqnarray}

\noindent Because the keys $\mathbf{S} \mathbf{W}_{0}^{k}$ and values $\mathbf{S} \mathbf{W}_{0}^{v}$ are independent of $h$, they are computed and stored only once, rather than redundantly for each head. This yields significant computational and memory bandwidth savings during inference, especially when the number of attention heads is large. Due to its efficiency, multi-query attention (and its variant, grouped-query attention) has been successfully integrated into modern large language models, including Llama 2~\citep{touvron-etal:2023llama2} and Falcon\footnote{https://falconllm.tii.ae/index.html}.

By extending the idea of sharing to more general situations, intermediate states can, in principle, be shared across a neural network. Reusing neuron activations, for instance, allows sub-components to be evaluated multiple times efficiently. In Transformers, this method can be applied directly to the self-attention matrices. Studies have empirically shown that attention maps across adjacent layers are often highly correlated in certain NLP tasks~\citep{xiao-etal:2019sharing}. As a result, a computationally efficient strategy is to compute the dense attention map once and reuse it across subsequent layers.

More broadly, the sharing mechanism can be viewed as a process by which we reuse previously computed results rather than recomputing them on the fly. It is thus possible to reuse the information across different runs of a neural network. A related example is \textbf{reversible residual networks}, in which activations of one layer can be recovered from the activations of the following layer~\citep{gomez-etal:2017reversible}. Hence we only keep the output of the latest layer in the forward pass. Then, in the backward pass of training, we reconstruct the output of each layer from its successor. One advantage of this reversible design is that the information produced in the forward pass is shared implicitly, and the model is memory-efficient~\citep{kitaev-etal:2020reformer}.

\subsection{Alternatives to Self-Attention}

\noindent We have seen that the use of self-attention is one of the main sources of high computational and memory costs in Transformers. It is natural to wonder if there are efficient alternatives to self-attention models. Here, we briefly present some Transformer variants in which self-attention sub-layers are replaced with other types of neural architectures.

\subsubsection{CNNs as a Replacement for Self-Attention}

\noindent CNNs are simple and widely used neural networks, and are considered potential alternatives to self-attention models. To apply CNNs to Transformers, all we need is to construct a convolutional sub-layer to replace the self-attention sub-layer in a Transformer block. While a CNN filter has a restricted receptive field and thus takes inputs from a ``local'' context window, large contexts can be easily modeled by stacking multiple convolutional sub-layers. One key advantage of CNNs is that their computational complexity scales linearly with the sequence length $n$, compared to the quadratic scaling of self-attention models. Furthermore, the availability of highly optimized CNN implementations makes them easy to apply to sequence modeling in practical systems. To further improve memory efficiency, we can utilize lightweight CNN variants, such as depthwise CNNs~\citep{wu-etal:2018pay}
\footnote{
Recall that in CNNs, a filter (or a set of filters) maps the input variables within its receptive field into an output variable (or a set of output variables) via a linear mapping. Suppose the input and output of a problem are represented as sequences of feature vectors. Given a standard filter with a $k \times d$ receptive field, we slide it along the sequence. At each step, the filter takes $k \times d$ input features and produces a single output feature. This procedure is typically expressed as:
\begin{eqnarray}
y & = & \mathrm{ReduceSum}(\mathbf{x} \odot \mathbf{W}),
\end{eqnarray}

\noindent where $\mathbf{x} \in \mathbb{R}^{k \times d}$ is the matrix representation of the input, $y \in \mathbb{R}$ is the output feature, and $\mathbf{W} \in \mathbb{R}^{k \times d}$ is the weight matrix. The function $\mathrm{ReduceSum}(\cdot)$ computes the sum of all element-wise products between $\mathbf{x}$ and $\mathbf{W}$. If we want the input and output to have the same number of features $d$, we must apply $d$ distinct filters, resulting in a total of $d \cdot (k \cdot d) = d^2 \cdot k$ parameters.
In depthwise CNNs, we decouple the spatial convolution from the feature dimensions. Instead of applying full $k \times d$ filters, each of the $d$ input feature channels is convolved independently with its own 1D spatial filter of size $k$. Thus, the number of unique parameters is drastically reduced to $d \cdot k$ (with each of the $d$ channels corresponding to a dedicated filter containing $k$ unique parameters).

}.

\subsubsection{Linear Attention}

\noindent As with many practical approaches to sequence modeling, there is also considerable interest in developing linear models in order to speed up the processing of long sequences. While there are many ways to define a linear model, one general form that is commonly used in sequence models is
\begin{eqnarray}
\mathbf{z}_{i} & = & f(a \cdot \mathbf{z}_{i-1} + b \cdot \mathbf{s}_{i}). \label{eq:linear-base-model-for-transformer}
\end{eqnarray}

\noindent Here $\mathbf{s}_i$ represents some intermediate states of the model at step $i$, and $\mathbf{z}_{i}$ represents the summary of the past states up to step $i$. It is easy to see that this is a recurrent model: the output at step $i$ depends only on the input at the current step and the output at the previous step. As with the popular design choices in neural network-based systems, the linear part is followed by a transformation $f(\cdot)$ which can be either an activation function or a feedforward neural network. Note that, Eq. (\ref{eq:linear-base-model-for-transformer}) defines a standard linear model only if $f(\cdot)$ is a linear function. The use of $f(\cdot)$ gives greater flexibility in modeling the problem, although the term \textit{linear model} may not be applied if $f(\cdot)$ chooses a non-linear form.

The above formula describes a linearly structured model which can be seen as an instance of a general family of mathematical models. Typically, it can be represented as a chain structure, or an ordered set of nodes. The model repeats the same computation process from the first node to the last, each time taking the information from the current and previous steps and producing an output vector that is used in the following time steps. As a result, the space and time cost of the model scales linearly with the length of the chain.

We can extend Eq. (\ref{eq:linear-base-model-for-transformer})  to a standard RNN model by simply applying a linear transformation of the current input and the previous state, that is, $\mathbf{z}_{i}  = f(\mathbf{z}_{i-1} \cdot \mathbf{W}_{z} + \mathbf{s}_{i} \cdot \mathbf{W}_{s})$. It is thus straightforward to apply RNNs and their variants to Transformer to obtain a hybrid model. For example, we can use LSTM and GRUs in building some of the Transformer layers to combine the merits of both recurrent models and self-attentive models~\citep{chen-etal:2018best}. As conventional recurrent models are well known, we skip the discussion of them here.

In fact, we may be more interested in developing linear attention models, so that we can obtain an efficient system, while still retaining the benefit of globally attentive sequence modeling. Part of the difficulty in doing this is that the form of self-attention is not linear. Let us take a moment to see how this difficulty arises. Recall that the result of self-attention can be written in the following form
\begin{eqnarray}
\mathrm{Att}_{\mathrm{self}} & = & \mathbf{A} \cdot \mathbf{V} \nonumber \\
                             & = & \psi(\mathbf{Q} \cdot \mathbf{K}^\top) \cdot \mathbf{V}.
\end{eqnarray}

\noindent Here $\psi(\cdot)$ is a function that is composed of scaling, exponentiation, masking, and normalization operations (i.e., $\psi(\mathbf{a}) = \mathrm{Normalize}(\mathrm{Mask}(\exp(\frac{\mathbf{a}}{\sqrt{d}})))$). Because $\psi(\cdot)$ is a complex non-linear function, there is no straightforward way to simplify the computation directly, and we typically need to calculate the two matrix multiplications separately (one inside $\psi(\cdot)$ and one outside $\psi(\cdot)$). As a consequence, we need to store all the key-value pairs explicitly, and visit each of them given a query. Not surprisingly, this leads to a model whose computational cost grows quadratically with the sequence length $n$.

Although in self-attention keys and values are coupled, they are used in separate steps. An elegant solution decouples the query from the key-value interaction so that the context information can be encoded independently of any specific query. The key insight is to bypass the non-linear softmax by applying a feature map $\phi(\cdot)$ to the queries and keys. As introduced in Section~\ref{sec:transformer-low-dim-models}, we transform $\mathbf{Q}$ and $\mathbf{K}$ into $\mathbf{Q}' = \phi(\mathbf{Q}) \in \mathbb{R}^{n \times d'}$ and $\mathbf{K}' = \phi(\mathbf{K}) \in \mathbb{R}^{n \times d'}$. The attention mechanism can then be reformulated as:
\begin{eqnarray}
\mathrm{Att}_{\mathrm{self}} & \equiv & \psi'(\mathbf{Q}' \cdot \mathbf{K}'^\top) \cdot \mathbf{V}  \nonumber \\
                             & = & \mathbf{D}^{-1} (\mathbf{Q}' \cdot \mathbf{K}'^\top) \cdot \mathbf{V} \nonumber \\
                             & = & \mathbf{D}^{-1} \Big( \mathbf{Q}' \cdot {\color{red} \big(} \mathbf{K}'^\top \cdot \mathbf{V} {\color{red} \big)} \Big),
\end{eqnarray}

\noindent where $\mathbf{D}$ is the diagonal normalization matrix, and $\psi'(\mathbf{Z}) = \mathbf{D}^{-1} \mathbf{Z}$\footnote{If $\mathbf{D}$ is a diagonal matrix, we occasionally denote  $\mathbf{D}^{-1} \mathbf{A}$ as $\frac{\mathbf{A}}{\mathbf{D}}$.}. In this transformed space, the query-key dot product does not require exponential softmax normalization. It corresponds to a normalization by multiplying with $\mathbf{D}^{-1}$. Because the core operation is now pure matrix multiplication, we can exploit matrix associativity to change the order of computation.

This yields a powerful autoregressive formulation: keys and values are first aggregated via $\mathbf{K}'^\top \cdot \mathbf{V}$, and then queries attend to this global representation. Given that $\mathbf{K}'^\top \cdot \mathbf{V} = \sum_{j=1}^{n} \mathbf{k'}_{j}^\top \cdot \mathbf{v}_{j}$, we can write $\mathbf{K}'^\top \cdot \mathbf{V}$ in the form of Eq. (\ref{eq:linear-base-model-for-transformer}), as follows
\begin{eqnarray}
\boldsymbol{\mu}_j & = & \boldsymbol{\mu}_{j-1} + \mathbf{k'}_{j}^\top \cdot \mathbf{v}_{j}. \label{eq:linear-attention-mu}
\end{eqnarray}

\noindent Here $\boldsymbol{\mu}_j \in \mathbb{R}^{d' \times d}$ is a variable that adds $\mathbf{k'}_{j}^\top \cdot \mathbf{v}_{j}$ at a time. Likewise, we can define another variable $\boldsymbol{\nu}_j \in \mathbb{R}^{d'}$
\begin{eqnarray}
\boldsymbol{\nu}_j & = & \boldsymbol{\nu}_{j-1} + \mathbf{k'}_{j}^\top. \label{eq:linear-attention-nu}
\end{eqnarray}

Then, the output of self-attention for the $j$-th query can be written as (see also Eq. (\ref{eq:context-vector-kernel-methods}))
\begin{eqnarray}
\mathrm{Att}_{\mathrm{self},j} & = & \frac{\mathbf{q}'_{j} \cdot \boldsymbol{\mu}_n}{\mathbf{q}'_{j} \cdot \boldsymbol{\nu}_n}.
\end{eqnarray}

Clearly, this yields a linear-time model, because $\boldsymbol{\mu}_n$ and $\boldsymbol{\nu}_n$ are linear with respect to $n$. In simple implementations of this model, only $\boldsymbol{\mu}_j$ and $\boldsymbol{\nu}_j$ are kept. Each time a new query is encountered, we update $\boldsymbol{\mu}_j$ and $\boldsymbol{\nu}_j$ using Eqs. (\ref{eq:linear-attention-mu}) and (\ref{eq:linear-attention-nu}), and then compute $\mathrm{Att}_{\mathrm{self},j} = \frac{\mathbf{q}'_{j} \cdot \boldsymbol{\mu}_j}{\mathbf{q}'_{j} \cdot \boldsymbol{\nu}_j}$\footnote{In autoregressive generation, we generate a sequence from left to right. In this case, we need not consider the keys and values for positions $>j$.}.

One straightforward extension to the linear attention model is to allow Eqs. (\ref{eq:linear-attention-mu}) and (\ref{eq:linear-attention-nu}) to combine different terms with different weights. For example, we can redefine $\boldsymbol{\mu}_j$ and $\boldsymbol{\nu}_j$ as
\begin{eqnarray}
    \boldsymbol{\mu}_j & = & a \cdot  \boldsymbol{\mu}_{j-1} + (1 - a) \cdot \mathbf{k'}_{j}^\top \cdot \mathbf{v}_{j}, \\
\boldsymbol{\nu}_j & = & a \cdot \boldsymbol{\nu}_{j-1} + (1 - a) \cdot \mathbf{k'}_{j}^\top
\end{eqnarray}

\noindent and train the parameter $a$ as usual. Also, we can treat $a$ as a gate and use another neural network to compute $a$~\citep{peng-etal:2021random}. Another model design is to add more terms to Eqs. (\ref{eq:linear-attention-mu}) and (\ref{eq:linear-attention-nu}) in order to provide a more expressive formulation of the linear attention approach~\citep{bello:2020lambdanetworks,schlag-etal:2021linear}.

We have seen a general idea of designing linear models for the attention mechanism. The key design choice of such models is to remove the softmax-based normalization, thereby yielding linear forms of representation based on various intermediate states of the models. This motivates several recently developed alternatives to self-attention in which efficient inference systems are developed on the basis of recurrent models of sequence modeling~\citep{peng-etal:2023rwkv,sun-etal:2023retentive}. While these systems have different architectures, the underlying models have a similar form, as described in Eq. (\ref{eq:linear-base-model-for-transformer}). Note that, by using the general formulation of recurrent models, we are not restricted to standard QKV attention. Instead we may give new meanings and forms to the queries, keys, and values.

The discussion here is also related to the memory models discussed in Section~\ref{sec:transformer-recurrent-and-memory-models}. From a memory perspective, the keys and values can be treated as encodings of the context. Therefore, in the linear attention model above we have a memory system in which two simple variables $\boldsymbol{\mu}_j$ and $\boldsymbol{\nu}_j$ are used to represent all the context information up to position $j$. This results in a fixed-length memory which is very useful in practice. There are also other linear approaches to encoding long sequences. For example, we can view the \textbf{moving average} model as an instance of Eq. (\ref{eq:linear-base-model-for-transformer}), and average a series of state vectors of a Transformer model, either weighted or unweighted.

\subsubsection{State-Space Models}

\noindent In control systems, \textbf{state-space models} (\textbf{SSMs}) are representations of a system whose input and output are related by some \textbf{state variables} (or states for short), and whose dynamics are described by first-order differential equations of these states. As a simple example, we consider a continuous-time linear time-invariant system which is given in the form of the state-space representation. We adopt a row-vector convention throughout this section, where all variables are represented as row vectors and linear transformations are applied on the right:
\begin{eqnarray}
\frac{d \mathbf{z}(t)}{d t} & = & \mathbf{z}(t) \cdot \mathbf{A} + \mathbf{s}(t) \cdot \mathbf{B}, \label{eq:ssm-state-equation} \\
\mathbf{o}(t) & = & \mathbf{z}(t) \cdot \mathbf{C} + \mathbf{s}(t) \cdot \mathbf{D}. \label{eq:ssm-output-equation}
\end{eqnarray}

\noindent Here, $\mathbf{s}(t)$, $\mathbf{o}(t)$, and $\mathbf{z}(t)$ represent the values of the input variable, output variable, and state variable at time $t$, respectively\footnote{We use boldface letters to emphasize that these variables are vectors.}. In a general setting, these variables may possess different dimensionalities. For simplicity, we assume $\mathbf{s}(t), \mathbf{o}(t) \in \mathbb{R}^{d}$ and $\mathbf{z}(t) \in \mathbb{R}^{d_z}$\footnote{In general state-space theory, these variables and matrices are defined over the field of complex numbers. Because models defined over the complex field generalize to real-valued state spaces, we restrict our discussion to the multi-dimensional real number field for simplicity.}. Eq. (\ref{eq:ssm-state-equation}) is called the \textbf{state equation}, where $\mathbf{A} \in \mathbb{R}^{d_z \times d_z}$ is the state matrix and $\mathbf{B} \in \mathbb{R}^{d \times d_z}$ is the input matrix. Eq. (\ref{eq:ssm-output-equation}) is called the \textbf{output equation}, where $\mathbf{C} \in \mathbb{R}^{d_z \times d}$ is the output matrix and $\mathbf{D} \in \mathbb{R}^{d \times d}$ is the feedforward matrix.

These equations describe a continuous mapping from the input $\mathbf{s}(t)$ to the output $\mathbf{o}(t)$ over time. They are, therefore, often used to deal with continuous time series data. To apply this model to sequence modeling, we need to modify the above equations to obtain a discrete form of the state-space representation. Suppose that $\{\mathbf{s}_0,\mathbf{s}_1,\cdots,\mathbf{s}_n\}$ is a sequence of input data points sampled from $\mathbf{s}(t)$ with time step $\Delta t$. Similarly, we define $\{\mathbf{z}_0,\mathbf{z}_1,\cdots,\mathbf{z}_n\}$ and $\{\mathbf{o}_0,\mathbf{o}_1,\cdots,\mathbf{o}_n\}$ as sequences of the state and output vectors. Given this notation, we now have a discretized version of the SSM, written as
\begin{eqnarray}
\mathbf{z}_t & = & \mathbf{z}_{t-1} \cdot \overline{\mathbf{A}} + \mathbf{s}_t \cdot \overline{\mathbf{B}}, \label{eq:ssm-state-equation-discrete} \\
\mathbf{o}_t & = & \mathbf{z}_t \cdot \overline{\mathbf{C}} + \mathbf{s}_t \cdot \overline{\mathbf{D}}. \label{eq:ssm-output-equation-discrete}
\end{eqnarray}

\noindent This formulation of the SSM defines an RNN with a residual connection. To be more precise, Eq. (\ref{eq:ssm-state-equation-discrete}) describes a recurrent unit that reads the input at step $t$ and the state at step $t-1$, without using any activation function. Eq. (\ref{eq:ssm-output-equation-discrete}) describes an output layer that sums the linear transformations of the state $\mathbf{z}_t$ and the residual mapping $\mathbf{s}_t$.

The parameters $\overline{\mathbf{A}}$, $\overline{\mathbf{B}}$, $\overline{\mathbf{C}}$, and $\overline{\mathbf{D}}$ can be induced from $\mathbf{A}$, $\mathbf{B}$, $\mathbf{C}$ and $\mathbf{D}$ in several different ways, depending on how Eq. (\ref{eq:ssm-state-equation}) is approximated by Eq. (\ref{eq:ssm-state-equation-discrete})\footnote{The discretization process can be interpreted as numerically solving the underlying differential equation. Note that Eq. (\ref{eq:ssm-state-equation}) is an ODE:
\begin{eqnarray}
\frac{d \mathbf{z}(t)}{d t} & = & g(\mathbf{z}(t),t), \label{eq:ssm-ode}
\end{eqnarray}

\noindent where
\begin{eqnarray}
g(\mathbf{z}(t),t) & = & \mathbf{z}(t) \cdot \mathbf{A} + \mathbf{s}(t) \cdot \mathbf{B}. \label{eq:ssm-ode-g}
\end{eqnarray}

There are many numerical methods to approximate the solution to an ODE. For example, the Euler method for solving ODEs can be expressed as (see Section~\ref{sec:transformer-numerical-methods})
\begin{eqnarray}
\mathbf{z}_t & = & \mathbf{z}_{t - 1} + \Delta t \cdot g(\mathbf{z}_{t-1},t). \label{eq:ssm-ode-euler-method}
\end{eqnarray}

Substituting Eq. (\ref{eq:ssm-ode-g}) into Eq. (\ref{eq:ssm-ode-euler-method}) yields
\begin{eqnarray}
\mathbf{z}_t & = & \mathbf{z}_{t - 1} + \Delta t (\mathbf{z}_{t - 1} \cdot \mathbf{A} + \mathbf{s}_{t} \cdot \mathbf{B}) \nonumber \\
             & = & \mathbf{z}_{t - 1} \cdot (\mathbf{I} + \Delta t \cdot \mathbf{A}) + \mathbf{s}_{t} \cdot (\Delta t \cdot \mathbf{B}).
\end{eqnarray}

This gives one of the simplest forms of the discretized state equations~\citep{gu-etal:2022ssms-blog}, that is,
\begin{eqnarray}
\overline{\mathbf{A}} & = & \mathbf{I} + \Delta t \cdot \mathbf{A}, \\
\overline{\mathbf{B}} & = & \Delta t \cdot \mathbf{B}
\end{eqnarray}

}.
One approach to time discretization is the \textbf{bilinear transform} (also known as Tustin's method). An alternative approach is to use the \textbf{Zero-Order-Hold} (\textbf{ZOH}) discretization. A detailed discussion of these approaches lies beyond the scope of this book, and we refer the interested reader to standard textbooks on control theory for further details~\citep{aastrom-and-wittenmark:2013computer}.

The recurrent form of Eq. (\ref{eq:ssm-state-equation-discrete}) makes it easy to compute the states and outputs over a sequence of discrete time steps. We can unroll $\mathbf{z}_t$ and $\mathbf{o}_t$ in a feedforward fashion

\vspace{0.5em}
\begin{center}
\begingroup
\renewcommand{\arraystretch}{1.2}
\begin{tabular}{l | l}
$\mathbf{z}_0 = \mathbf{s}_{0} \cdot \overline{\mathbf{B}}$
&
$\mathbf{o}_0 = \mathbf{s}_{0} \cdot \overline{\mathbf{B}} \cdot \overline{\mathbf{C}} + \mathbf{s}_0 \cdot \overline{\mathbf{D}}$
\\
$\mathbf{z}_1 = \mathbf{s}_{0} \cdot \overline{\mathbf{B}} \cdot \overline{\mathbf{A}} + \mathbf{s}_1 \cdot \overline{\mathbf{B}}$
&
$\mathbf{o}_1 = \mathbf{s}_{0} \cdot \overline{\mathbf{B}} \cdot \overline{\mathbf{A}} \cdot \overline{\mathbf{C}} + \mathbf{s}_{1} \cdot \overline{\mathbf{B}} \cdot \overline{\mathbf{C}} + \mathbf{s}_1 \cdot \overline{\mathbf{D}}$
\\
$\mathbf{z}_2 = \mathbf{s}_{0} \cdot \overline{\mathbf{B}} \cdot \overline{\mathbf{A}}^2 + \mathbf{s}_1 \cdot \overline{\mathbf{B}} \cdot \overline{\mathbf{A}} + \mathbf{s}_2 \cdot \overline{\mathbf{B}}$
&
$\mathbf{o}_2 = \mathbf{s}_{0} \cdot \overline{\mathbf{B}} \cdot \overline{\mathbf{A}}^2 \cdot \overline{\mathbf{C}} + \mathbf{s}_{1} \cdot \overline{\mathbf{B}} \cdot \overline{\mathbf{A}} \cdot \overline{\mathbf{C}} + $
\\
&
\hspace{2em} $\mathbf{s}_{2} \cdot \overline{\mathbf{B}} \cdot \overline{\mathbf{C}} + \mathbf{s}_2 \cdot \overline{\mathbf{D}}$
\\
......
&
......
\end{tabular}
\endgroup
\end{center}
\vspace{0.5em}

It is easy to write
\begin{eqnarray}
\mathbf{z}_t & = & \sum_{i=0}^{t} \mathbf{s}_{i} \cdot \overline{\mathbf{B}} \cdot \overline{\mathbf{A}}^{t-i}, \\
\mathbf{o}_t & = & \sum_{i=0}^{t} \mathbf{s}_{i} \cdot \overline{\mathbf{B}} \cdot \overline{\mathbf{A}}^{t-i} \cdot \overline{\mathbf{C}} + \mathbf{s}_{t} \cdot \overline{\mathbf{D}}. \label{eq:ssm-o-t-unrolled}
\end{eqnarray}

\noindent Clearly, the right-hand side of Eq. (\ref{eq:ssm-o-t-unrolled}) can be interpreted as the sum of a convolutional term and a linear term. Given that
\begin{eqnarray}
\sum_{i=0}^{t} \mathbf{s}_{i} \cdot \overline{\mathbf{B}} \cdot \overline{\mathbf{A}}^{t-i} \cdot \overline{\mathbf{C}} & = &
\begin{bmatrix} \mathbf{s}_0 & \mathbf{s}_1 & \cdots & \mathbf{s}_t \end{bmatrix} \cdot \nonumber \\
& & \begin{bmatrix} \overline{\mathbf{B}} \cdot \overline{\mathbf{A}}^{t} \cdot \overline{\mathbf{C}} & \overline{\mathbf{B}} \cdot \overline{\mathbf{A}}^{t - 1} \cdot \overline{\mathbf{C}} & \cdots & \overline{\mathbf{B}} \cdot \overline{\mathbf{C}} \end{bmatrix}^\top, \label{eq:ssm-convolutional-part}
\end{eqnarray}

\noindent we define a convolutional filter with parameters
\begin{eqnarray}
\mathbf{W}_{\mathrm{ssm}} & = & \begin{bmatrix} \overline{\mathbf{B}} \cdot \overline{\mathbf{A}}^{n_{\mathrm{max}}} \cdot \overline{\mathbf{C}} & \overline{\mathbf{B}} \cdot \overline{\mathbf{A}}^{n_{\mathrm{max}} - 1} \cdot \overline{\mathbf{C}} & \cdots & \overline{\mathbf{B}} \cdot \overline{\mathbf{C}} \end{bmatrix},
\end{eqnarray}

\noindent where $n_{\mathrm{max}}$ is the maximum length of the sequence\footnote{Here $\mathbf{W}_{\mathrm{ssm}}$ can be represented as an $n_{\mathrm{max}} \times d \times d$ tensor.}. Then, the output of the state-space model for a sequence $\mathbf{S} = \begin{bmatrix} \mathbf{s}_0 \\ \vdots \\ \mathbf{s}_n \end{bmatrix}$ can be expressed as
\begin{eqnarray}
\mathbf{O} & = & \mathrm{Conv}(\mathbf{S},\mathbf{W}_{\mathrm{ssm}}) + \mathrm{Linear}(\mathbf{S},\overline{\mathbf{D}}),
\end{eqnarray}

\noindent where $\mathrm{Conv}(\cdot)$ is the convolution operation, and $\mathrm{Linear}(\cdot)$ is the linear transformation operation. Such a treatment of the state-space model enables the system to be efficiently implemented using fast parallel convolution algorithms.

However, naive implementations of this model often perform poorly. As with many deep neural architectures, careful parameter initialization is crucial. Specifically, restricting the state matrix $\mathbf{A}$ to particular structures (e.g., HiPPO matrices) has proven essential for memorizing and generalizing over long sequences~\citep{gu-etal:2022parameterization}.

Another computational bottleneck is the repeated matrix multiplication required to compute $\overline{\mathbf{A}}^{n}$. For large $n$, this operation is computationally expensive and numerically unstable. A common approach in modern SSMs is \textbf{diagonalization}. The idea is to apply a basis transformation to the state space such that $\mathbf{A}$ (or $\overline{\mathbf{A}}$) becomes diagonal. Given an SSM parameterized by $(\mathbf{A}, \mathbf{B}, \mathbf{C}, \mathbf{D})$, we can define an equivalent model $(\mathbf{U} \mathbf{A} \mathbf{U}^{-1}, \mathbf{B} \mathbf{U}^{-1}, \mathbf{U} \mathbf{C}, \mathbf{D})$ via an invertible transformation matrix $\mathbf{U}$. It can be shown that both representations are mathematically equivalent under a change of basis\footnote{A state-space transformation maps the state variables into a new coordinate system: $\mathbf{z}'(t) = \mathbf{z}(t) \cdot \mathbf{U}^{-1}$.}.

By applying this transformation, and assuming $\overline{\mathbf{A}}$ can be diagonalized into the canonical form $\mathbf{P}^{-1} \Lambda \mathbf{P}$ (where $\Lambda$ is diagonal), we enforce the constraint that the state matrix is diagonal, giving rise to \textbf{diagonal state-space models}. To illustrate, consider the filter used in the convolutional representation (Eq. (\ref{eq:ssm-convolutional-part})). Substituting $\overline{\mathbf{A}} = \mathbf{P}^{-1} \Lambda \mathbf{P}$, we can rewrite the term $\overline{\mathbf{B}} \cdot \overline{\mathbf{A}}^{t} \cdot \overline{\mathbf{C}}$ as:
\begin{eqnarray}
\overline{\mathbf{B}} \cdot \overline{\mathbf{A}}^{t} \cdot \overline{\mathbf{C}} & = & \overline{\mathbf{B}} \cdot (\mathbf{P}^{-1} \Lambda \mathbf{P})^{t} \cdot \overline{\mathbf{C}} \nonumber \\
& = & \overline{\mathbf{B}} \cdot (\mathbf{P}^{-1} \Lambda \mathbf{P}) \cdot (\mathbf{P}^{-1} \Lambda \mathbf{P}) \cdots (\mathbf{P}^{-1} \Lambda \mathbf{P}) \cdot \overline{\mathbf{C}} \nonumber \\
& = & (\overline{\mathbf{B}} \cdot \mathbf{P}^{-1}) \cdot \Lambda^t \cdot (\mathbf{P} \cdot \overline{\mathbf{C}}).
\end{eqnarray}

Since $\Lambda$ is a diagonal matrix, we can efficiently compute $\Lambda^t$ by simply raising all the entries of $\Lambda$ to the $t$-th power. We then have a computationally cheaper model, in which
\begin{eqnarray}
\overline{\mathbf{A}}' & = & \Lambda, \\
\overline{\mathbf{B}}' & = & \overline{\mathbf{B}} \cdot \mathbf{P}^{-1}, \\
\overline{\mathbf{C}}' & = & \mathbf{P} \cdot \overline{\mathbf{C}}, \\
\overline{\mathbf{D}}' & = & \overline{\mathbf{D}}.
\end{eqnarray}

\noindent More detailed discussions of diagonal state-space models in sequence modeling can be found in \citet{gu-etal:2021efficiently}'s work.

The application of state-space models to Transformers is simple. Each self-attention sub-layer is replaced in this case by an SSM sub-layer as described in Eqs. (\ref{eq:ssm-state-equation-discrete}) and (\ref{eq:ssm-output-equation-discrete}). As we have seen there is a close relationship between state-space models and both CNNs and RNNs. For sequence modeling, we can deal with a sequence of tokens either sequentially as in RNNs, or in parallel as in CNNs. This leads to a new paradigm that takes both the sequential view and the parallel view of the sequence modeling problem --- for training, the system operates like CNNs to make use of fast parallel training algorithms; for prediction, the problem is re-cast as a sequential update problem which can be efficiently solved by using RNN-like models.

While the formalism of state-space models is different from that we discussed in this chapter, it provides a unifying framework for sequence modeling. By viewing the problem through complementary perspectives (parallel and sequential), we can optimize for different hardware and latency constraints. Several recent architectures are motivated by this duality, yielding systems that effectively combine the parallelizability of Transformers with the inference efficiency of RNNs~\citep{orvieto-etal:2023resurrecting,sun-etal:2023retentive}.

\subsection{Conditional Computation}
\label{sec:transformer-conditional-computation}

\noindent So far in our discussion of efficient Transformer models, we have assumed that the model architecture is determined before training and remains fixed throughout. We now turn to the case of learning efficient model architectures. We can write a model in the form
\begin{eqnarray}
\mathbf{y} & = & \mathrm{Model}(\mathbf{x},g(\mathbf{x})),
\end{eqnarray}

\noindent where $\mathbf{x}$ and $\mathbf{y}$ are the input and output of the model. $g(\mathbf{x})$ is a function that returns the model architecture and corresponding parameters for the given input $\mathbf{x}$. In general, we adopt the convention prevalent in learning problems of using a fixed model architecture and learning only the parameters, say, $g(\mathbf{x}) = \theta$. In this case, the goal is to learn optimal parameter values given the architecture and training data. On test data, we make predictions using the same model architecture along with the optimized parameters.

A natural extension of this approach is to consider the learning of both the model architecture and parameters. In architecture learning, we would like to find a function $\hat{g}(\mathbf{x})$ that produces the optimal model architecture and parameter values given the input $\mathbf{x}$. However, searching the entire hypothesis space of all possible combinations of architectures and parameters is intractable, and so practical methods are required. Two classes of methods can be applied:

\begin{itemize}
\item \vspace{0.5em} \textbf{Neural Architecture Search} (\textbf{NAS}). In \textbf{automated machine learning} (\textbf{AutoML}), neural architecture search is the process of exploring a space of neural networks to find one that best fits some criteria~\citep{zoph-and-Le:2016neural,elsken-etal:2019neural}. Once the optimal neural network is determined, its parameters are trained as usual, and then applied to new data. In order to make search tractable, several additional techniques, such as search space pruning and fast search algorithms, are typically used. Applying neural architecture search to the development of efficient neural networks is straightforward~\citep{howard-etal:2019searching,tan-etal:2019efficientnet}. We need only incorporate efficiency measures into the performance estimation of neural networks, for example, the search can be guided by a criterion that penalizes neural networks with high latency or excessive memory requirements.
\item \vspace{0.3em} \textbf{Dynamic Neural Networks}. The key idea of dynamic neural networks is to adapt a neural network to various inputs dynamically~\citep{gupta-etal:2004static,han-etal:2021dynamic}. Ideally, we would like to learn $\hat{g}(\cdot)$, and then, for any input $\mathbf{x}_{\mathrm{new}}$, we apply the model $\mathrm{Model}(\mathbf{x}_{\mathrm{new}},\hat{g}(\mathbf{x}_{\mathrm{new}}))$. As a result, at test time we may have different model structures and/or different parameters for different inputs. However, it is infeasible to develop a function $\hat{g}(\cdot)$ that can model arbitrary neural networks. In practice, $\hat{g}(\cdot)$ is often considered to represent a family of sub-networks of a super-network. The problem is therefore reframed as a simpler problem to learn to choose which sub-network is used for a given input.
\end{itemize}
\vspace{0.3em}

From a machine learning perspective, the approaches to neural architecture search are general and can be applied to any neural network. On the other hand, from a practical perspective, it is still difficult to find an efficient neural network that is sufficiently powerful and generalizes well. While neural architecture search provides interesting ideas for developing efficient Transformer models, we do not discuss it in depth here. Instead, the reader can refer to the above papers to have a general idea of it, and refer to \citet{so-etal:2019evolved}, \citet{wang-etal:2020hat}, and \citet{hu-etal:2021ranknas}'s work for its application to Transformers.

In this subsection, we focus on a particular family of approaches to dynamic neural networks, called \textbf{conditional computation}. This concept was originally motivated by the dynamic selection of neurons of a neural network~\citep{bengio-etal:2013estimating,bengio-etal:2015conditional}. More recently, it has often been used to refer to the process of dynamically selecting parts of a neural network. A narrow view of conditional computation is to see $g(\cdot)$ as an adaptive neural network which dynamically reduces or grows the number of computation units (such as neurons and layers). As a result, computation can adapt to changing conditions, and we can seek a good accuracy-latency trade-off by this adaptation mechanism.

A common way to achieve this is to learn how to skip some computation steps so that we can work with a subset of the network~\citep{xu-and-mcauley:2023survey}. One of the simplest methods, sometimes called \textbf{early exiting}, is to stop the computation at some point during encoding or decoding a sequence. This technique is often used in practical sequence generation applications where a low latency is required. Suppose $y_1 \cdots y_{n_{\mathrm{max}}}$ is the longest sequence that the system can generate, and $\mathbf{s}_1\cdots\mathbf{s}_{n_{\mathrm{max}}}$ is the corresponding sequence of the hidden states of the top Transformer layer. Then we develop a model $f_{\mathrm{stop}}(\cdot)$ that takes one hidden state $\mathbf{s}_i$ at a time and produces a distribution of a binary variable $c \in \{\mathrm{stop},\mathrm{nonstop}\}$:
\begin{eqnarray}
\Pr(c|\mathbf{s}_i) & = & f_{\mathrm{stop}}(\mathbf{s}_i). \label{eq:early-stop-classification}
\end{eqnarray}

The generation process terminates if $\Pr(\mathrm{stop}|\mathbf{s}_i)$ is sufficiently large, for example
\begin{eqnarray}
\Pr(\mathrm{stop}|\mathbf{s}_i) & \ge & \Pr(\mathrm{nonstop}|\mathbf{s}_i) + \theta_{\mathrm{stop}}, \label{eq:conditional-computation-simple-early-stop}
\end{eqnarray}

\noindent where $\theta_{\mathrm{stop}}$ denotes the minimal margin for distinguishing the two actions\footnote{An equivalent form of Eq. (\ref{eq:conditional-computation-simple-early-stop}) is $\Pr(\mathrm{stop}|\mathbf{s}_i) \ge \frac{1 + \theta_{\mathrm{stop}}}{2}$.}. This formulation is also related to the stopping criterion problem that is frequently discussed in search algorithms for sequence generation. $f_{\mathrm{stop}}(\cdot)$ can be designed in several different ways. For example, in many practical applications, the stopping criterion is based on simple heuristics. Alternatively, we can define the function $f_{\mathrm{stop}}(\cdot)$ as a neural network and train it using labeled data.

\begin{figure}[!htp]
\centering
\input{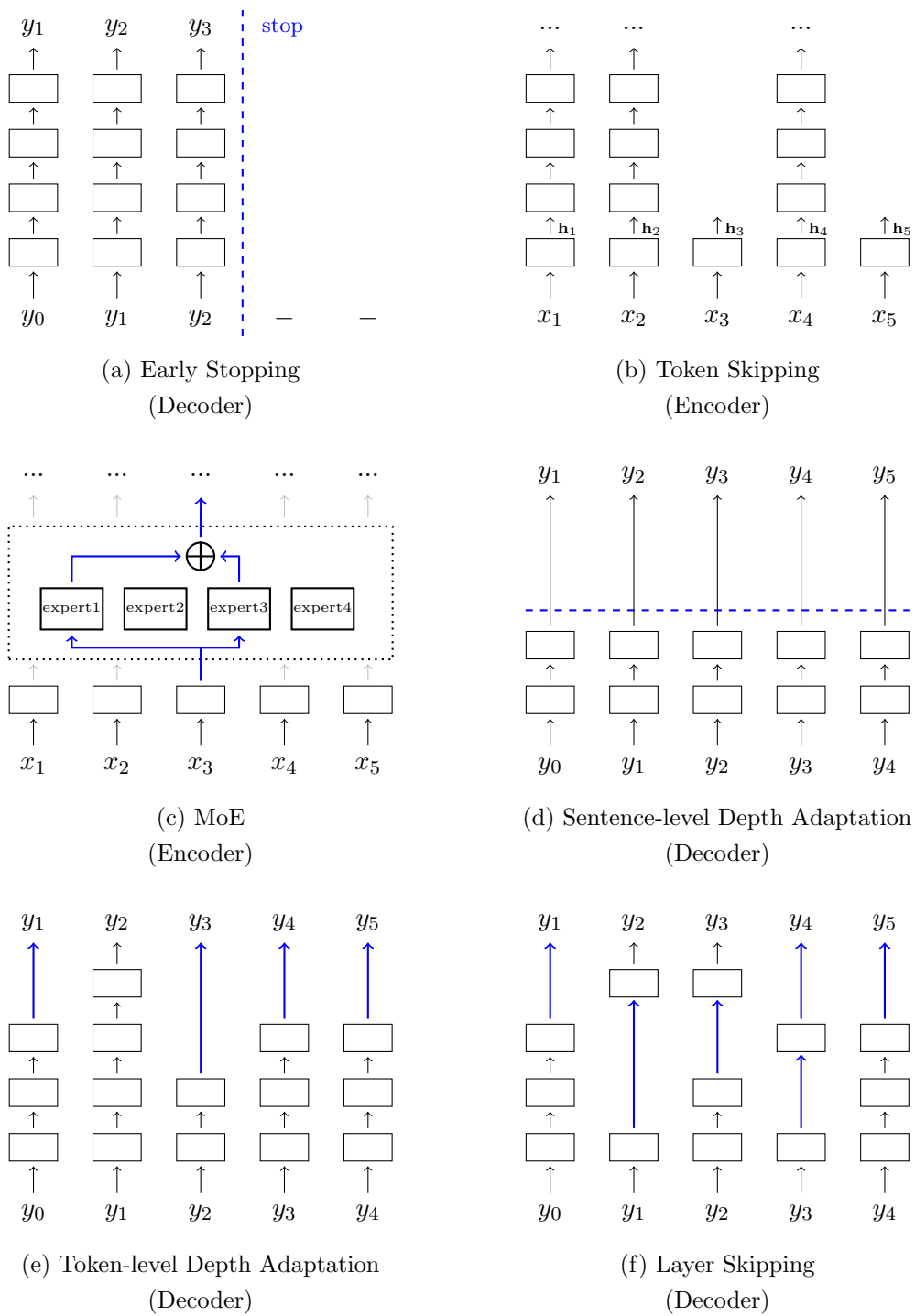}
\caption{Methods of conditional computation, including early exiting, token skipping, MoE, sentence-level depth adaptation, token-level depth adaptation, and layer skipping. While these methods are illustrated using either the encoding or decoding process, most of them can be applied to both Transformer encoders and decoders.}
\label{fig:transformer-conditional-computation}
\end{figure}

The above approach can be easily extended to handle situations in which some of the tokens are skipped. This learning-to-skip approach is typically used in the encoding stage in which all input tokens are given in advance. Let $\mathbf{h}_1 \cdots\mathbf{h}_m$ be low-level representations of a sequence $x_1 \cdots x_m$. Like Eq. (\ref{eq:early-stop-classification}), we can develop a model $\Pr(c|\mathbf{s}_i)$ ($c \in \{\mathrm{skip},\mathrm{nonskip}\}$) to determine whether the token $x_i$ can be skipped. Figure~\ref{fig:transformer-conditional-computation} (a) and (b) show illustrations of early exiting and skipping. Note that the learning-to-skip method overlaps with other lines of research on training neural networks. For example, erasing some of the input tokens in training is found to be useful for achieving higher generalization of Transformer models~\citep{shen-etal:2020simple,kim-and-cho:2021length}. This method is also related to the downsampling methods which will be discussed in Section~\ref{sec:transformer-downsampling}.

A second approach to conditional computation is to resort to \textbf{sparse expert models}, or its popular instance --- MoE~\citep{yuksel-etal:2012twenty}. In deep learning, a model of this kind is typically built from a number of experts which are neural networks having the same structure but with different parameters. In this way, we can construct a large model by simply increasing the number of experts. When running this model, during either training or prediction, we activate only a small number of the experts by some routing algorithms (see Figure~\ref{fig:transformer-conditional-computation} (c)). An MoE model is an adaptive network since each time we have a new input, the model routes it to different experts. In Section~\ref{sec:transformer-wide-models}, we presented the basic form of MoE, and showed how Transformer models can be scaled up by this sparse method. For a comprehensive review of the recent advances in MoE, we refer the interested reader to \citet{fedus-etal:2022review}'s work.

A third approach that can be used to adapt a Transformer model to changing input is to dynamically shrink the number of layers. Several methods have been proposed to do this in an attempt to improve inference efficiency. The simplest of these is to exit at some hidden layers by which we can still make accurate predictions for the input (see Figure~\ref{fig:transformer-conditional-computation} (d) and (e)). To do this, we can either determine the appropriate depth for the entire sequence (call it a \textbf{sentence-level depth-adaptive model}), or use an adaptive depth for each token (call it a \textbf{token-level depth-adaptive model}). Here we consider token-level depth-adaptive models but the methods can be easily extended to sequence-level depth-adaptive models.

Suppose there are $L$ stacked layers at position $i$\footnote{A layer is a standard Transformer block consisting of a few sub-layers.}. We would ideally like to find a layer in the stack, which can be used as the last hidden layer for making predictions, and whose depth is as low as possible. However, we cannot simply use the $L$-th layer of the stack as the oracle for this problem, because we never know in advance what the last layer generates during inference. Instead, we need to determine whether the network should stop growing at depth $i$, considering the layers generated so far.

Now suppose we have a Transformer decoder which produces a distribution over a vocabulary $V$ at each step. As usual, we denote the output of the $l$-th layer at step $i$ by $\mathbf{s}_i^{l}$. For each $\mathbf{s}_i^{l}$, we create an output layer that produces a distribution $\mathbf{p}_i^{l}$ over the vocabulary (call it an \textbf{early exit classifier}), given by
\begin{eqnarray}
\mathbf{p}_i^{l} & = & \mathrm{Softmax}(\mathbf{s}_i^{l} \cdot \mathbf{W}_{\mathrm{o}}^{l}),
\end{eqnarray}

\noindent where $\mathbf{W}_{\mathrm{o}}^{l} \in \mathbb{R}^{d \times |V|}$ is the parameter matrix. Hence we have $L-1$ additional output layers, each corresponding to a hidden layer from depth 1 to $L-1$. At training time, we consider the cross-entropy losses of $\{\mathbf{p}_i^{1},\cdots,\mathbf{p}_i^{L-1}\}$, and train these layers together with the Transformer model. At test time, the depth of the network grows as usual, and we use $\{\mathbf{p}_i^{1},\cdots,\mathbf{p}_i^{l}\}$ and/or $\{\mathbf{s}_i^{1},\cdots,\mathbf{s}_i^{l}\}$ to determine whether we should exit at the $l$-th layer. There are several exit criteria, for example,

\begin{itemize}
\item \vspace{0.5em} Common criteria are based on measures of the confidence of predictions. A simple method is to compute the entropy of $\mathbf{p}_i^{l}$, and exit if this entropy is above a pre-defined value.
\item \vspace{0.3em} Alternatively, one can view the maximum probability of the entries of $\mathbf{p}_i^{l}$ as the confidence of the prediction.
\item \vspace{0.3em} Instead of considering the output of a single layer, we can also examine the change in the outputs or hidden states over a number of layers. For example, we can measure the similarity between $\mathbf{p}_i^{l-1}$ and $\mathbf{p}_i^{l}$ or between $\mathbf{s}_i^{l-1}$ and $\mathbf{s}_i^{l}$. If the similarity is above a given threshold, then we say that the output of the neural network tends to converge and the number of layers can stop growing.
\item \vspace{0.3em} The above methods can be extended to examine the change in the predictions made by the classifiers associated with the layers. For example, the model can choose to exit if the predictions made by the classifiers remain unchanged for a number of layers.
\end{itemize}
\vspace{0.3em}

Discussions of these criteria can be found in the related papers~\citep{xin-etal:2020deebert,zhou-etal:2020bert,schuster-etal:2022confident}. There are a variety of ways to improve these early exit methods. One is to explore other forms of the prediction for each layer. For example, we can develop a model that directly predicts how many layers we need to model the input~\citep{elbayad-etal:2020depth}. Another line of research on early exit focuses on better training for these models, for example, we can consider various loss functions for training the classifiers~\citep{schwartz-etal:2020right,schuster-etal:2022confident}. In addition, there is also interest in learning the combination of the outputs of multiple layers so that we can make predictions by using multiple levels of representation~\citep{zhou-etal:2020bert,liao-etal:2021global}.

A problem with token-level adaptive-depth models is that the representations at certain depths may be absent in the previous steps. In this case, standard self-attention is not directly applicable, because we may not attend to the previous tokens in the same level of representation. For training, this can be addressed by using all the $L$ layers of the full model. For inference, we can either duplicate the layer from which we exit to fill up the layer stack, or modify the self-attention model to enable it to attend to the representations of the previous tokens at different depths.

It is also possible to select any subset of the layers for constructing a shallow network. The adaptive models therefore can be generalized to skipping models (see Figure~\ref{fig:transformer-conditional-computation} (f)). As with the early exit problem, the skipping problem can be framed as a learning task, in which a classifier is trained to decide whether a layer should be dropped. The learning-to-skip problem has been studied in the field of computer vision~\citep{wang-etal:2018skipnet,wu-etal:2018blockdrop}. However, learning a skipping model for large-scale, deep neural networks is difficult. For practical systems, it still seems reasonable to use heuristics or cheap models to obtain a neural network having skipped layers, which has been discussed in recent pre-trained NLP models~\citep{wang-etal:2022skipbert,del-etal:2023skipdecode}.

\subsection{Model Transfer and Pruning}
\label{sec:transformer-model-transfer-and-pruning}

\noindent Many large Transformer models have been successfully developed to address NLP problems. A common question is: can we transform a large, well-trained model into a smaller one that allows for more efficient inference? At a high level, this can be thought of as a \textbf{transfer learning} problem in which the knowledge is transferred from one model to another. However, we will not discuss this general topic, which spans a broad range of issues and models, many of which fall outside the scope of this chapter. Instead, we narrow our discussion to two approaches that are widely used for training smaller neural networks from larger ones.

\subsubsection{Knowledge Distillation}

\noindent \textbf{Knowledge distillation} is a process of compressing the knowledge in a large neural network (or an ensemble of neural networks) into a smaller neural network~\citep{hinton-etal:2015distilling}. In the supervised learning of neural networks, objective functions are generally designed to quantify the error between the predicted and true answers. Hence, by minimizing this loss, models are trained to output the correct answers. While models are typically optimized on the training data in this manner, what we really want is to generalize to unseen data. This is, however, difficult because training only with ground-truth labels provides no direct supervision signal regarding generalization. In knowledge distillation, instead of forcing a model to match the ground-truth output, we train it to mimic the behavior (or output distributions) of a larger model. To achieve this, we directly transfer the knowledge of a pre-trained model to the model we wish to train.

A frequently used approach to knowledge distillation is \textbf{teacher-student training}. A teacher model is typically a relatively large neural network that has already been trained and can generalize well. A student model is a smaller neural network, such as one with fewer layers, to which we transfer the knowledge. A simple way to distill the knowledge from the teacher model into the student model is to use the output of the teacher model as the ``correct'' answer for training the student model. Suppose we have a teacher Transformer model that can generate a sequence of distributions $\{\Pr(\cdot|y_0,\mathbf{x}),\cdots,\Pr(\cdot|y_0 \cdots y_{n-1},\mathbf{x})\}$ for the input $\mathbf{x}$. To keep the notation simple, we denote the distribution $\Pr(\cdot|y_0 \cdots y_{i-1},\mathbf{x})$ as $\widetilde{\mathbf{p}}_i$. Similarly, we denote the output of the student Transformer model for the same input as $\mathbf{p}_i$. As usual, we consider a loss function $\mathrm{Loss}(\widetilde{\mathbf{p}}_i,\mathbf{p}_i)$ (such as the cross-entropy function) for computing the distance between $\widetilde{\mathbf{p}}_i$ and $\mathbf{p}_i$. Then, we can define the loss over the entire sequence as
\begin{eqnarray}
\mathcal{L}(\mathbf{x},\theta) & = & \frac{1}{n} \sum_{i=1}^{n} \mathrm{Loss}(\widetilde{\mathbf{p}}_i,\mathbf{p}_i),
\end{eqnarray}

\noindent where $\theta$ denotes the parameters of the student model\footnote{We omit the parameters of the teacher model because they are fixed throughout the training process.}. Using this loss, we can optimize $\theta$ for any given set of source sequences $\{\mathbf{x}_1, \cdots, \mathbf{x}_K\}$ by minimizing the objective $\sum_{k=1}^{K} \mathcal{L}(\mathbf{x}_k,\theta)$.

Several extensions to this basic method have been developed to model the problem of knowledge transfer between two models. A simple way is to use the hidden states instead of the output probabilities as the training targets~\citep{romero-etal:2014fitnets}. In this case, the objective is to minimize the difference between some hidden states of the teacher model and the corresponding states of the student model. Rather than using the outputs of various layers as the targets for training the student model, another technique is to model the relations between samples and train the student model by minimizing the difference between the relation encodings of the teacher and student models~\citep{park-etal:2019relational,peng-etal:2019correlation}. For example, we can develop a relation encoding model based on the Transformer architecture. The goal is then to optimize the student model so that its corresponding relation encoding of a group of samples is as close as possible to that of the teacher model.

For sequence generation problems, a special case of knowledge distillation, which can be viewed as a means of \textbf{data augmentation}, is often used for developing lightweight models~\citep{kim-etal:2016sequence}. For example, consider the problem of transferring the translation ability of a well-developed machine translation model (i.e., the teacher model) to a new model (i.e., the student model). Given a set of source-side sentences $\{\mathbf{x}_1,\cdots,\mathbf{x}_K\}$, we can use the teacher model to translate each $\mathbf{x}_k$ to a target-side sentence $\widetilde{\mathbf{y}}_k$. Then, by treating $\mathbf{x}_k$ and $\widetilde{\mathbf{y}}_k$ as paired sentences, we obtain a bilingual dataset consisting of $\{(\mathbf{x}_1,\widetilde{\mathbf{y}}_1),\cdots,(\mathbf{x}_K,\widetilde{\mathbf{y}}_K)\}$. We can use this bilingual dataset as the labeled dataset to train the student model as usual. One advantage of this data augmentation method is that it is architecture-agnostic. It does not require access to the internal architectures of either the teacher or the student models. Hence, it can be applied even if the teacher model is used as a black-box model. More detailed discussions on knowledge distillation can be found in the surveys by \citet{gou-etal:2021knowledge} and \citet{wang-and-yoon:2021knowledge}.

\subsubsection{Structured Pruning}

\noindent Pruning is among the most popular model compression methods and has been applied to a broad range of systems. One common approach is \textbf{unstructured pruning}, in which we retain only a sparse subset of connections between neurons. However, as with most sparse models, networks pruned in this way typically require specialized implementations and hardware support, which in turn limits their efficiency gains in certain applications. A simpler yet more aggressive alternative is \textbf{structured pruning}. In deep learning, structured pruning is a technique that removes entire groups of neurons or connections simultaneously. For example, we can remove an entire layer of neurons from a neural network to obtain a shallower model. Given their multi-layer and multi-head architecture, Transformers are naturally suited to structured pruning, and we can prune a Transformer network in several different ways. For instance, we can prune specific heads in the multi-head attention mechanism~\citep{voita-etal:2019analyzing,michel-etal:2019sixteen}, or entire layers within the stack~\citep{hou-etal:2020dynabert,kim-and-awadalla:2020fastformers}.

Formally, we can represent a neural network as a set of parameter groups $\{\theta_1, \cdots, \theta_{R}\}$, each corresponding to a specific component or sub-module of the network. Our goal is to find a subset of $\{\theta_1, \cdots, \theta_{R}\}$ that allows us to build a model yielding strong performance while maintaining significantly lower computational complexity. However, an exhaustive search for such an optimal subset is infeasible because there is a combinatorial number of possible candidates, and evaluating all of them is computationally prohibitive.

One approach to structured pruning is to randomly discard components of a model. We can repeat this random pruning process multiple times to generate a pool of model candidates, and select the best-performing one. Another approach relies on heuristics to determine which components are redundant and can be safely removed. Common measures for evaluating the importance of a parameter group $\theta_r$ include various metrics based on the norms of its weights or gradients~\citep{santacroce-etal:2023matters}. We can then prune $\theta_r$ if these metrics fall below or exceed predefined thresholds. A third approach frames the pruning problem as a continuous optimization task by introducing trainable gates that indicate the presence or absence of different components~\citep{mccarley-etal:2019structured,wang-etal:2020structured,lagunas-etal:2021block}. The final pruned model can be directly derived using these trained gates. It is worth noting that, in many cases, pruning is no longer merely a post-processing step for a fully trained model, but rather an integral part of the training process itself.

\subsection{Sequence Compression}
\label{sec:transformer-downsampling}

\noindent In sequence modeling and generation problems, time and space complexities are strongly influenced by the length of the input or output sequences. Therefore, shorter sequences are generally preferred. This is particularly important for Transformer models, as their time and space complexities scale quadratically with sequence length, making memory footprint and latency significant computational bottlenecks for very long sequences. While previous subsections discussed modifications to the Transformer architecture to handle long sequences, in this section, we instead explore methods for compressing sequences to more manageable lengths.

One simple approach is to map the input sequence to a fixed-size representation. For example, using the recurrent models discussed in Section~\ref{sec:transformer-recurrent-and-memory-models}, we can encode a sequence of vectors into a single vector. This method can be easily extended to generate a larger representation so that this representation can retain more information from the original input. For example, we can select a fixed number of the hidden states over the sequence to form a new sequence of fixed length. Another way to represent a variable-length sequence as a fixed-length sequence is to perform cross-attention between the input vectors and a set of learnable hidden representations. In the work of \citet{jaegle-etal:2021perceiver}, this is done by introducing $r$ hidden representations $\{\mathbf{u}_{1},\cdots,\mathbf{u}_{r}\}$, and then applying cross-attention between the input vectors ${\mathbf{x}_1,\cdots,\mathbf{x}_m}$ and these hidden representations. The attention model can be a standard QKV attention model in which we view $\{\mathbf{u}_{1},\cdots,\mathbf{u}_{r}\}$ as queries and $\{\mathbf{x}_1,\cdots,\mathbf{x}_m\}$ as keys and values. The output of this model is a sequence of $r$ vectors, which can be used as fixed-length input to downstream systems.

A second approach is to use downsampling to compress the sequence into a shorter one. A typical method of downsampling is strided convolution, which has been widely used in computer vision and speech processing. For example, suppose there is a sequence of $m$ vectors in $\mathbb{R}^d$. We can define a filter with a width of $2$ and a stride of $2$. By taking the sequence as input, the filter produces a sequence of $\frac{m}{2}$ new vectors in $\mathbb{R}^d$, and so we have a reduction factor of $2$. Also, we can stack multiple convolutional layers or pooling layers to achieve a desired level of length reduction, called \textbf{progressive downsampling}. However, it seems inevitable that downsampling will lead to information loss~\citep{han-etal:2020contextnet,burchi-and-vielzeuf:2021efficient}. We need to consider a trade-off between the compressed sequence length and the performance of downstream systems~\citep{xu-etal:2023bridging}.

In NLP, the problem of sequence compression is also closely related to the problem of tokenizing input strings. Therefore, tokenization is a practical approach that can be taken to address the length issue. Segmenting a string into small tokens (such as characters) generally reduces the sparsity of the data, which makes it easier to learn the embeddings of these tokens, but such approaches often lead to a long sequence. By contrast, we will have a shorter sequence if we segment the input string into larger units, but this may suffer from data sparsity. In deterministic tokenization methods, which produce tokenization results using statistics collected from the entire dataset, the sequence length can be controlled to some extent by adjusting specific hyperparameters, for example, in byte pair encoding~\citep{sennrich:2016neural}, increasing the size of the vocabulary generally reduces the number of resulting tokens. Another way to obtain an appropriate sequence of tokens is to use a model for choosing among tokenization candidates~\citep{kudo:2018subword,provilkov-etal:2020bpe}. As with many probabilistic models for text generation, in this case, we can incorporate priors to the criterion for tokenization selection so that we can express a preference for shorter sequences over longer sequences.

A fourth approach to sequence compression is to drop some of the tokens in the sequence. For instance, in many practical applications, sequences are simply truncated when their lengths exceed a predefined threshold. We can relate this to the early exiting and skipping approaches in conditional computation. Thus the methods discussed in Section~\ref{sec:transformer-conditional-computation} can be directly applied. The token dropping methods can also be viewed as pruning methods, called \textbf{token pruning}. By discarding tokens that are less important for representing the entire sequence, token pruning can significantly reduce the sequence length while maintaining the performance of NLP systems on downstream tasks~\citep{kim-etal:2023full}.

\subsection{High Performance Computing Methods}

\noindent So far in this section, we have discussed efficient Transformer models from the perspective of deep learning. However, we have not considered their hardware efficiency. As modern hardware provides a variety of execution paradigms, the practical time and memory footprint savings generally depend on the specifications of the hardware systems. One line of research on the efficient use of computing resources explores parallel computing methods. There have been many attempts to develop large-scale Transformer models by using a cluster of machines. Typically, scaling Transformers to models with billions or even tens of billions of parameters requires a careful design of parallelism strategies for sharding the big networks. More efficient implementations of such systems also require consideration of communication overhead in the cluster, as well as the utilization of sparse models that activate only a small subset of the parameters for each sample, thereby enabling the use of very large models. Most of these methods have been extensively studied in the literature~\citep{lepikhing-etal:2021shard,barham-etal:2022pathways,fedus-etal:2022switch}. The results of these studies are foundational to many follow-on works investigating the \textbf{scaling laws} for large language models~\citep{brown-etal:2020language,chowdhery-etal:2022palm}. Since large-scale distributed models are generic and not specialized to the case of Transformers, we skip their discussion here. Interested readers can refer to the above papers for more detailed discussions.

In this subsection, we consider hardware-aware methods to seek greater computational efficiency for Transformer models. We first consider a simple but widely used method that aims to represent and execute neural networks using lower- or mixed-precision formats~\citep{gholami-etal:2022survey}. Conventional neural networks are typically based on single-precision and/or double-precision floating-point representations of data. While single-precision floating-point data types provide a sufficiently precise way to represent parameters and intermediate states in most cases, in some applications, they are not essential. As an alternative, one can use half-precision (or even lower-precision) formats for representing floating-point numbers in neural networks. The size of the resulting model is thus half the size of the original model. One advantage of using half-precision floating-point representations is that, although processing such data types requires new APIs for linear algebra operations and hardware support, it does not change the model architecture, requiring only slight modifications to the system. For example, half-precision floating-point representations can be applied to either the training or inference of Transformers, or both.

Recently, the deployment of large Transformer models has been further improved by quantizing these models. In signal processing, quantization is a process of mapping continuous values (i.e., floating-point representations) to a set of discrete values (i.e., fixed-point representations). This process is generally implemented using a system called a quantizer. In the context of neural networks, a quantizer involves two functions --- the quantization function and the de-quantization function. The quantization function maps a floating-point number to a (lower-bit) integer. A simple quantization function is given by
\begin{eqnarray}
Q(x) & = & \lfloor \frac{x}{s} \rceil,
\end{eqnarray}

\noindent where $\lfloor \cdot \rceil$ is a rounding function\footnote{$\lfloor a \rceil$ returns the integer closest to $a$.}, $x$ is the real-valued input, and $s$ is the quantization step size that controls the level of quantization. The quantization function is coupled with a de-quantization function
\begin{eqnarray}
D(r) & = & s \cdot r.
\end{eqnarray}

\noindent With this notation, the quantizer can be expressed as $D(Q(x)) = s \cdot \lfloor \frac{x}{s} \rceil$. The difference between $D(Q(x))$ and $x$ is called the quantization error. A smaller value of $s$ typically reduces the quantization error. In practice, however, we wish to choose an appropriate value of $s$ in order to spread the possible values of $Q(x)$ evenly across the integer representation range. For example, $s = \frac{\max\{D(r)\}}{2^p-1}$, where $p$ is the number of bits used to represent an integer and $\max\{D(r)\}$ is the maximum representable value after de-quantization. The equations above show one of the simplest cases of quantization. More general discussions of quantization can be found in books on digital signal processing and related surveys~\citep{oppenheim-and-schafer:1975digital,rabiner-gold:1975theory,gray-etal:1998quantization}.

Applying quantization to Transformers is relatively straightforward. The idea is that we quantize the inputs and model parameters using $Q(x)$ and feed them to a quantized Transformer model in which all layers operate on integer-valued tensors. In other words, we implement the model using integer-only arithmetic. However, the price to be paid for this compressed model, as with many approximation approaches to deep learning, is that its predictions are less accurate than those of the standard Transformer. Using integer operations to approximate continuous-valued operations generally leads to approximation errors. These errors will accumulate if the quantized neural network is deep. Furthermore, Transformer models involve components (such as self-attention sub-layers) that require relatively complex linear algebra operations. Simply applying quantization to these modules will lead to a significant loss in accuracy. One solution is to simplify the model architecture and develop new sub-models that are more suited for quantization. Alternatively, a more common paradigm in quantized neural networks is to add de-quantization functions to the neural networks so that the outputs of a layer are floating-point tensors and can be used as usual in the following steps. Consider a simple example where we multiply a real-valued input matrix $\mathbf{a}$ with a real-valued parameter matrix $\mathbf{A}$. We first quantize $\mathbf{a}$ and $\mathbf{A}$, and multiply them using integer-based matrix multiplication. The result is then de-quantized to a real-valued matrix. In this way, we obtain an approximation $D(Q(\mathbf{a}) \cdot Q(\mathbf{A}))$ to $\mathbf{a} \cdot \mathbf{A}$ at a very low computational cost.

However, sandwiching each layer between $Q(\cdot)$ and $D(\cdot)$ will lead to the additional cost of running $Q(\cdot)$ and $D(\cdot)$. In some practical applications, the computational overhead introduced by $Q(\cdot)$ and $D(\cdot)$ is even larger than the time savings of performing integer-based operations. In general, the benefits of quantizing neural networks are greater than the costs only if the neural networks are sufficiently large. Therefore, in practice, it is common to perform quantized computation only for operations whose computational costs are high. For example, in recent large language models, quantization is primarily applied to the multiplication of large matrices, yielding significant time and memory savings.

While quantization approaches can be used in both training and inference, a widely used approach is to quantize Transformer models after training (called \textbf{post-training quantization}). In this approach, quantization is performed on well-trained floating-point-based neural networks, resulting in fewer quantization-related errors. However, these errors cannot be easily compensated for because they are introduced after training is complete. A more promising approach is to incorporate quantization during training so that the model can learn to compensate for quantization-related errors~\citep{jacob-etal:2018quantization,nagel-etal:2021white}. There have been several attempts to apply quantization-aware training to Transformers~\citep{bondarenko-etal:2021understanding,stock-etal:2021training,yang-etal:2023quantization}. In addition to computational efficiency, another important consideration for high-performance systems is the constraints of the memory hierarchy. In general, better system design requires considering the speeds and capacities of different memory levels. The problem is even more complicated when training large Transformer models on modern hardware where both GPUs and CPUs are utilized. A general principle of system design is that memory transfers between different memory levels should be minimized. While we would ideally like to have a large, high-level memory on which we can store all the data we need to process, in many practical situations, the size of fast, on-chip memory is orders of magnitude smaller than the total size of the data. In this case, we can re-order the memory access in the algorithms so that data used in adjacent computation steps can be loaded into high-speed memory all at once. This idea motivates the development of many fast linear algebra libraries. For example, there are matrix multiplication algorithms that are highly optimized for different shapes of input matrices.

It is relatively straightforward to use these optimized linear algebra algorithms to build a Transformer-based system, but the modules of this system are not jointly optimized for efficiency. For example, a self-attention sub-layer involves a series of scaling, normalization, and matrix multiplication operations. Although each of these operations is implemented in supported and efficient linear algebra libraries, successive calls to them still require multiple memory transfers when switching from one operation to another. In practice, a better approach would be to keep some of the intermediate states in the on-chip memory and reuse them in subsequent computation steps instead of fetching them again from slower memory. For example, on modern GPUs, a simple way to achieve this is to merge multiple operations into a single operation, a technique known as \textbf{kernel fusion}. For Transformer models, a general idea is to design data partitioning and layout strategies that maximize the computation on each data block loaded into high-performance memory while simultaneously minimizing memory transfers. There have been several attempts to use these strategies to improve the attention models in Transformers~\citep{ivanov-etal:2021data,pope-etal:2023efficiently}. Some of these methods, such as FlashAttention and PagedAttention, have been successfully incorporated into recent large language models~\citep{dao-etal:2022flashattention,kwon-etal:2023efficient}.


\section{Applications}
\label{sec:transformer-applications}

\noindent Transformers have a wide range of applications across NLP tasks. While the Transformer model introduced by \citet{Vaswani-etal:2017Transformer} is based on a standard encoder-decoder architecture, it is primarily used in three different configurations:

\begin{itemize}
\item \vspace{0.5em} \textbf{Decoder-only Models}. If the cross-attention sub-layers are removed, the decoder becomes a standard language model. Hence, this decoder-only model can be applied to text generation problems. For example, given a sequence of left-context tokens, we use the model to predict subsequent tokens.
\item \vspace{0.3em} \textbf{Encoder-only Models}. Transformer encoders can be treated as sequence models that take a sequence of tokens as input and produce a sequence of representations, each corresponding to an input token. These representations can be viewed as a form of encoding of the input sequence, and are often taken as input to a prediction model. This encoder-predictor architecture forms the basis of many NLP systems, such as systems for sentence classification and sequence labeling. Pre-trained Transformer encoders can also be used to map texts into the same vector space so that we can compute the distance or similarity between any two texts.
\item \vspace{0.3em} \textbf{Encoder-Decoder Models}. Encoder-decoder models are typically used to model sequence-to-sequence problems. Applications include many tasks in NLP and related fields, such as machine translation and image captioning.
\end{itemize}
\vspace{0.5em}

Note that while most Transformer-based systems can be categorized into the above three types, the same NLP problem can generally be addressed using different types of models. For example, recent decoder-only models have demonstrated strong performance on a broad range of problems by framing them as text generation tasks, even though many of these problems were historically addressed using encoder-decoder or encoder-only models. To illustrate how the above models are applied, this section considers a few applications where Transformers are chosen as backbone models.

\subsection{Language Modeling}
\label{sec:transformer-app-language-modeling}

\noindent Language modeling is an NLP task in which we predict the next token given its preceding tokens. This is generally formulated as a problem of estimating the distribution of tokens at position $i+1$ given tokens at positions $0 \sim i$. Here we denote this distribution by $\Pr(\cdot \mid x_0, \ldots, x_i)$, where $x_0, \ldots, x_i$ are the observed tokens. The predicted token is obtained by maximizing the probability:
\begin{eqnarray}
\hat{x}_{i+1} & = & \argmax_{x_{i+1} \in V} \Pr(x_{i+1}|x_0,\cdots,x_{i}),
\end{eqnarray}

\noindent where $V$ is the vocabulary. The prediction can be extended to the tokens following $\hat{x}_{i+1}$

\begin{eqnarray}
\hat{x}_{k+1} & = & \argmax_{x_{k+1} \in V} \Pr(x_{k+1}|x_0,\cdots,x_{i},\hat{x}_{i+1},\cdots,\hat{x}_{k}) \label{eq:transformer-app-lm-long}.
\end{eqnarray}

\noindent This model forms the basis of many systems for text generation: given the context tokens $x_0 \cdots x_{i}$, we generate the remaining tokens $\hat{x}_{i+1} \cdots \hat{x}_{k+1}$ to make the sequence complete and coherent.

As discussed in Section~\ref{sec:transformer-architecture}, Transformer decoders are essentially language models. A key difference between the problem of decoding in an encoder-decoder Transformer and the problem of language modeling is that the Transformer decoder makes predictions conditioned on the context tokens on both the encoder and decoder sides, rather than being conditioned only on preceding tokens. To modify the Transformer decoder to implement a standard language model, the cross-attention sub-layers are simply removed and a Transformer decoding block can be expressed as
\begin{eqnarray}
\mathbf{S}^{l} & = & \mathrm{Layer}_{\mathrm{ffn}}(\mathbf{S}_{\mathrm{self}}^{l}), \\
\mathbf{S}_{\mathrm{self}}^{l} & = & \mathrm{Layer}_{\mathrm{self}}(\mathbf{S}^{l-1}).
\end{eqnarray}

\noindent Here, $\mathbf{S}^{l}$ denotes the output of the block at depth $l$. $\mathrm{Layer}_{\mathrm{self}}(\cdot)$ denotes the self-attention sub-layer, and $\mathrm{Layer}_{\mathrm{ffn}}(\cdot)$ denotes the FFN sub-layer. We see that this decoding block has the same form as an encoding block. The difference between the decoding and encoding blocks arises from the masking strategies adopted during training: the former masks the attention from a position $i$ to any right-context position $k > i$, whereas the latter has no such restriction. A softmax layer is stacked on top of the last block and is used to produce the distribution over the vocabulary at each position. For inference, the Transformer decoder works in an autoregressive manner, as described in Eq. (\ref{eq:transformer-app-lm-long}).

Training these models follows a standard paradigm. We learn the model by repeatedly updating the parameters, based on the gradients of the loss on the training samples. This paradigm can be extended to the training of large Transformer-based language models, which have been widely applied in generative AI. However, training Transformer models at scale, including decoder-only, encoder-only, and encoder-decoder models, may lead to new difficulties, such as training instabilities. 

\subsection{Text Encoding}
\label{sec:transformer-text-encoding}

\noindent For many NLP problems, a widely used paradigm is to first learn a representation of an input sequence, and then make predictions for downstream tasks based on this representation. As a result, we separate sequence representation from specific NLP tasks. One of the advantages of this paradigm is that we can train a sequence model that is not specialized to particular problems, thereby enabling it to generalize well.

Clearly, Transformer encoders are a type of sequence model and can be used as text encoders. Consider a Transformer encoder with $L$ encoding layers. The output of the last encoding layer can be viewed as the sequence representation. Here, we prepend a special token $x_0$ to each sequence, indicating the beginning of the sequence (often written as $\langle \mathrm{SOS} \rangle$ or $[\mathrm{CLS}]$). Given a sequence of $m+1$ input tokens $x_0, x_1,\cdots, x_m$, the output of the encoder will be a sequence of $m+1$ vectors $\mathbf{h}_0^{L}, \mathbf{h}_1^{L},\cdots, \mathbf{h}_m^{L}$. Since $x_0$ does not correspond to an actual word and has a fixed positional embedding, it serves as a global representation token for collecting information from other positions via the self-attention mechanism. Hence, $\mathbf{h}_0^{L}$ acts as a representation of the entire sequence, without being biased toward any specific tokens or positions. In many cases, we require a single representation of a sequence to serve as input for downstream components of the system. For example, we can construct a sentence classification system based on a single vector generated from $\{\mathbf{h}_0^{L}, \dots, \mathbf{h}_m^{L}\}$. In this scenario, we can simply use $\mathbf{h}_0^{L}$ as the sequence representation. A more general approach is to add a pooling layer to the encoder, which allows us to explore various pooling methods to generate the sequence embedding from $\{\mathbf{h}_0^{L}, \cdots, \mathbf{h}_m^{L}\}$.

In text encoding, token sequences are represented by real-valued vectors, often referred to as sentence representations or sentence embeddings, which can be interpreted as points in a multi-dimensional space~\citep{hill-etal:2016learning}. Consequently, another use of text encoding is to compute the semantic or syntactic similarity of token sequences based on their relative proximity in this space. A straightforward method for this is to compute the Euclidean distance or cosine similarity between sequence embeddings. The shorter the distance between two sequences, the more similar they are considered to be. There are many distance metrics available, and it is possible to combine them to obtain a more robust measure of sequence similarity. Such similarity computations are applied in areas such as textual entailment, information retrieval, and translation evaluation~\citep{cer-etal:2018universal,reimers-etal:2019sentence}. Additionally, they are frequently used to assess the inherent quality of text encoding models.

\begin{figure}[!htp]
\centering
\input{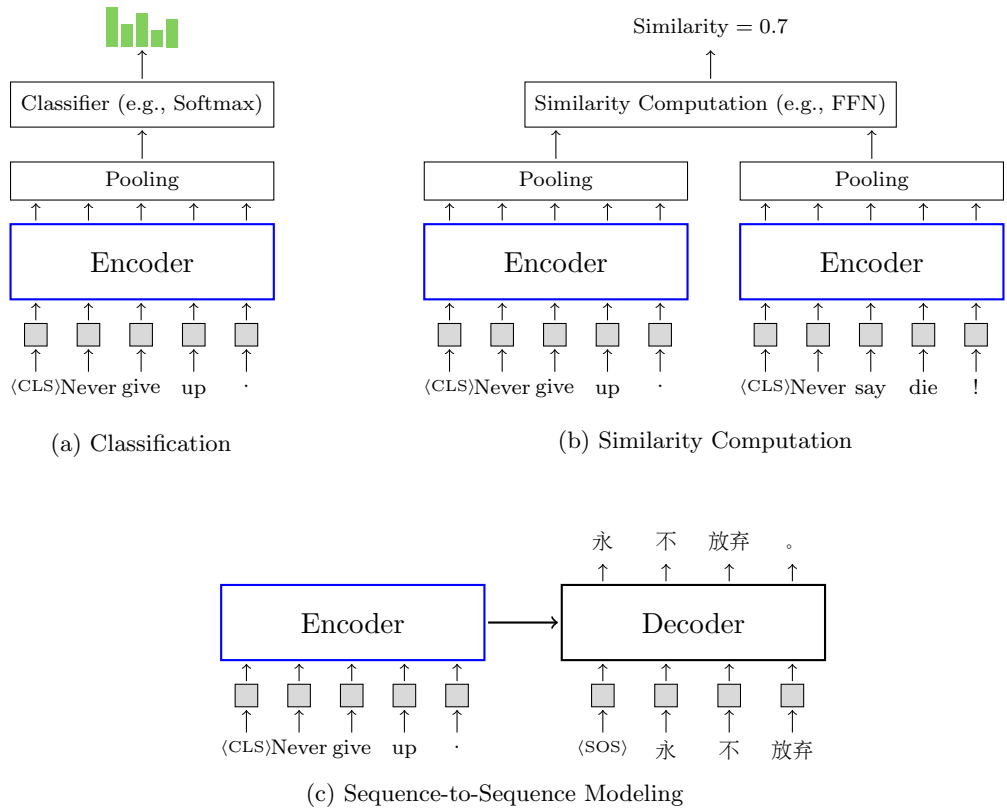}
\caption{Integrating Transformer encoders as components of different systems. A common approach is to feed the output of the encoder (with pooling) into a classifier to obtain a sequence classification system. Another way to utilize Transformer encoders is to compute the similarity between two sequences. We use the same encoder to represent the two sequences, and then build a neural network over the two representations to produce a similarity score between them. Furthermore, Transformer encoders can also be used in encoder-decoder systems to model sequence-to-sequence problems.}
\label{fig:transformer-encoders}
\end{figure}

Text encoding is also a crucial component of sequence-to-sequence models. In sequence-to-sequence tasks, we can develop a separate Transformer encoder for source-side sequence modeling within an encoder-decoder system (see Figure~\ref{fig:transformer-encoders}). For instance, we can pre-train a Transformer encoder on a large-scale corpus of source-side texts and subsequently use it as the encoder in a downstream encoder-decoder model. It is worth noting that while the encoder is based on the Transformer architecture, the decoder is not restricted to this architecture. This flexibility enables us to incorporate pre-trained Transformer encoders into hybrid sequence-to-sequence architectures, such as systems that combine a Transformer encoder with an LSTM decoder.

In supervised learning scenarios, training a Transformer encoder is straightforward. We can treat it as a regular component of the target model and train this model on task-specific labeled data. However, such a method requires the encoder to be optimized on each task, and the resulting encoder might not always generalize well to other tasks, especially given that labeled data is scarce in most cases. A more prevalent approach is to frame the training of text encoders as an independent task in which supervision signals are derived solely from raw text. This led researchers to develop self-supervised Transformer encoders, such as BERT, which make use of large-scale unlabeled text, and these encoders were found to generalize well across many downstream tasks. Further discussions of pre-trained Transformer encoders can be found in the literature on pre-trained models.

\subsection{Speech Translation}

\noindent As illustrated in Section~\ref{sec:transformer-the-basic-model}, the standard encoder-decoder Transformer model was proposed to model sequence-to-sequence problems. Here we consider the problem of translating speech in one language to text in another language --- a problem that is conventionally addressed using both automatic speech recognition (ASR) and machine translation techniques. Instead of cascading an automatic speech recognition system and a machine translation system, we can use Transformer models to build an end-to-end speech-to-text (S2T) translation system to directly translate the input speech to the output text.

To simplify the discussion, we assume that the input to an S2T translation system is a sequence of source-side acoustic feature vectors, denoted by $\mathbf{a}_1 \cdots \mathbf{a}_m$, and the output of the system is a sequence of target-side tokens, denoted by $y_1 \cdots y_n$.\footnote{In order to obtain the input sequence to the system, we need to discretize continuous speech into signals represented by feature vectors. This process is typically nontrivial, requiring either a feature extractor based on a variety of signal processing operations or a neural network that learns feature mappings in an end-to-end manner. But we will not dive into the details of these methods and simply treat the input feature extractor as an upstream system.} Mapping $\mathbf{a}_1 \cdots \mathbf{a}_m$ to $y_1 \cdots y_n$ is a sequence-to-sequence problem. Thus it is straightforward to model the problem using an encoder-decoder Transformer model, and the training and inference of this model are standard, as in neural machine translation.

In S2T translation, however, we have to deal with sequence mappings across both modalities and languages. This poses new challenges compared to conventional machine translation problems and influences the design of S2T translation models. There have been several improvements to Transformers to adapt them better for S2T translation tasks. Some improvements focus on the design of Transformer layers~\citep{di-etal:2019enhancing}. For example, in \citet{gulati-etal:2020conformer}'s system, a CNN sub-layer and relative positional embeddings are integrated into each Transformer layer, enabling the model to efficiently capture both local and global features.

Another line of research on S2T translation focuses on improving the encoder-decoder architecture. This involves modifying the encoder, the decoder, or both. To illustrate, Figure~\ref{fig:transformer-st-architectures} shows the architectures of three S2T translation models. All of them are based on Transformers but have different encoder architectures. As shown in the figure, the standard encoder-decoder architecture has one Transformer encoder for reading the source-side input $\mathbf{a}_1, \cdots, \mathbf{a}_m$ and one Transformer decoder for producing the target-side output $y_1, \cdots, y_n$. By contrast, the decoupled encoder model decomposes the encoder into two stacked encoders --- one for acoustic modeling (referred to as the \textbf{speech encoder}) and one for textual modeling (referred to as the \textbf{text encoder})~\citep{liu-etal:2020bridging,xu-etal:2021stacked}. This design reflects a modeling hierarchy in which representations at different levels of the network are concerned with different aspects of the problem. For example, the speech encoder models low-level features when mapping acoustic embeddings into larger language units, and the text encoder models the semantic or syntactic features when representing the entire input sequence. An advantage of separating out the text encoder is that the encoding process follows our prior knowledge: we need to first transcribe the speech input and then translate the transcript into the target language. Therefore, we can train the speech encoder in a manner similar to how we train an ASR system. This enables us to pre-train the speech encoder and the text encoder on unlabeled data and subsequently incorporate the pre-trained encoders into S2T translation systems.

\begin{figure}[!htp]
\centering

\begin{tikzpicture}

\def\boxsep{0.6cm}
\def\lsep{0.3cm}

\tikzstyle{enode} = [minimum width=2.5cm,minimum height=1.2cm,draw,thick];


\begin{scope}

\node [enode,anchor=west,align=center,draw=blue] (encoder) at (0,0) {Encoder\\ (Speech) };
\node [anchor=north] (input) at ([yshift=-\lsep]encoder.south) {\small{Speech (Source)}};
\draw [->] ([yshift=-2pt]input.north) -- ([yshift=-2pt]encoder.south);

\node [enode,anchor=west] (decoder) at ([xshift=1.5*\boxsep]encoder.east) {Decoder};
\node [anchor=south] (output) at ([yshift=\lsep]decoder.north) {\small{Text (Target)}};
\draw [->] ([yshift=2pt]decoder.north) -- ([yshift=2pt]output.south);

\draw [->,thick] ([xshift=1pt]encoder.east) -- ([xshift=-1pt]decoder.west);

\node [anchor=north west] (caption) at ([yshift=-1.2cm,xshift=3.0cm]encoder.south west) {\small{(a) Single Encoder + Single Decoder}};

\end{scope}


\begin{scope}[yshift=-4.5cm]

\node [enode,anchor=west,align=center,draw=blue] (encoder) at (0,0) {Encoder\\(Speech)};
\node [anchor=north] (input) at ([yshift=-\lsep]encoder.south) {\small{Speech (Source)}};
\draw [->] ([yshift=-2pt]input.north) -- ([yshift=-2pt]encoder.south);

\node [enode,anchor=west,align=center,draw=red] (tencoder) at ([xshift=1.5*\boxsep]encoder.east) {Encoder\\(Text)};
\node [anchor=north] (tinput) at ([yshift=-\lsep]tencoder.south) {\small{Transcript (Source)}};

\node [enode,anchor=west] (decoder) at ([xshift=1.5*\boxsep]tencoder.east) {Decoder};
\node [anchor=south] (output) at ([yshift=\lsep]decoder.north) {\small{Text (Target)}};
\draw [->] ([yshift=2pt]decoder.north) -- ([yshift=2pt]output.south);

\draw [->,thick] ([yshift=1pt]encoder.north) -- ([yshift=\lsep]encoder.north) -- ([yshift=\lsep,xshift=0.75*\boxsep]encoder.north east) -- ([yshift=-\lsep,xshift=0.75*\boxsep]encoder.south east) -- ([yshift=-\lsep]tencoder.south) -- ([yshift=-1pt]tencoder.south);
\draw [->,dashed,thick] ([yshift=\lsep,xshift=0.75*\boxsep]encoder.north east) -- ([yshift=\lsep,xshift=0.75*\boxsep]tencoder.north east) -- ([yshift=0.2cm,xshift=0.75*\boxsep]tencoder.east) -- ([yshift=0.2cm,xshift=-1pt]decoder.west);
\draw [->,thick] ([xshift=1pt]tencoder.east) -- ([xshift=-1pt]decoder.west);

\node [anchor=north west] (caption) at ([yshift=-1.2cm,xshift=3.0cm]encoder.south west) {\small{(b) Decoupled Encoder + Single Decoder}};

\end{scope}


\begin{scope}[yshift=-9cm]

\node [enode,anchor=west,align=center,draw=blue] (encoder) at (0,0) {Encoder\\(Speech)};
\node [anchor=north] (input) at ([yshift=-\lsep]encoder.south) {\small{Speech (Source)}};
\draw [->] ([yshift=-2pt]input.north) -- ([yshift=-2pt]encoder.south);

\node [enode,anchor=west,align=center,draw=red] (tencoder) at ([xshift=1.5*\boxsep]encoder.east) {Encoder\\(Text)};
\node [anchor=north] (tinput) at ([yshift=-\lsep]tencoder.south) {\small{Transcript (Source)}};
\draw [->] ([yshift=-2pt]tinput.north) -- ([yshift=-2pt]tencoder.south);

\node [enode,anchor=west,align=center,draw=ugreen] (sencoder) at ([xshift=1.5*\boxsep]tencoder.east) {Shared\\Encoder};

\node [enode,anchor=west] (decoder) at ([xshift=1.5*\boxsep]sencoder.east) {Decoder};
\node [anchor=south] (output) at ([yshift=\lsep]decoder.north) {\small{Text (Target)}};
\draw [->] ([yshift=2pt]decoder.north) -- ([yshift=2pt]output.south);

\draw [->,thick] ([yshift=1pt]encoder.north) -- ([yshift=\lsep]encoder.north) -- ([yshift=\lsep,xshift=0.75*\boxsep]encoder.north east) -- ([yshift=\lsep,xshift=0.75*\boxsep]tencoder.north east) -- ([yshift=0.2cm,xshift=0.75*\boxsep]tencoder.east) -- ([yshift=0.2cm,xshift=-1pt]sencoder.west);
\draw [->,thick] ([xshift=1pt]tencoder.east) -- ([xshift=-1pt]sencoder.west);
\draw [<->,thick,dotted] ([xshift=1pt]encoder.east) -- ([xshift=-1pt]tencoder.west);
\draw [->,thick] ([xshift=1pt]sencoder.east) -- ([xshift=-1pt]decoder.west);

\node [anchor=north west] (caption) at ([yshift=-1.2cm,xshift=3.0cm]encoder.south west) {\small{(c) Two-stream Encoder + Single Decoder}};

\end{scope}

\end{tikzpicture}
\caption{Architectures of speech-to-text translation models based on Transformers. In addition to the standard encoder-decoder architecture, we can explicitly model the acoustic and textual (semantic) information using two separate encoders, called the speech encoder and the text encoder. In the decoupled encoder architecture, the two encoders are stacked, that is, text encoding is a subsequent process after speech encoding. In the two-stream encoder architecture, the two encoders work in parallel, and their outputs are merged using an additional encoder, called the shared encoder. The dotted line indicates the potential for interaction between the two encoders. For example, we could define a loss function to minimize the difference between their outputs, thereby guiding the model towards more aligned representations.}
\label{fig:transformer-st-architectures}
\end{figure}

An alternative encoding architecture is the two-stream architecture, as shown in Figure~\ref{fig:transformer-st-architectures}(c). Like the decoupled encoder architecture, this architecture has a speech encoder and a text encoder, but the two encoders work in parallel rather than in sequence~\citep{ye-etal:2021end}. The speech encoder takes acoustic features as input, and the text encoder takes tokens (or their embeddings) as input. A third encoder, called the \textbf{shared encoder}, integrates the outputs from both the speech and text encoders, merging the representations from the two modalities. This two-stream architecture is flexible because it provides multiple ways to train S2T translation models. A common approach is to train each branch individually. For example, if we mask the speech encoder, the model will transform into a machine translation model which can be trained using bilingual texts. Conversely, if we mask the text encoder, we can train the model as a standard S2T translation model. For inference, the text encoder can be dropped, and the speech input is modeled using the speech encoder and the shared encoder.

In deep learning, training is often intertwined with architecture design. Because we are dealing with data in two modalities and two languages, we can develop multiple supervision signals for the multi-task learning of S2T translation models. A widely used method is to introduce an ASR-related loss into the training of speech encoders. For example, in the decoupled encoder model, a classifier can be constructed based on the output from the speech encoder. By minimizing the connectionist temporal classification (CTC) loss for this classifier, the speech encoder can be optimized in a manner similar to ASR. In general, training S2T translation models is challenging because speech-to-text aligned data is scarce. Typical responses to this challenge include data augmentation, pre-training, knowledge distillation with machine translation, and so on. However, an in-depth discussion of these methods goes beyond the scope of this section on Transformers. Interested readers can refer to a recent survey on speech translation for more detailed information~\citep{xu-etal:2023recent}.

\subsection{Vision Models}

\noindent While Transformers were first used in NLP, their application to other domains has been a prominent research topic. In computer vision, for instance, there is a notable trend of shifting from CNNs to Transformers as backbone models. In this subsection, we consider the \textbf{Vision Transformer} (\textbf{ViT}) --- an interesting application of Transformers to image classification~\citep{dosovitskiy-etal:2020image}. The Vision Transformer is a milestone model that opened the door to purely Transformer-based vision models. Here, we consider the basic structure of the Vision Transformer to keep this section focused and coherent, although extensive literature exists on the Vision Transformer and its variants. More detailed discussions of vision models can be found in recent surveys~\citep{han-etal:2022survey,liu-etal:2023survey}.

The core idea behind the Vision Transformer is to transform an image into a sequence of visual tokens, and input this sequence into a Transformer encoder to generate a representation of the image. The Transformer encoder is standard, and so we will not discuss it here, given the introduction to Transformers we have presented so far in this chapter. Mapping a 2D image into a sequence of tokens requires additional processing. Suppose we have an image represented as an $H \times W \times C$ feature map, where $H$ is the height, $W$ is the width, and $C$ is the number of channels. The first step is to partition this image into a set of \textbf{patches}. Suppose all patches are squares of side length $P$. Then, the resulting patches can be represented by feature maps of shape $P \times P \times C$. By ordering these patches in a specific manner, we obtain a sequence of $\frac{HW}{P^2}$ patches, with each patch being treated as a ``token.''

\begin{figure}[!htp]
\centering
\input{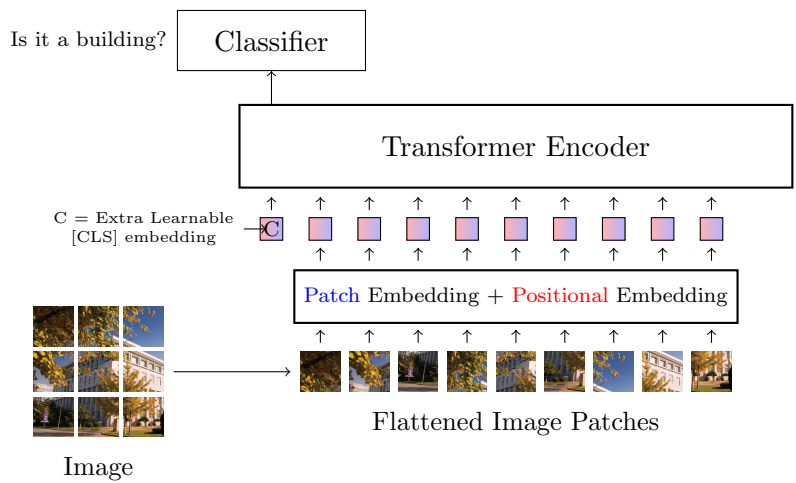}
\caption{Illustration of the Vision Transformer for image classification~\citep{dosovitskiy-etal:2020image}. There are three steps. In the first step, the input image is partitioned into patches, which are then flattened and mapped into embeddings. In the second step, a Transformer encoder is employed to process the sequence of embeddings, representing the image as a real-valued vector (e.g., the output of the encoder at the first position). In the last step, a classifier is built on top of this image representation.}
\label{fig:vision-transformer}
\end{figure}

Given this patch sequence, the subsequent steps are straightforward. For the patch at each position, we obtain a $d$-dimensional embedding via a linear transformation of the flattened patch. The input to the Transformer encoder is a sequence of $d$-dimensional vectors, each of which is the sum of the corresponding patch embedding and positional embedding. Figure~\ref{fig:vision-transformer} illustrates the patching and embedding steps in the Vision Transformer.

Once we have a sequence of vectors for representing the image, we can employ the Transformer encoder to encode the sequence. The encoding process is exactly the same as that in text encoding, as discussed in Section~\ref{sec:transformer-text-encoding}. For classification problems, we need only a single representation of the input. It is convenient to take the output of the encoder at position 0 (denoted by $\mathbf{h}_0^{L}$) and feed it into a classifier. Given that the first token ($[\mathrm{CLS}]$) serves as a special global token that is attended to by all other tokens, $\mathbf{h}_0^{L}$ can be viewed as an aggregated representation of the entire sequence.

A standard way to train the Vision Transformer is to minimize a task-specific loss on labeled data, such as ImageNet. More recently, inspired by self-supervised learning in BERT-like models, there have been successful attempts to train Transformer-based image encoders on large-scale unlabeled data~\citep{caron-etal:2021emerging,bao-etal:2021beit,he-etal:2022masked}. Note that one of the most significant contributions of the Vision Transformer is that it unifies the representation models for different modalities. This suggests that if an object, whether an image or text, is represented as a sequence of embeddings, it can be easily modeled using the Transformer architecture.

\subsection{Multimodal Models}

\noindent The above discussion of the Vision Transformer offers the possibility of unifying the representations from multiple modalities using the same Transformer architecture. In fact, many recent multimodal systems draw heavy inspiration from Transformers~\citep{xu-etal:2023multimodal}. Such systems convert objects from different modalities into vector sequences and feed these vectors into a single Transformer model. The output is a fused representation of all inputs, which can then be used in downstream systems.

As a simple example, consider the task of encoding a pair consisting of text and its corresponding image. First, we represent both the text and the image as sequences of embeddings that have the same dimensionality. This is a common step in sequence modeling, which we have encountered many times so far. We can do this by using either a simple embedding model (e.g., a word or patch embedding model) or a pre-trained model (e.g., a vision model). Then, these two sequences are concatenated into a long sequence comprising both textual and visual embeddings. The subsequent step is standard: a Transformer encoder takes the concatenated sequence of embeddings as input and produces representations of the text and image as output. Note that concatenating textual and visual sequences is one of the simplest methods for vision-text modeling. There are several alternative ways to merge information from different modalities. For example, we can feed visual representations into the attention layers of a text encoder or decoder~\citep{li-etal:2022blip,alayrac-etal:2022flamingo}.

The above multimodal encoder can be used in both encoder-only and encoder-decoder systems. For encoder-only systems, consider an example where, given an image and its description, we predict the class of the image using a classifier built on top of the encoder~\citep{kim-etal:2021vilt}. For encoder-decoder systems, we pair the encoder with a decoder, as in sequence-to-sequence modeling~\citep{cho-etal:2021unifying}. For example, we might employ a Transformer decoder to generate text based on the output of the encoder. A common application of this architecture is \textbf{visual question answering} (\textbf{VQA}), where an image and a question about the image are provided, and the system is tasked with generating an answer~\citep{antol-etal:2015vqa}. The architectures of these models are illustrated in Figure~\ref{fig:vision-text-models} (a-b).

\begin{figure}[!htp]
\centering
\input{./Figures/figure-multi-modal-transformer}
\caption{Vision-text models. Blue boxes represent word+position embeddings, and red boxes represent image patch+position embeddings.}
\label{fig:vision-text-models}
\end{figure}

Large language models have also been extended to handle both textual and other modalities, such as images, video, and audio, leading to new advances in multimodal processing~\citep{liu-etal:2023visual,yin-etal:2023survey}. By representing all inputs as a sequence of token embeddings, the problem becomes straightforward: we predict the next token given its context. This can be done using decoder-only systems, as shown in Figure~\ref{fig:vision-text-models} (c).


\section{Summary}

\noindent Transformer models have been widely adopted since the architecture was first proposed by \citet{Vaswani-etal:2017Transformer}. This adoption has accelerated the development of a wide range of algorithms, systems, and concepts. A thorough discussion of Transformers requires broad coverage, and so it is impossible to address all topics or provide an exhaustive list of references. Figure~\ref{fig:overview-of-transformers} provides a high-level overview of the Transformer landscape. Note that these models and related techniques can be categorized in various ways, and we present one such perspective. To summarize, we highlight the following points:

\begin{figure}[!htp]
\centering
\includegraphics[scale=0.48]{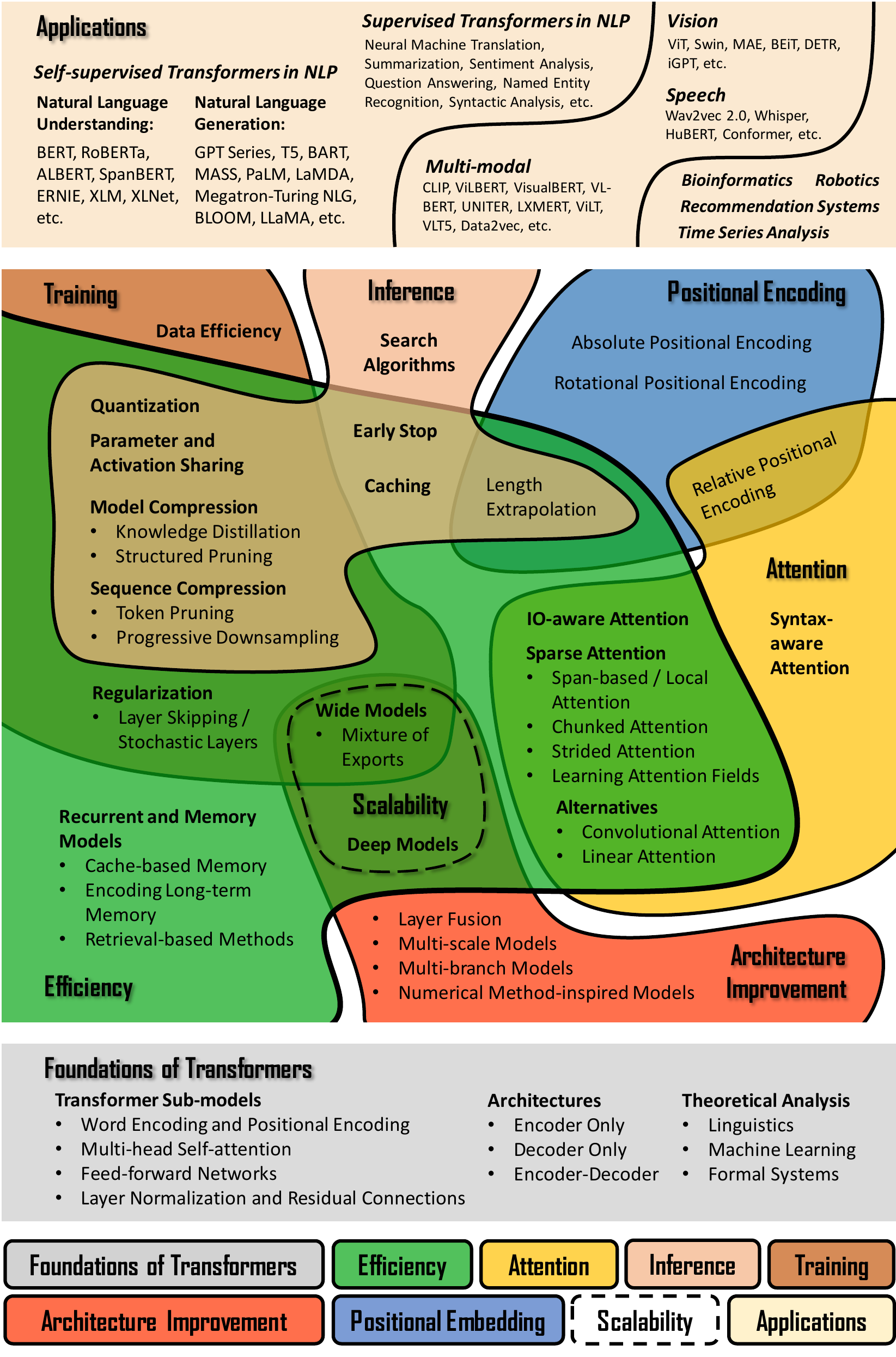}
\caption{An overview of Transformers.}
\label{fig:overview-of-transformers}
\end{figure}

\begin{itemize}
\item \vspace{0.5em} \textbf{Foundations of Transformers}. Although the impact of Transformers has been revolutionary, they are not completely ``new'' models. From a deep learning perspective, Transformers are composed of common building blocks, including word and positional embeddings~\citep{Bengio-et-al:2003,Mikolov-et-al:2013distributed,gehring-etal:2017convolutional}, attention mechanisms~\citep{bahdanau2014neural,luong-etal:2015effective}, residual connections~\citep{he-etal:2016identity}, layer normalization~\citep{Ba-etal:2016layer}, and so on. Many of these components were presented in earlier systems, for example, similar ideas with QKV attention can be found in memory networks~\citep{sukhbaatar-etal:2015end} and hierarchical attention networks~\citep{yang-etal:2016hierarchical}. Transformers offer a novel approach to integrating these components, resulting in a unique architecture. For example, in Transformers, the combination of multi-head attention and dot-product QKV attention, along with the incorporation of layer normalization and residual connections, gives rise to a distinctive neural network block, specifically a self-attention sub-layer. This design has since become a de facto standard in many follow-on sequence modeling systems.

\item \vspace{0.3em} \textbf{Attention Models}. The success of Transformers on NLP tasks has largely been attributed to the use of multi-head self-attention for sequence modeling. This has led to a surge of interest in enhancing the attention mechanisms within Transformers. While it is impossible to detail every attention model, there are several notable research directions. One prominent direction involves modifying the forms of QKV attention and multi-head attention for improved performance. The scope of this direction is vast, as there are numerous aspects to consider when enhancing Transformers~\citep{lin-etal:2022survey}. For example, one may add new components to self-attention sub-layers to adapt them to specific tasks, resulting in various Transformer variants. A second direction is to incorporate prior knowledge into the design of attention models. This makes sense, because much of the emphasis in traditional NLP has been on using linguistic insights to guide system design, and we generally want NLP systems to be linguistically explainable. For example, many Transformer-based systems take syntactic parses as input in various forms and make use of syntax in sequence modeling. A third direction is to develop efficient attention models~\citep{Tay-etal:2020efficient}. Self-attention has long been criticized for its quadratic time complexity and dependency on all previous  tokens for each new token. In response, many researchers have focused on simplifying the structure of self-attention, or on approximating it using sparse or recurrent models. This concern for efficiency also motivates the development of alternatives to self-attention, such as attention models with linear time complexity. In addition to exploring stronger and more efficient attention models, it is natural to examine what knowledge is learned by such models. Interestingly, researchers have found that the underlying structure of languages can be learned by multi-head self-attention models, although these models are not trained to represent such knowledge~\citep{manning-etal:2020emergent}.

\item \vspace{0.3em} \textbf{Word and Positional Embeddings}. Transformers represent each input word (or token) through a combination of word and positional embeddings. Learning these word embeddings is not unique to Transformers; one can either utilize pre-trained embeddings, such as Word2Vec or GloVe, or treat them as learnable parameters within the model. A closely related issue is the tokenization of input sequences. Generally, tokenization determines the number of resulting tokens and affects the difficulty of learning their corresponding embeddings. Therefore, selecting an appropriate tokenization method is crucial for many applications. Furthermore, positional embeddings play a vital role in Transformers, as the underlying attention mechanisms are order-insensitive by design~\citep{dufter-etal:2022position}. While positional encoding is a general research topic, much effort has been dedicated to tailoring it for Transformers, leading to various architectural modifications~\citep{shaw-etal:2018self,huang-etal:2018music}. Additionally, studies show that for sequences significantly longer than those encountered during training, better extrapolation can be achieved by replacing sinusoidal embeddings with rotary positional embeddings (RoPE) or by scaling attention weights with positional scalars~\citep{raffel-etal:2020exploring,su-etal:2021roformer,press-etal:2021train}.

\item \vspace{0.3em} \textbf{Training and Model Scaling}. In the era of deep learning, high-performance systems are typically built using large-scale neural networks. A straightforward approach to increasing Transformer capacity is to stack more layers or enlarge the hidden representation size. Indeed, deep and wide Transformer models consistently outperform their smaller counterparts. However, training extremely large models poses significant challenges, particularly when applying gradient descent over vast datasets, which demands substantial computational resources. Engineering solutions involve distributing training across computer clusters~\citep{lepikhing-etal:2021shard,chowdhery-etal:2022palm}. Although distributed training is a general paradigm not restricted to Transformers, it heavily influences architectural design; for instance, sparse mixture-of-experts (MoE) models can facilitate training with massive parameter sets and serve as the foundation for many expansive systems. Scaling also enables the study of scaling laws --- quantifying how performance scales with model size, data volume, and compute budget~\citep{hestness-etal:2017deep,kaplan-etal:2020scaling}. This scaling is occasionally accompanied by ``emergence''~\citep{wei-etal:2022emergent}, where models manifest unpredictable new abilities. In recent research, the acquisition of such emergent abilities has been viewed as a hallmark of powerful large language models.

\item \vspace{0.3em} \textbf{Efficient Models}. There are different goals for efficiency. A system may prioritize memory efficiency for memory-bound problems or low latency when inference speed is critical. Balancing these requirements leads to diverse optimization strategies. In Transformers, many optimizations involve modifying attention mechanisms; for instance, several variants reduce the memory footprint when processing long sequences~\citep{Tay-etal:2020efficient}, while others minimize computation to decrease latency. Furthermore, Transformers can benefit from architecture-independent optimizations typical of neural networks, such as conditional computation, knowledge distillation, structured pruning, and sequence compression. Efficiency can also be addressed from a computer architecture perspective~\citep{kim-etal:2023full}. In sequence-to-sequence tasks, for example, the encoding process is typically compute-intensive, whereas the autoregressive decoding process is often IO-intensive. Therefore, we can employ different optimization methods for different components of Transformers.

\item \vspace{0.3em} \textbf{Inference}. The inference problem is commonly discussed in sequence generation. In NLP, we often need to find the ``best'' hypothesis in a space involving sequences of hundreds or even thousands of tokens over a vocabulary. As a classic search problem in artificial intelligence, various algorithms are applicable, including breadth-first, depth-first, and A* search. In many practical applications of NLP, the efficiency of the search systems is an important consideration. As a result, optimized search algorithms are required. Most of these algorithms have been explored in machine translation and ASR, and are directly applicable to neural text generation models like Transformers. Beyond traditional search, recent optimizations such as speculative decoding are specifically tailored to the Transformer architecture~\citep{leviathan-etal:2023fast}. Moreover, the efficient attention models mentioned previously are now widely deployed to accelerate large language models and neural machine translation systems~\citep{heafield-etal:2021findings,dao-etal:2023flashdecoding}.

\item \vspace{0.3em} \textbf{Applications}. Applications of Transformers cover a wide variety of NLP problems. During the development of Transformers, they were at first used to build supervised models that perform particular tasks. Later, a greater success was achieved by using them as backbone networks for large scale self-supervised learning of foundation models~\citep{bommasani-etal:2021foundation}. This shift markedly changed the NLP paradigm: a model is first pre-trained on vast amounts of text to acquire general linguistic knowledge, and then adapted to downstream tasks through efficient methods such as fine-tuning or prompting. Beyond NLP, we have witnessed an explosion of Transformer applications in fields like computer vision, speech processing, and bioinformatics. The core idea is to represent diverse data types as sequences of tokens, extending Transformers into general-purpose representation models capable of handling multi-modal data.

\item \vspace{0.3em} \textbf{Large Language Models as Foundation Models}. Transformers form the basis of recent large language models, such as the GPT series, which show surprising breakthroughs in NLP~\citep{bubeck-etal:2023sparks,yang-etal:2023dawn}. Much of the research in large language models is more or less related to Transformers. As discussed in Section~\ref{sec:transformer-app-language-modeling}, training these language models is often synonymous with training large-scale Transformer decoders, and architectural refinements to the latter directly benefit the former. On the other hand, the rapid development of large language models has spurred further innovation in Transformer-related techniques, particularly in developing efficient and low-cost adaptation methods for massive models.

\item \vspace{0.3em} \textbf{Theoretical Analysis}. While Transformers dominate empirically, their theoretical foundations have historically received less focus compared to engineering improvements --- an empirical-first trend common across the broader machine learning community. In response, researchers have begun to analyze Transformers more deeply. One approach is to view Transformers as deep neural networks and interpret them via mathematical tools. For instance, the residual connections within Transformers can be viewed as Euler discretizations of ODEs. This equivalence suggests that we can leverage insights from numerical ODE methods to inform model design. Another promising avenue aims to develop a theoretical understanding of the self-attention mechanism, which distinguishes Transformers from earlier deep learning models. Studies have interpreted self-attention through various theoretical lenses, including data compression~\citep{yu-etal:2023white}, optimization~\citep{li-etal:2022theoretical}, and function approximation~\citep{yun-etal:2019transformers}. Moreover, Transformers can be mapped to formal systems, such as Turing machines~\citep{perez-etal:2018turing}, counter machines~\citep{bhattamishra-etal:2020ability}, regular and context-free languages~\citep{hahn-etal:2020theoretical}, Boolean circuits~\citep{hao-etal:2022formal,merrill-etal:2022saturated}, programming languages~\citep{weiss-etal:2021thinking}, and first-order logic~\citep{chiang-etal:2023tighter}. These connections provide theoretical tools to rigorously study the expressivity of Transformers. It is worth noting, however, that while we can analyze Transformers through these diverse perspectives, no unified theory yet explains the fundamental nature of these models. Addressing this theoretical gap remains a central challenge in the field, yet it is essential for developing complex neural systems with explainable and predictable behaviors.

\end{itemize}

\vspace{0.3em}


\section*{Acknowledgements}

\noindent We would like to thank Yongyu Mu, Chenglong Wang, Bei Li, Weiqiao Shan, Yuchun Fan, Kaiyan Chang, Tong Zheng, and Huiwen Bao for their suggestions on improving the early version of this work.

\appendix
 \renewcommand{\thesection}{\Alph{section}}
\renewcommand{\thesubsection}{\Alph{section}.\arabic{subsection}}
\renewcommand{\thesubsubsection}
  {\Alph{section}.\arabic{subsection}.\arabic{subsubsection}}

\addappheadtotoc

\section{Sinusoidal Positional Encoding}
\label{sec:appendix-sinusoidal-positional-encoding}

\noindent In Transformers, positions are represented as vectors. Although vectorizing the representations of positions sounds complicated, a simple idea is to use a carrying system which describes how a natural number is expressed by a polynomial with respect to a base \citep{kernes:2021master}. For example, $i$ can be written as
\begin{eqnarray}
i & = & \sum_{k=0}^{k_{\mathrm{max}}} a(i,k) b^k \label{eq:pos-sin-encoding-carry-math}
\end{eqnarray}

\noindent where $a(i,k)$ is the $k$-th digit, $k_{\mathrm{max}} + 1$ is the maximum number of digits, and $b$ is the base of the system. The carrying occurs when $a(i,k)$ reaches $b$: we increase $a(i,k+1)$ by 1 and roll back $a(i,k)$ to 0. In this way we can change $a(i,k)$ with a period of $b^k$, that is, $a(i,0)$ changes with a period of $b^0$, $a(i,1)$ changes with a period of $b^1$, $a(i,2)$ changes with a period of $b^2$, and so on.

Using this system, $i$ can be represented as a vector
\begin{eqnarray}
\mathrm{PE}(i) & = & \begin{bmatrix} a(i,0) & a(i,1) & ... & a(i,k_{\mathrm{max}})\end{bmatrix} \label{eq:pos-sin-encoding-carrying-vector}
\end{eqnarray}

\noindent For example, when $b=2$, $\mathrm{PE}(11) = \begin{bmatrix} 1 & 1 & 0 & 1\end{bmatrix}$. However, in Eq. (\ref{eq:pos-sin-encoding-carrying-vector}), $\mathrm{PE}(i)$ is still a discrete function. We may want a continuous vector representation that can describe intermediate states between discrete events. Considering $a(i,k)$ as a periodic function, a common choice is the sine function. Thus $a(i,k)$ can be re-defined, as follows
\begin{eqnarray}
a(i,k) & = & \mathrm{sin}(i \cdot \omega_k) \label{eq:eq:pos-sin-encoding-sin}
\end{eqnarray}

\noindent This function has an amplitude of 1 and a period of $\frac{2 \pi}{\omega_k}$. Using an analogous form of periods to that used in Eq. (\ref{eq:pos-sin-encoding-carry-math}), we define $\omega_k$ as
\begin{eqnarray}
\omega_k & = & \frac{1}{(b_{\textrm{model}})^{k/d_{\textrm{model}}}}
\end{eqnarray}

\noindent where $b_{\textrm{model}} > 0$ and $d_{\textrm{model}}>0$ are hyper-parameters of the model. Obviously, we have $\frac{2 \pi}{\omega_0} < \frac{2 \pi}{\omega_1} < ... < \frac{2 \pi}{\omega_{k_{\mathrm{max}}}}$.

Similarly, we can define $a(i,k)$ via the cosine function
\begin{eqnarray}
a(i,k) & = & \mathrm{cos}(i \cdot \omega_k) \label{eq:eq:pos-sin-encoding-cos}
\end{eqnarray}

Taking both Eqs. (\ref{eq:eq:pos-sin-encoding-sin}) and (\ref{eq:eq:pos-sin-encoding-cos}), we create a new representation of $i$, as follows
\begin{eqnarray}
\mathrm{PE}(i) & = & \begin{bmatrix} \mathrm{sin}(i \cdot \omega_0) & \mathrm{cos}(i \cdot \omega_0) & ... &  \mathrm{sin}(i \cdot \omega_{k_{\mathrm{max}}}) & \mathrm{cos}(i \cdot \omega_{k_{\mathrm{max}}}) \end{bmatrix}
\end{eqnarray}

\citet{Vaswani-etal:2017Transformer} instantiated the above form by setting $b_{\textrm{model}} = 10,000$. Let $\mathrm{PE}(i,k)$ be the $k$-th dimension of $\mathrm{PE}(i)$. \citet{Vaswani-etal:2017Transformer}'s version of positional encoding is written as
\begin{eqnarray}
\mathrm{PE}(i,2k) & = & \mathrm{sin}(i \cdot \frac{1}{10000^{2k/d_{\mathrm{model}}}}) \label{eq:eq:pos-sin-encoding-sin-transformer} \\
\mathrm{PE}(i,2k+1) & = & \mathrm{cos}(i \cdot \frac{1}{10000^{2k/d_{\mathrm{model}}}}) \label{eq:eq:pos-sin-encoding-cos-transformer}
\end{eqnarray}

\begin{figure}[!t]
\centering
\includegraphics[scale=0.33]{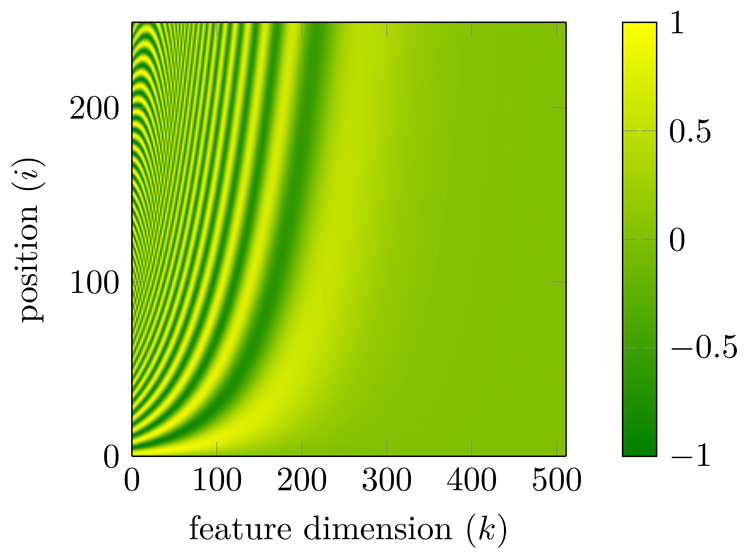}
\caption{A heat map of the positional embedding model of Eqs. (\ref{eq:eq:pos-sin-encoding-sin-transformer}) and (\ref{eq:eq:pos-sin-encoding-cos-transformer}) ($b_{\textrm{model}} = 10,000$ and $d_{\textrm{model}} = 512$). Consider a position $i$ (i.e., the $i$-th row), then move another position $j$ from $i$ upwards or downwards. Intuitively, when $i$ and $j$ are closer, the corresponding row vectors are more similar. By contrast, when $j$ moves away from $i$, the similarity is not that obvious. This property helps explain the idea behind the positional embedding model: the ``distance'' between two positions is implicitly modeled by comparing their multi-dimensional representations. }
\label{fig:sinusoidal-pos-encoding-heat-map}
\end{figure}

Choosing $b_{\textrm{model}} = 10,000$ is empirical. One can adjust it for specific tasks. Figure \ref{fig:sinusoidal-pos-encoding-heat-map} plots the positional encoding for different positions. We see that, when $k$ becomes larger,  the change of the color follows a larger period.

Note that Eqs. (\ref{eq:eq:pos-sin-encoding-sin-transformer}) and (\ref{eq:eq:pos-sin-encoding-cos-transformer}) have a useful property that $\mathrm{PE}(i+\mu)$ can be easily expressed by a linear function of $\mathrm{PE}(i)$ for a given offset $\mu$\footnote{One can derive these by taking
\begin{eqnarray}
\mathrm{sin}(\alpha+\beta) & = & \mathrm{sin} (\alpha) \cdot \mathrm{cos} (\beta) + \mathrm{cos} (\alpha) \cdot \mathrm{sin} (\beta) \\
\mathrm{cos}(\alpha+\beta) & = & \mathrm{cos} (\alpha) \cdot \mathrm{cos} (\beta) - \mathrm{sin} (\alpha) \cdot \mathrm{sin} (\beta)
\end{eqnarray}
}
\begin{eqnarray}
\mathrm{PE}(i+\mu,2k) & = & \mathrm{PE}(i,2k) \cdot \mathrm{PE}(\mu,2k+1) + \nonumber \\
& & \mathrm{PE}(i,2k+1) \cdot \mathrm{PE}(\mu,2k) \\
\mathrm{PE}(i+\mu,2k+1) & = & \mathrm{PE}(i,2k+1) \cdot \mathrm{PE}(\mu,2k+1) - \nonumber \\
& & \mathrm{PE}(i,2k) \cdot \mathrm{PE}(\mu,2k)
\end{eqnarray}

\noindent The resulting benefit is that the encoding can somewhat model relative positions. That is, the state at position $i + \mu$ can be described by starting with $i$ and then appending it with the offset $\mu$.

When applying the sinusoidal positional encoding, one way is to concatenate $\mathbf{x}_i$ and $\mathrm{PE}(i)$. In \citet{Vaswani-etal:2017Transformer}'s work, they instead assume $\mathrm{PE}(i)$ to be a vector of the same size as $\mathbf{x}_i$ (i.e., $|\mathrm{PE}(i)|= |\mathbf{x}_i| = d_e$), and add $\mathrm{PE}(i)$ to $\mathbf{x}_i$, like this
\begin{eqnarray}
\mathbf{xp}_i & = & \mathbf{x}_i + \mathrm{PE}(i)
\end{eqnarray}

\noindent This sinusoidal addictive model has been the basis of many positional encoding approaches \citep{dehghani-etal:2018universal,likhomanenko-etal:2021cape,su-etal:2021roformer}.

\vskip 0.2in
\bibliography{bibli}
\end{CJK}

\end{document}